\documentclass[10pt,twocolumn,letterpaper]{article}

\usepackage{cvpr}              

\usepackage{times}
\usepackage{epsfig}
\usepackage{graphicx}
\usepackage{amsmath}
\usepackage{amssymb}

\usepackage{xcolor}         
\usepackage{float}
\usepackage{caption}
\usepackage{amsfonts,amssymb}
\usepackage{mathrsfs} 
\usepackage{tabularx}
\usepackage{wrapfig}
\usepackage{verbatim}
\usepackage{dsfont}
\usepackage{tablefootnote}
\usepackage[stable]{footmisc}
\usepackage{booktabs}
\usepackage{multicol}
\usepackage{multirow}
\usepackage{color}
\usepackage{subcaption}
\usepackage{pythonhighlight}

\newcommand{\rulesep}{\color{black} \unskip\ \vrule\ }

\definecolor{todocolor}{RGB}{255,0,00}

\definecolor{jiapeng}{rgb}{0.2, 0.4,0.9}

\newcommand{\expnumber}[2]{{#1}\mathrm{e}{#2}}
\DeclareMathOperator{\IoU}{IoU}

\DeclareMathOperator{\CD}{CD}

\DeclareMathOperator{\KL}{KL}

\definecolor{cvprblue}{rgb}{0.21,0.49,0.74}
\usepackage[pagebackref,breaklinks,colorlinks,citecolor=cvprblue]{hyperref}

\def\paperID{3704} 
\def\confName{CVPR}
\def\confYear{2024}

\begin{document}

\title{DiffuScene: Denoising Diffusion Models for Generative Indoor Scene Synthesis}

\author{
    Jiapeng Tang$^1$ \quad Yinyu Nie$^1$ \quad Lev Markhasin$^2$ \quad Angela Dai$^1$ \quad Justus Thies$^3$ \quad Matthias Nie{\ss}ner$^1$ \\
    \\
     $^{1}$ Technical University of Munich \quad
     $^{2}$ Sony Europe RDC Stuttgart \\ 
     $^{3}$ Technical University of Darmstadt \\
     \url{https://tangjiapeng.github.io/projects/DiffuScene} \\
}

\twocolumn[{%
	\renewcommand\twocolumn[1][]{#1}%
	\maketitle
	\begin{center}
            \vspace{-6mm}
		\centerline{
                 \includegraphics[width=\linewidth]{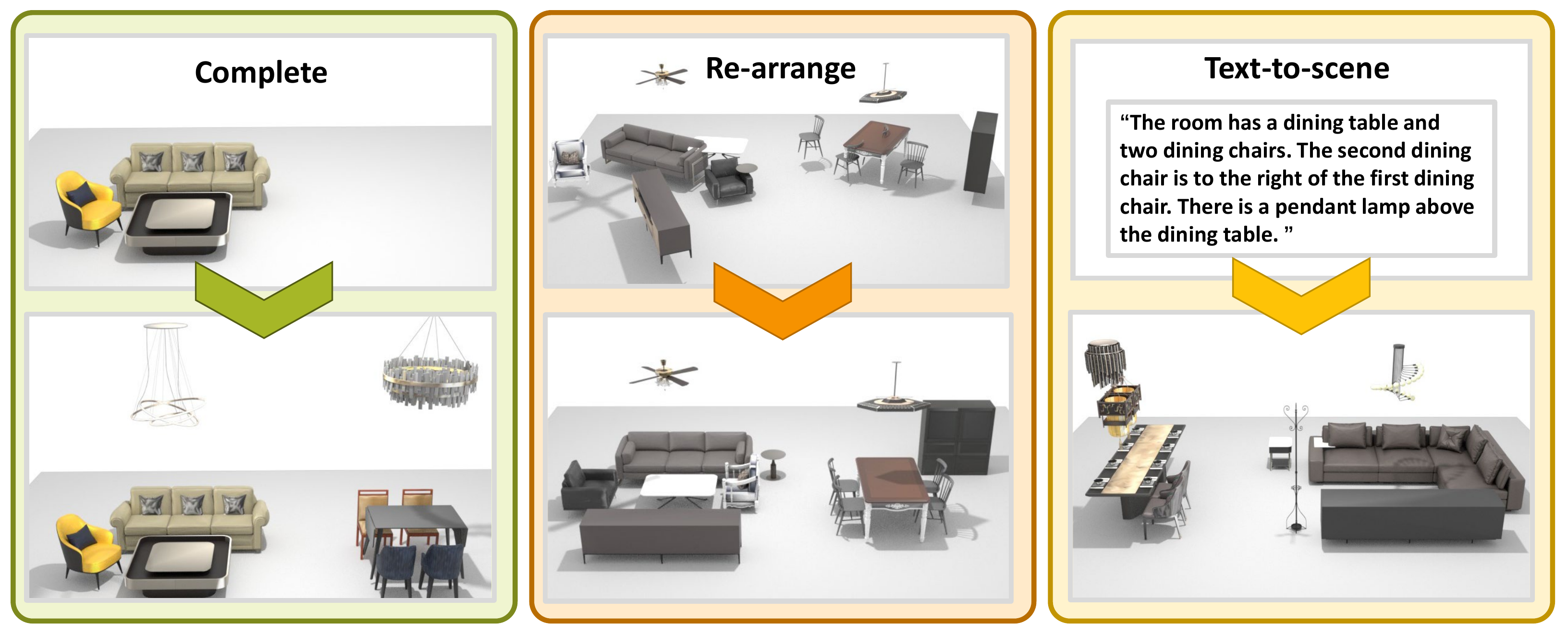}
                
		}
        \captionof{figure}{We present \emph{DiffuScene}, a diffusion model for diverse and realistic indoor scene synthesis. It facilitates various downstream applications: scene completion from partial scenes (left); scene arrangements of given objects (middle); scene generation from a text prompt describing partial scene configurations. (right).} 
        \label{fig:teaser}
	\end{center}
}]

\begin{abstract}
We present DiffuScene for indoor 3D scene synthesis based on a novel scene configuration denoising diffusion model. It generates 3D instance properties stored in an unordered object set and retrieves the most similar geometry for each object configuration, which is characterized as a concatenation of different attributes, including location, size, orientation, semantics, and geometry features.
%
%
%
%
We introduce a diffusion network to synthesize a collection of 3D indoor objects by denoising a set of unordered object attributes.
Unordered parametrization simplifies and eases the joint distribution approximation. The shape feature diffusion facilitates natural object placements, including symmetries.
Our method enables many downstream applications, including scene completion, scene arrangement, and text-conditioned scene synthesis.
Experiments on the 3D-FRONT dataset show that our method can synthesize more physically plausible and diverse indoor scenes than state-of-the-art methods. 
Extensive ablation studies verify the effectiveness of our design choice in scene diffusion models.

\vspace{-0.5cm}

\end{abstract}


\section{Introduction}
\label{SecIntro}
%
Synthesizing 3D indoor scenes that are realistic, semantically meaningful, and diverse is a long-standing problem in computer graphics.  It can significantly reduce costs in game development, CGI for films, and virtual reality. Furthermore, scene synthesis has practical applications in virtual interior design, enabling virtual rearrangement based on existing furniture or textual descriptions.  It also serves as a fundamental component in data-driven approaches for 3D scene understanding and reconstruction, necessitating large-scale 3D datasets with ground-truth labels.

Traditional scene modeling and synthesis formulate this as an optimization problem. With pre-defined scene prior constraints defined by room design rules such as layout guidelines ~\cite{merrell2011interactive,yeh2012synthesizing}, object category frequency distributions \cite{chang2014learning,chang2017sceneseer,fisher2010context}, affordance maps from human-object interactions~\cite{fisher2015activity,fu2017adaptive,jiang2012learning}, or scene arrangement examples~\cite{fisher2012example,fu2017adaptive}, they initially sample an initial scene and subsequently refine scene configurations through iterative optimization. However, defining precise rules is time-consuming and demands significant artistic expertise. The scene optimization stage is often laborious and computationally inefficient. Additionally, predefined design rules may limit the expression of complex and diverse scene compositions.

To automate the scene synthesis, some approaches~\cite{wang2018deep,li2019grains,ritchie2019fast,wang2019planit,zhang2020deep, purkait2020sg, wang2021sceneformer,yang2021indoor,yang2021scene,paschalidou2021atiss,nie2022learning} resort to deep generative models to learn scene priors from large-scale datasets. GAN-based methods~\cite{yang2021indoor} implicitly fit the scene distribution via adversarial training, yielding favorable results. However, they often lack diversity due to limited mode coverage and are prone to mode collapse. 
VAE-based methods~\cite{purkait2020sg,yang2021scene} explicitly approximate the scene distribution, offering better generative diversity but with lower-fidelity results. Recent auto-regressive models~\cite{wang2021sceneformer,paschalidou2021atiss,nie2022learning} progressively predict object properties sequentially. However, the sequential process may not accurately capture inter-object relationships and can accumulate prediction errors.

%
To capture more complicated scene configuration patterns for diverse scene synthesis, we strive to design a diffusion model for 3D scene synthesis. Diffusion models offer a compelling balance between diversity and realism and are relatively easier to train compared to other generative models~\cite{kingma2013auto,goodfellow2020generative,graves2013generating, rezende2015variational, chen2018neural,van2016conditional, van2017neural, razavi2019generating, esser2021taming}.
%
%
In this work, we represent a scene as a set of unordered objects, with each element comprising a concatenation of various attributes, including location, size, orientation, semantics, and geometry features.
Compared to other scene representations like multi-view images~\cite{dai20183dmv,han2019deep}, voxel grids~\cite{choy20163d,wu2016learning}, and neural fields~\cite{park2019deepsdf, mescheder2019occupancy, chen2019learning, mildenhall2021nerf,tang2021sa}, our representation is more compact and lightweight, making it suitable for learning through diffusion models.
Rather than representing a scene as an ordered object sequence and diffusing them sequentially~\cite{wang2021sceneformer, paschalidou2021atiss}, unordered set diffusion simplifies and eases the approximation of joint distribution of object instances. 
To this end, we design a denoising diffusion model~\cite{ho2020denoising, song2020score, ho2022classifier} to estimate object attributes to determine the placements and types of 3D instances and then perform shape retrieval to obtain final surface geometries.
%
%
%
The scene diffusion priors are learned through iterative transitions between noisy and clean object sets, allowing for generating a diverse range of physically plausible scenes. During denoising, we simultaneously refine the properties of all objects within a scene, explicitly leveraging spatial relationships through an attention mechanism~\cite{vaswani2017attention}.
Different from previous works~\cite{wang2021sceneformer, yang2021scene, paschalidou2021atiss} that only predict object bounding boxes, we diffuse semantics, oriented bounding boxes, and geometry features together to promote a holistic understanding of composition structure and surface geometries. The synthesized shape codes for geometry retrieval can produce more natural object arrangements, such as symmetric relations commonly seen in the real world.
We show compelling results in the unconditional and conditional settings against state-of-the-art scene generation models and provide extensive ablation studies to verify the design choices of our method.

\medskip
\noindent
Our contributions can be summarized as follows.
\begin{itemize}
    \item We introduce 3D scene denoising diffusion models for diverse indoor scene synthesis, which learn holistic scene configurations of object semantics, placements, and geometries.
    \item We introduce shape latent feature diffusion for geometry retrieval, which exploits accurate inter-object relationships for symmetry formation.
    \item based on this proposed model we facilitate completion from partial scenes, object re-arrangement in an existing scene, as well as text-conditioned scene synthesis.
\end{itemize}
\begin{figure*}
    \centering
    \includegraphics[width=.92\textwidth]{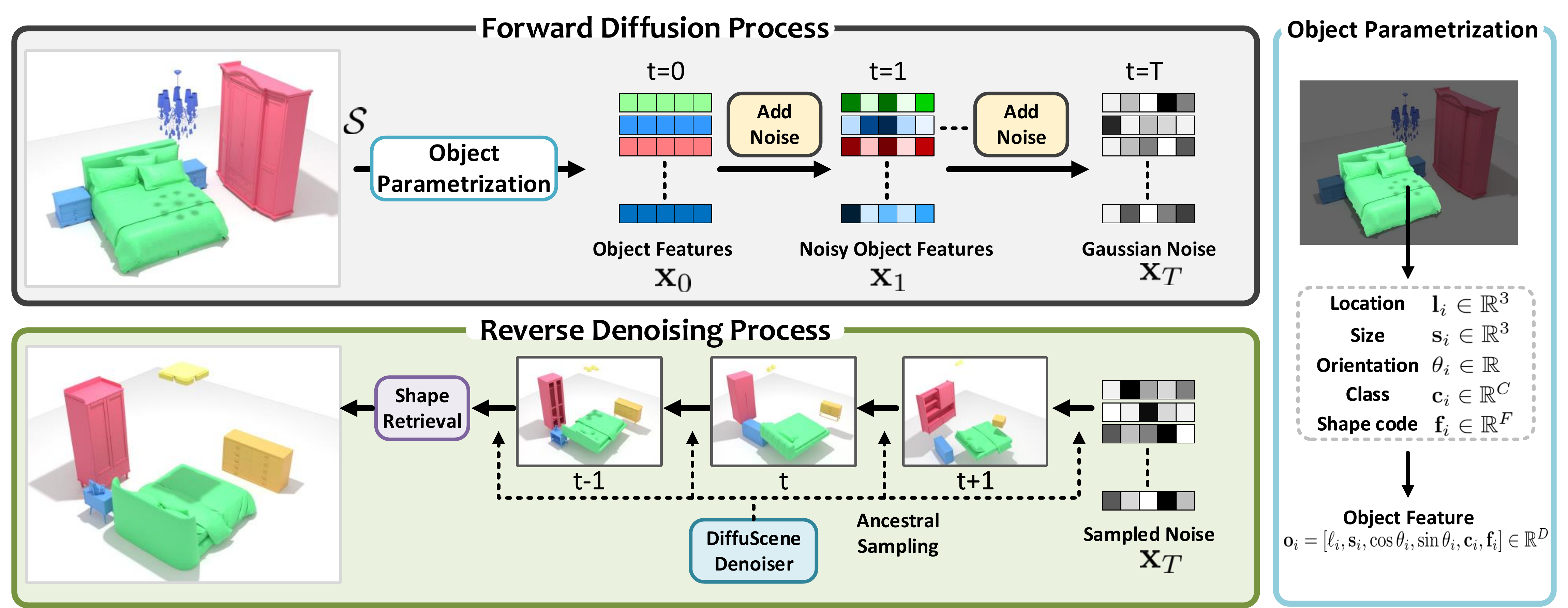} 
    \caption{\textbf{Overview.} Given a 3D scene $\mathcal{S}$ of $N$ objects, we represent it as an unordered set $\Vec{x}_0=\{ \Vec{o}_i \}_{i=1}^{N}$, by parametrizing each object $\Vec{o}_i$ as a vector storing all object attributes~\ie, location $\Vec{l}_i$, size $\Vec{s}_i$, orientation $\theta_i$, class label $\Vec{c}_i$, and latent shape code $\Vec{f}_i$. Based on a set of all possible $\Vec{x}_0$, we propose \emph{DiffuScene}, a denoising diffusion probabilistic model for 3D scene generation. In the forward process, we gradually add noise to $\Vec{x}_0$ until we obtain a standard Gaussian noise $\Vec{x}_T$. In the reverse process i.e. generative process, a denoising network iteratively cleans the noisy scene using ancestral sampling. Finally, we use the denoised class labels and shape latent codes to perform shape retrieval, and place object geometries through denoised locations, sizes, and orientations.} 
    \label{fig:pipeline}
\end{figure*}

\section{Related work}
\label{SecRelate}
\paragraph{Traditional Scene Modeling and Synthesis} 
%
Traditional methods usually formulate this problem into a data-driven optimization task.
To synthesize plausible 3D scenes, prior knowledge of reasonable configurations is required to drive scene optimization. Scene priors were often defined by following guidelines of interior design~\cite{merrell2011interactive,yeh2012synthesizing}, object frequency distributions (e.g., co-occurrence map of object categories)~\cite{chang2014learning,chang2017sceneseer,fisher2010context}, affordance maps from human motions~\cite{fisher2015activity,fu2017adaptive,jiang2012learning,ma2016action,qi2018human}, or scene arrangement examples~\cite{fisher2012example,fu2017adaptive}.
%
%
Constrained by scene priors, a new scene can be sampled from the formulation using different optimization methods, 
e.g., iterative methods~\cite{fisher2015activity,fu2017adaptive}, non-linear optimization~\cite{chang2014learning,qi2018human,xu2013sketch2scene,yeh2012synthesizing,yu2011make,fisher2012example}, or
manual interaction~\cite{chang2017sceneseer,merrell2011interactive,savva2017scenesuggest}. Unlike them, we learn complicated scene composition patterns from datasets, avoiding human-defined constraints and iterative optimization processes.
\paragraph{Learning-based Generative Scene Synthesis}
3D deep learning reforms this task by learning scene priors in a fully automatic, end-to-end, and differentiable manner. The capacity to process large-scale datasets dramatically increases the inference ability in synthesizing diverse object arrangements.
Existing generative models for 3D scene synthesis are usually based on feed-forward networks~\cite{zhang2020deep,wu2022targf}, VAEs~\cite{purkait2020sg,yang2021scene} , GANs~\cite{yang2021indoor}, or Autoregressive models~\cite{paschalidou2021atiss,nie2022learning,wang2021sceneformer}. 
GAN methods generate high-quality results rapidly but often lack mode coverage and diversity. VAEs offer better mode coverage but face challenges in generating faithful samples~\cite{xiao2021tackling}.
Recurrent networks~\cite{li2019grains,paschalidou2021atiss,nie2022learning,ritchie2019fast,wang2019planit,wang2018deep,wang2021sceneformer} including autoregressive models predict each new object conditioned on the previously generated objects. 
In contrast, we approach scene generation as an unordered object-set diffusion process where we explicitly model the joint distribution of object compositions. Multiple object properties are denoised synchronously, enhancing inter-object relationships and object composition plausibility.
%
\paragraph{3D Diffusion Models}
%
%
Recently, diffusion models~\cite{sohl2015deep,song2019generative,song2020improved,song2020denoising,ho2020denoising} have shown impressive visual quality in generative tasks, especially in various applications of 2D image synthesis~\cite{ho2020denoising, meng2021sdedit, kim2022diffusionclip, nichol2021glide, avrahami2022blended, saharia2022image, ho2022cascaded, dhariwal2021diffusion, rombach2022high,lugmayr2022repaint, ho2022imagen, cong2024flatten}  and single shape generation~\cite{luo2021diffusion,zhou20213d,zeng2022lion,zhang20233dshape2vecset,hui2022neural,tang2019skeleton,tang2022neural,tang2024dphms, cao2024motion2vecsets,zhang2023functional}
%
%
However, diffusion models in the 3D scene receive much less attention.
%
%
A concurrent work of LEGO-Net~\cite{wei2023lego} aims to predict 2D object locations and orientations, taking the input of a floor plane, object semantics, and geometries. 
Meanwhile, CommonScene~\cite{zhai2024commonscenes} generates 3D indoor scenes conditioned on scene graphs.
%
In contrast, DiffuScene is a scene-generative model that predicts 3D instance properties from random noise, including 3D locations and orientations, semantics, and geometries. 
Our method is more generic and versatile, which can benefit scene completion and conditioned scene synthesis from multi-modal signals like texts. 
In terms of implementation, our approach is based on a denoising diffusion model~\cite{ho2020denoising}, while LEGO-Net uses a Langevin Dynamics scheme based on a score-based method~\cite{song2019generative}.
We use a UNet-1D with attention as a denoiser rather than a transformer in LEGO-Net. 
These implementation differences contribute to our model's ability to acquire more natural scene arrangements, as evidenced by the discovery of more symmetric pairs in our method.
\section{DiffuScene}
\label{SecApp}
We introduce DiffuScene, a scene denoising diffusion model aiming at learning the distribution of 3D indoor scenes which includes semantic classes, surface geometries, and placements of multiple objects.
Specifically, we assume indoor scenes to be located in a world coordinate system with the origin at the floor center, and each scene $\mathcal{S}$ is a composition of at most $N$ objects $\{ \Vec{o} \}_{i=1}^{N}$.
%
%
We represent each scene as an unordered set with $N$ objects, each object in a scene set is defined by its class category $\Vec{c} \in \mathbb{R}^C$, object size $\Vec{s}\in \mathbb{R}^3$, location $\Vec{\ell}\in \mathbb{R}^3$, rotation angle around the vertical axis $\Vec{\theta}\in \mathbb{R}$, and shape code $\Vec{f}\in \mathbb{R}^F$ extracted from object surfaces in the canonical system through a pre-trained shape auto-encoder~\cite{yang2018foldingnet}.
Since the number of objects varies across different scenes, we define an additional `empty' object and pad it into scenes to have a fixed number of objects across scenes.
%
%
As proposed in~\cite{yin2021center}, we represent the object rotation angle by parametrizing a 2-d vector of cosine and sine values.
In summary, each object $\Vec{o}_i$ is characterized by the concatenation of all attributes,~\ie $\Vec{o}_i = [ \Vec{\ell}_i , \Vec{s}_i , \cos\Vec{\theta}_i  , \sin\Vec{\theta}_i , \Vec{c}_i , \Vec{f}_i ] \in \mathbb{R}^D$, where $D$ is the dimension of concatenated attributes.
%
%
%
Based on this representation, we design our denoising diffusion model in Sec.~\ref{SubSecSceDiff}, which supports many different downstream applications like scene completion, scene re-arrangement, and text-conditioned scene synthesis in Sec.~\ref{SubSecApp}.
%
\subsection{Object Set Diffusion}
\label{SubSecSceDiff}
An overview of our approach is shown in Fig.~\ref{fig:pipeline}.
We design a denoising diffusion model that employs Gaussian noise corruptions and removals on object attributes to transition between noisy and clean scene distributions.
%
\paragraph{Diffusion process.} 
The (forward) diffusion process is a pre-defined discrete-time Markov chain in the data space $\mathcal{X}$ spanning all possible scene configurations represented as 2D tensors of fixed size $\Vec{x} \in \mathbb{R}^{N \times D}$, which are the concatenations of $N$ object properties $\{ \Vec{o}_i \}_{i=1}^{N}$ within a scene $\mathcal{S}$.
Given a clean scene configuration $\Vec{x}_0$ from the underlying distribution  $q(\Vec{x}_0)$, we gradually add Gaussian noise to $\Vec{x}_0$, obtaining a series of intermediate scene variables $\Vec{x}_1, ..., \Vec{x}_T$ with the same dimensionality as $\Vec{x}_0$, according to a pre-defined, linearly increased noise variance schedule $\beta_1, ..., \beta_T$ (where $\beta_1 < ... < \beta_T$).
The joint distribution $q (\Vec{x}_{1:T} | \Vec{x}_{0} )$ of the diffusion process can be expressed as:
\vspace{-3mm}
\begin{equation}
	\label{Equadiffusion}
	q (\Vec{x}_{1:T} | \Vec{x}_{0} ) :=  \prod_{t=1}^{T} q(\Vec{x}_{t} | \Vec{x}_{t-1}),
	\vspace{-3mm}
\end{equation}
where the diffusion step at time $t$ is defined as:
\vspace{-1mm}
\begin{equation}
	\label{Equadiffusion_each}
	q (\Vec{x}_{t} | \Vec{x}_{t-1} ) :=  \mathcal{N}(\Vec{x}_{t}; \sqrt{ 1-\beta_{t} } \Vec{x}_{t-1}, \beta_{t} \Vec{I} ).
	\vspace{-1mm}
\end{equation}
A helpful property of diffusion processes is that we can directly sample $\Vec{x}_t$ from $\Vec{x}_0$ via the conditional distribution:
\vspace{-1mm}
\begin{equation}
	\label{Equadiffusion_02t}
	q( \Vec{x}_{t} | \Vec{x}_{0} ) :=  \mathcal{N}( \Vec{x}_{t}; \sqrt{\bar{\alpha_t}}\Vec{x}_{0}, (1-\bar{\alpha_{t}}) \Vec{I} ) ,
\end{equation}
where $\Vec{x}_t = \sqrt{\bar{\alpha}_t} \Vec{x}_0 + \sqrt{1-\bar{\alpha}_t} \Vec{\epsilon}$ where $\alpha_t := 1 - \beta_t$  , $\bar{\alpha}_t := \prod_{r=1}^{t} \alpha_s$, and $\Vec{\epsilon}$ is the noise used to corrupt $\Vec{x}_t$. 
\paragraph{Generative process.}
The generative (\ie denoising) process is parameterized as a Markov chain of learnable reverse Gaussian transitions.
Given a noisy scene from a standard multivariate Gaussian distribution $\Vec{x}_T\sim\mathcal{N}(\mathbf{0}, \Vec{I})$ as the initial state, it corrects $\Vec{x}_{t}$ to obtain a cleaner version $\Vec{x}_{t-1}$ at each time step by using a learned Gaussian transition $p_\Vec{\phi} (\Vec{x}_{t-1} | \Vec{x}_{t} )$ which is parameterized by a learnable network $\Vec{\phi}$.
By repeating this reverse process until the maximum number of steps $T$, we can reach the final state $\Vec{x}_0$, the clean scene configuration we aim to obtain.
Specifically, the joint distribution of the generative process $p_\Vec{\phi} (\Vec{x}_{0:T} )$ is formulated as:
\vspace{-3mm}
\begin{equation}
	\label{Equadenoise}
	p_\Vec{\phi} (\Vec{x}_{0:T} ) :=  p(\Vec{X}_T) \prod_{t=1}^{T} p_\Vec{\phi}( \Vec{x}_{t-1} | \Vec{x}_{t} ).
	\vspace{-1mm}
\end{equation}
%
\begin{equation}
	\label{Equadenoise_each}
	p_\Vec{\phi} (\Vec{x}_{t-1} | \Vec{x}_{t} ) :=  \mathcal{N}(\Vec{x}_{t-1}; \Vec{\mu}_{\Vec{\phi}}(\Vec{x}_{t}, t), \Vec{\Sigma}_{\Vec{\phi}}(\Vec{x}_{t}, t)),
\end{equation}
where $\Vec{\mu_{\phi}}(\Vec{x}_{t})$ and $\Vec{\Sigma}_\Vec{\phi}(\Vec{x}_{t})$ are the predicted mean and covariance of the Gaussian $\Vec{x}_{t-1}$ by feeding $\Vec{x}_{t}$ into the denoising network $\Vec{\phi}$.
For simplicity, we pre-define the constants of $\Sigma_{\Vec{\phi}}(\Vec{x}_{t}) := \sigma_t := \frac{1-\bar{\alpha}_{t-1}}{ 1-\bar{\alpha}_t } \beta_t$, although Song et al. has shown that learnable covariances can increase generation quality in DDIM~\cite{song2020improved}.
Ho et al. empirically found in DDPM~\cite{ho2020denoising} that rather than directly predicting $\Vec{\mu}_{\Vec{\phi}}(\Vec{x}_{t}, t)$, we can synthesize more high-frequent details by estimating the noise $\Vec{\epsilon}_{\Vec{\phi}}(\Vec{x}_{t}, t)$ applied to perturb $\Vec{x}_{t}$.
Then $\Vec{\mu}_{\Vec{\phi}}(\Vec{x}_{t})$ can be re-parametrized by subtracting the predicted noise according to Bayes's theorem:
\vspace{-2mm}
\begin{equation}
	\label{Equareparam}
	\Vec{\mu}_{\Vec\phi}(\Vec{x}_{t}, t) :=  \frac{1}{\sqrt{\alpha_{t}}} (\Vec{x}_{t} - \frac{\beta_t}{\sqrt{1-\bar{\alpha}_t}} \Vec{\epsilon}_{\Vec{\phi}}(\Vec{x}_{t}, t)).
	\vspace{-8mm}
\end{equation}
\begin{figure}[t]
	\centering
	\includegraphics[width=.5\textwidth]{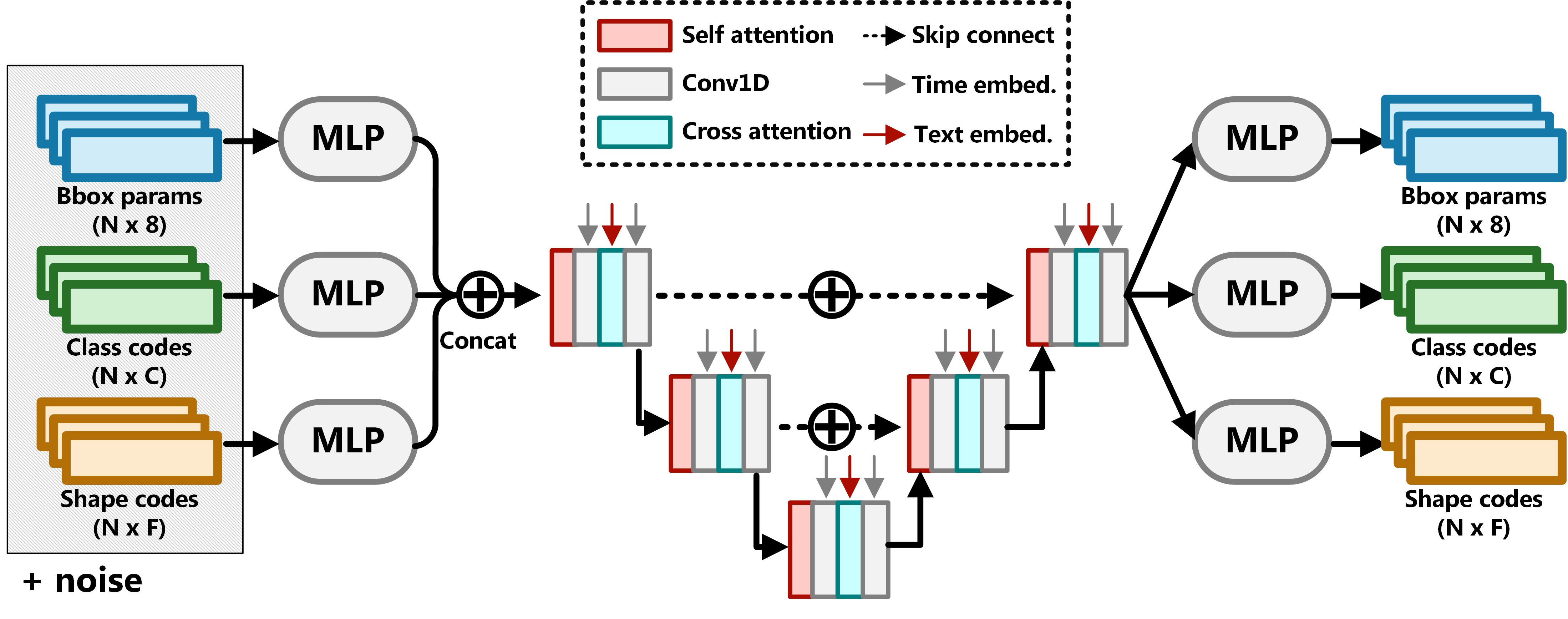}
	\caption{The denoising network architecture takes the attributes of multiple objects (bounding box, object class, geometry code) as input and denoises them using 1D convolutions with skip connections and attention blocks.}
	\label{fig:denoise_net}
	\vspace{-5mm}
\end{figure}
\paragraph{Denoising network.}
As shown in Fig.~\ref{fig:denoise_net}, the denoiser in our method is based on 1D convolution with skip connections, where convolution blocks are interleaved with attention blocks~\cite{vaswani2017attention} to aggregate the features of different objects, exploiting the inter-object relationships and capturing the global scene context.
%
%
%
\paragraph{Training objective.}
The goal of training the reverse diffusion process is to find optimal denoising network parameters $\Vec{\phi}$ that can generate natural and plausible scenes.
Our training objective is composed of two parts:
i) A loss $L_{\text{sce}}$ to constrain that the generated object set can approximate the underlying data distribution, 
and ii) a regularization term $L_{\text{iou}}$ to penalize the object intersections.
The $L_{\text{sce}}$ is derived by maximizing the negative log-likelihood of the last denoised scene $\mathbb{E}[ -\log p_{\Vec{\phi}}(\Vec{x}_0) ]$, which is yet not intractable to optimize directly.
Thus, we can instead choose to maximize its variational upper bound:
\vspace{-2mm}
\begin{equation}
	\label{equaNLL}
	L_{\text{sce}} :=  \mathbb{E}_q [ -\log \frac{p_{\phi}(\Vec{x}_{0:T})} {q(\Vec{x}_{1:T}|\Vec{x}_0)} ] \ge \mathbb{E}[ -\log p_{\Vec{\phi}}(\Vec{x}_0) ].
	\vspace{-1mm}
\end{equation}
By surrogating variables, we can further simplify $L_{\text{sce}}$ as the sum of KL divergence between posterior $p_{\Vec{\phi}} (\Vec{x}_{t-1}  | \Vec{x}_{t}, \Vec{x}_{0})$ and conditional distribution $q(\Vec{x}_{t} | \Vec{x}_{t-1})$ at each $t$ :
\vspace{-3mm}
\begin{equation}
	\label{equaDecompose}
	\begin{aligned}
		L_{\text{sce}} 
		:=  \mathbb{E}_q [ -\log p(\Vec{x}_T) - \sum_{t=1}^{T} \log \frac{ p_{\Vec{\phi}} (\Vec{x}_{t-1} | \Vec{x}_{t}, \Vec{x}_{0}) } {q(\Vec{x}_{t} | \Vec{x}_{t-1})} ] ,
		\vspace{-5mm}
	\end{aligned}
\end{equation}
where $-\log p(\Vec{x}_T)$ is a fixed constant since $\Vec{x}_T \sim \mathcal{N}(0, \Vec{I})$.
Here, we refer to DDPM~\cite{ho2020denoising} for the details of the derivation process.
Moreover, we can re-write $L_{\text{sce}}$ into a simple and intuitive version that constrains the correct prediction of the corrupted noise on $\Vec{x}_t$:
\vspace{-2mm}
\begin{equation}
	\label{equaNoisypred}
	\begin{aligned}
		L_{\text{sce}} := \mathbb{E}_{\Vec{x}_0, \Vec{\epsilon}, t} [ \| \Vec{\epsilon} - \Vec{\epsilon}_{\Vec{\phi}} (\Vec{x}_t, t) \|^2 ] \quad \quad \quad \quad \quad \quad \\
		:= \mathbb{E}_{\Vec{\phi}} [\| \Vec{\epsilon} - \Vec{\epsilon}_{\Vec{\phi}} ( \sqrt{\bar{\alpha}_t} \Vec{x}_0 +  \sqrt{1-\bar{\alpha}_t} \Vec{\epsilon}, t) \|^2] .
	\end{aligned}
	\vspace{-2mm}
\end{equation}
Based on Eq.~\ref{Equareparam}, we can obtain the approximation of clean scene $\tilde{\Vec{x}}_0^t$.
Thus, we can compute $L_{\text{iou}}$ as the IoU summation of arbitrary two bounding boxes:
\vspace{-2mm}
\begin{equation}
	\label{equaIoU}
	L_{\text{iou}} :=  \sum_{t=1}^{T} 0.1 * \bar{\alpha}_t  * \sum_{\Vec{o}_i, \Vec{o}_j \in \tilde{\Vec{x}}_0^t} \IoU(\Vec{o}_i, \Vec{o}_j).
	\vspace{-2mm}
\end{equation}
%
%
\subsection{Applications}
\label{SubSecApp}
Based on our diffusion model above, we can support various downstream tasks (see Fig.~\ref{fig:teaser}) with few modifications. 
\vspace{-2mm}
\paragraph{Scene completion.}
Assuming a partial scene with $M (\le N)$ objects, \ie $\Vec{y} \in \mathbb{R}^{M \times D}$, we utilize the learned scene priors from diffusion models to complement novel  $\hat{\Vec{x}}_0)$ into $\Vec{y}_0$ to obtain a complete object set  $\Vec{x}_0 = (\Vec{y}, \hat{\Vec{x}}_0)$.
%
We keep the already known elements and only hallucinate the missing ones through learnable reverse Gaussian transitions $q_\Vec{\phi}$ conditioning on $\Vec{y}$.
%
%
The complemented scene $\hat{\Vec{x}}_t$ at time step $t$ is generated by:
\vspace{-1mm}
\begin{equation}
	\begin{aligned}
		\label{EquanCompletion}
		p_{\Vec{\phi}}( \hat{\Vec{x}}_{t-1} | \hat{\Vec{x}}_t  ) := \mathcal{N}(\mu_{\Vec{\phi}}(\Vec{x}_t, t, \Vec{y}), \sigma_t^2\Vec{I}) . \quad \quad \quad 
	\end{aligned}
	\vspace{-2mm}
\end{equation}
\paragraph{Scene re-arrangement.}
Given a set of objects with random spatial positions, we can leverage the priors of our diffusion model to rearrange reasonable object placements by estimating their locations and orientations.
We denote the noisy scene initialization as $\hat{\Vec{x}}_0 = [\hat{\Vec{u}}_0, \Vec{v}]$, where $\hat{\Vec{u}}_0 = \{  [\Vec{l}_i, \cos\theta_i, \sin\theta_i] \}_{i=1}^N$ is the concatenation of $N$ objects' locations and orientations, and $\Vec{v} = \{ [\Vec{s}_i, \Vec{c}_i, \Vec{f}] \}_{i=1}^N$ is the concatenation of $N$ objects' sizes, category classes, and shape codes.
The intermediate scenes during the arrangement diffusion process can be expressed as:
\vspace{-1mm}
\begin{equation}
	\begin{aligned}
		\label{EquanRearrange}
		p_{\Vec{\phi}}( \hat{\Vec{u}}_{t-1} | \hat{\Vec{u}}_t ) := \mathcal{N}(\mu_{\Vec{\phi}}(\hat{\Vec{u}}_t , t, \Vec{v}), \sigma_t^2\Vec{I}) , \quad \quad \quad
	\end{aligned}
\end{equation}
where we iteratively update the object locations and orientations $\Vec{u}_t$ via $p_\Vec{\phi}$ conditioned on $\Vec{v}$.
\vspace{-2mm}
\paragraph{Text-conditioned scene synthesis.}
Given a list of sentences describing the desired object classes and inter-object spatial relationship as conditional inputs, we can employ a pre-trained BERT encoder~\cite{devlin2018bert} to extract word embeddings $\Vec{z} \in \mathbb{R}^{48 \times 768}$, then we utilize cross attention layers to inject the language guidance into the denoising network that predicts out noise via $\Vec{\epsilon}_\Vec{\phi}(\Vec{x}_t, t, \Vec{z})$, as depicted in Fig.~\ref{fig:denoise_net}.

\begin{figure*}[!ht]
    \vspace{-3mm}
	\centering
 	\begin{subfigure}[t]{0.23\textwidth}
		\includegraphics[width=\textwidth]
		{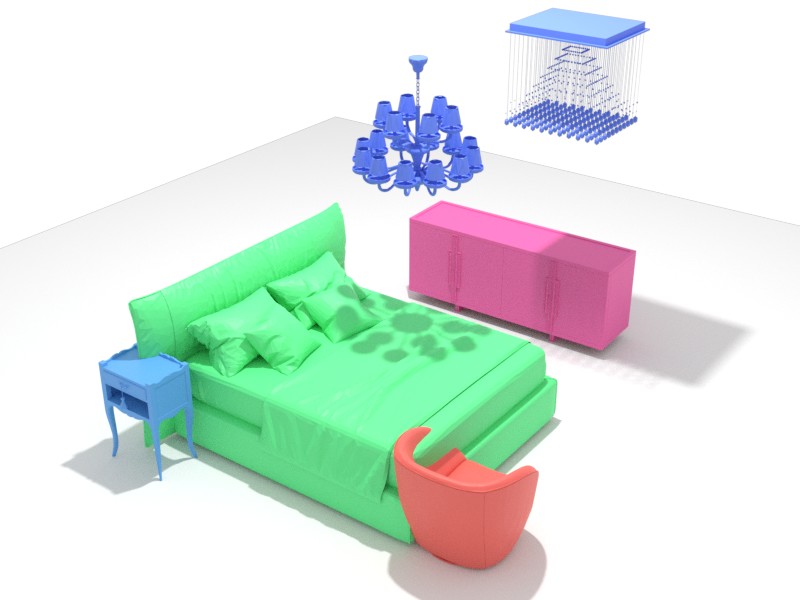}
            %
		\includegraphics[width=\textwidth]
		{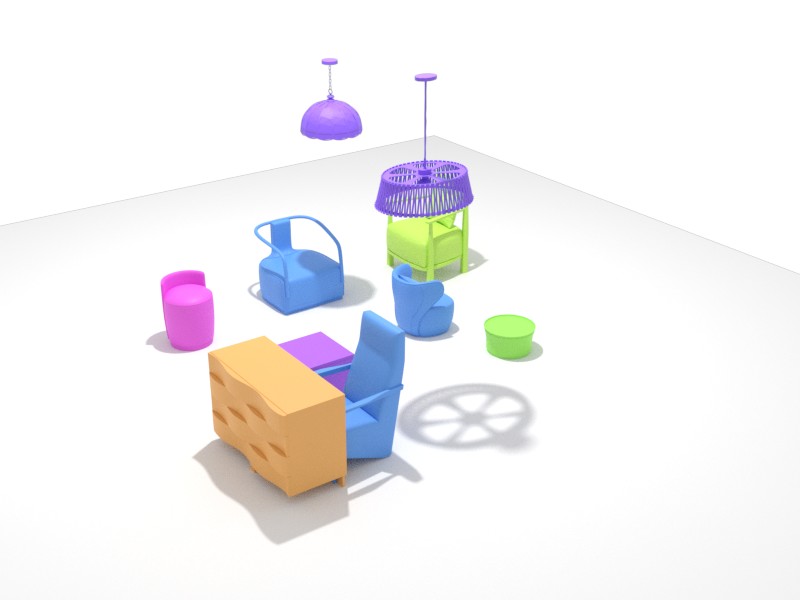}
		\includegraphics[width=\textwidth]
		{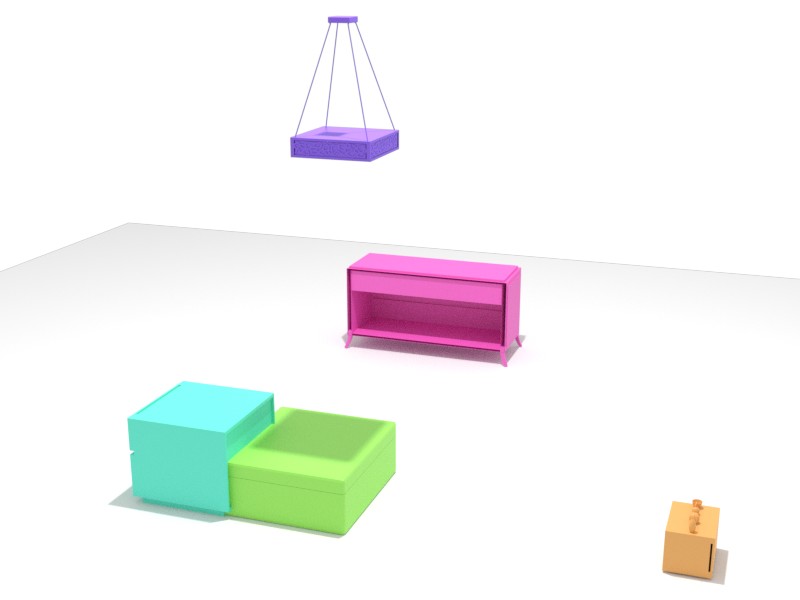}
		\includegraphics[width=\textwidth]
		{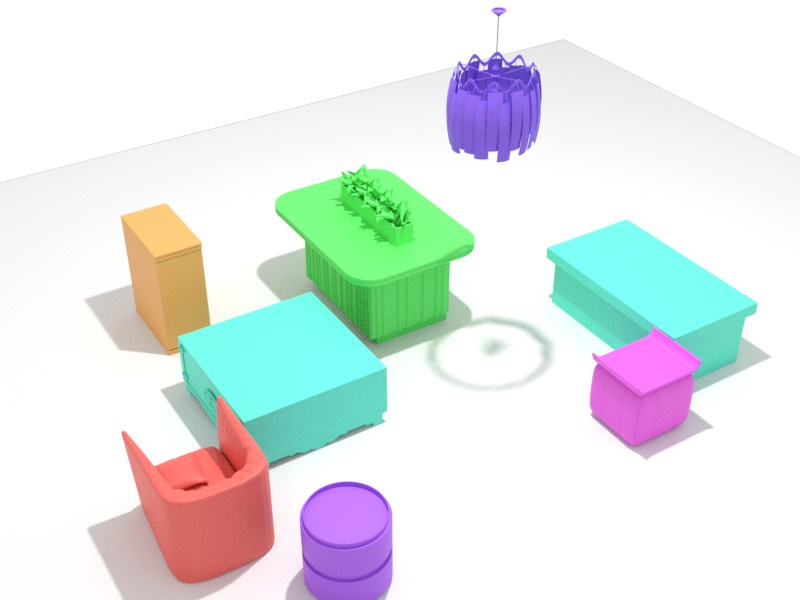}
		\caption{DepthGAN~\cite{yang2021indoor}}
	\end{subfigure}
	\rulesep
	\begin{subfigure}[t]{0.23\textwidth}
		\includegraphics[width=\textwidth]
		{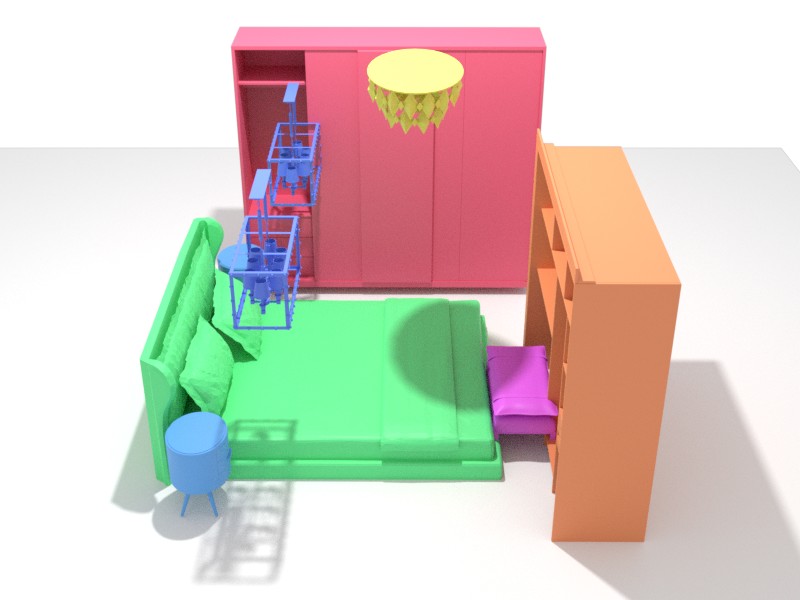}
            %
            \includegraphics[width=\textwidth]
		{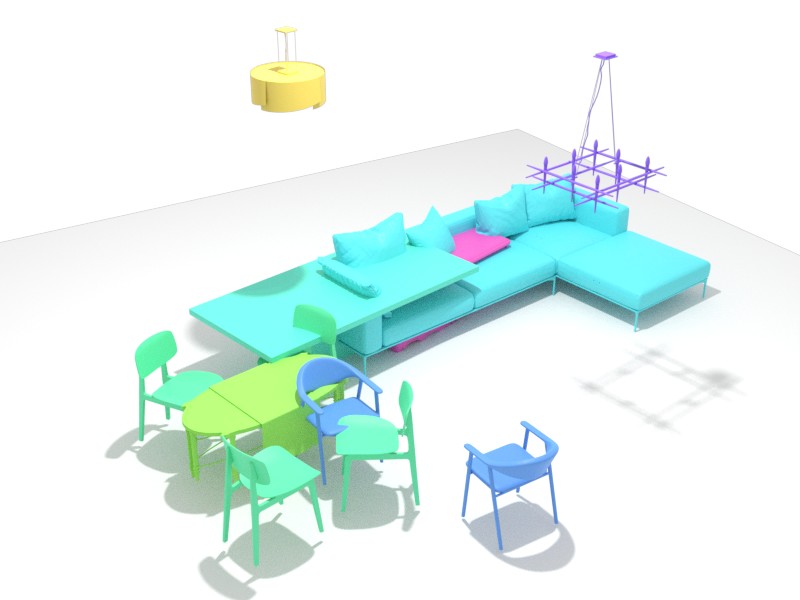}
		\includegraphics[width=\textwidth]
		{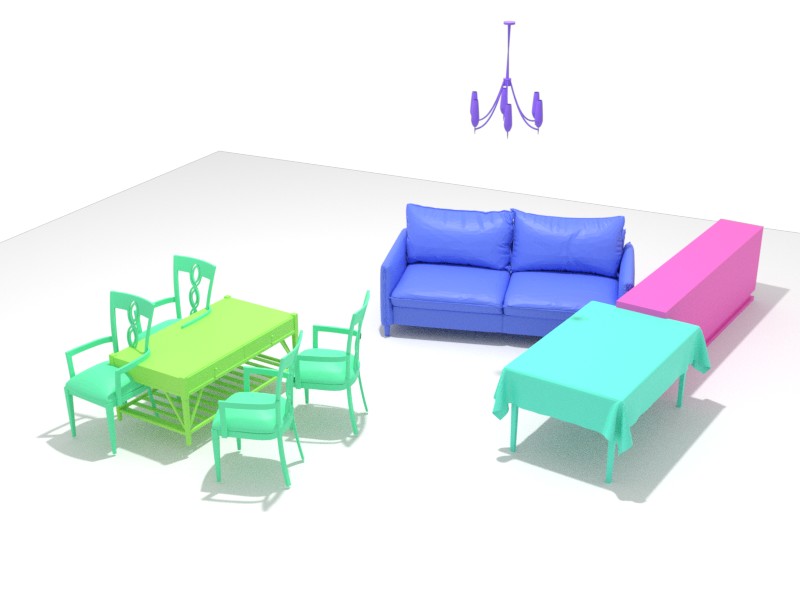}
            \includegraphics[width=\textwidth]
		{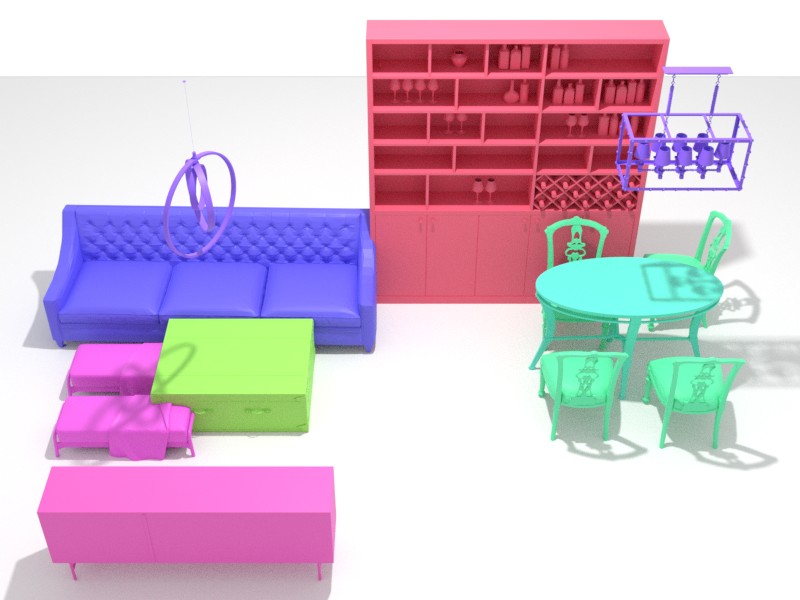}
		\caption{Sync2Gen~\cite{yang2021scene}}
	\end{subfigure}
	\rulesep
        \begin{subfigure}[t]{0.23\textwidth}
		\includegraphics[width=\textwidth]
		{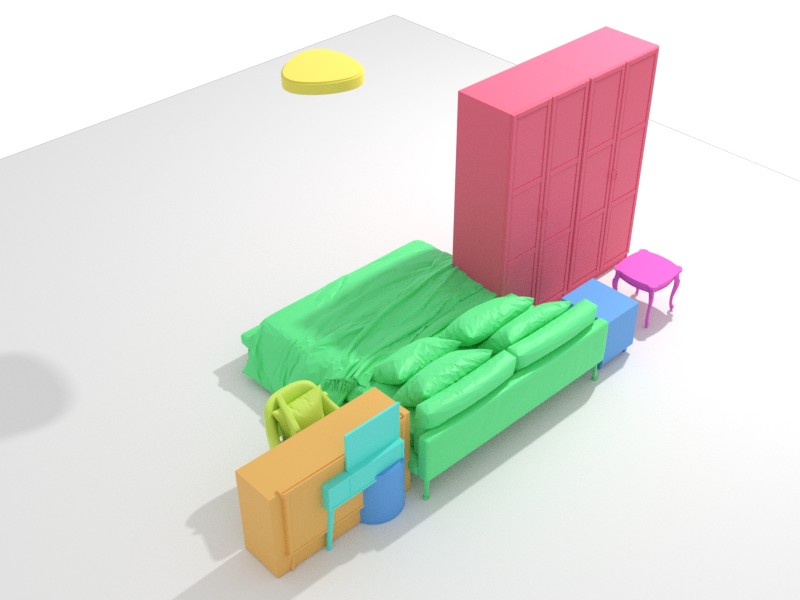}
		\includegraphics[width=\textwidth]
		{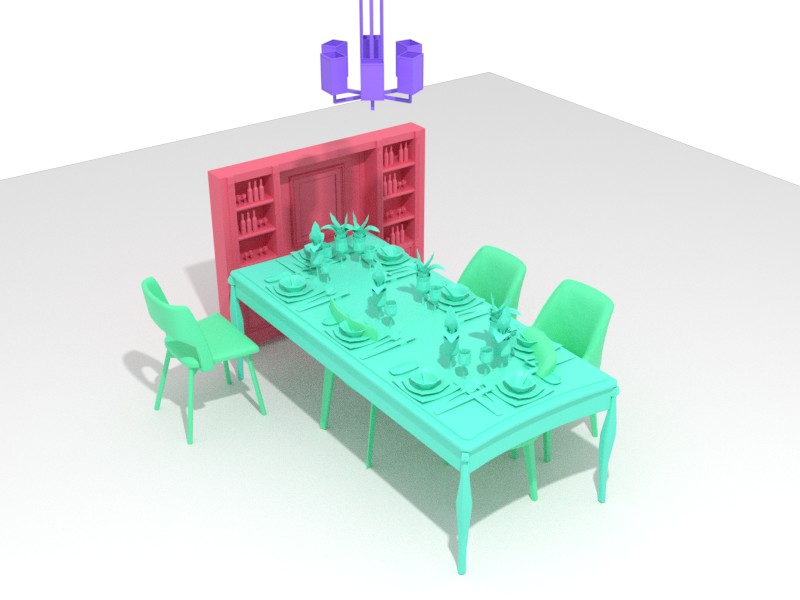}
		\includegraphics[width=\textwidth]
		{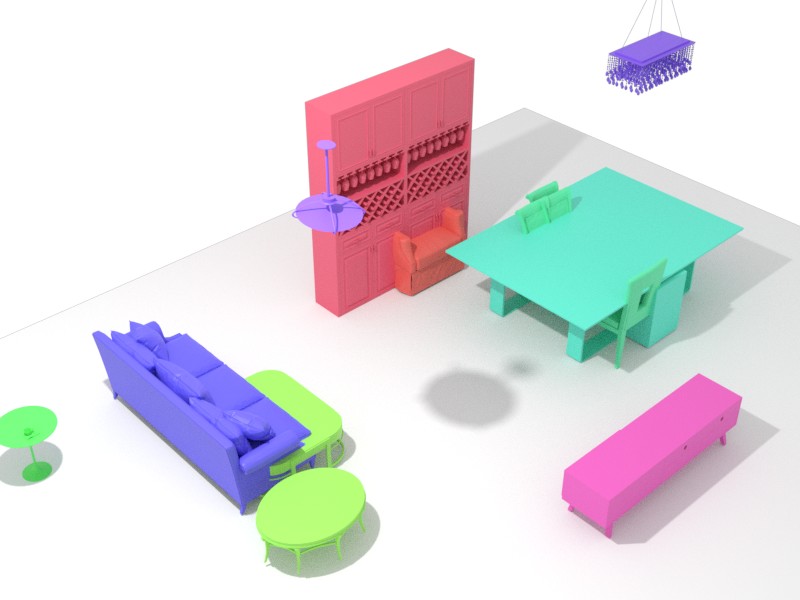}
		\includegraphics[width=\textwidth]
		{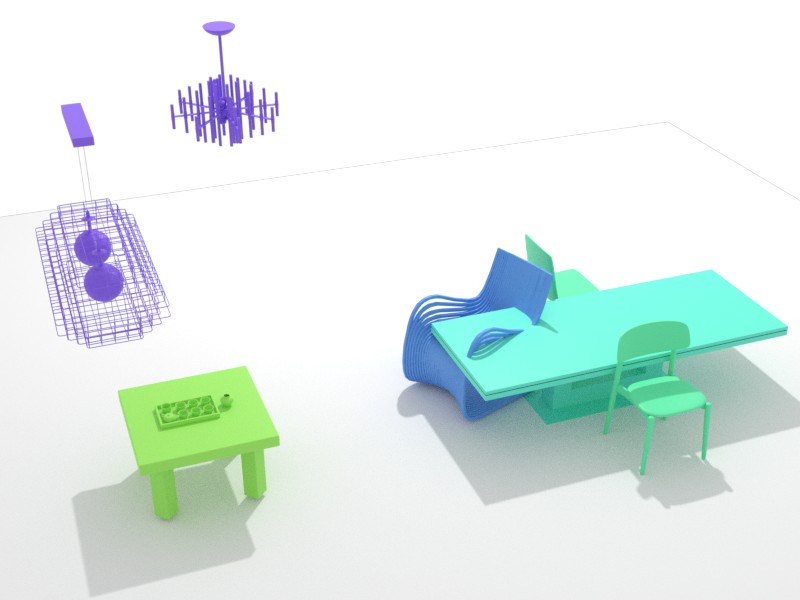}
		\caption{ATISS~\cite{paschalidou2021atiss}}
	\end{subfigure}
	\rulesep
	\begin{subfigure}[t]{0.23\textwidth}
		\includegraphics[width=\textwidth]
		{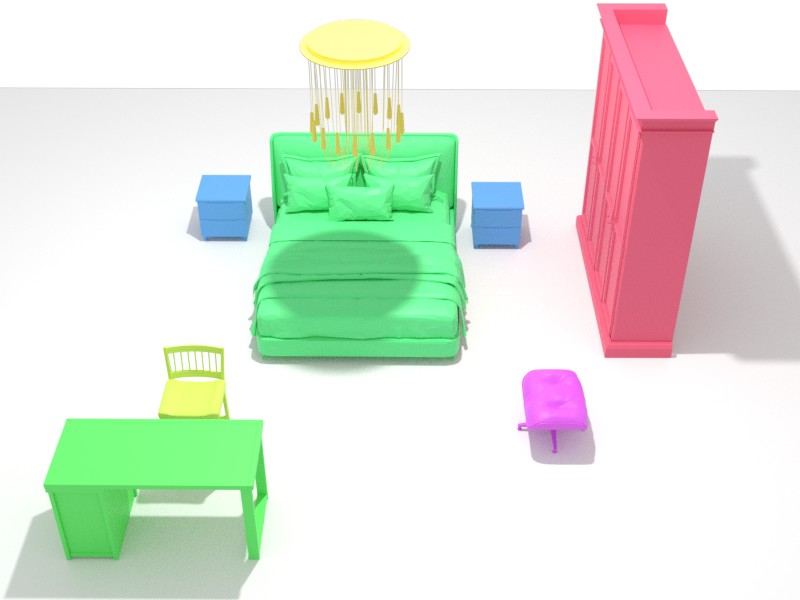}
            \includegraphics[width=\textwidth]
		{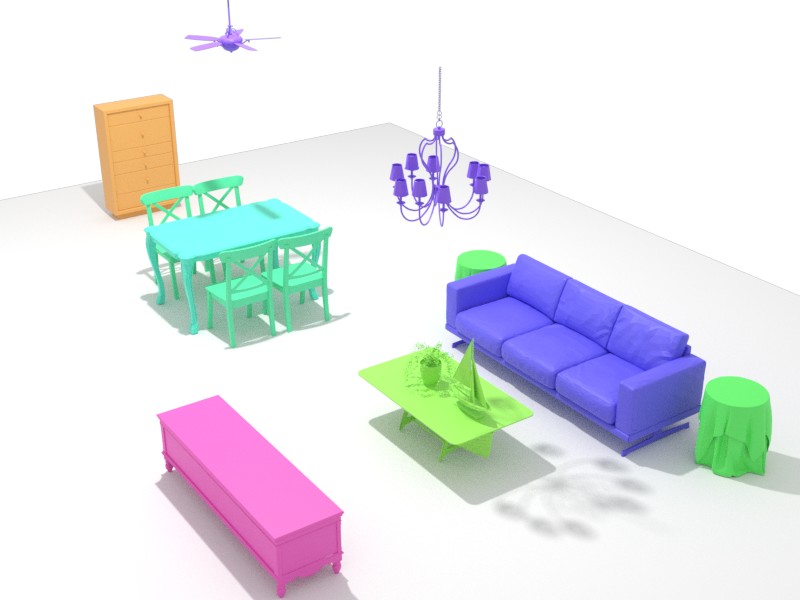}
		\includegraphics[width=\textwidth]
		{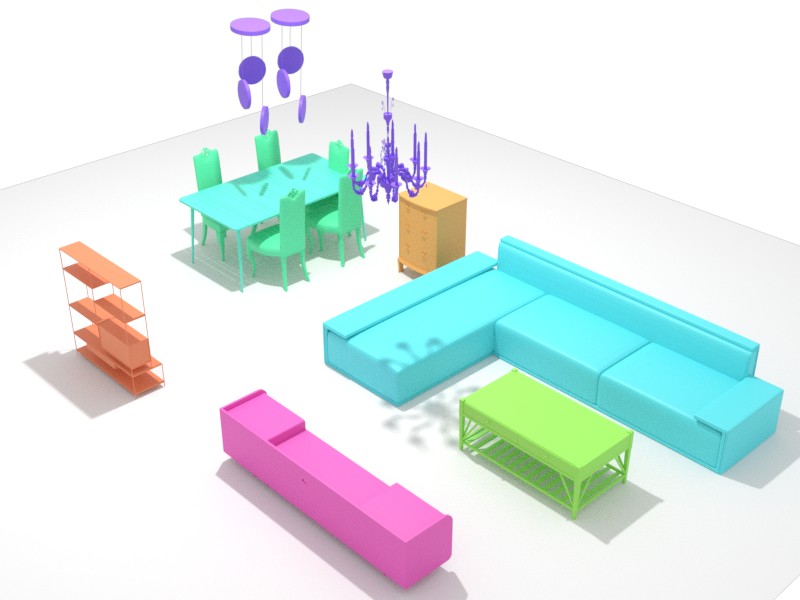}
            \includegraphics[width=\textwidth]
		{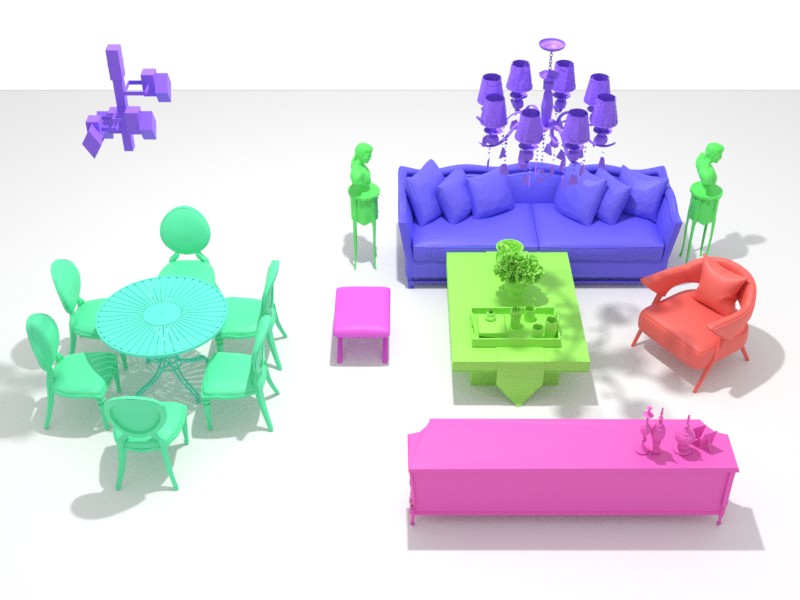}
		\caption{Ours}
	\end{subfigure}
	\caption{\textbf{Unconditional scene synthesis}. We compare our method with the state-of-the-art by generating from random noises, where our results present higher diversity and better plausibility with fewer penetration issues and more symmetric pairs.}
    \label{fig:uncond_comparison}
    \vspace{-3mm}
\end{figure*}
\section{Experiments}
\label{SecExp}
\paragraph{Datasets}
For experimental comparisons, we use the large-scale 3D indoor scene dataset 3D-FRONT~\cite{fu20213d} as the benchmark. 3D-FRONT is a synthetic dataset composed of 6,813 houses with 14,629 rooms, where each room is arranged by a collection of high-quality 3D furniture objects from the 3D-FUTURE dataset~\cite{fu20213dm}.  Following ATISS~\cite{paschalidou2021atiss}, we use three types of indoor rooms for training and evaluation, including 4,041 bedrooms, 900 dining rooms, and 813 living rooms.
For each room type, we use $80\%$ of rooms as the training sets, while the remaining are for testing.
\paragraph{Baselines}
We compare against state-of-the-art scene synthesis approaches using various generative models, including: 
1)  DepthGAN~\cite{yang2021indoor}, learning a volumetric generative adversarial network from multi-view semantic-segmented depth maps;
2) Sync2Gen~\cite{yang2021scene}, learning a latent space through a variational auto-encoder of scene object arrangements represented by a sequence of 3D object attributes; A Bayesian optimization stage based on the relative attributes prior model further regularized and refined the results.
3) ATISS~\cite{paschalidou2021atiss}, an autoregressive model to sequentially predict the 3D object bounding box attributes.
\paragraph{Implementation}
We train our scene diffusion models on different types of indoor rooms respectively.
They are trained on a single RTX 3090 with a batch size of 128 for $T=100,000$ epochs. The learning rate is initialized to $lr=\expnumber{2}{-4}$ and then gradually decreases with the decay rate of 0.5 in every 15,000 epochs.
For the diffusion processes, we use the default settings from the denoising diffusion probabilistic models (DDPM)~\cite{ho2020denoising}, where the noise intensity is linearly increased from 0.0001 to 0.02 with 1,000-time steps. During inference, we first use the ancestral sampling strategy to obtain the object properties and then retrieve the most similar CAD model in the 3D-FUTURE~\cite{fu20213dm} for each object based on generated shape codes.
\setlength{\tabcolsep}{2pt}
\begin{table*}[!hbt]
	\renewcommand\arraystretch{1.2}
	\begin{center}
		\begin{tabular}{*{13}{c}}
			\toprule
			\multirow{2}*{Method} & \multicolumn{4}{c}{Bedroom}   & \multicolumn{4}{c}{Dining room}  & \multicolumn{4}{c}{Living room} \\
			\cmidrule(lr){2-5} \cmidrule(lr){6-9} \cmidrule(lr){10-13}
			 & FID $\downarrow$ & KID $\downarrow$ & SCA $\%$ & CKL $\downarrow$   & FID $\downarrow$ & KID $\downarrow$ & SCA $\%$ & CKL $\downarrow$ & FID $\downarrow$ & KID $\downarrow$ & SCA  $\%$ & CKL $\downarrow$ \\ 
			\midrule
			\midrule

            DepthGAN~\cite{yang2021indoor}
                  & 40.15  & 18.54 & 96.04 & 5.04 
                  & 81.13  & 50.63 & 98.59 & 9.72
                  & 88.10  & 63.81 & 97.85 & 7.95 \\
                  
            Sync2Gen*
                  & 33.59  & 13.78  & 87.11  & 2.67 
                  & 48.79  & 12.01  & 91.43  & 5.03
                  & 47.14  & 11.42  & 86.71 & 1.60\\

            Sync2Gen~\cite{yang2021scene}
                  & 31.07  & 11.21  & 82.97 & 2.24 
                  & 46.05  & 8.74   & 88.02 & 4.96
                  & 48.45  & 12.31  & 84.57 & 7.52\\


            ATISS~\cite{paschalidou2021atiss}
                  & 18.60 & 1.72 & 61.71  & 0.78
                  & 38.66 & 5.62  & 71.34 &  0.64
                  & 40.83 & 5.18  & 72.66  & 0.69 \\
                  
            Ours
                 & \textbf{17.21} & \textbf{0.70}  & \textbf{52.15}  & \textbf{0.35} 
                   & \textbf{32.60} & \textbf{0.72} & \textbf{55.50}  & \textbf{0.22} 
                    & \textbf{36.18} & \textbf{0.88} & \textbf{57.81}  & \textbf{0.21}  \\

        \bottomrule
        \end{tabular}
        \caption{Quantitative comparisons on the task of \textbf{unconditional scene synthesis}. The Sync2Gen* is a variant of Sync2Gen~\cite{yang2021scene} without Bayesian optimization. Note that for the Scene Classification Accuracy (SCA), the score closer to $50\%$ is better. }
        \label{tab:uncond}
        \end{center}
        \vspace{-7mm}
\end{table*}
\begin{table}[!htbp]
	\renewcommand\arraystretch{1.2}
	\begin{center}
        \footnotesize
		\begin{tabular}{*{10}{c}}
                \toprule
			\multirow{2}*{Method} & \multicolumn{3}{c}{Bedroom}  & \multicolumn{3}{c}{Dining} & \multicolumn{3}{c}{Living} \\ \cmidrule(lr){2-4}  \cmidrule(lr){5-7} \cmidrule(lr){8-10} 
             & Obj  & Sym & PIoU  & Obj  & Sym & PIoU   & Obj  & Sym & PIoU  \\ 
                \midrule
                DepthGAN
                       & 5.12 & 0.03 & 0.35  
                       & 9.64 & 0.19 & 0.17  
                       & 6.70 & 0.01 & 0.14 \\ 
                Sync2Gen
                       & 6.25 &  0.85  & 0.51
                       & 8.65  & 2.85 & \bf{0.55} 
                       & 9.03 & 2.27 &  \bf{0.39} \\ 
                ATISS
                        & 5.47 & 0.33 &  0.50 
                        & 11.96 & 2.75 & 1.61 
                        & 10.81 & 1.42 & 1.10 \\ 
                Ours
                       & \bf{4.99} & \bf{0.72} & \bf{0.43}  
                       & \bf{10.95} & \bf{4.47}  & 0.65  
                       & \bf{11.85} & \bf{3.47}  & \bf{0.39} \\
                \midrule
                GT
                       & 5.00 & 0.71 & 0.43
                       & 10.80  & 4.22 & 0.48
                       & 11.70 & 3.59  & 0.30 \\ 
                  \bottomrule
        \end{tabular}
        \vspace{-2mm} 
        \caption{The average of object numbers (Obj.), symmetric object pairs (Sym.), and pairwise box IoU (PIoU) in unconditionally generated scenes. The closer to the statistics of GT, the better.} 
        \label{tab:num_obj}
        \end{center}
        \vspace{-8mm}
\end{table}

\paragraph{Evaluation Metrics}
Following previous works~\cite{wang2021sceneformer, yang2021scene, paschalidou2021atiss}, we use Fr\'echet inception distance (FID)~\cite{heusel2017gans}, Kernel inception distance~\cite{binkowski2018demystifying} (KID $\times$ 0.001), scene classification accuracy (SCA), and Category KL divergence (CKL $\times$ 0.01) to measure the plausibility and diversity of 1,000 synthesized scenes. For FID, KID, and SCA, we render the generated and ground-truth scenes into 256$\times$256 semantic maps through top-down orthographic projections, where the texture of each object is uniquely determined by the associate color of its semantic class. We use a unified camera and rendering setting for all methods to ensure fair comparisons.  For CKL, we calculate the KL divergence between the semantic class distributions of synthesized scenes and ground-truth scenes. For FID, KID, and CKL, the lower number denotes a better approximation of the data distribution. FID and KID can also manifest the result diversity. For the SCA, a score close to $50\%$ represents that the generated scenes are indistinguishable from real scenes.  Additionally, we delve into scene complexity, symmetry, and object interactions using the following metrics:
Number of objects (Obj): This metric quantifies the average object count per scene.
Number of symmetric object pairs (Sym): It measures the average number of symmetric object pairs in each scene.
Pair-wise object bounding box intersection over union (PIoU $\times$ 0.01) assesses the intersection over union between pairwise object bounding boxes. This metric provides insights into object interactions and intersections. 
 The proximity of Obj, Sym, and PIoU to the ground truth statistics indicates closeness in scene configuration patterns.
\subsection{Unconditional Scene Synthesis}
\label{SubSecExpSceSyn}
Fig.~\ref{fig:uncond_comparison} visualizes the qualitative comparisons of different scene synthesis methods. We observe that both DepthGAN~\cite{yang2021indoor} and Sync2Gen~\cite{yang2021scene} are vulnerable to object intersections. While ATISS~\cite{paschalidou2021atiss} can alleviate the penetration issue by autoregressive scene priors, it cannot always generate reasonable scene results. However, our scene diffusion can synthesize natural and diverse scene arrangements. Tab.~\ref{tab:uncond} presents the quantitative comparisons under various evaluation metrics. Our method consistently outperforms others in all metrics, which clearly demonstrates that our method can generate more diverse and plausible scenes.
\subsection{Ablation Studies}
\label{SubSecExpSceAbla}
\setlength{\tabcolsep}{3.5pt}
\begin{table}[!htbp]
    \vspace{-3mm}
        \setlength{\tabcolsep}{1.5pt}
	\begin{center}
		\begin{tabular}{*{8}{c}}
			\toprule
                Method & FID $\downarrow$ & KID $\downarrow$ & SCA $\%$  & CKL $\downarrow$   & Obj  & Sym & PIoU  \\ 
			\midrule
			\midrule
        


            C1
                    & 29.08  & 4.59 & 73.63  & 0.76  & 5.10 & 0.70 & 0.46 \\

            C2
                  & 19.78  & 2.07 & 54.53  & 0.69 & 5.03 & 0.63 & 0.38\\

            
            C3
                    & 17.93  & 1.29  & 55.14 & 0.46 & 5.02 & 0.64 & 0.47 \\ 
                    
            C4
                   & 18.40  & 1.55  & 55.42 & 0.66  & 4.97 & 0.50 & 0.52 \\

            C5
                    & \textbf{17.21} & \textbf{0.70}  & \textbf{52.15} & \textbf{0.35}   & \textbf{4.99} & \textbf{0.72} & \textbf{0.43} \\

        \bottomrule
        \end{tabular}
        \vspace{-2mm}
        \caption{Quantitative ablation studies on the task of unconditional scene synthesis on the 3D-FRONT bedrooms.} 
        \label{tab:ablation}
        \end{center}
        \vspace{-6mm}
\end{table}
We conduct detailed ablation studies to verify the effectiveness of each design in our scene diffusion models.
The quantitative results are provided in Tab.~\ref{tab:ablation}. We refer to the supplementary material for more detailed explanations.
%

%

\vspace{2mm}
\noindent \textbf{What is the effect of UNet-1D+Attention as the denoiser? (C1 vs. C5)} 
We investigate the different choices of denoising networks. The performances degrade when we use the transformer in DALLE-2~\cite{ramesh2022hierarchical}. 

\vspace{2mm}
\noindent \textbf{What is the effect of multiple prediction heads in the denoiser? (C2 vs. C5)} 
In the denoiser, we use three different encoding and prediction heads for respective object properties,~\eg bounding box parameter, semantic class labels, and geometry codes. 
Multiple diffusion heads with individual losses for attributes 
can prevent biasing towards one attribute in a single encoding and prediction head.

\vspace{2mm}
\noindent \textbf{What is the effect of the IoU loss? (C3 vs. C5)}
The IoU loss can penalize object intersections, promote more reasonable placements, and preserve symmetries. This is reflected by consistent improvement in each metric.

\vspace{2mm}
\noindent \textbf{What is the effect of geometry feature diffusion? (C4 vs. C5)}
\begin{figure}[!h]
    \vspace{-0.3cm}
    \centering
\includegraphics[width=\linewidth]{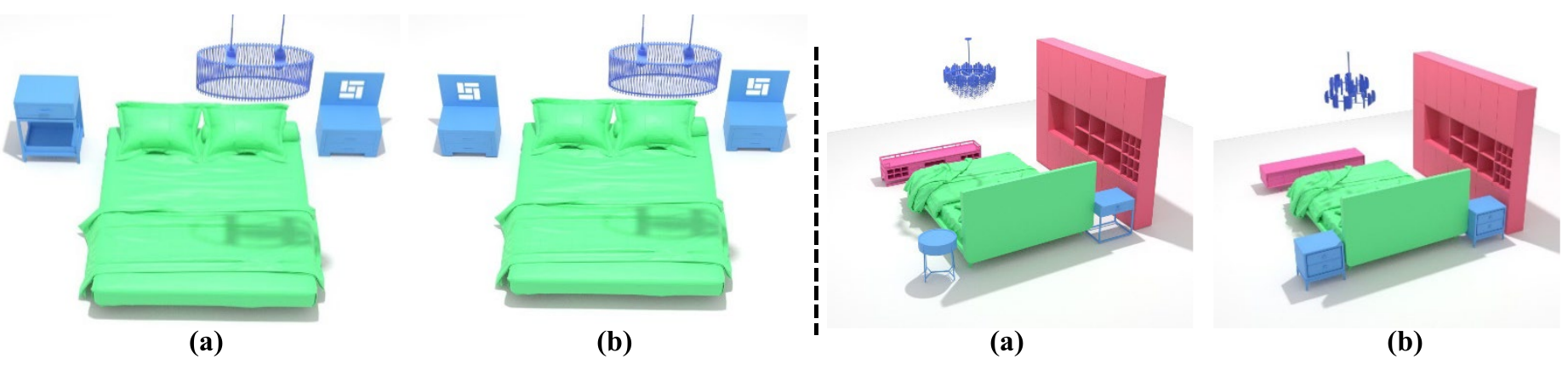}
\vspace{-0.3cm}
    \caption{
    (b) w/ shape diffusion captures symmetries vs. (a) w/o. The shape latent diffusion promotes symmetry discovery.
    }
    \label{fig:shape_diffusion}
\vspace{-0.3cm}
\end{figure}
The geometry feature enables better capture of symmetric placements and semantically coherent arrangements. Fig.~\ref{fig:shape_diffusion} shows that our model can find symmetric nightstands by beds due to the geometry awareness of the diffusion process and shape retrieval. This is supported by  Sym: 0.72 (w/ shape diffusion) vs. 0.50 (w/o shape diffusion) in Tab.~\ref{tab:ablation}.
More plausible synthesis results improve FID, KID, and SCA. Besides, the decrease in CKL can manifest that the joint diffusion of geometry code and object layout can learn more similar object class distribution.

\begin{figure}[!h]
    \vspace{-0.5cm}
    \centering
\includegraphics[width=\linewidth]{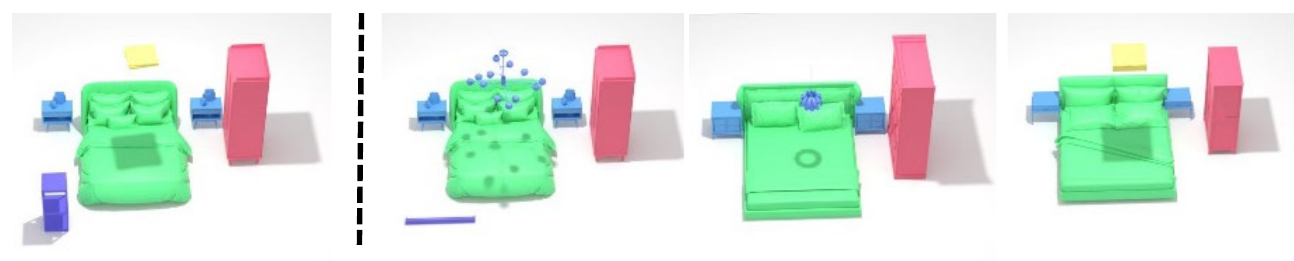}
\vspace{-0.5cm}
    \caption{
    \textbf{Left:} Ours. \textbf{Right:} top-3 nearest scenes in the train set.}
    \label{fig:scene_retrieval}
\vspace{-0.4cm}
\end{figure}

\noindent \textbf{Can DiffuScene generate novel scenes?}
In Fig.~\ref{fig:scene_retrieval}, We retrieve the three most similar training scenes for a generated scene using the Chamfer distance. Our result reveals unique object compositions, highlighting our method's ability to generate novel scenes rather than reproducing training data.
%
%
\begin{figure*}[!ht]
    \vspace{-2mm}
	\centering
	\begin{subfigure}[t]{0.14\textwidth}
            \includegraphics[width=\textwidth]{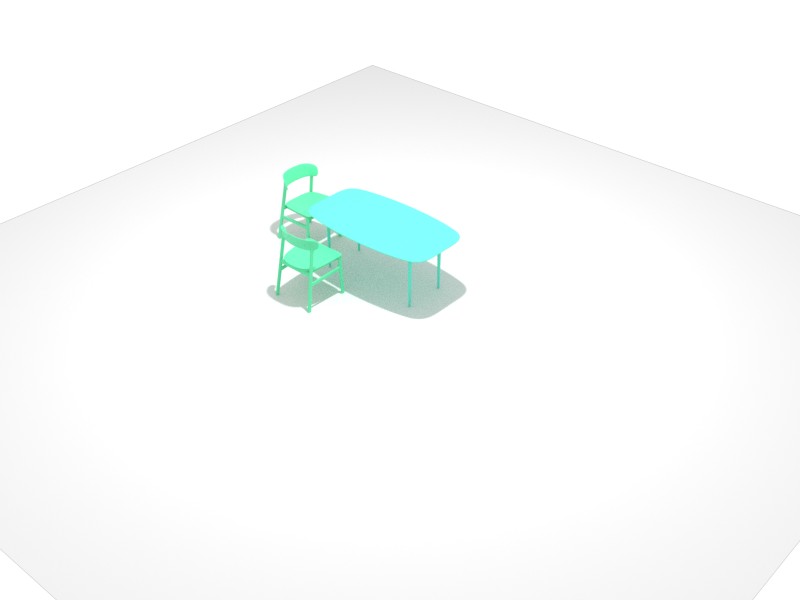}
        \caption{Partial Scenes}
	\end{subfigure}
        \rulesep
        \begin{subfigure}[t]{0.41\textwidth}
    	\includegraphics[width=0.33\textwidth]{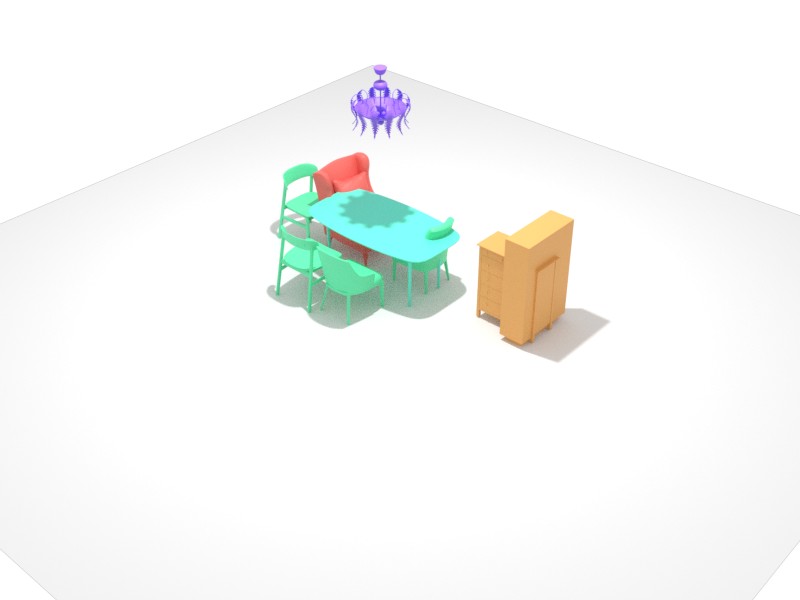}%
            \hfill
            \includegraphics[width=0.33\textwidth]{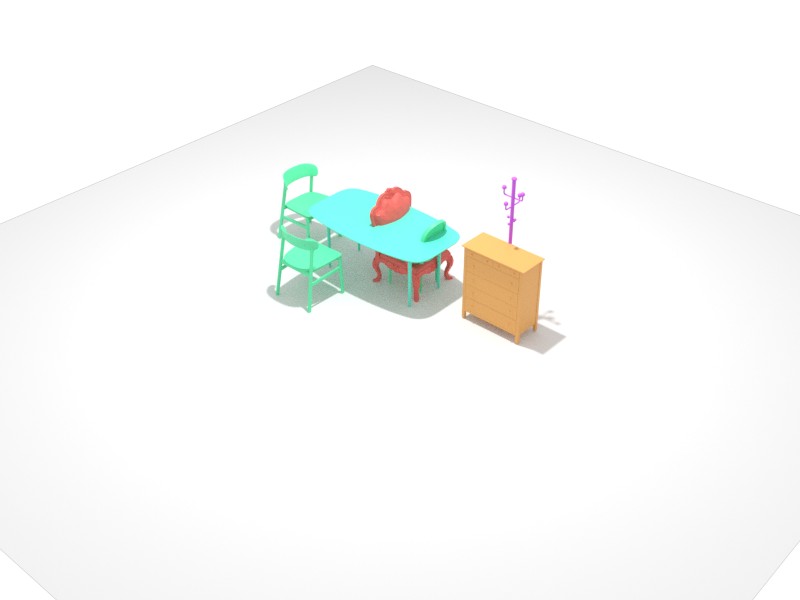}%
    	\hfill
            \includegraphics[width=0.33\textwidth]{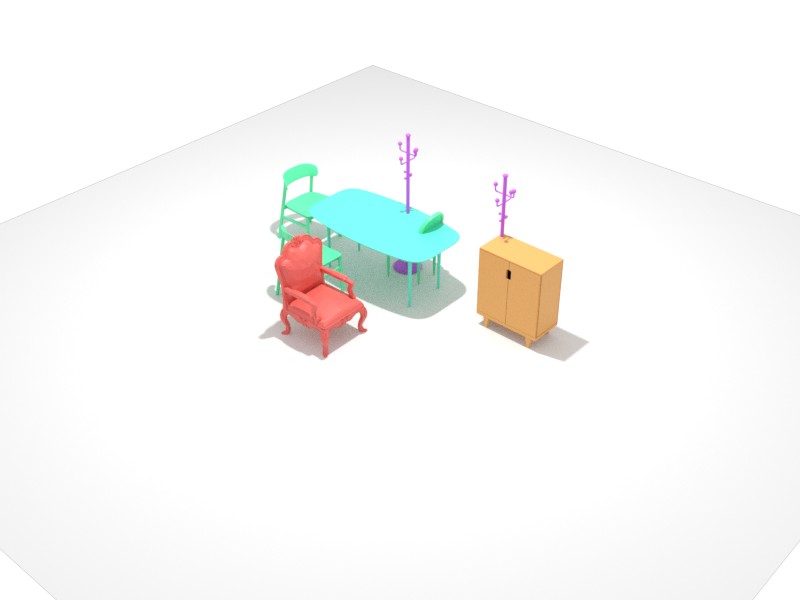}
        \caption{ATISS~\cite{paschalidou2021atiss}}
	\end{subfigure}
        \rulesep
	\begin{subfigure}[t]{0.41\textwidth}
    	\includegraphics[width=0.33\textwidth]{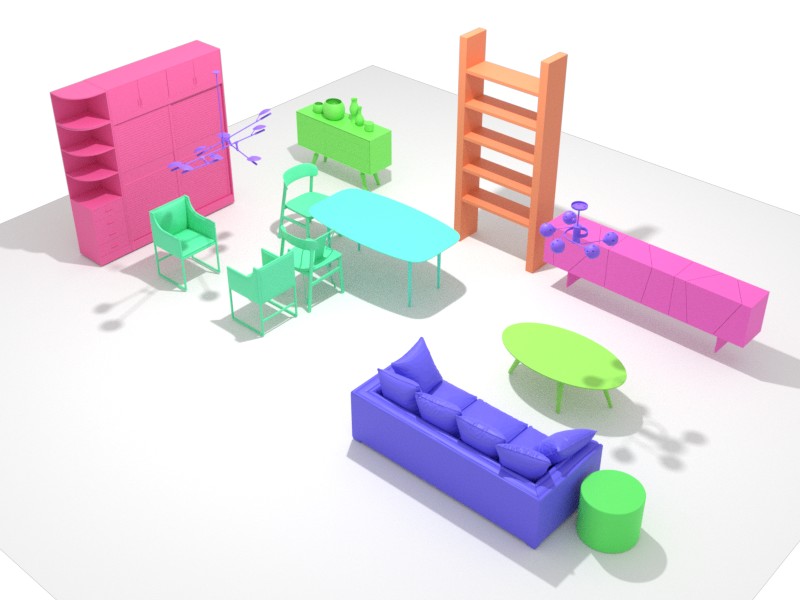}%
            \hfill
            \includegraphics[width=0.33\textwidth]{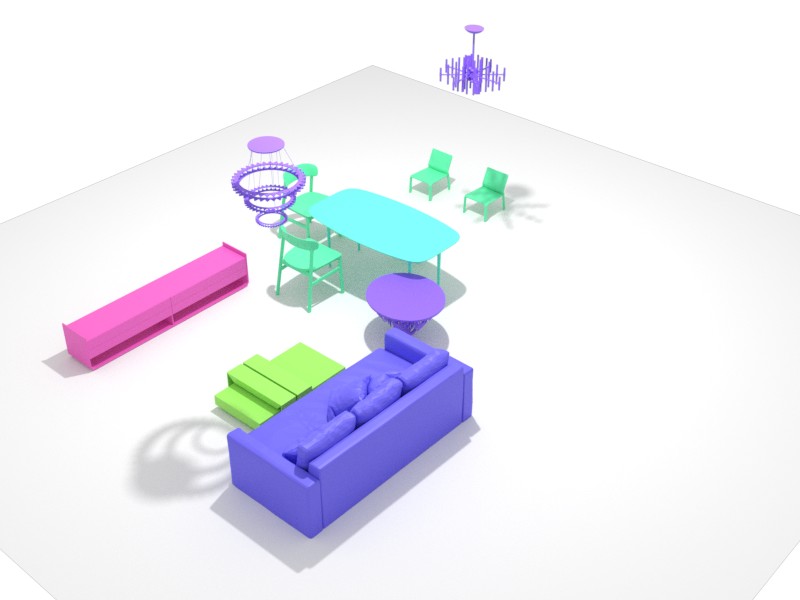}%
    	\hfill
    	\includegraphics[width=0.33\textwidth]{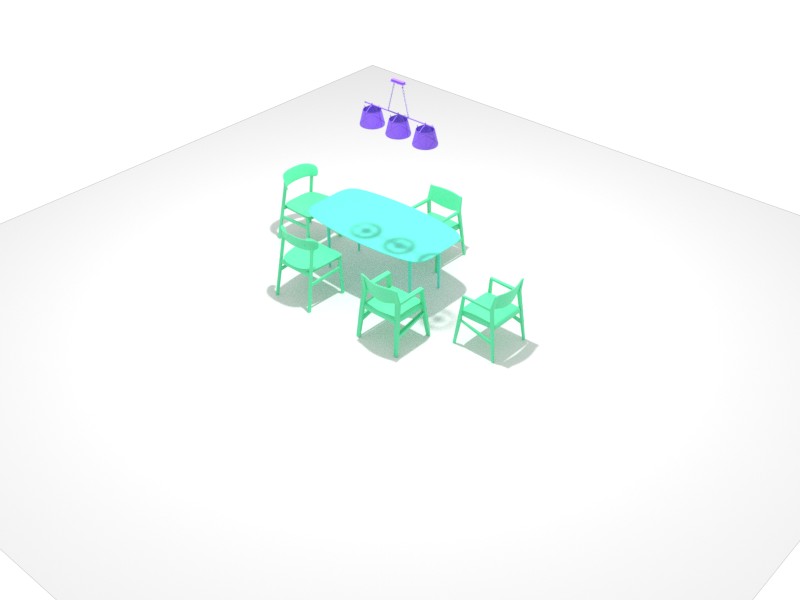}
		\caption{Ours}
	\end{subfigure}
	\caption{Scene completion from partial scenes with only 3 objects given as inputs. Compared to ATISS, our diffusion-based method produces more diverse completion results with higher fidelity, fewer intersections, and more symmetries.}
    \label{fig:completion}
    \vspace{-2mm}
\end{figure*}

\begin{figure*}[!ht]
	\centering
    	\begin{subfigure}[t]{0.23\textwidth}
            \includegraphics[width=\textwidth]{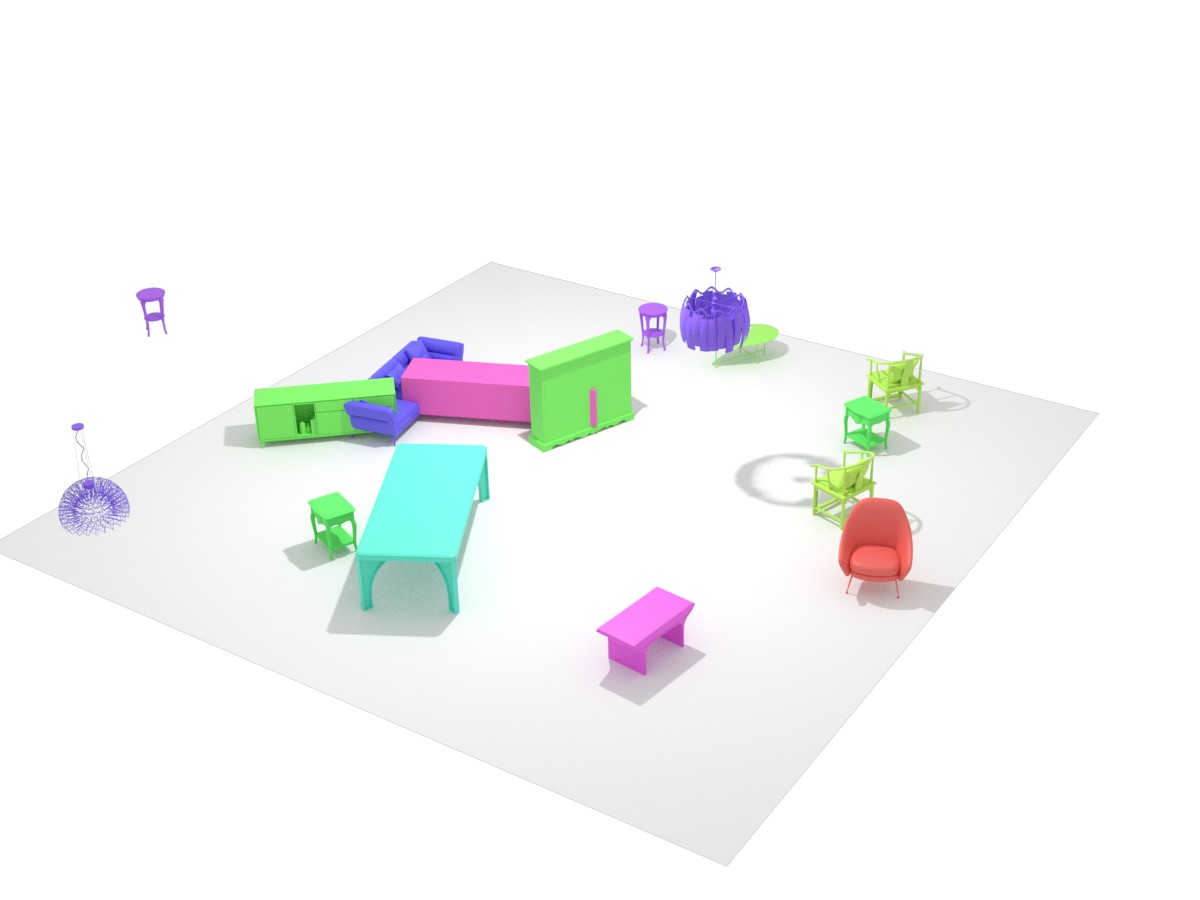}
            \caption{Noisy Scenes}
    	\end{subfigure}
        \rulesep
        \begin{subfigure}[t]{0.23\textwidth}
            \includegraphics[width=\textwidth]{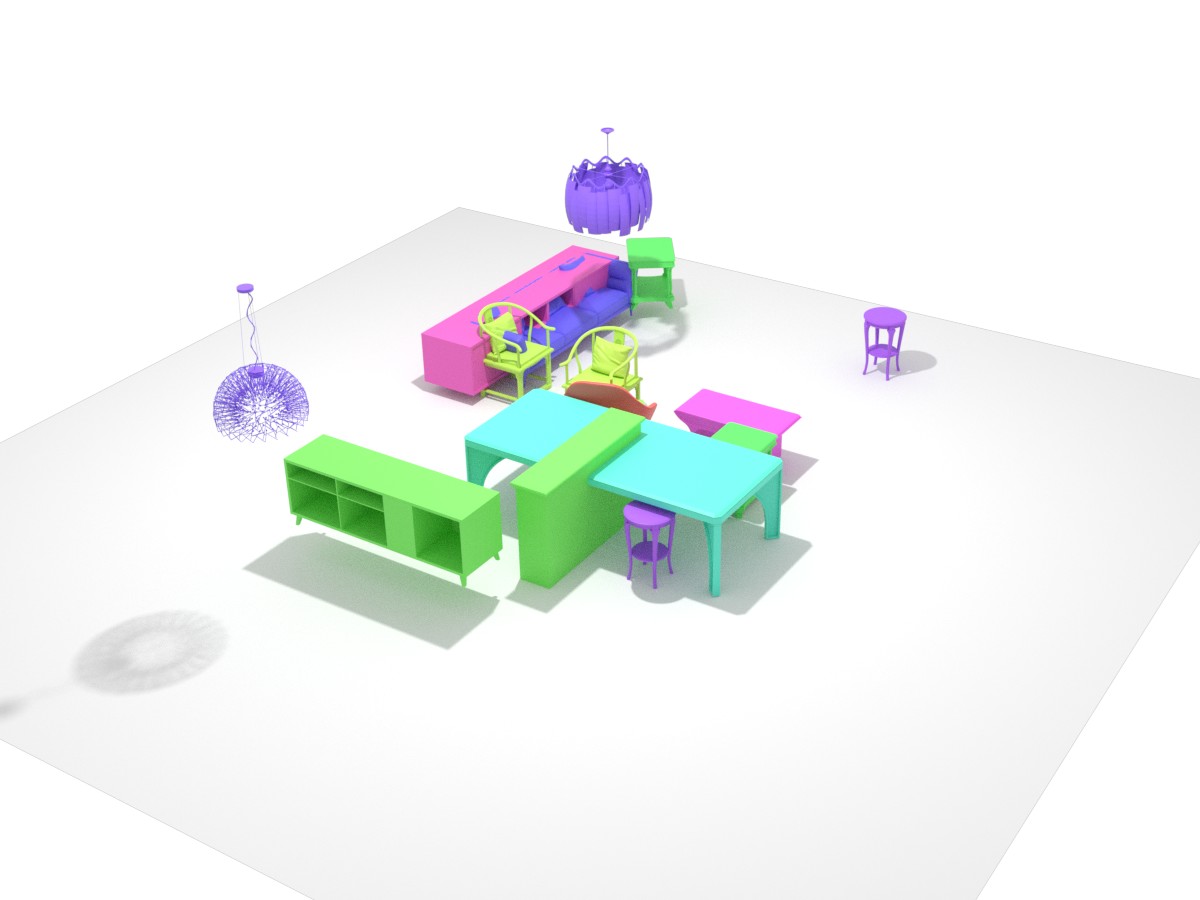}
            \caption{ATISS~\cite{paschalidou2021atiss}}
	   \end{subfigure}
        \rulesep
    	\begin{subfigure}[t]{0.23\textwidth}
            \includegraphics[width=\textwidth]{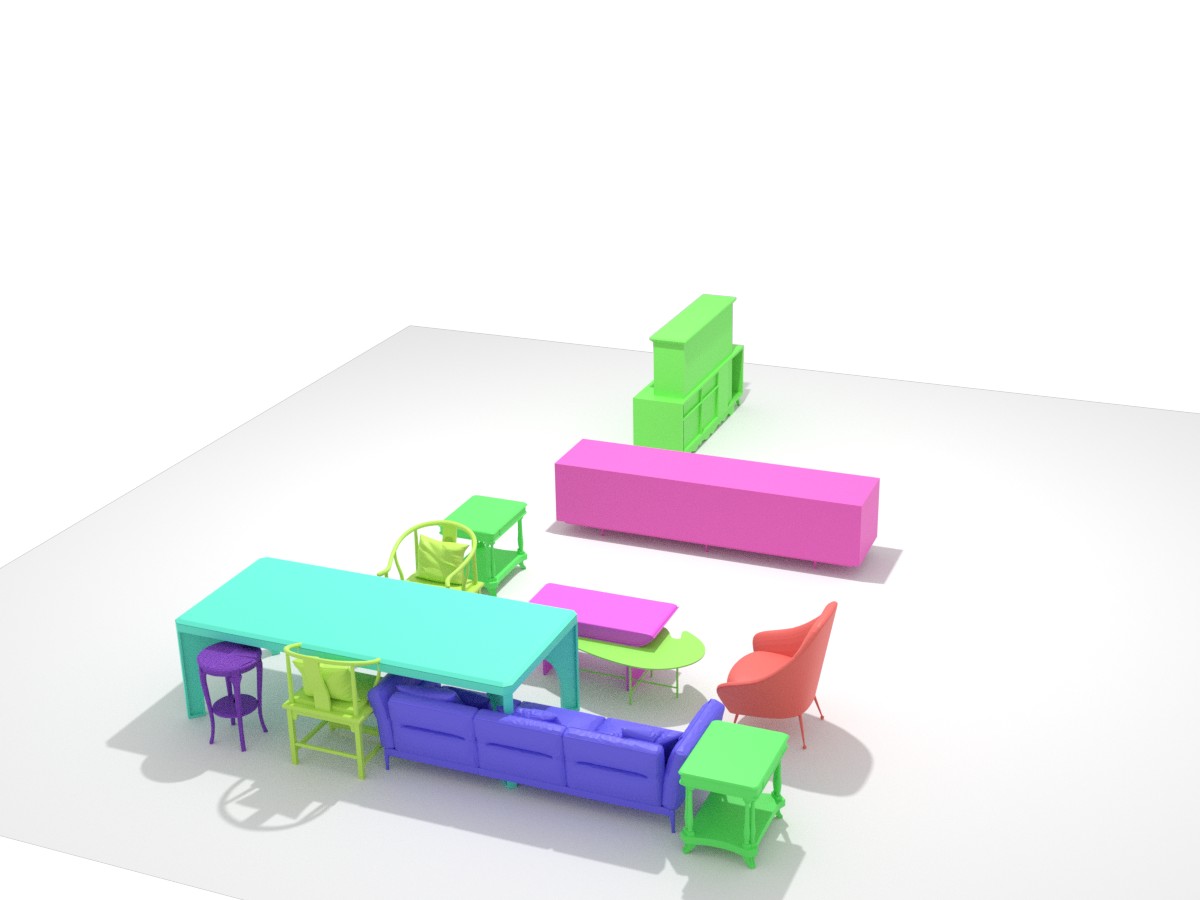}
    		\caption{LEGO~\cite{wei2023lego}}
    	\end{subfigure}
     \rulesep
     \begin{subfigure}[t]{0.23\textwidth}
            \includegraphics[width=\textwidth]{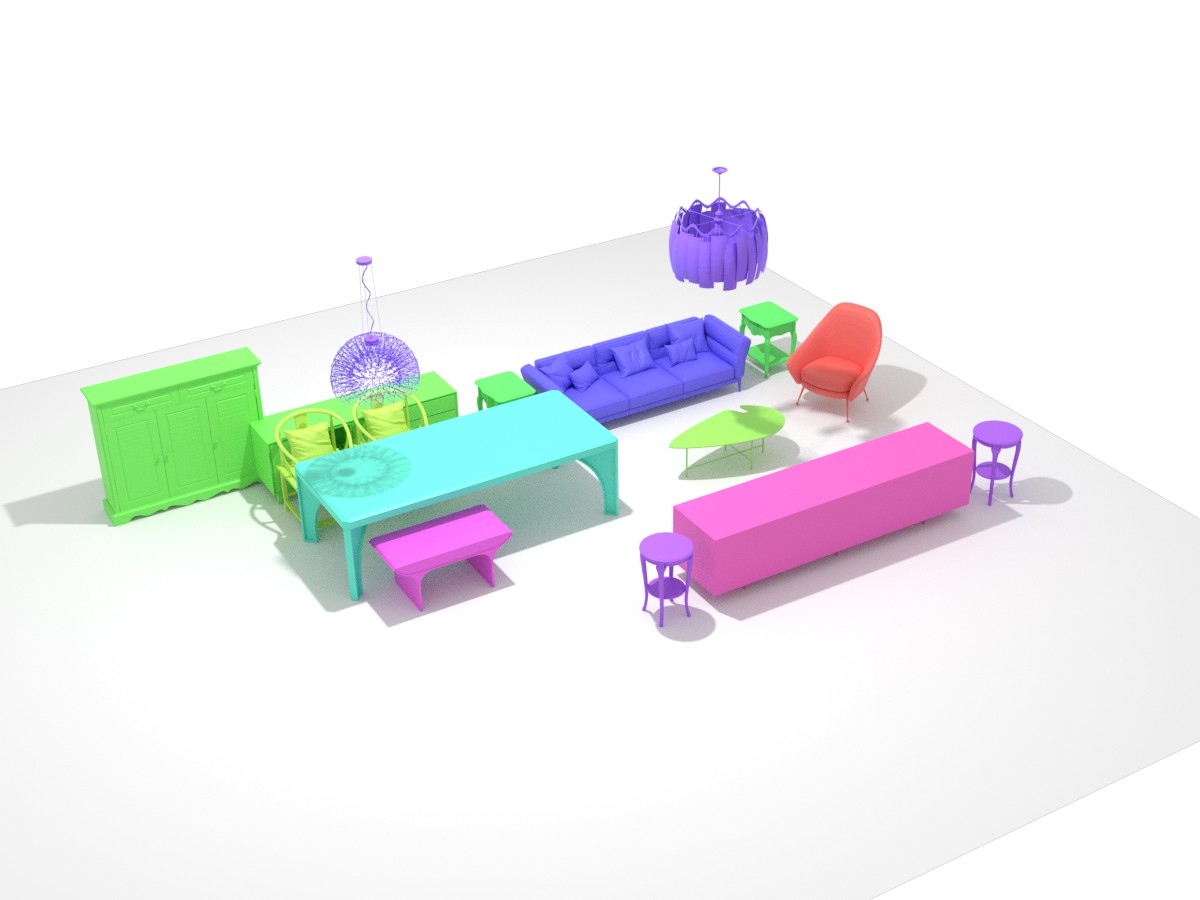}
    		\caption{Ours}
    	\end{subfigure}
	\caption{Scene re-arrangements of collections of random objects. Compared to ATISS and LEGO, our method generates more favourable object placements with more symmetric pairs.}
    \label{fig:arrangement}
    \vspace{-4mm}
\end{figure*}

\subsection{Applications}
\paragraph{Scene Completion}
We compare against ATISS~\cite{paschalidou2021atiss} on the task of scene completion. As shown in Fig.~\ref{fig:completion}, our method can produce more diverse completion results with high fidelity, fewer intersections, and more symmetries. 

\paragraph{Scene Re-arrangement}
\vspace{-6mm}
\begin{table}[!hbt]
        \setlength{\tabcolsep}{2.4pt}
	\begin{center}
		\begin{tabular}{*{6}{c}}
			\toprule
			{Room} & {Method} & FID $\downarrow$ & KID $\downarrow$ 
                   & \#Sym. & PIoU \\  
			\midrule
			\midrule
            \multirow{3}*{Bedroom} &  ATISS &  27.14  & 1.56  
                                        &  0.01 & 0.84  \\  
                                        
                                  & LEGO & 23.73 & 4.70 
                                        & 0.45 & 0.89 \\ 
                                  
                                  &  Ours   & \textbf{22.16} & \textbf{1.02}  
                                      & \textbf{0.70} & \textbf{0.61} \\ 

        \midrule
        \multirow{3}*{Living room} &  ATISS & 44.94  & 5.41 
                              &  1.42  & 1.73 \\ 
                              
                              &  LEGO & 45.40  & 9.57 
                              & 2.50 & 1.63 \\ 
                              
                              &  Ours & \textbf{41.15} &\textbf{2.24}   
                              & \textbf{3.69} & \textbf{0.95} \\ 
        \bottomrule
        \end{tabular}
        \vspace{-2mm}
        \caption{Quantitative comparisons on the task of \textbf{scene arrangement} on the 3D-FRONT bedrooms and dining rooms. Given a collection of objects as inputs, we predict their locations and orientations to obtain object placements.}
        \label{tab:arrange}
        \end{center}
        \vspace{-6mm}
\end{table}
We also conduct comparisons with ATISS~\cite{paschalidou2021atiss} on the application of scene re-arrangement. As depicted in Fig.~\ref{fig:arrangement}, our method generates more favorable object placements and more symmetric relations compared to ATISS~\cite{paschalidou2021atiss} and LEGO~\cite{wei2023lego}.
\paragraph{Text-conditioned Scene Synthesis}
Given a text prompt describing a partial scene configuration, we aim to synthesize a whole scene satisfying the input.
We conduct a perceptual user study for the text-conditioned scene synthesis.
Given a text prompt and a ground-truth scene as a reference, we ask the attendance two questions for each pair of results from ATISS and ours: which of the synthesized scenes is closely matched with the input text, and which one is more realistic and reasonable. We collect the answers of 225 scenes from 45 users. 
62$\%$ of users prefer our method to ATISS in realism. 
55$\%$ of users are in favor of us in the matching score. This illustrates that our text-conditioned model generates more realistic scenes while capturing more accurate object relationships described in the text prompt.
Please refer to the supplementary material for more details. 

\subsection{Limitations}
Although we have shown impressive scene synthesis results, our method still has some limitations. 
First, the shape retrieval searches the closest shape with the same semantics within defined classes of CAD models. Thus, the retrieved model could fail to match the style of desired scene. 
Second, the object textures are from the provided 3D CAD model dataset via shape retrieval. An interesting direction is to integrate texture diffusion into our model.  
Third, we only consider single-room generation and train our model on a specific room type. Thus, our method cannot synthesize large-scale scenes with multiple rooms. 
Finally, we rely on 3D labeled scenes to drive the learning of scene diffusion. Leveraging scene datasets with only 2D labels to learn scene diffusion priors is also a promising direction. We leave these mentioned limitations as our future efforts.
\begin{figure}[!ht]
        \vspace{-4mm}
	\centering
	\begin{subfigure}[t]{0.23\textwidth}
            \includegraphics[width=\textwidth]{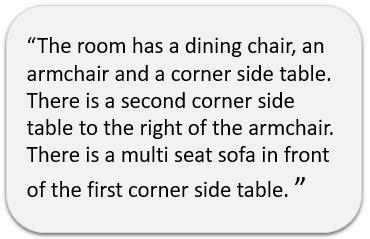}
        \caption{Input text}
	\end{subfigure}%
        \hfill
 	\begin{subfigure}[t]{0.23\textwidth}
            \includegraphics[width=\textwidth]{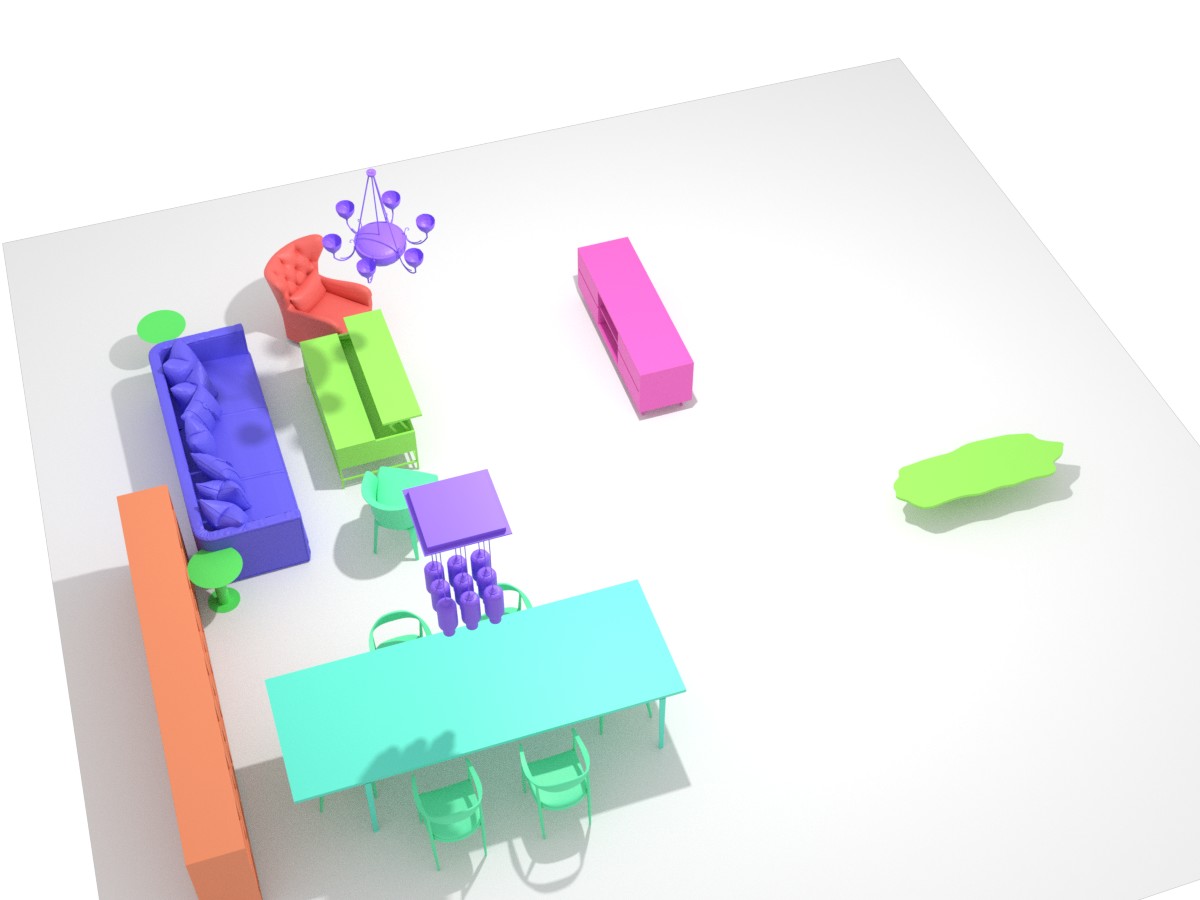}
        \caption{Reference}
	\end{subfigure}
 	\begin{subfigure}[t]{0.23\textwidth}
            \includegraphics[width=\textwidth]{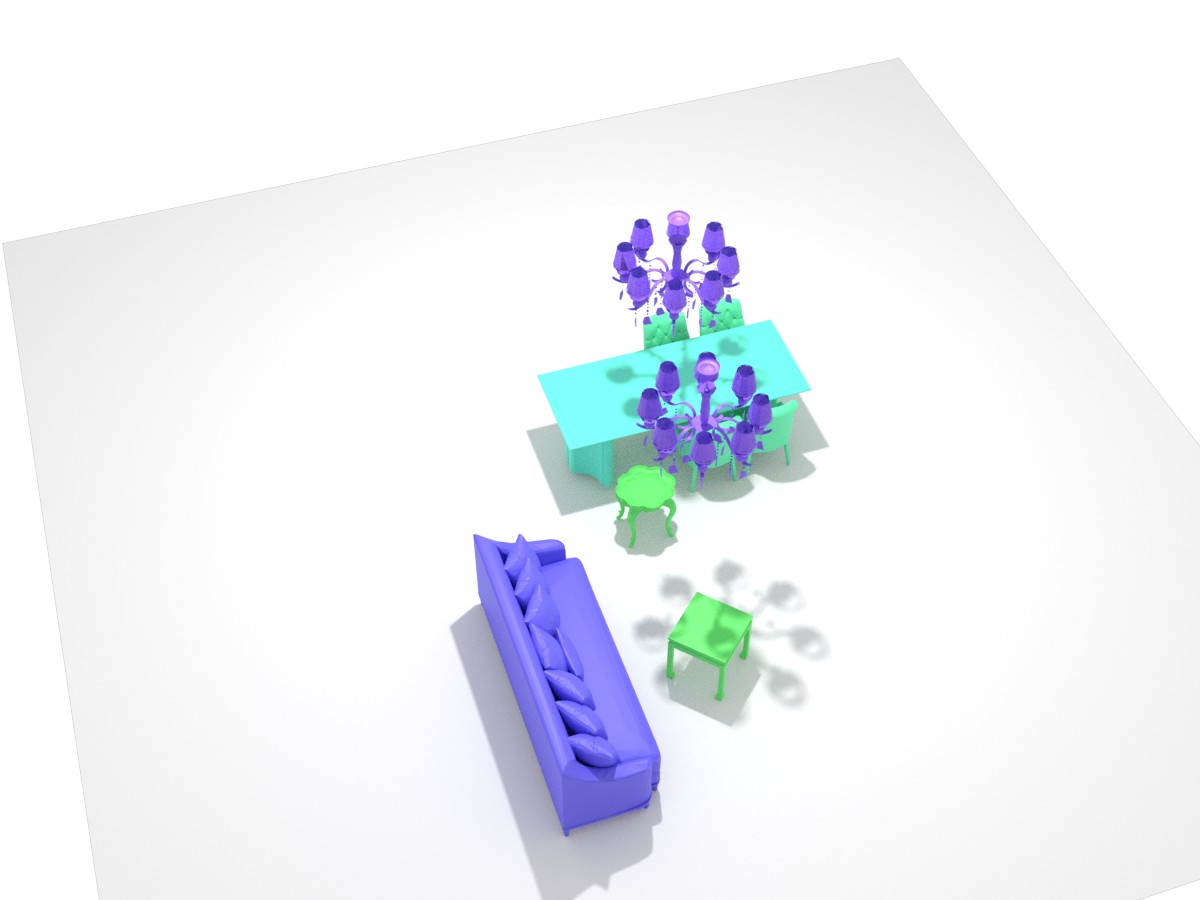}
        \caption{ATISS~\cite{paschalidou2021atiss}}
	\end{subfigure}%
        \hfill
 	\begin{subfigure}[t]{0.23\textwidth}
            \includegraphics[width=\textwidth]{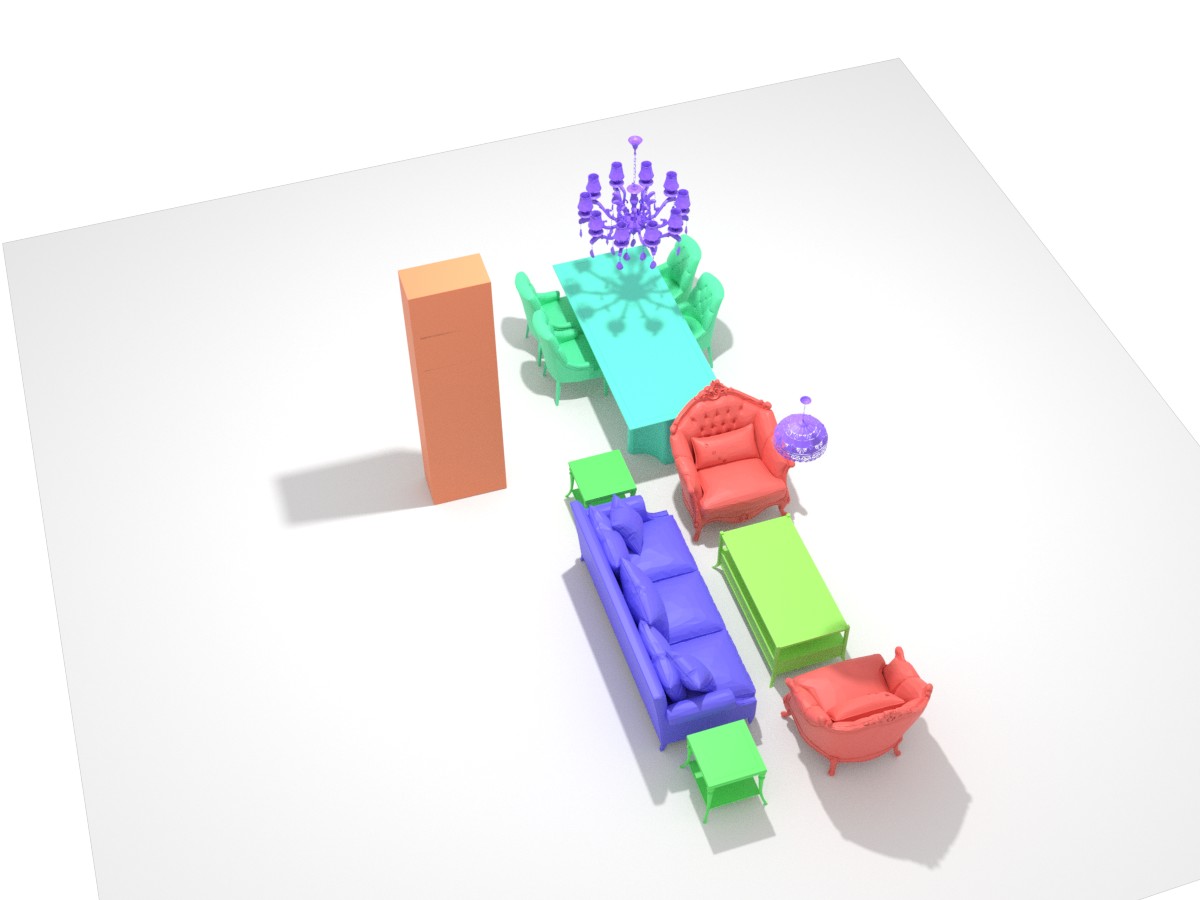}
        \caption{Ours}
	\end{subfigure}
	\caption{Text-conditioned scene synthesis. The input text only describes a partial scene configuration. Our method generates a more plausible scene matching the input text.}
    \label{fig:text2scene}
    \vspace{-5mm}
\end{figure}

\section{Conclusion}
\label{SecConclu}
In this work, we introduced DiffuScene, a novel method for generative indoor scene synthesis based on a denoising diffusion probabilistic model that learns holistic scene configuration priors in the full set diffusion process of object semantics, bounding boxes, and geometry features.
%
We applied our method to several downstream applications, namely scene completion, scene re-arrangement, and text-conditioned scene synthesis. Compared to prior state-of-the-art methods. Our approach can synthesize more plausible and diverse indoor scenes as has been measured by different metrics and confirmed in a user study.
Our method is an important piece in the puzzle of 3D generative modeling and we hope that it will inspire research in denoising diffusion-based 3D synthesis.

\paragraph{Acknowledgement.}
This work is supported by a TUM-IAS Rudolf M{\"{o}}{\ss}bauer Fellowship, the ERC
Starting Grant Scan2CAD (804724), and Sony Semiconductor Solutions.

{\small
\bibliographystyle{ieee_fullname}
\bibliography{egbib}

\begin{thebibliography}{87}
\providecommand{\natexlab}[1]{#1}
\providecommand{\url}[1]{\texttt{#1}}
\expandafter\ifx\csname urlstyle\endcsname\relax
  \providecommand{\doi}[1]{doi: #1}\else
  \providecommand{\doi}{doi: \begingroup \urlstyle{rm}\Url}\fi

\bibitem[Avrahami et~al.(2022)Avrahami, Lischinski, and
  Fried]{avrahami2022blended}
Omri Avrahami, Dani Lischinski, and Ohad Fried.
\newblock Blended diffusion for text-driven editing of natural images.
\newblock In \emph{Proceedings of the IEEE/CVF Conference on Computer Vision
  and Pattern Recognition}, pages 18208--18218, 2022.

\bibitem[Bi{\'n}kowski et~al.(2018)Bi{\'n}kowski, Sutherland, Arbel, and
  Gretton]{binkowski2018demystifying}
Miko{\l}aj Bi{\'n}kowski, Danica~J Sutherland, Michael Arbel, and Arthur
  Gretton.
\newblock Demystifying mmd gans.
\newblock \emph{arXiv preprint arXiv:1801.01401}, 2018.

\bibitem[Cao et~al.(2024)Cao, Luo, Zhang, Nie{\ss}ner, and
  Tang]{cao2024motion2vecsets}
Wei Cao, Chang Luo, Biao Zhang, Matthias Nie{\ss}ner, and Jiapeng Tang.
\newblock Motion2vecsets: 4d latent vector set diffusion for non-rigid shape
  reconstruction and tracking.
\newblock In \emph{Proceedings of the ieee/cvf conference on computer vision
  and pattern recognition}, 2024.

\bibitem[Chang et~al.(2014)Chang, Savva, and Manning]{chang2014learning}
Angel Chang, Manolis Savva, and Christopher~D Manning.
\newblock Learning spatial knowledge for text to 3d scene generation.
\newblock In \emph{Proceedings of the 2014 conference on empirical methods in
  natural language processing (EMNLP)}, pages 2028--2038, 2014.

\bibitem[Chang et~al.(2017)Chang, Eric, Savva, and Manning]{chang2017sceneseer}
Angel~X Chang, Mihail Eric, Manolis Savva, and Christopher~D Manning.
\newblock Sceneseer: 3d scene design with natural language.
\newblock \emph{arXiv preprint arXiv:1703.00050}, 2017.

\bibitem[Chen et~al.(2018)Chen, Rubanova, Bettencourt, and
  Duvenaud]{chen2018neural}
Ricky~TQ Chen, Yulia Rubanova, Jesse Bettencourt, and David~K Duvenaud.
\newblock Neural ordinary differential equations.
\newblock \emph{Advances in neural information processing systems}, 31, 2018.

\bibitem[Chen and Zhang(2019)]{chen2019learning}
Zhiqin Chen and Hao Zhang.
\newblock Learning implicit fields for generative shape modeling.
\newblock In \emph{CVPR}, 2019.

\bibitem[Choy et~al.(2016)Choy, Xu, Gwak, Chen, and Savarese]{choy20163d}
Christopher~B Choy, Danfei Xu, JunYoung Gwak, Kevin Chen, and Silvio Savarese.
\newblock 3d-r2n2: A unified approach for single and multi-view 3d object
  reconstruction.
\newblock In \emph{Computer Vision--ECCV 2016: 14th European Conference,
  Amsterdam, The Netherlands, October 11-14, 2016, Proceedings, Part VIII 14},
  pages 628--644. Springer, 2016.

\bibitem[Cong et~al.(2024)Cong, Xu, christian simon, Chen, Ren, Xie, Perez-Rua,
  Rosenhahn, Xiang, and He]{cong2024flatten}
Yuren Cong, Mengmeng Xu, christian simon, Shoufa Chen, Jiawei Ren, Yanping Xie,
  Juan-Manuel Perez-Rua, Bodo Rosenhahn, Tao Xiang, and Sen He.
\newblock {FLATTEN}: optical {FL}ow-guided {ATTEN}tion for consistent
  text-to-video editing.
\newblock In \emph{The Twelfth International Conference on Learning
  Representations}, 2024.

\bibitem[Dai and Nie{\ss}ner(2018)]{dai20183dmv}
Angela Dai and Matthias Nie{\ss}ner.
\newblock 3dmv: Joint 3d-multi-view prediction for 3d semantic scene
  segmentation.
\newblock In \emph{Proceedings of the European Conference on Computer Vision
  (ECCV)}, pages 452--468, 2018.

\bibitem[Devlin et~al.(2018)Devlin, Chang, Lee, and Toutanova]{devlin2018bert}
Jacob Devlin, Ming-Wei Chang, Kenton Lee, and Kristina Toutanova.
\newblock Bert: Pre-training of deep bidirectional transformers for language
  understanding.
\newblock \emph{arXiv preprint arXiv:1810.04805}, 2018.

\bibitem[Dhariwal and Nichol(2021)]{dhariwal2021diffusion}
Prafulla Dhariwal and Alexander Nichol.
\newblock Diffusion models beat gans on image synthesis.
\newblock \emph{Advances in Neural Information Processing Systems},
  34:\penalty0 8780--8794, 2021.

\bibitem[Esser et~al.(2021)Esser, Rombach, and Ommer]{esser2021taming}
Patrick Esser, Robin Rombach, and Bjorn Ommer.
\newblock Taming transformers for high-resolution image synthesis.
\newblock In \emph{Proceedings of the IEEE/CVF conference on computer vision
  and pattern recognition}, pages 12873--12883, 2021.

\bibitem[Fisher and Hanrahan(2010)]{fisher2010context}
Matthew Fisher and Pat Hanrahan.
\newblock Context-based search for 3d models.
\newblock In \emph{ACM SIGGRAPH Asia 2010 papers}, pages 1--10. 2010.

\bibitem[Fisher et~al.(2012)Fisher, Ritchie, Savva, Funkhouser, and
  Hanrahan]{fisher2012example}
Matthew Fisher, Daniel Ritchie, Manolis Savva, Thomas Funkhouser, and Pat
  Hanrahan.
\newblock Example-based synthesis of 3d object arrangements.
\newblock \emph{ACM Transactions on Graphics (TOG)}, 31\penalty0 (6):\penalty0
  1--11, 2012.

\bibitem[Fisher et~al.(2015)Fisher, Savva, Li, Hanrahan, and
  Nie{\ss}ner]{fisher2015activity}
Matthew Fisher, Manolis Savva, Yangyan Li, Pat Hanrahan, and Matthias
  Nie{\ss}ner.
\newblock Activity-centric scene synthesis for functional 3d scene modeling.
\newblock \emph{ACM Transactions on Graphics (TOG)}, 34\penalty0 (6):\penalty0
  1--13, 2015.

\bibitem[Fu et~al.(2021{\natexlab{a}})Fu, Cai, Gao, Zhang, Wang, Li, Zeng, Sun,
  Jia, Zhao, et~al.]{fu20213d}
Huan Fu, Bowen Cai, Lin Gao, Ling-Xiao Zhang, Jiaming Wang, Cao Li, Qixun Zeng,
  Chengyue Sun, Rongfei Jia, Binqiang Zhao, et~al.
\newblock 3d-front: 3d furnished rooms with layouts and semantics.
\newblock In \emph{Proceedings of the IEEE/CVF International Conference on
  Computer Vision}, pages 10933--10942, 2021{\natexlab{a}}.

\bibitem[Fu et~al.(2021{\natexlab{b}})Fu, Jia, Gao, Gong, Zhao, Maybank, and
  Tao]{fu20213dm}
Huan Fu, Rongfei Jia, Lin Gao, Mingming Gong, Binqiang Zhao, Steve Maybank, and
  Dacheng Tao.
\newblock 3d-future: 3d furniture shape with texture.
\newblock \emph{International Journal of Computer Vision}, 129:\penalty0
  3313--3337, 2021{\natexlab{b}}.

\bibitem[Fu et~al.(2017)Fu, Chen, Wang, Wen, Zhou, and Fu]{fu2017adaptive}
Qiang Fu, Xiaowu Chen, Xiaotian Wang, Sijia Wen, Bin Zhou, and Hongbo Fu.
\newblock Adaptive synthesis of indoor scenes via activity-associated object
  relation graphs.
\newblock \emph{ACM Transactions on Graphics (TOG)}, 36\penalty0 (6):\penalty0
  1--13, 2017.

\bibitem[Goodfellow et~al.(2020)Goodfellow, Pouget-Abadie, Mirza, Xu,
  Warde-Farley, Ozair, Courville, and Bengio]{goodfellow2020generative}
Ian Goodfellow, Jean Pouget-Abadie, Mehdi Mirza, Bing Xu, David Warde-Farley,
  Sherjil Ozair, Aaron Courville, and Yoshua Bengio.
\newblock Generative adversarial networks.
\newblock \emph{Communications of the ACM}, 63\penalty0 (11):\penalty0
  139--144, 2020.

\bibitem[Graves(2013)]{graves2013generating}
Alex Graves.
\newblock Generating sequences with recurrent neural networks.
\newblock \emph{arXiv preprint arXiv:1308.0850}, 2013.

\bibitem[Han et~al.(2019)Han, Zhang, Du, Yang, Yu, Pan, Yang, Liu, Xiong, and
  Cui]{han2019deep}
Xiaoguang Han, Zhaoxuan Zhang, Dong Du, Mingdai Yang, Jingming Yu, Pan Pan, Xin
  Yang, Ligang Liu, Zixiang Xiong, and Shuguang Cui.
\newblock Deep reinforcement learning of volume-guided progressive view
  inpainting for 3d point scene completion from a single depth image.
\newblock In \emph{Proceedings of the IEEE/CVF Conference on Computer Vision
  and Pattern Recognition}, pages 234--243, 2019.

\bibitem[Heusel et~al.(2017)Heusel, Ramsauer, Unterthiner, Nessler, and
  Hochreiter]{heusel2017gans}
Martin Heusel, Hubert Ramsauer, Thomas Unterthiner, Bernhard Nessler, and Sepp
  Hochreiter.
\newblock Gans trained by a two time-scale update rule converge to a local nash
  equilibrium.
\newblock \emph{Advances in neural information processing systems}, 30, 2017.

\bibitem[Ho and Salimans(2022)]{ho2022classifier}
Jonathan Ho and Tim Salimans.
\newblock Classifier-free diffusion guidance.
\newblock \emph{arXiv preprint arXiv:2207.12598}, 2022.

\bibitem[Ho et~al.(2020)Ho, Jain, and Abbeel]{ho2020denoising}
Jonathan Ho, Ajay Jain, and Pieter Abbeel.
\newblock Denoising diffusion probabilistic models.
\newblock \emph{Advances in Neural Information Processing Systems},
  33:\penalty0 6840--6851, 2020.

\bibitem[Ho et~al.(2022{\natexlab{a}})Ho, Chan, Saharia, Whang, Gao, Gritsenko,
  Kingma, Poole, Norouzi, Fleet, et~al.]{ho2022imagen}
Jonathan Ho, William Chan, Chitwan Saharia, Jay Whang, Ruiqi Gao, Alexey
  Gritsenko, Diederik~P Kingma, Ben Poole, Mohammad Norouzi, David~J Fleet,
  et~al.
\newblock Imagen video: High definition video generation with diffusion models.
\newblock \emph{arXiv preprint arXiv:2210.02303}, 2022{\natexlab{a}}.

\bibitem[Ho et~al.(2022{\natexlab{b}})Ho, Saharia, Chan, Fleet, Norouzi, and
  Salimans]{ho2022cascaded}
Jonathan Ho, Chitwan Saharia, William Chan, David~J Fleet, Mohammad Norouzi,
  and Tim Salimans.
\newblock Cascaded diffusion models for high fidelity image generation.
\newblock \emph{J. Mach. Learn. Res.}, 23:\penalty0 47--1, 2022{\natexlab{b}}.

\bibitem[Hui et~al.(2022)Hui, Li, Hu, and Fu]{hui2022neural}
Ka-Hei Hui, Ruihui Li, Jingyu Hu, and Chi-Wing Fu.
\newblock Neural wavelet-domain diffusion for 3d shape generation.
\newblock In \emph{SIGGRAPH Asia 2022 Conference Papers}, pages 1--9, 2022.

\bibitem[Jiang et~al.(2012)Jiang, Lim, and Saxena]{jiang2012learning}
Yun Jiang, Marcus Lim, and Ashutosh Saxena.
\newblock Learning object arrangements in 3d scenes using human context.
\newblock \emph{arXiv preprint arXiv:1206.6462}, 2012.

\bibitem[Kim et~al.(2022)Kim, Kwon, and Ye]{kim2022diffusionclip}
Gwanghyun Kim, Taesung Kwon, and Jong~Chul Ye.
\newblock Diffusionclip: Text-guided diffusion models for robust image
  manipulation.
\newblock In \emph{Proceedings of the IEEE/CVF Conference on Computer Vision
  and Pattern Recognition}, pages 2426--2435, 2022.

\bibitem[Kingma and Welling(2013)]{kingma2013auto}
Diederik~P Kingma and Max Welling.
\newblock Auto-encoding variational bayes.
\newblock \emph{arXiv preprint arXiv:1312.6114}, 2013.

\bibitem[Kynk{\"a}{\"a}nniemi et~al.(2019)Kynk{\"a}{\"a}nniemi, Karras, Laine,
  Lehtinen, and Aila]{kynkaanniemi2019improved}
Tuomas Kynk{\"a}{\"a}nniemi, Tero Karras, Samuli Laine, Jaakko Lehtinen, and
  Timo Aila.
\newblock Improved precision and recall metric for assessing generative models.
\newblock \emph{Advances in Neural Information Processing Systems}, 32, 2019.

\bibitem[Li et~al.(2019)Li, Patil, Xu, Chaudhuri, Khan, Shamir, Tu, Chen,
  Cohen-Or, and Zhang]{li2019grains}
Manyi Li, Akshay~Gadi Patil, Kai Xu, Siddhartha Chaudhuri, Owais Khan, Ariel
  Shamir, Changhe Tu, Baoquan Chen, Daniel Cohen-Or, and Hao Zhang.
\newblock Grains: Generative recursive autoencoders for indoor scenes.
\newblock \emph{ACM Transactions on Graphics (TOG)}, 38\penalty0 (2):\penalty0
  1--16, 2019.

\bibitem[Lugmayr et~al.(2022)Lugmayr, Danelljan, Romero, Yu, Timofte, and
  Van~Gool]{lugmayr2022repaint}
Andreas Lugmayr, Martin Danelljan, Andres Romero, Fisher Yu, Radu Timofte, and
  Luc Van~Gool.
\newblock Repaint: Inpainting using denoising diffusion probabilistic models.
\newblock In \emph{Proceedings of the IEEE/CVF Conference on Computer Vision
  and Pattern Recognition}, pages 11461--11471, 2022.

\bibitem[Luo and Hu(2021)]{luo2021diffusion}
Shitong Luo and Wei Hu.
\newblock Diffusion probabilistic models for 3d point cloud generation.
\newblock In \emph{Proceedings of the IEEE/CVF Conference on Computer Vision
  and Pattern Recognition}, pages 2837--2845, 2021.

\bibitem[Ma et~al.(2016)Ma, Li, Zou, Liao, Tong, and Zhang]{ma2016action}
Rui Ma, Honghua Li, Changqing Zou, Zicheng Liao, Xin Tong, and Hao Zhang.
\newblock Action-driven 3d indoor scene evolution.
\newblock \emph{ACM Trans. Graph.}, 35\penalty0 (6):\penalty0 173--1, 2016.

\bibitem[Meng et~al.(2021)Meng, He, Song, Song, Wu, Zhu, and
  Ermon]{meng2021sdedit}
Chenlin Meng, Yutong He, Yang Song, Jiaming Song, Jiajun Wu, Jun-Yan Zhu, and
  Stefano Ermon.
\newblock Sdedit: Guided image synthesis and editing with stochastic
  differential equations.
\newblock In \emph{International Conference on Learning Representations}, 2021.

\bibitem[Merrell et~al.(2011)Merrell, Schkufza, Li, Agrawala, and
  Koltun]{merrell2011interactive}
Paul Merrell, Eric Schkufza, Zeyang Li, Maneesh Agrawala, and Vladlen Koltun.
\newblock Interactive furniture layout using interior design guidelines.
\newblock \emph{ACM transactions on graphics (TOG)}, 30\penalty0 (4):\penalty0
  1--10, 2011.

\bibitem[Mescheder et~al.(2019)Mescheder, Oechsle, Niemeyer, Nowozin, and
  Geiger]{mescheder2019occupancy}
Lars Mescheder, Michael Oechsle, Michael Niemeyer, Sebastian Nowozin, and
  Andreas Geiger.
\newblock Occupancy networks: Learning 3d reconstruction in function space.
\newblock In \emph{CVPR}, 2019.

\bibitem[Mildenhall et~al.(2021)Mildenhall, Srinivasan, Tancik, Barron,
  Ramamoorthi, and Ng]{mildenhall2021nerf}
Ben Mildenhall, Pratul~P Srinivasan, Matthew Tancik, Jonathan~T Barron, Ravi
  Ramamoorthi, and Ren Ng.
\newblock Nerf: Representing scenes as neural radiance fields for view
  synthesis.
\newblock \emph{Communications of the ACM}, 65\penalty0 (1):\penalty0 99--106,
  2021.

\bibitem[Nichol et~al.(2021)Nichol, Dhariwal, Ramesh, Shyam, Mishkin, McGrew,
  Sutskever, and Chen]{nichol2021glide}
Alex Nichol, Prafulla Dhariwal, Aditya Ramesh, Pranav Shyam, Pamela Mishkin,
  Bob McGrew, Ilya Sutskever, and Mark Chen.
\newblock Glide: Towards photorealistic image generation and editing with
  text-guided diffusion models.
\newblock \emph{arXiv preprint arXiv:2112.10741}, 2021.

\bibitem[Nie et~al.(2022)Nie, Dai, Han, and Nie{\ss}ner]{nie2022learning}
Yinyu Nie, Angela Dai, Xiaoguang Han, and Matthias Nie{\ss}ner.
\newblock Learning 3d scene priors with 2d supervision.
\newblock \emph{arXiv preprint arXiv:2211.14157}, 2022.

\bibitem[Park et~al.(2019)Park, Florence, Straub, Newcombe, and
  Lovegrove]{park2019deepsdf}
Jeong~Joon Park, Peter Florence, Julian Straub, Richard Newcombe, and Steven
  Lovegrove.
\newblock Deepsdf: Learning continuous signed distance functions for shape
  representation.
\newblock In \emph{CVPR}, 2019.

\bibitem[Paschalidou et~al.(2021)Paschalidou, Kar, Shugrina, Kreis, Geiger, and
  Fidler]{paschalidou2021atiss}
Despoina Paschalidou, Amlan Kar, Maria Shugrina, Karsten Kreis, Andreas Geiger,
  and Sanja Fidler.
\newblock Atiss: Autoregressive transformers for indoor scene synthesis.
\newblock \emph{Advances in Neural Information Processing Systems},
  34:\penalty0 12013--12026, 2021.

\bibitem[Purkait et~al.(2020)Purkait, Zach, and Reid]{purkait2020sg}
Pulak Purkait, Christopher Zach, and Ian Reid.
\newblock Sg-vae: Scene grammar variational autoencoder to generate new indoor
  scenes.
\newblock In \emph{Computer Vision--ECCV 2020: 16th European Conference,
  Glasgow, UK, August 23--28, 2020, Proceedings, Part XXIV 16}, pages 155--171.
  Springer, 2020.

\bibitem[Qi et~al.(2017)Qi, Su, Mo, and Guibas]{qi2017pointnet}
Charles~R Qi, Hao Su, Kaichun Mo, and Leonidas~J Guibas.
\newblock Pointnet: Deep learning on point sets for 3d classification and
  segmentation.
\newblock In \emph{Proceedings of the IEEE conference on computer vision and
  pattern recognition}, pages 652--660, 2017.

\bibitem[Qi et~al.(2018)Qi, Zhu, Huang, Jiang, and Zhu]{qi2018human}
Siyuan Qi, Yixin Zhu, Siyuan Huang, Chenfanfu Jiang, and Song-Chun Zhu.
\newblock Human-centric indoor scene synthesis using stochastic grammar.
\newblock In \emph{Proceedings of the IEEE Conference on Computer Vision and
  Pattern Recognition}, pages 5899--5908, 2018.

\bibitem[Ramesh et~al.(2022)Ramesh, Dhariwal, Nichol, Chu, and
  Chen]{ramesh2022hierarchical}
Aditya Ramesh, Prafulla Dhariwal, Alex Nichol, Casey Chu, and Mark Chen.
\newblock Hierarchical text-conditional image generation with clip latents.
\newblock \emph{arXiv preprint arXiv:2204.06125}, 2022.

\bibitem[Razavi et~al.(2019)Razavi, Van~den Oord, and
  Vinyals]{razavi2019generating}
Ali Razavi, Aaron Van~den Oord, and Oriol Vinyals.
\newblock Generating diverse high-fidelity images with vq-vae-2.
\newblock \emph{Advances in neural information processing systems}, 32, 2019.

\bibitem[Rezende and Mohamed(2015)]{rezende2015variational}
Danilo Rezende and Shakir Mohamed.
\newblock Variational inference with normalizing flows.
\newblock In \emph{International conference on machine learning}, pages
  1530--1538. PMLR, 2015.

\bibitem[Ritchie et~al.(2019)Ritchie, Wang, and Lin]{ritchie2019fast}
Daniel Ritchie, Kai Wang, and Yu-an Lin.
\newblock Fast and flexible indoor scene synthesis via deep convolutional
  generative models.
\newblock In \emph{Proceedings of the IEEE/CVF Conference on Computer Vision
  and Pattern Recognition}, pages 6182--6190, 2019.

\bibitem[Rombach et~al.(2022)Rombach, Blattmann, Lorenz, Esser, and
  Ommer]{rombach2022high}
Robin Rombach, Andreas Blattmann, Dominik Lorenz, Patrick Esser, and Bj{\"o}rn
  Ommer.
\newblock High-resolution image synthesis with latent diffusion models.
\newblock In \emph{Proceedings of the IEEE/CVF Conference on Computer Vision
  and Pattern Recognition}, pages 10684--10695, 2022.

\bibitem[Saharia et~al.(2022)Saharia, Ho, Chan, Salimans, Fleet, and
  Norouzi]{saharia2022image}
Chitwan Saharia, Jonathan Ho, William Chan, Tim Salimans, David~J Fleet, and
  Mohammad Norouzi.
\newblock Image super-resolution via iterative refinement.
\newblock \emph{IEEE Transactions on Pattern Analysis and Machine
  Intelligence}, 2022.

\bibitem[Savva et~al.(2017)Savva, Chang, and Agrawala]{savva2017scenesuggest}
Manolis Savva, Angel~X Chang, and Maneesh Agrawala.
\newblock Scenesuggest: Context-driven 3d scene design.
\newblock \emph{arXiv preprint arXiv:1703.00061}, 2017.

\bibitem[Sohl-Dickstein et~al.(2015)Sohl-Dickstein, Weiss, Maheswaranathan, and
  Ganguli]{sohl2015deep}
Jascha Sohl-Dickstein, Eric Weiss, Niru Maheswaranathan, and Surya Ganguli.
\newblock Deep unsupervised learning using nonequilibrium thermodynamics.
\newblock In \emph{International Conference on Machine Learning}, pages
  2256--2265. PMLR, 2015.

\bibitem[Song et~al.(2020{\natexlab{a}})Song, Meng, and
  Ermon]{song2020denoising}
Jiaming Song, Chenlin Meng, and Stefano Ermon.
\newblock Denoising diffusion implicit models.
\newblock \emph{arXiv preprint arXiv:2010.02502}, 2020{\natexlab{a}}.

\bibitem[Song and Ermon(2019)]{song2019generative}
Yang Song and Stefano Ermon.
\newblock Generative modeling by estimating gradients of the data distribution.
\newblock \emph{Advances in Neural Information Processing Systems}, 32, 2019.

\bibitem[Song and Ermon(2020)]{song2020improved}
Yang Song and Stefano Ermon.
\newblock Improved techniques for training score-based generative models.
\newblock \emph{Advances in neural information processing systems},
  33:\penalty0 12438--12448, 2020.

\bibitem[Song et~al.(2020{\natexlab{b}})Song, Sohl-Dickstein, Kingma, Kumar,
  Ermon, and Poole]{song2020score}
Yang Song, Jascha Sohl-Dickstein, Diederik~P Kingma, Abhishek Kumar, Stefano
  Ermon, and Ben Poole.
\newblock Score-based generative modeling through stochastic differential
  equations.
\newblock \emph{arXiv preprint arXiv:2011.13456}, 2020{\natexlab{b}}.

\bibitem[Tang et~al.(2019)Tang, Han, Pan, Jia, and Tong]{tang2019skeleton}
Jiapeng Tang, Xiaoguang Han, Junyi Pan, Kui Jia, and Xin Tong.
\newblock A skeleton-bridged deep learning approach for generating meshes of
  complex topologies from single rgb images.
\newblock In \emph{Proceedings of the IEEE/CVF Conference on Computer Vision
  and Pattern Recognition}, pages 4541--4550, 2019.

\bibitem[Tang et~al.(2021)Tang, Lei, Xu, Ma, Jia, and Zhang]{tang2021sa}
Jiapeng Tang, Jiabao Lei, Dan Xu, Feiying Ma, Kui Jia, and Lei Zhang.
\newblock Sa-convonet: Sign-agnostic optimization of convolutional occupancy
  networks.
\newblock In \emph{Proceedings of the IEEE/CVF International Conference on
  Computer Vision}, pages 6504--6513, 2021.

\bibitem[Tang et~al.(2022)Tang, Markhasin, Wang, Thies, and
  Nie{\ss}ner]{tang2022neural}
Jiapeng Tang, Lev Markhasin, Bi Wang, Justus Thies, and Matthias Nie{\ss}ner.
\newblock Neural shape deformation priors.
\newblock In \emph{Advances in Neural Information Processing Systems}, 2022.

\bibitem[Tang et~al.(2024)Tang, Dai, Nie, Markhasin, Thies, and
  Niessner]{tang2024dphms}
Jiapeng Tang, Angela Dai, Yinyu Nie, Lev Markhasin, Justus Thies, and Matthias
  Niessner.
\newblock Dphms: Diffusion parametric head models for depth-based tracking.
\newblock 2024.

\bibitem[Van~den Oord et~al.(2016)Van~den Oord, Kalchbrenner, Espeholt,
  Vinyals, Graves, et~al.]{van2016conditional}
Aaron Van~den Oord, Nal Kalchbrenner, Lasse Espeholt, Oriol Vinyals, Alex
  Graves, et~al.
\newblock Conditional image generation with pixelcnn decoders.
\newblock \emph{Advances in neural information processing systems}, 29, 2016.

\bibitem[Van Den~Oord et~al.(2017)Van Den~Oord, Vinyals, et~al.]{van2017neural}
Aaron Van Den~Oord, Oriol Vinyals, et~al.
\newblock Neural discrete representation learning.
\newblock \emph{Advances in neural information processing systems}, 30, 2017.

\bibitem[Vaswani et~al.(2017)Vaswani, Shazeer, Parmar, Uszkoreit, Jones, Gomez,
  Kaiser, and Polosukhin]{vaswani2017attention}
Ashish Vaswani, Noam Shazeer, Niki Parmar, Jakob Uszkoreit, Llion Jones,
  Aidan~N Gomez, {\L}ukasz Kaiser, and Illia Polosukhin.
\newblock Attention is all you need.
\newblock \emph{Advances in neural information processing systems}, 30, 2017.

\bibitem[Wang et~al.(2018)Wang, Savva, Chang, and Ritchie]{wang2018deep}
Kai Wang, Manolis Savva, Angel~X Chang, and Daniel Ritchie.
\newblock Deep convolutional priors for indoor scene synthesis.
\newblock \emph{ACM Transactions on Graphics (TOG)}, 37\penalty0 (4):\penalty0
  1--14, 2018.

\bibitem[Wang et~al.(2019{\natexlab{a}})Wang, Lin, Weissmann, Savva, Chang, and
  Ritchie]{wang2019planit}
Kai Wang, Yu-An Lin, Ben Weissmann, Manolis Savva, Angel~X Chang, and Daniel
  Ritchie.
\newblock Planit: Planning and instantiating indoor scenes with relation graph
  and spatial prior networks.
\newblock \emph{ACM Transactions on Graphics (TOG)}, 38\penalty0 (4):\penalty0
  1--15, 2019{\natexlab{a}}.

\bibitem[Wang et~al.(2021)Wang, Yeshwanth, and
  Nie{\ss}ner]{wang2021sceneformer}
Xinpeng Wang, Chandan Yeshwanth, and Matthias Nie{\ss}ner.
\newblock Sceneformer: Indoor scene generation with transformers.
\newblock In \emph{2021 International Conference on 3D Vision (3DV)}, pages
  106--115. IEEE, 2021.

\bibitem[Wang et~al.(2019{\natexlab{b}})Wang, Sun, Liu, Sarma, Bronstein, and
  Solomon]{wang2019dynamic}
Yue Wang, Yongbin Sun, Ziwei Liu, Sanjay~E Sarma, Michael~M Bronstein, and
  Justin~M Solomon.
\newblock Dynamic graph cnn for learning on point clouds.
\newblock \emph{Acm Transactions On Graphics (tog)}, 38\penalty0 (5):\penalty0
  1--12, 2019{\natexlab{b}}.

\bibitem[Wei et~al.(2023)Wei, Ding, Park, Sajnani, Poulenard, Sridhar, and
  Guibas]{wei2023lego}
Qiuhong~Anna Wei, Sijie Ding, Jeong~Joon Park, Rahul Sajnani, Adrien Poulenard,
  Srinath Sridhar, and Leonidas Guibas.
\newblock Lego-net: Learning regular rearrangements of objects in rooms.
\newblock \emph{arXiv preprint arXiv:2301.09629}, 2023.

\bibitem[Wu et~al.(2016)Wu, Zhang, Xue, Freeman, and Tenenbaum]{wu2016learning}
Jiajun Wu, Chengkai Zhang, Tianfan Xue, Bill Freeman, and Josh Tenenbaum.
\newblock Learning a probabilistic latent space of object shapes via 3d
  generative-adversarial modeling.
\newblock \emph{Advances in neural information processing systems}, 29, 2016.

\bibitem[Wu et~al.(2022)Wu, Zhong, Xia, and Dong]{wu2022targf}
Mingdong Wu, Fangwei Zhong, Yulong Xia, and Hao Dong.
\newblock Targf: Learning target gradient field for object rearrangement.
\newblock \emph{arXiv preprint arXiv:2209.00853}, 2022.

\bibitem[Xiao et~al.(2021)Xiao, Kreis, and Vahdat]{xiao2021tackling}
Zhisheng Xiao, Karsten Kreis, and Arash Vahdat.
\newblock Tackling the generative learning trilemma with denoising diffusion
  gans.
\newblock \emph{arXiv preprint arXiv:2112.07804}, 2021.

\bibitem[Xu et~al.(2013)Xu, Chen, Fu, Sun, and Hu]{xu2013sketch2scene}
Kun Xu, Kang Chen, Hongbo Fu, Wei-Lun Sun, and Shi-Min Hu.
\newblock Sketch2scene: Sketch-based co-retrieval and co-placement of 3d
  models.
\newblock \emph{ACM Transactions on Graphics (TOG)}, 32\penalty0 (4):\penalty0
  1--15, 2013.

\bibitem[Yang et~al.(2021{\natexlab{a}})Yang, Zhang, Yan, Huang, Ma, Zheng,
  Bajaj, and Huang]{yang2021scene}
Haitao Yang, Zaiwei Zhang, Siming Yan, Haibin Huang, Chongyang Ma, Yi Zheng,
  Chandrajit Bajaj, and Qixing Huang.
\newblock Scene synthesis via uncertainty-driven attribute synchronization.
\newblock In \emph{Proceedings of the IEEE/CVF International Conference on
  Computer Vision}, pages 5630--5640, 2021{\natexlab{a}}.

\bibitem[Yang et~al.(2021{\natexlab{b}})Yang, Guo, Zhou, and
  Tong]{yang2021indoor}
Ming-Jia Yang, Yu-Xiao Guo, Bin Zhou, and Xin Tong.
\newblock Indoor scene generation from a collection of semantic-segmented depth
  images.
\newblock In \emph{Proceedings of the IEEE/CVF International Conference on
  Computer Vision}, pages 15203--15212, 2021{\natexlab{b}}.

\bibitem[Yang et~al.(2018)Yang, Feng, Shen, and Tian]{yang2018foldingnet}
Yaoqing Yang, Chen Feng, Yiru Shen, and Dong Tian.
\newblock Foldingnet: Point cloud auto-encoder via deep grid deformation.
\newblock In \emph{Proceedings of the IEEE conference on computer vision and
  pattern recognition}, pages 206--215, 2018.

\bibitem[Yeh et~al.(2012)Yeh, Yang, Watson, Goodman, and
  Hanrahan]{yeh2012synthesizing}
Yi-Ting Yeh, Lingfeng Yang, Matthew Watson, Noah~D Goodman, and Pat Hanrahan.
\newblock Synthesizing open worlds with constraints using locally annealed
  reversible jump mcmc.
\newblock \emph{ACM Transactions on Graphics (TOG)}, 31\penalty0 (4):\penalty0
  1--11, 2012.

\bibitem[Yin et~al.(2021)Yin, Zhou, and Krahenbuhl]{yin2021center}
Tianwei Yin, Xingyi Zhou, and Philipp Krahenbuhl.
\newblock Center-based 3d object detection and tracking.
\newblock In \emph{Proceedings of the IEEE/CVF conference on computer vision
  and pattern recognition}, pages 11784--11793, 2021.

\bibitem[Yu et~al.(2011)Yu, Yeung, Tang, Terzopoulos, Chan, and
  Osher]{yu2011make}
Lap~Fai Yu, Sai~Kit Yeung, Chi~Keung Tang, Demetri Terzopoulos, Tony~F Chan,
  and Stanley~J Osher.
\newblock Make it home: automatic optimization of furniture arrangement.
\newblock \emph{ACM Transactions on Graphics (TOG)-Proceedings of ACM SIGGRAPH
  2011, v. 30,(4), July 2011, article no. 86}, 30\penalty0 (4), 2011.

\bibitem[Zeng et~al.(2022)Zeng, Vahdat, Williams, Gojcic, Litany, Fidler, and
  Kreis]{zeng2022lion}
Xiaohui Zeng, Arash Vahdat, Francis Williams, Zan Gojcic, Or Litany, Sanja
  Fidler, and Karsten Kreis.
\newblock Lion: Latent point diffusion models for 3d shape generation.
\newblock \emph{arXiv preprint arXiv:2210.06978}, 2022.

\bibitem[Zhai et~al.(2024)Zhai, {\"O}rnek, Wu, Di, Tombari, Navab, and
  Busam]{zhai2024commonscenes}
Guangyao Zhai, Evin~P{\i}nar {\"O}rnek, Shun-Cheng Wu, Yan Di, Federico
  Tombari, Nassir Navab, and Benjamin Busam.
\newblock Commonscenes: Generating commonsense 3d indoor scenes with scene
  graphs.
\newblock \emph{Advances in Neural Information Processing Systems}, 36, 2024.

\bibitem[Zhang and Wonka(2024)]{zhang2023functional}
Biao Zhang and Peter Wonka.
\newblock Functional diffusion.
\newblock In \emph{Proceedings of the IEEE/CVF Conference on Computer Vision
  and Pattern Recognition}, 2024.

\bibitem[Zhang et~al.(2023)Zhang, Tang, Niessner, and
  Wonka]{zhang20233dshape2vecset}
Biao Zhang, Jiapeng Tang, Matthias Niessner, and Peter Wonka.
\newblock 3dshape2vecset: A 3d shape representation for neural fields and
  generative diffusion models.
\newblock \emph{arXiv preprint arXiv:2301.11445}, 2023.

\bibitem[Zhang et~al.(2020)Zhang, Yang, Ma, Luo, Huth, Vouga, and
  Huang]{zhang2020deep}
Zaiwei Zhang, Zhenpei Yang, Chongyang Ma, Linjie Luo, Alexander Huth, Etienne
  Vouga, and Qixing Huang.
\newblock Deep generative modeling for scene synthesis via hybrid
  representations.
\newblock \emph{ACM Transactions on Graphics (TOG)}, 39\penalty0 (2):\penalty0
  1--21, 2020.

\bibitem[Zhou et~al.(2021)Zhou, Du, and Wu]{zhou20213d}
Linqi Zhou, Yilun Du, and Jiajun Wu.
\newblock 3d shape generation and completion through point-voxel diffusion.
\newblock In \emph{Proceedings of the IEEE/CVF International Conference on
  Computer Vision}, pages 5826--5835, 2021.

\end{thebibliography}
}

\cleardoublepage
\appendix
\section*{Appendix}
In this supplemental material, we provide details for our implementation in Sec.~\ref{SecImple}, dataset pre-processing and text prompt generation in Sec.~\ref{SecData}, baseline implementations in Sec.~\ref{SecBaseline}, additional results in Sec.~\ref{SecAddRes}, and user studies in Sec.~\ref{SecUser}.

\section{Implementations}
\label{SecImple}

\subsection{Shape Auto-Encoder}
\label{SubSecShapeAE}

\begin{figure}[!htp]
    \centering
    \includegraphics[width=\linewidth]{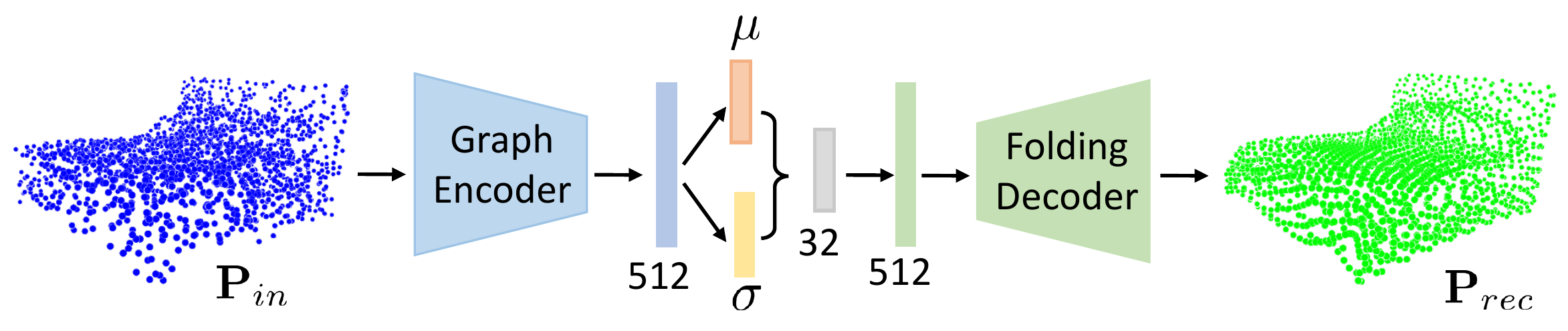}
    \caption{\textbf{Shape Auto-encoder.}}
    \label{fig:shapeae}
\end{figure}

We adopt a pre-trained shape auto-encoder to extract a set of latent shape codes for CAD models from the 3D-FUTURE~\cite{fu20213dm} dataset. The network architecture of the shape auto-encoder is shown in Fig.~\ref{fig:shapeae}. It is a variational auto-encoder, similar to FoldingNet~\cite{yang2018foldingnet}.
Specifically, a point cloud $\mathbf{P}_{in}$ of size 2,048 is fed into a graph encoder based on PointNet~\cite{qi2017pointnet} with graph convolutions~\cite{wang2019dynamic} to extract a global latent code of dimension 512, which is used to predict the mean $\mathbf{\mu}$ and variance $\mathbf{\sigma}$ of a low-dimensional latent space of size 32.
Subsequently, a compressed latent is sampled from $\mathcal{N}(\mathbf{\mu}, \mathbf{\sigma})$.
Finally, the compressed latent is mapped back to the original space and passed to the FoldingNet decoder to recover a point cloud $\mathbf{P}_{rec}$ of size 2,025.
The used training objective is a weighted combination of Chamfer distance (\ie CD) and KL divergence.
\begin{equation}
    \label{EquaShapeAE}
        L_{vae} = \CD(\mathbf{P}_{in}, \mathbf{P}_{rec}) + \omega_{kl} *\KL(\mathcal{N}(\mathbf{\mu}, \mathbf{\sigma}) || \mathcal{N}(\mathbf{0}, \mathbf{I})) ,
\end{equation}
where $\omega_{kl}$ is set to 0.001.
The latent compression and KL regularization leads to a compact and structured latent space, focusing on global shape structures.
The shape autoencoder is trained on a single RTX 2080 with a batch size of 16 for 1,000 epochs.
The learning rate is initialized to $lr=\expnumber{1}{-4}$ and then gradually decreases with the decay rate of 0.1 in every 400 epochs.

\subsection{Shape Code Diffusion}
\label{SubSecShapeDiffu}

We use the extracted latent codes to train shape code diffusion.
While we apply KL regularization, the value range of latent codes is still unbound.
To make it easier to diffuse, we scale the latent codes to $[-1, 1]$ by using the statistical minimum and maximum feature values over the whole set.
During inference, we rescale generated shape codes.

\subsection{Shape Retrieval}
\label{SubSecRetrieval}

During inference, we use shape retrieval as the post-processing procedure to acquire object surface geometries for generated scenes.
Concretely, for each instance, we perform the nearest neighbor search in the 3D-FUTURE~\cite{fu20213dm} dataset to find the CAD model with the same class label and the closest geometry feature.


\section{Dataset}
\label{SecData}

\paragraph{Preprocessing}
The dataset preprocessing is based on the setting of ATISS~\cite{paschalidou2021atiss}.
We start by filtering out those scenes with problematic object arrangements such as severe object intersections or incorrect object class labels, e.g., beds are misclassified as wardrobes in some scenes.
Then, we remove those scenes with unnatural sizes. The floor size of a natural room is within $6m \times 6m$ and its height is less than $4m$. Subsequently, we ignore scenes that have too few or many objects.
The number of objects in valid bedrooms is between 3 and 13. As for dining and living rooms, the minimum and maximum numbers are set to 3 and 21 respectively. Thus, the number of objects is $N=13$ in bedrooms and $N=21$ in dining and living rooms. In addition, we delete scenes that have objects out of pre-defined categories. After pre-processing, we obtained 4,041 bedrooms, 900 dining rooms, and 813 living rooms.

For the semantic class diffusion, we have an additional class of  `empty' to define the existence of an object. Combining with the object categories that appeared in each room type, we have $L=22$ object categories for bedrooms, and
$L=25$ object categories for dining and living rooms in total. The category labels 
are listed as follows.

\begin{python}
# 22 3D-Front bedroom categories
['empty', 'armchair', 'bookshelf', 'cabinet',
'ceiling_lamp', 'chair', 'children_cabinet',
'coffee_table', 'desk', 'double_bed',
'dressing_chair', 'dressing_table', 'kids_bed',
'nightstand', 'pendant_lamp', 'shelf',
'single_bed', 'sofa', 'stool', 'table',
'tv_stand', 'wardrobe']

# 25 3D-Front dining or living room categories
['empty', 'armchair', 'bookshelf', 'cabinet', 
'ceiling_lamp', 'chaise_longue_sofa', 
'chinese_chair', 'coffee_table', 'console_table',  
'corner_side_table',  'desk', 'dining_chair', 
'dining_table', 'l_shaped_sofa', 'lazy_sofa', 
'lounge_chair', 'loveseat_sofa', 
'multi_seat_sofa', 'pendant_lamp', 
'round_end_table', 'shelf', 'stool', 
'tv_stand', 'wardrobe', 'wine_cabinet']
\end{python}

\paragraph{Text Prompt Generation}
We follow the SceneFormer~\cite{wang2021sceneformer} to generate text prompts describing partial scene configurations. Each text prompt contains one to three sentences. We explain the details of text formulation process by using the text prompt 'The room has a dining table, a pendant lamp, and a lounge chair. The pendant lamp is above the dining table. There is a stool to the right of the lounge chair.` as an example. First, we randomly select three objects from a scene, get their class labels, and then count the number of appearances of each selected object category. As such, we can get the first sentence. Then, we find all valid object pairs associated with the selected three objects. An object pair is valid only if the distance between two objects is less than a certain threshold that is set to 1.5 in our method. Next, we calculate the relative orientations and translations, from which we can determine the relationship type of the valid object pair from the candidate pool: 'is above to`, 'is next to`, 'is left of`, 'is right of`, ' surrounding`, 'inside`, 'behind`, 'in front of`, and 'on`. In this way, we can acquire some relation-describing sentences like the second and third sentences in the example. Finally, we randomly sampled zero to two relation-describing sentences.

\section{Baselines}
\label{SecBaseline}

\paragraph{DepthGAN} 
DepthGAN~\cite{yang2021indoor} adopts a generative adversary network to train 3D scene synthesis using both semantic maps and depth images. The generator network is built with 3D convolution layers, which decode a volumetric scene with semantic labels. A differentiable projection layer is applied to project the semantic scene volume into depth images and semantic maps under different views, where a multi-view discriminator is designed to distinguish the synthesized views from ground-truth semantic maps and depth images during the adversarial training.

\paragraph{Sync2Gen} 
Sync2Gen~\cite{yang2021scene} represents a scene arrangement as a sequence of 3D objects characterized by different attributes (e.g., bounding box, class category, shape code). The generative ability of their method relies on a variational auto-encoder network, where they learn objects' relative attributes. Besides, a Bayesian optimization stage is used as a post-processing step to refine object arrangements based on the learned relative attribute priors.

\paragraph{ATISS}
ATISS~\cite{paschalidou2021atiss} considers a scene as an unordered set of objects and then designs a novel autoregressive transformer architecture to model the scene synthesis process. During training, based on the previously known object attributes, ATISS utilizes a permutation-invariant transformer to aggregate their features and  predicts the location, size, orientation, and class category of the next possible object conditioned on the fused feature. 
The original version of ATISS~\cite{paschalidou2021atiss} is conditioned on a 2D room mask from the top-down orthographic projection of the 3D floor plane of a scene. To ensure fair comparisons, we train an unconditional ATISS without using a 2D room mask as input, following the same training strategies and hyperparameters as the original ATISS.

\section{Ablation Studies}
\label{SecAbla}
In main paper, we investigated the effectiveness of each design in our DiffuScene, including network architecture, loss function, and geometry feature diffusion. We present more implementation details of each method variant.

\noindent \textbf{What is the effect of UNet-1D+Attention as the denoiser? } 
We advocate the use of UNet-1D with attention layers as the denoising network. 
The self-attention layers within this architecture effectively aggregate all object features and explore inter-object relationships, facilitating the learning of a global context that aids in distinguishing different objects within the scene.
An alternative choice is to use a pure transformer network, like the one adopted in DALLE-2~\cite{ramesh2022hierarchical}. However, our comparisons revealed a marginal degradation in performance metrics such as FID, KID, SCA, and CKL. It demonstrates that UNet-1D with attention layers is more adept at capturing accurate scene distributions than networks solely composed of transformation layers.

\noindent \textbf{What is the effect of multiple prediction heads in the denoiser?} 
In our denoiser architecture, we employ three distinct encoding and prediction heads tailored for specific object properties, including bounding box parameters, semantic class labels, and geometry codes. By utilizing multiple diffusion heads with individual loss functions for each attribute (e.g., bbox, class, geometry), we mitigate the risk of bias towards any single attribute within a single encoding and prediction head. This approach ensures that our denoiser effectively captures and processes diverse object properties without favoring one over the others. The consistent improvement in each evaluation metric verifies the effectiveness of multiple prediction heads.

\noindent \textbf{What is the effect of the IoU loss? }
In scene diffusion models, we employ noise prediction loss as the primary supervision, focusing on attribute denoising of individual object instances. However, this loss does not address object intersections within a scene. To alleviate the issue, we augment it with pair-wise bounding box IoU loss. Quantitative comparisons indicate that incorporating IoU loss results in the synthesis of scenes with improved symmetry and enhanced plausibility, as evidenced by lower FID, KID, SCA, PIoU and higher Sym.

\noindent \textbf{What is the effect of geometry feature diffusion?}
To evaluate our method's performance without geometry feature diffusion, we eliminate the geometry feature encoding and prediction heads from our denoiser network. Consequently, this method only produces bounding boxes and class labels for objects within a scene.
During inference, 
 for each generated object, we conduct shape retrieval in the 3D-FUTURE~\cite{fu20213dm} dataset to find the CAD model with the same class label and the closest 3D bounding box sizes.
Fig. 5 of the main paper shows that our model can find symmetric nightstands by beds due to the geometry awareness of the diffusion process and shape retrieval. Table 3 in the main paper presents the comparison in the formation of symmetric pairs: 0.72 (w/ shape diffusion) vs. 0.50 (w/o shape diffusion).
This highlights the effectiveness of geometry feature diffusion in achieving symmetric placements and semantically coherent arrangements. Improved plausibility in synthesis results is reflected in lower FID, KID, and SCA evaluations. Additionally, the decrease in CKL suggests that the joint diffusion of geometry code and object layout facilitates learning more similar object class distributions.





\section{Additional Results}
\label{SecAddRes}

\paragraph{Diversity Analysis.} 
The qualitative comparisons in Fig. 7 of the main paper and Fig.~\ref{fig:completion_supple} 
illustrate that our diffusion-based method can produce more diverse results than the baseline methods.
Following ATISS and LEGO, we use FID and KID to quantitatively evaluate the result diversity.
We compare both the mean and covariance of generated and reference scene distribution.
Additionally, we include Precision / Recall commonly used to evaluate generative models~\cite{kynkaanniemi2019improved}. 
Precision is the probability that a randomly generated scene falls within the support of real scene distribution.
Recall is the probability that a random scene from the datasets falls within the generated scene distribution.
Tab.~\ref{tab:pre_rec} shows that our approach outperfoms all baselines in both metrics, which demonstrates better diversity, plausiblity, and mode coverage. 
\begin{table}[!htbp] 
	\renewcommand\arraystretch{1.2}
        \setlength{\tabcolsep}{2.4pt}
	\begin{center}
        \resizebox{0.98\linewidth}{!}{
		\begin{tabular}{*{7}{c}}
                \toprule
			\multirow{2}*{Method} & \multicolumn{2}{c}{Bedroom}  & \multicolumn{2}{c}{Dining} & \multicolumn{2}{c}{Living} \\ \cmidrule(lr){2-3}  \cmidrule(lr){4-5} \cmidrule(lr){6-7} 
              & Precision  & Recall   &  Precision  & Recall     & Precision  & Recall   \\
                \midrule
                DepthGAN
                       & 58.05 & 31.66
                       & 70.16 & 15.77
                       & 81.30  & 12.08  \\ 
                       
                Sync2Gen
                       & 59.00 & 67.74
                       & 76.15 & 33.19
                       & 77.77 & 48.79 \\ 
                
                Sync2Gen*
                       & 55.10 & 67.57
                       & 70.90 & 47.16
                       & 75.20 & 52.01 \\ 
                
                ATISS
                        & 72.80 & 77.08 
                        & 77.70 & 64.17  
                        & 76.50 & 62.64  \\ 
                Ours
                       & \bf{82.31} & \bf{77.93} 
                       & \bf{82.80} & \bf{78.83} 
                       & \bf{79.30} & \bf{70.53} \\ 
                  \bottomrule
        \end{tabular}
        }
        \caption{
        The Precision [\%] of generated scenes and Recall [\%] of reference scenes. For both metrics, the higher the better.}  
        \label{tab:pre_rec}
        \end{center}
\end{table}

\begin{figure}[!htp]
    \centering
\includegraphics[width=.86\linewidth]{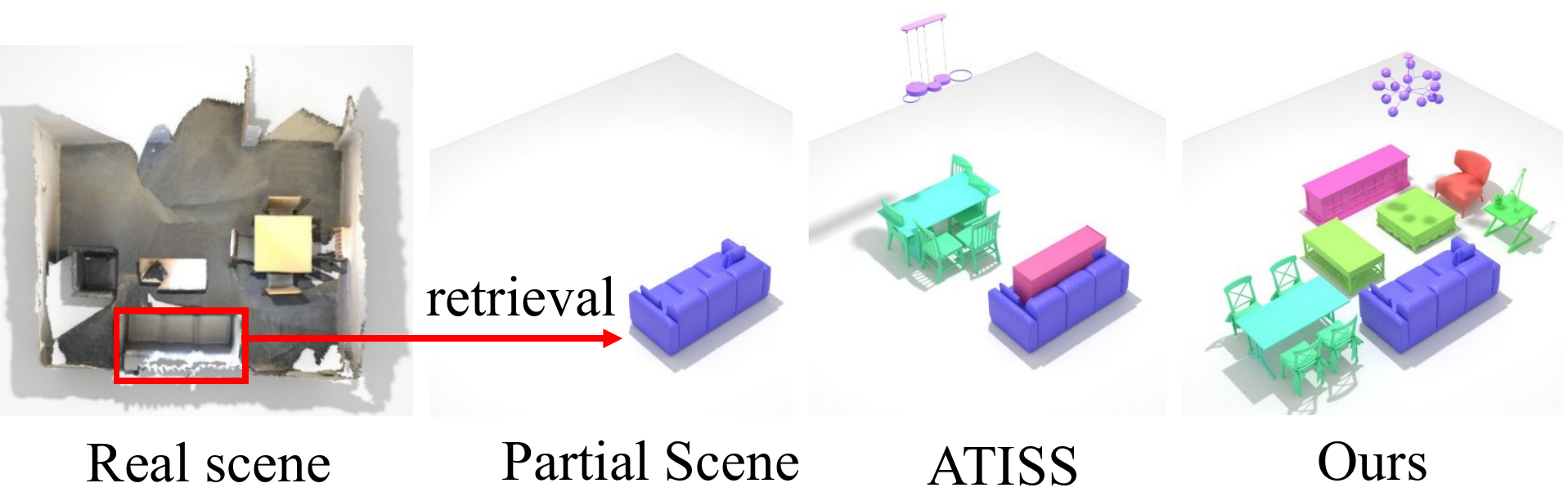}
    \caption{
    Scene completion of a real scene. We select an sofa and perform CAD retrieval to obtain a partial scene as input.
    }
    \label{fig:real_world_completion}
\end{figure}
\begin{figure}[!htp]
    \centering
\includegraphics[width=.9\linewidth]{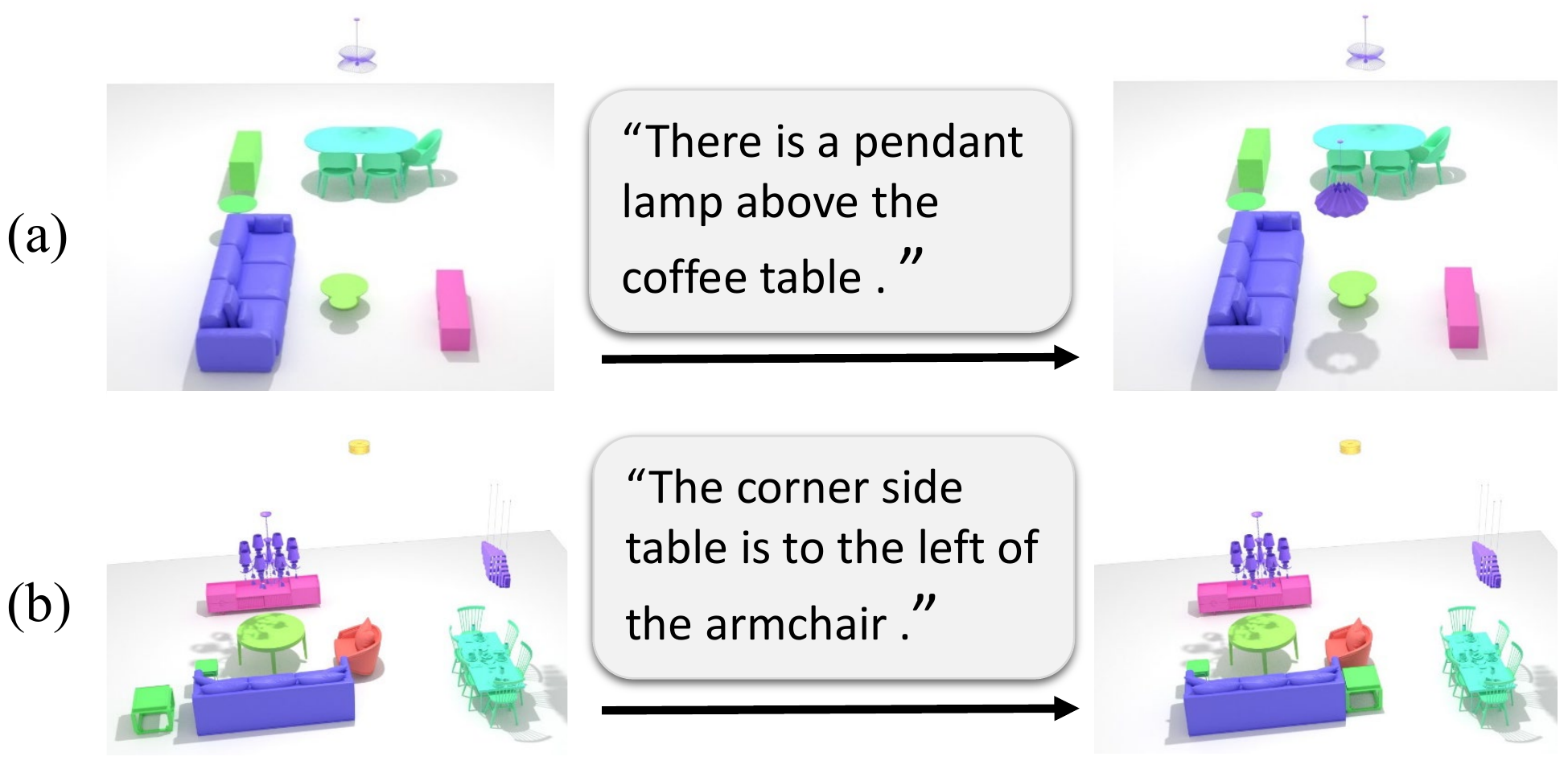}
    \caption{
    Text-guided (a) object suggestion (b) scene editing.
    }
    \label{fig:text_editing}
\end{figure}

\paragraph{Unconditional Scene Synthesis}
\begin{figure*}[!ht]
	\centering
 	\begin{subfigure}[t]{0.23\textwidth}
            \includegraphics[width=\textwidth]{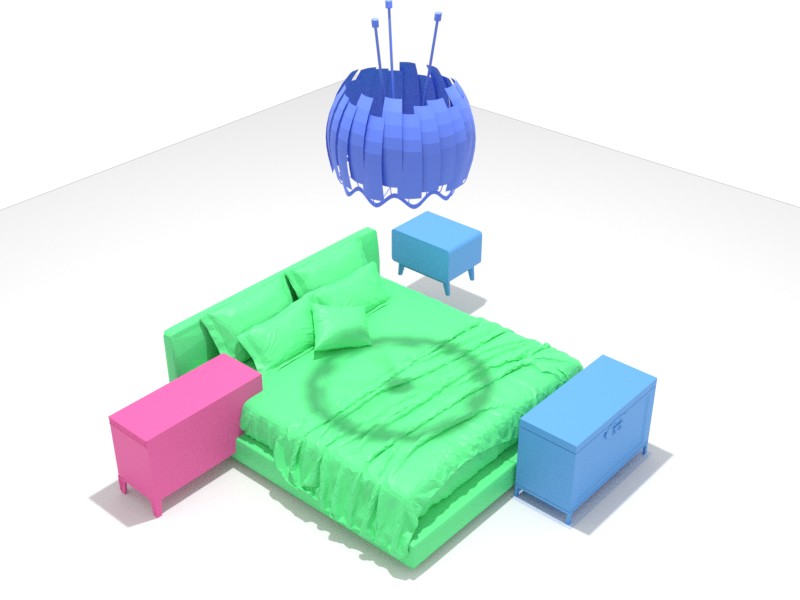}
            \includegraphics[width=\textwidth]{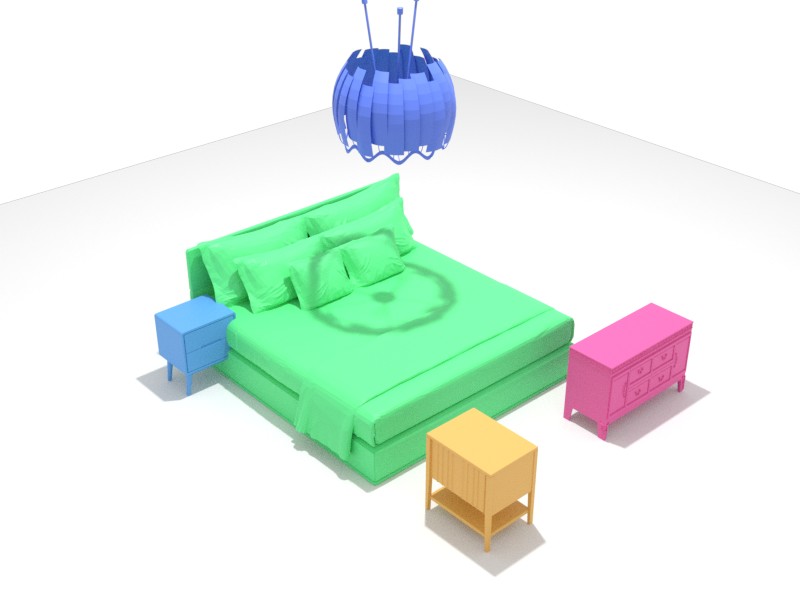}
            \includegraphics[width=\textwidth]{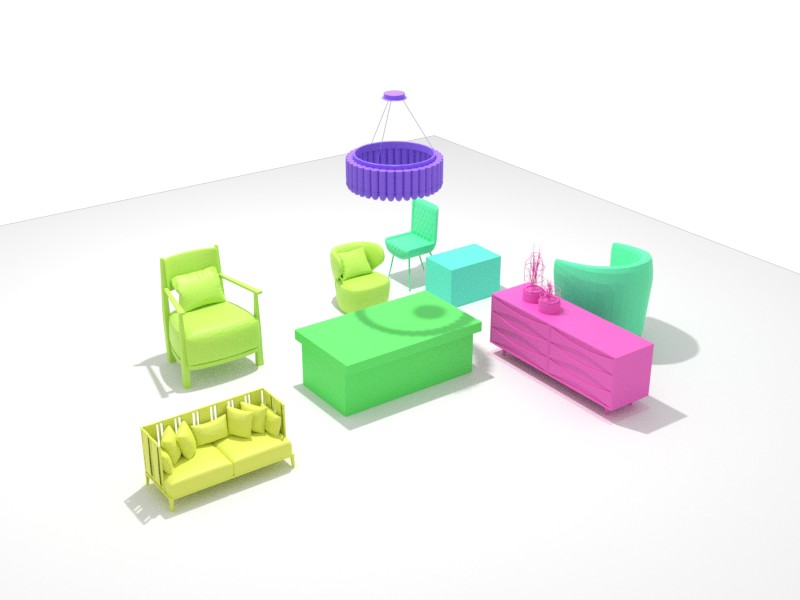}
            \includegraphics[width=\textwidth]{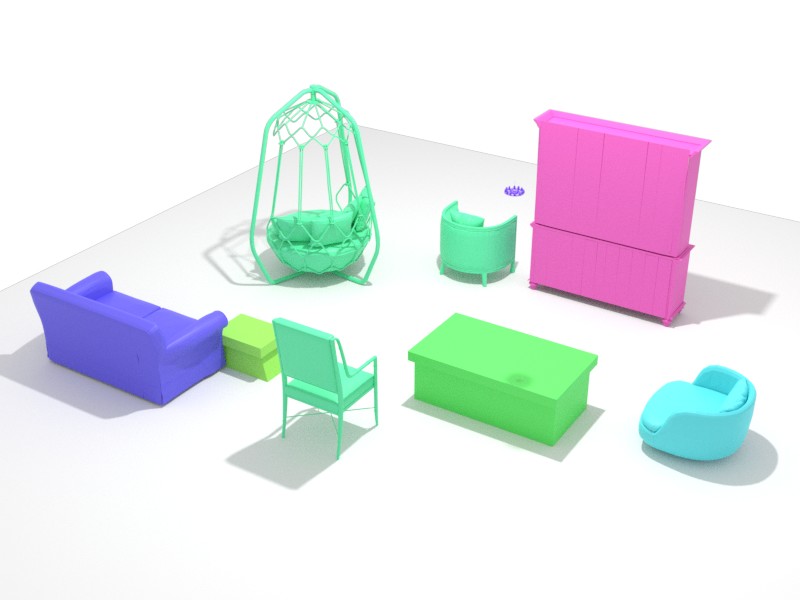}
		\includegraphics[width=\textwidth]{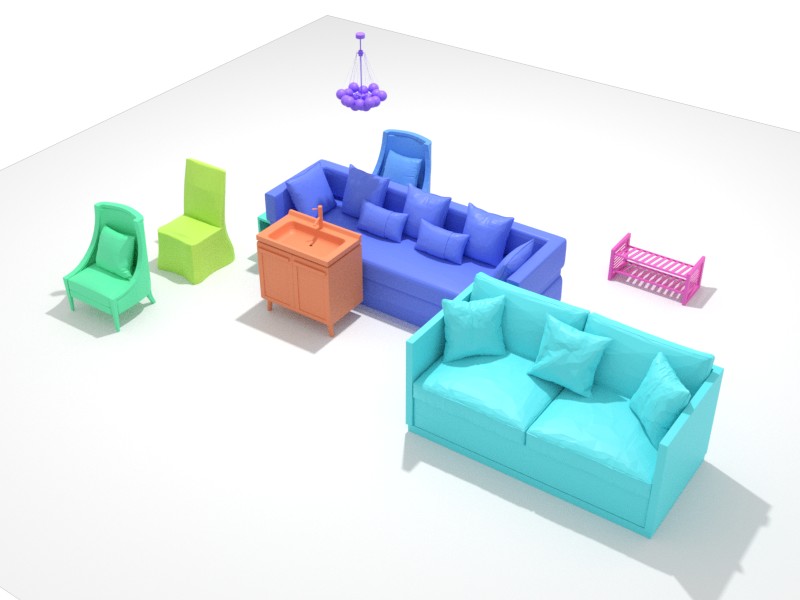}
            \includegraphics[width=\textwidth]{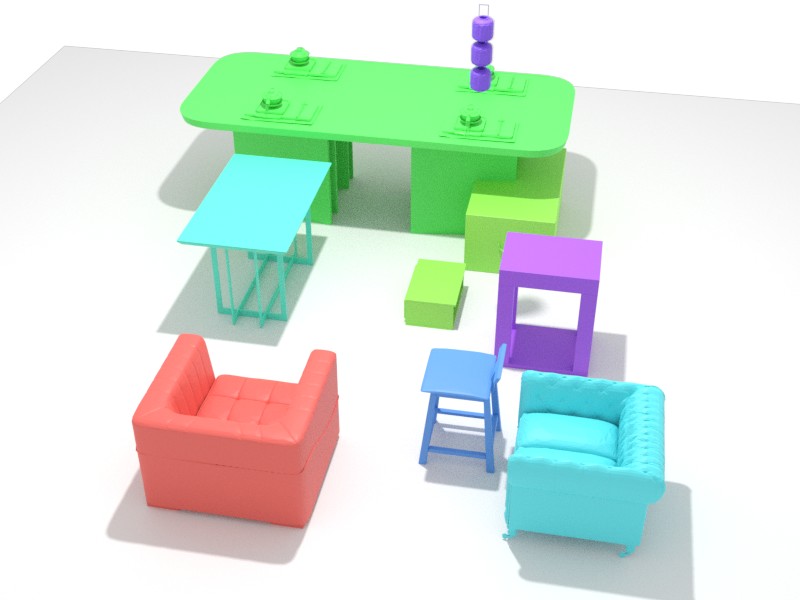}
            \includegraphics[width=\textwidth]{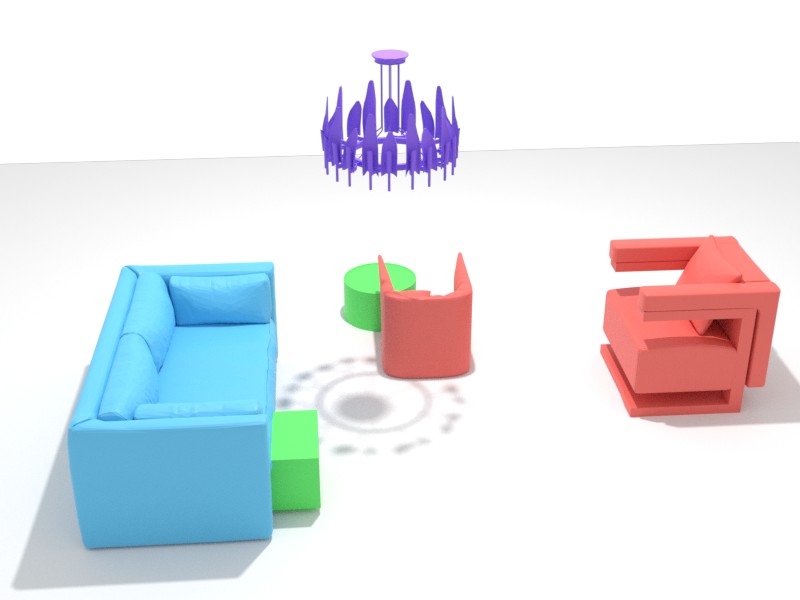}
		\caption{DepthGAN~\cite{yang2021indoor}}
	\end{subfigure}
	\rulesep
	\begin{subfigure}[t]{0.23\textwidth}
		\includegraphics[width=\textwidth] {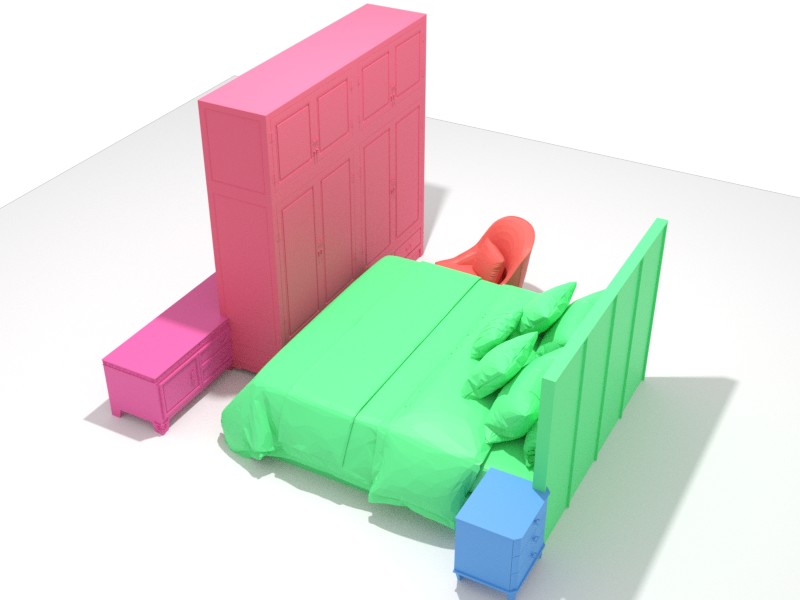}
            \includegraphics[width=\textwidth]{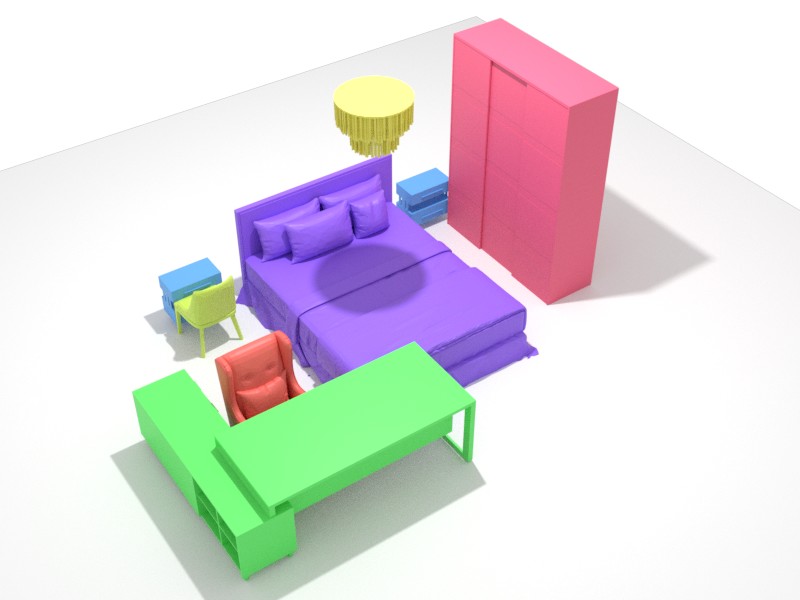}
            \includegraphics[width=\textwidth]{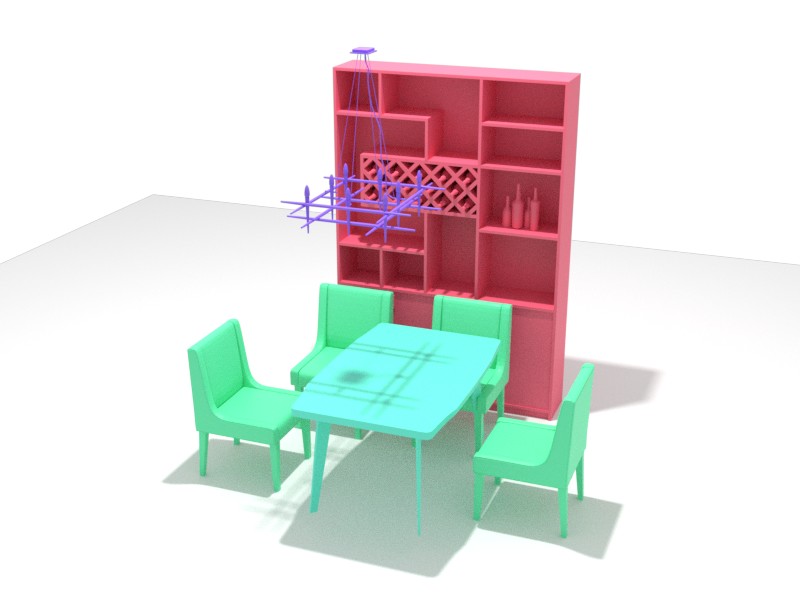}
		\includegraphics[width=\textwidth]{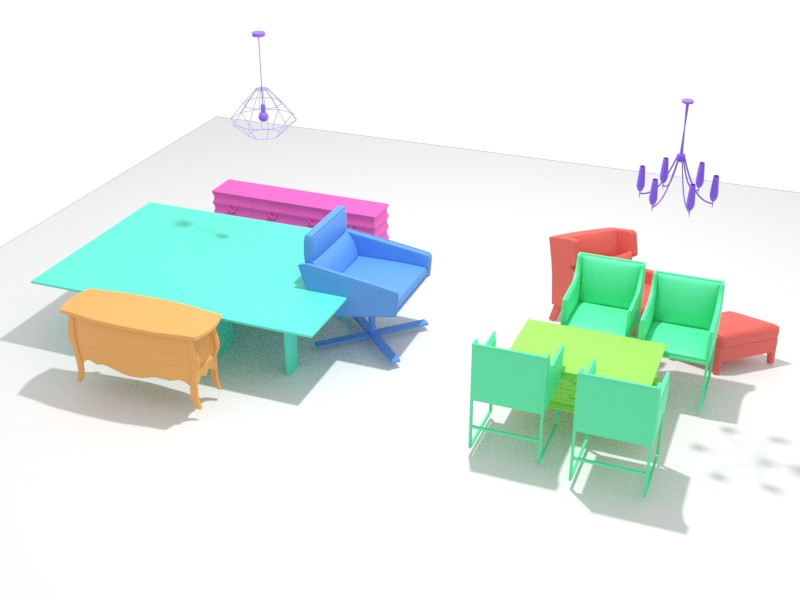}
		\includegraphics[width=\textwidth]{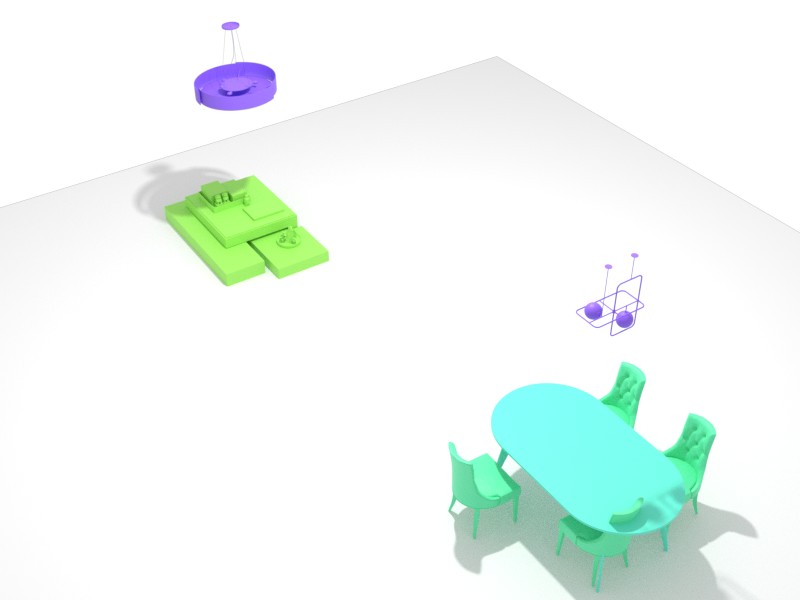}
            \includegraphics[width=\textwidth]{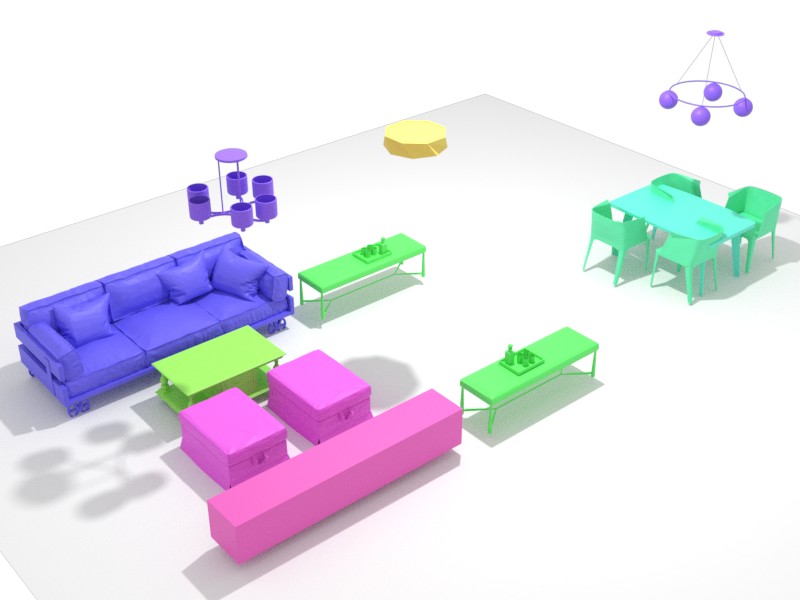}
		\includegraphics[width=\textwidth]{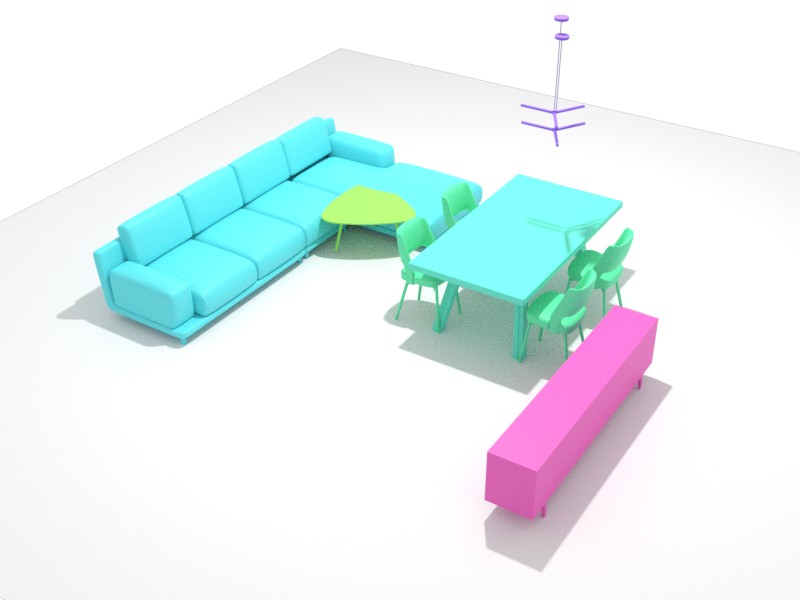}
		\caption{Sync2Gen~\cite{yang2021scene}}
	\end{subfigure}
	\rulesep
        \begin{subfigure}[t]{0.23\textwidth}
            \includegraphics[width=\textwidth]{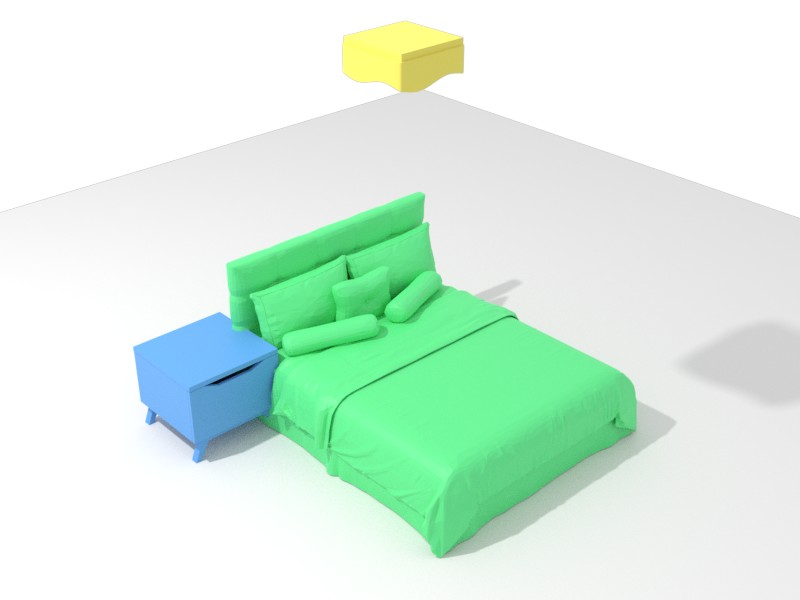}
		\includegraphics[width=\textwidth]{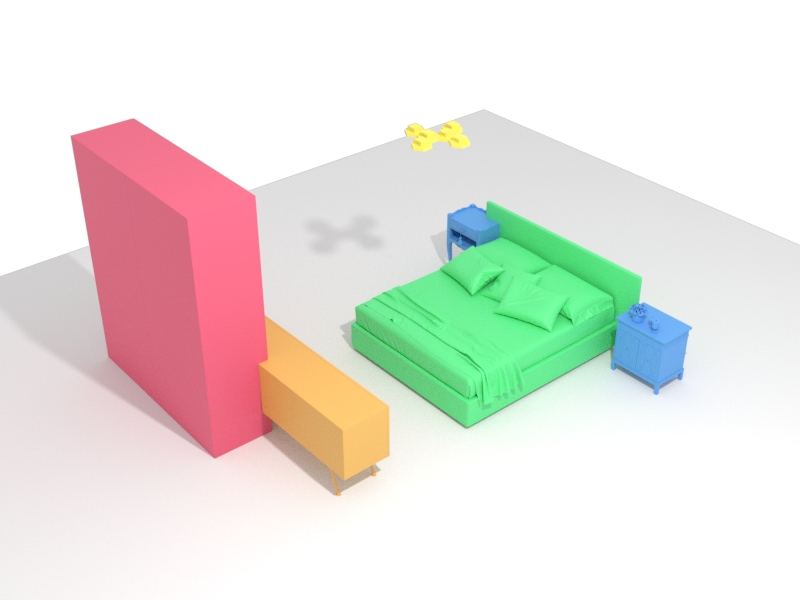}
            \includegraphics[width=\textwidth]{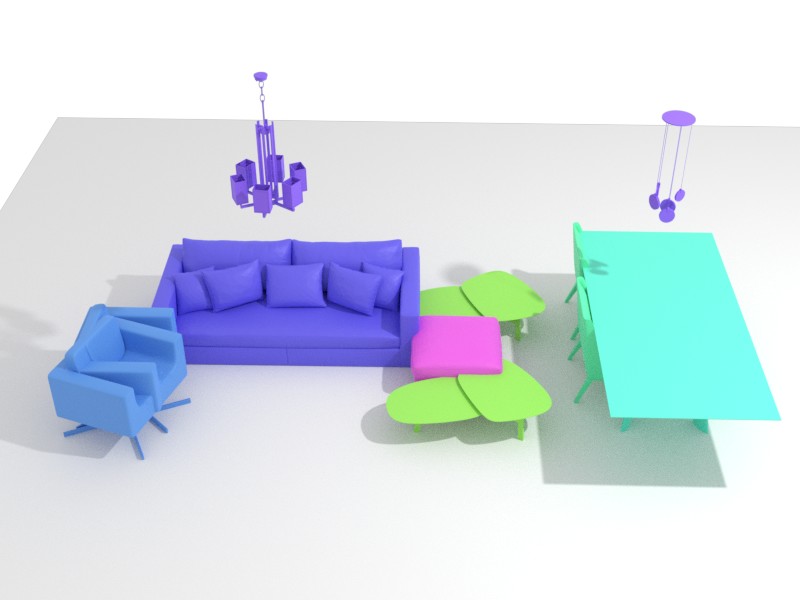}
            \includegraphics[width=\textwidth]{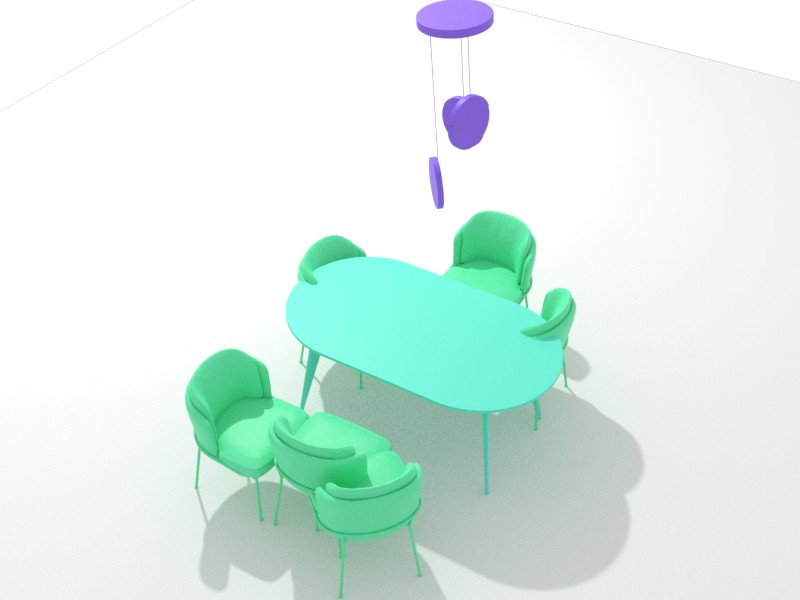}
            \includegraphics[width=\textwidth]{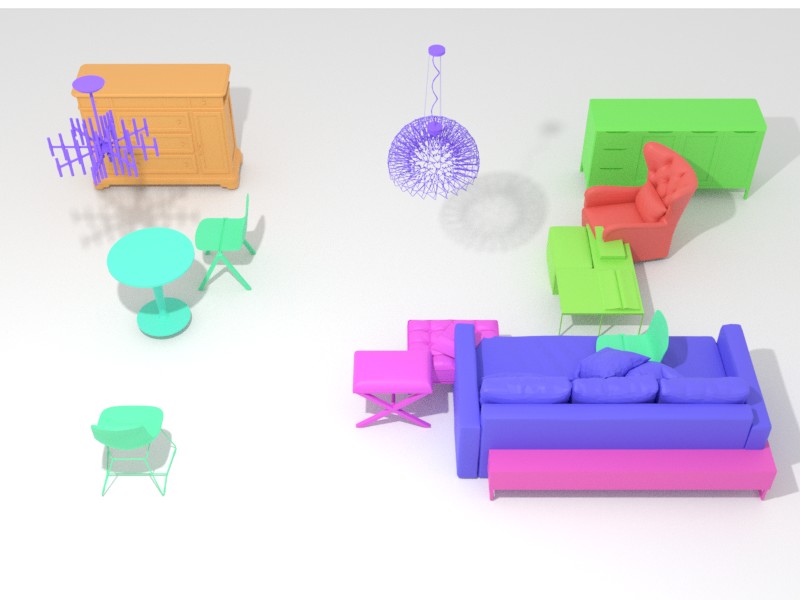}
            \includegraphics[width=\textwidth]{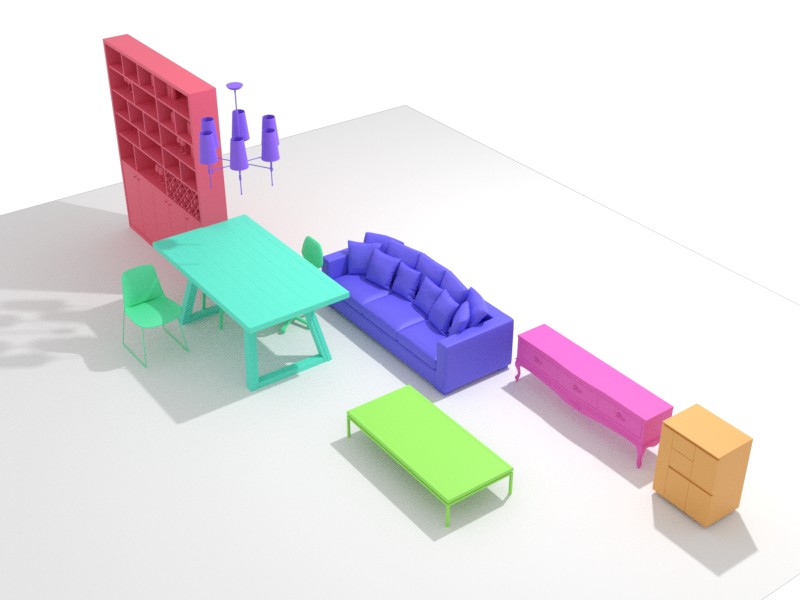}
		\includegraphics[width=\textwidth]{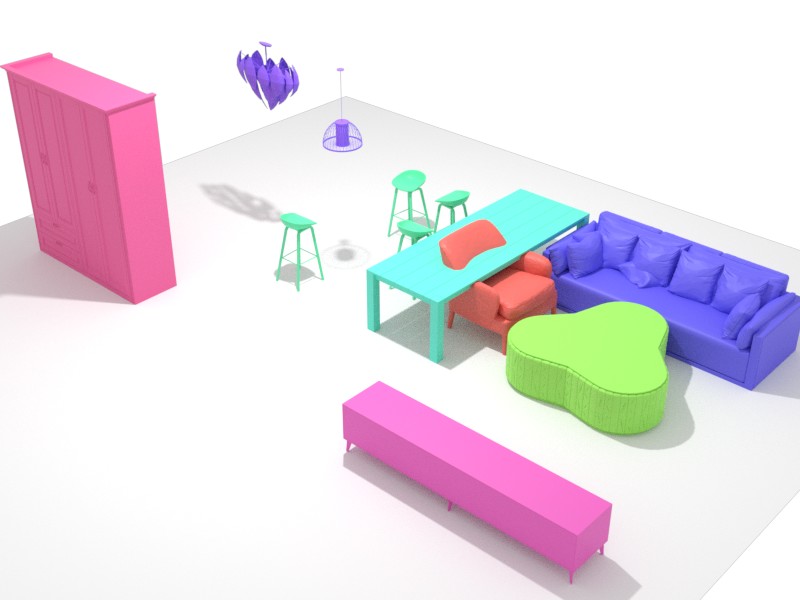}
		\caption{ATISS~\cite{paschalidou2021atiss}}
	\end{subfigure}
	\rulesep
	\begin{subfigure}[t]{0.23\textwidth}
        \includegraphics[width=\textwidth]{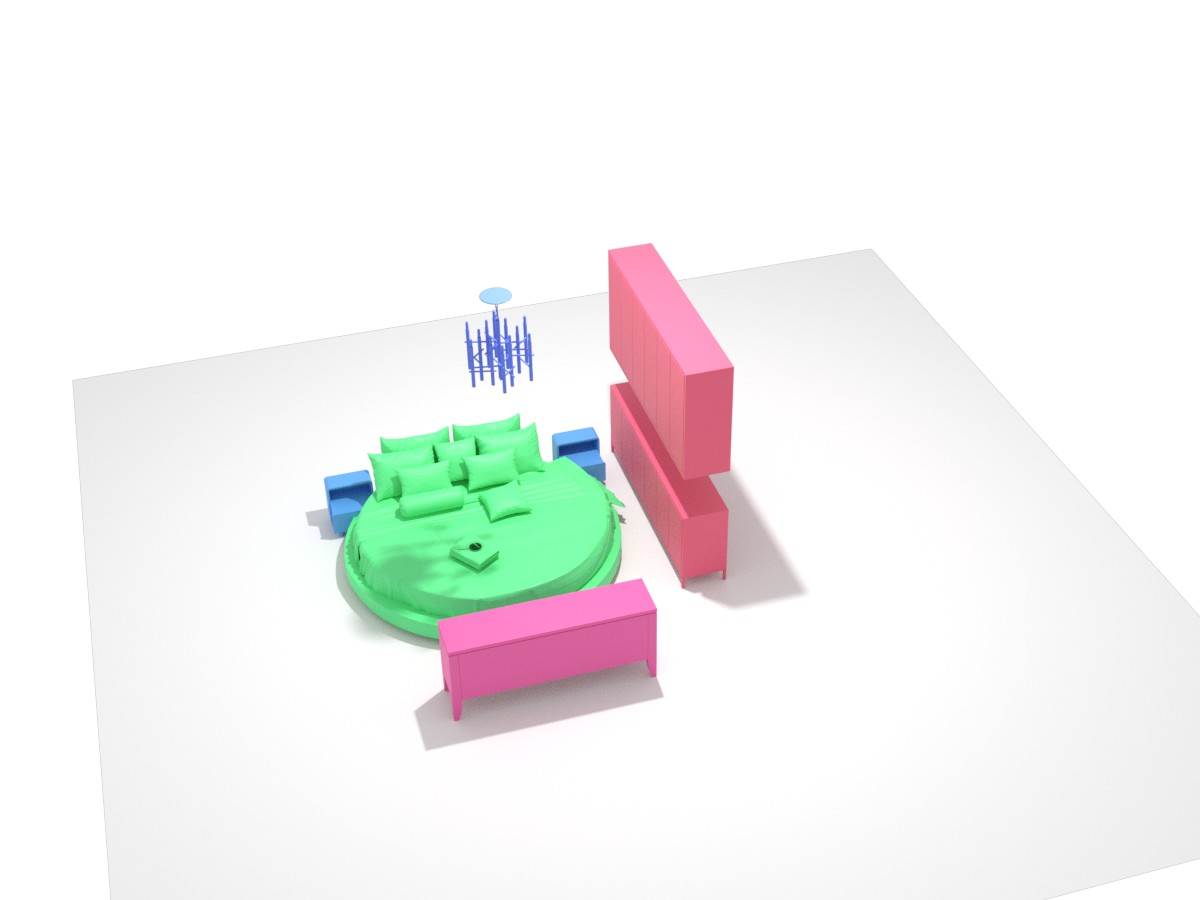}
        \includegraphics[width=\textwidth]{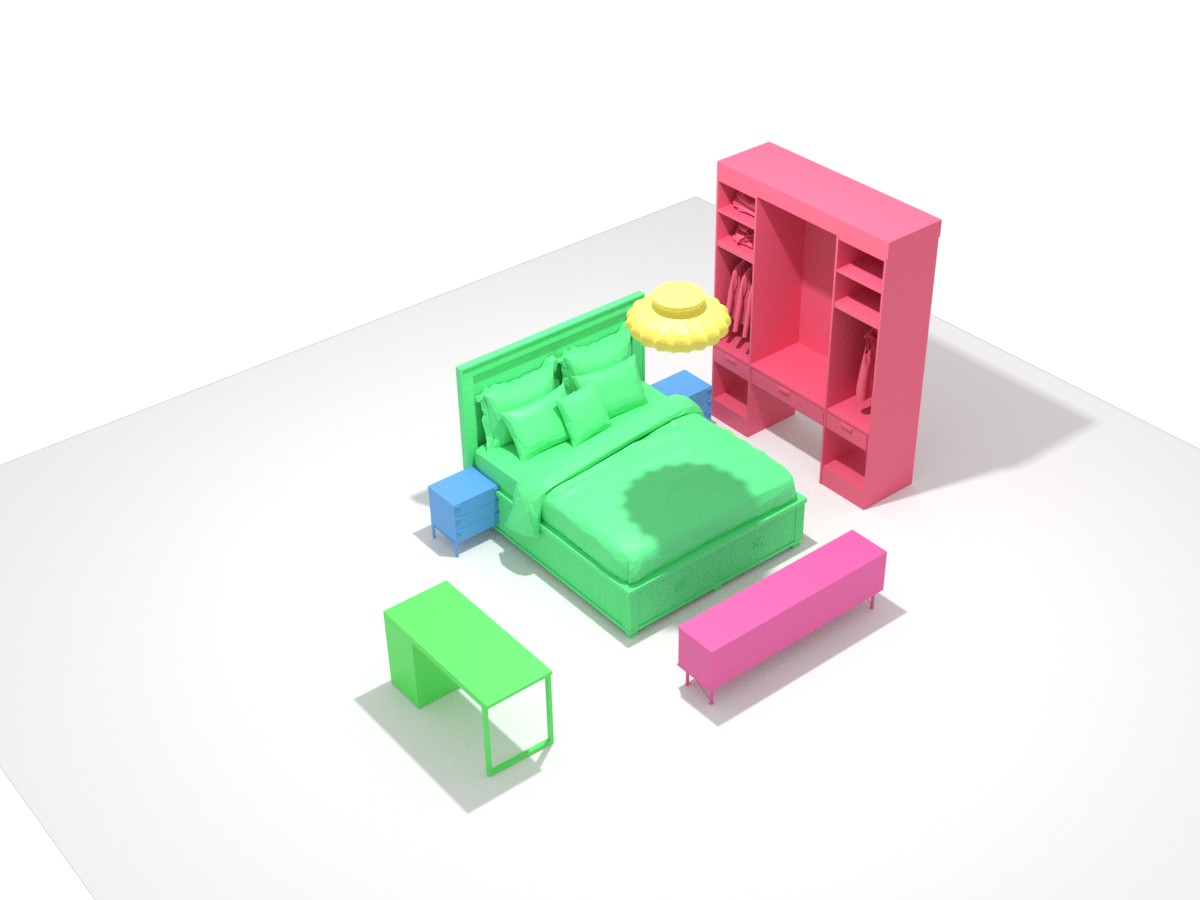}
        \includegraphics[width=\textwidth]{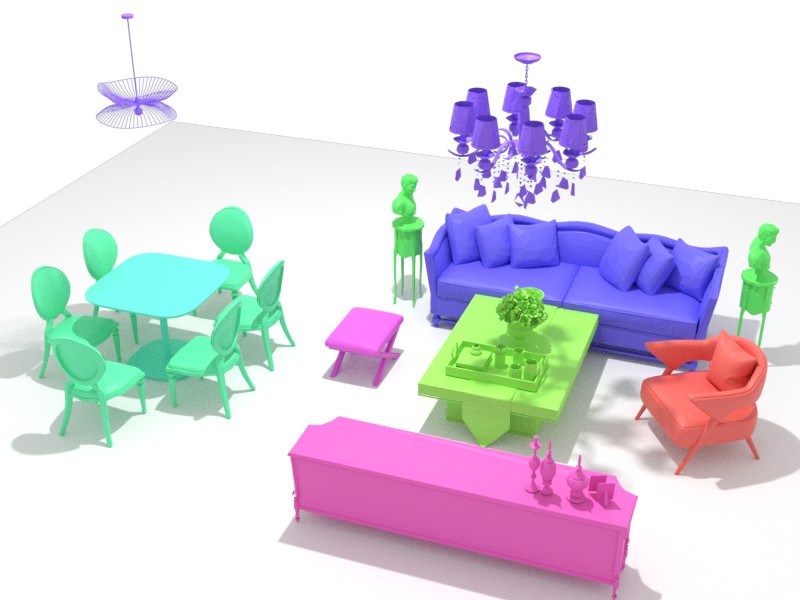}
		\includegraphics[width=\textwidth]{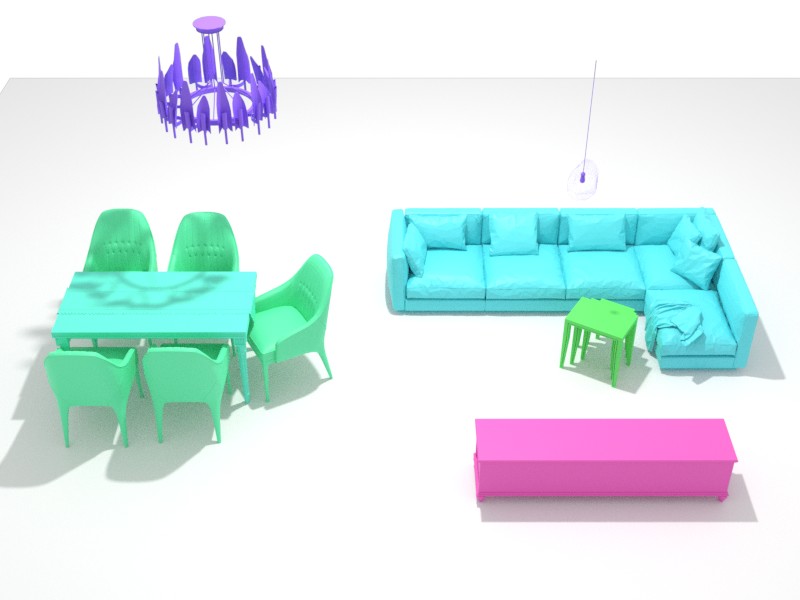}
		\includegraphics[width=\textwidth]{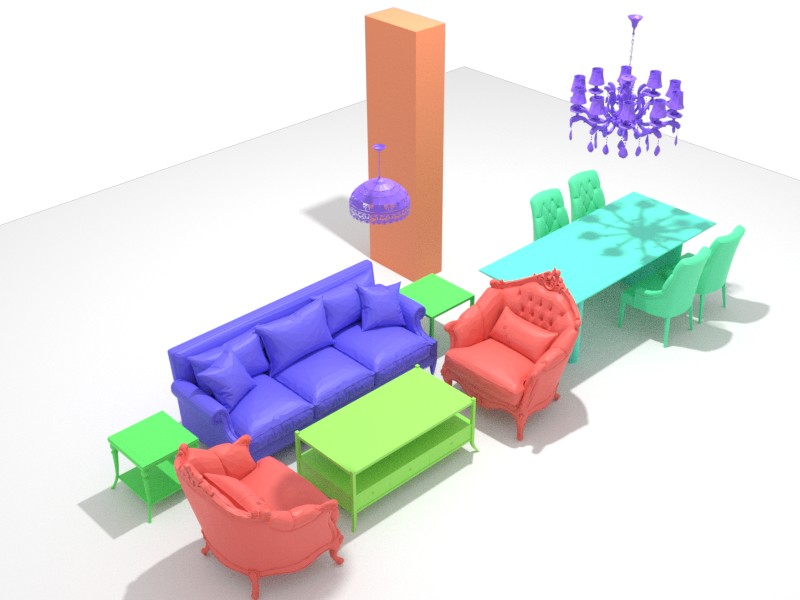}
            \includegraphics[width=\textwidth]{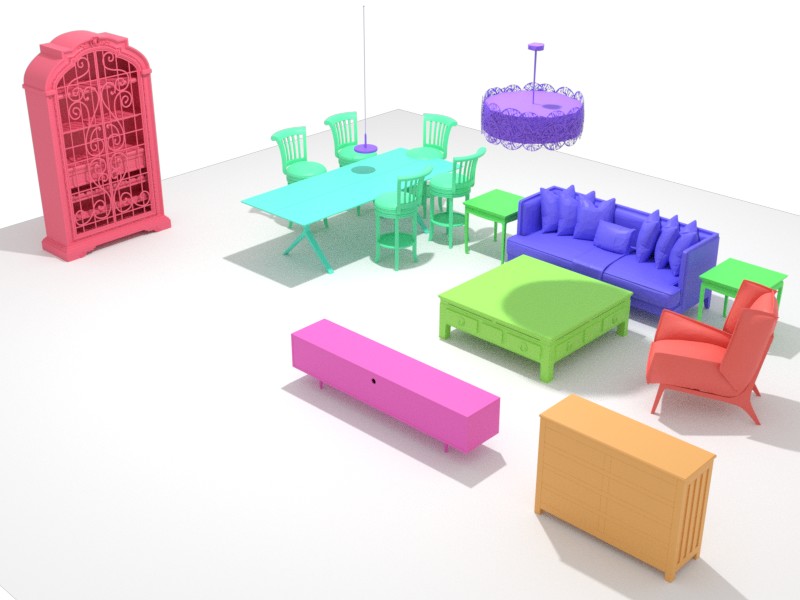}
		\includegraphics[width=\textwidth]{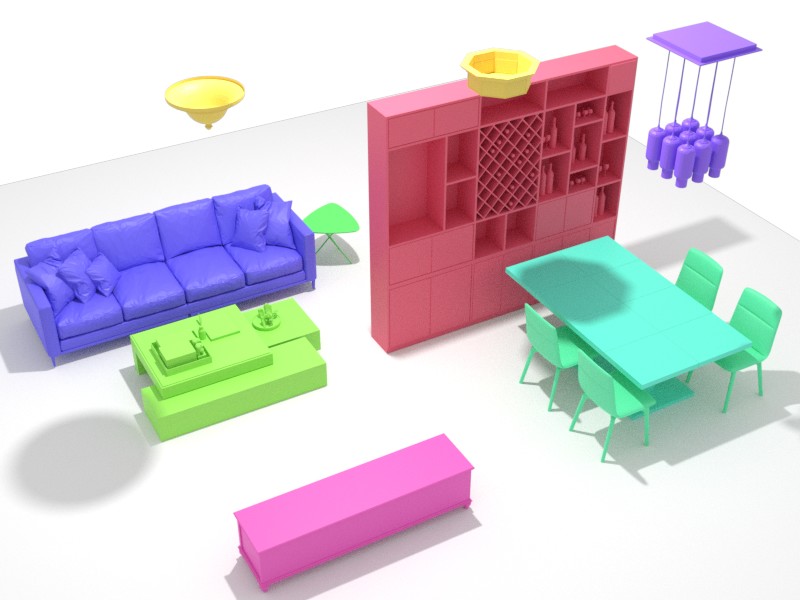}
		\caption{Ours}
	\end{subfigure}
	\caption{\textbf{Additional results of unconditional scene synthesis}. We compare our method with the state-of-the-art by generating from random noises, where our results present higher diversity and better plausibility with fewer penetration issues and more symmetric pairs.}
    \label{fig:uncond_comparison_supple}
\end{figure*}
\begin{figure*}[!htbp]
	\centering
 	\begin{subfigure}[t]{0.23\textwidth}
		\includegraphics[width=\textwidth]{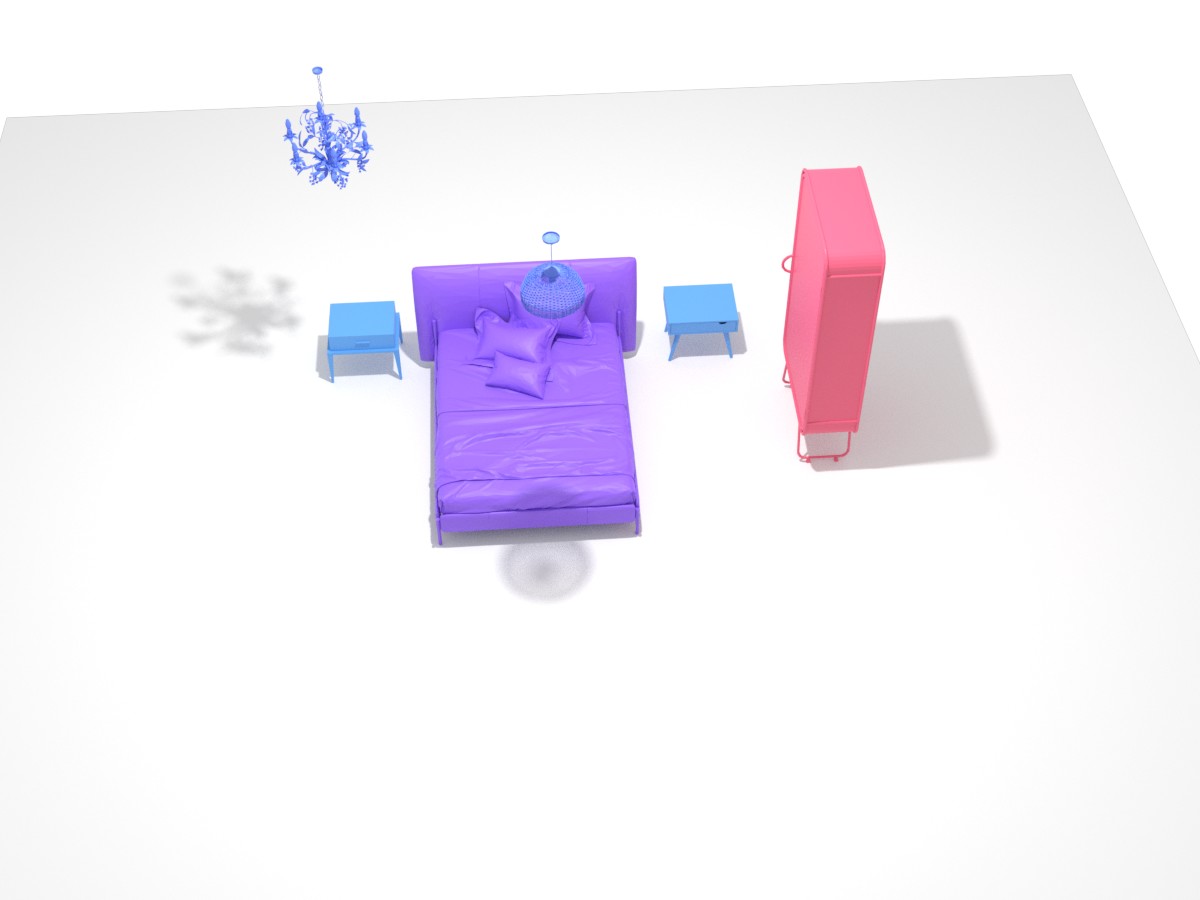}
 
  	\includegraphics[width=\textwidth]{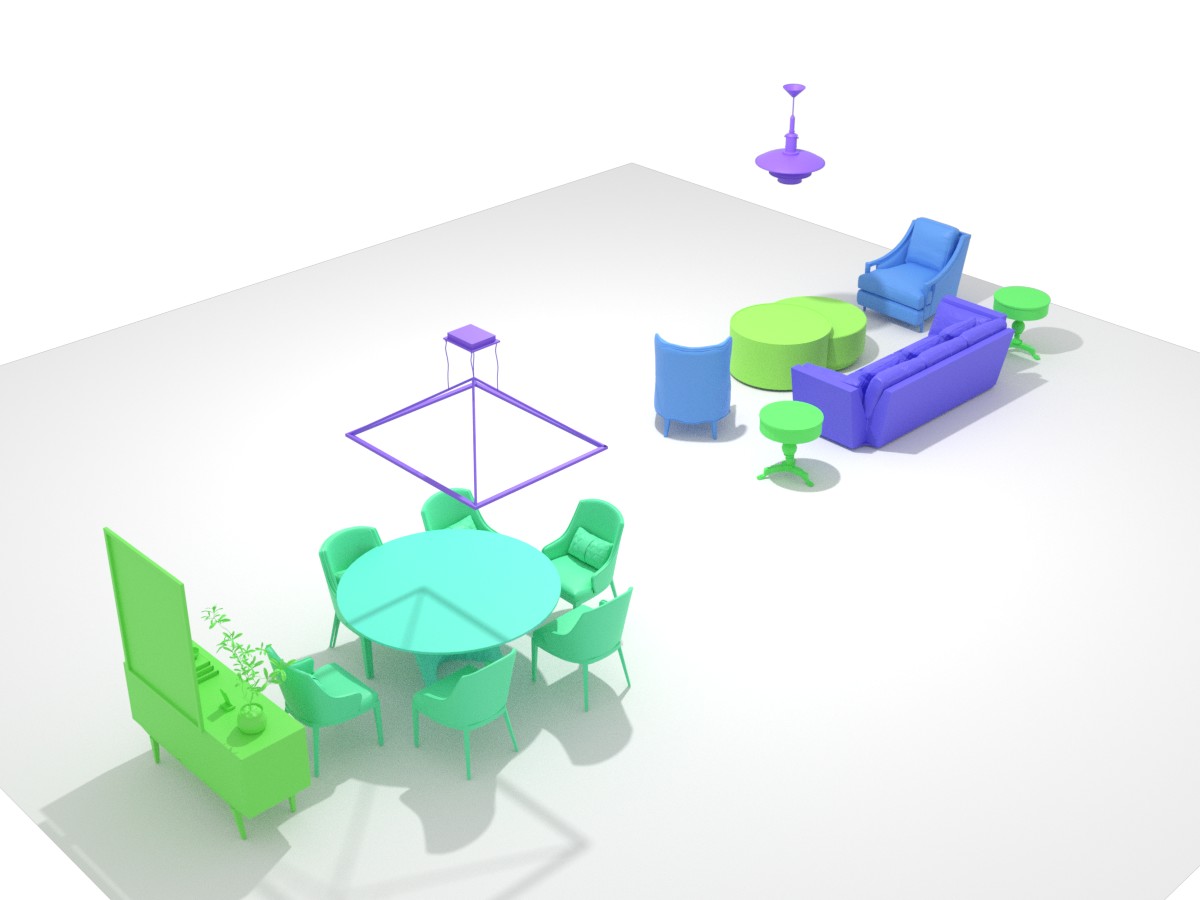}
   
		\includegraphics[width=\textwidth]{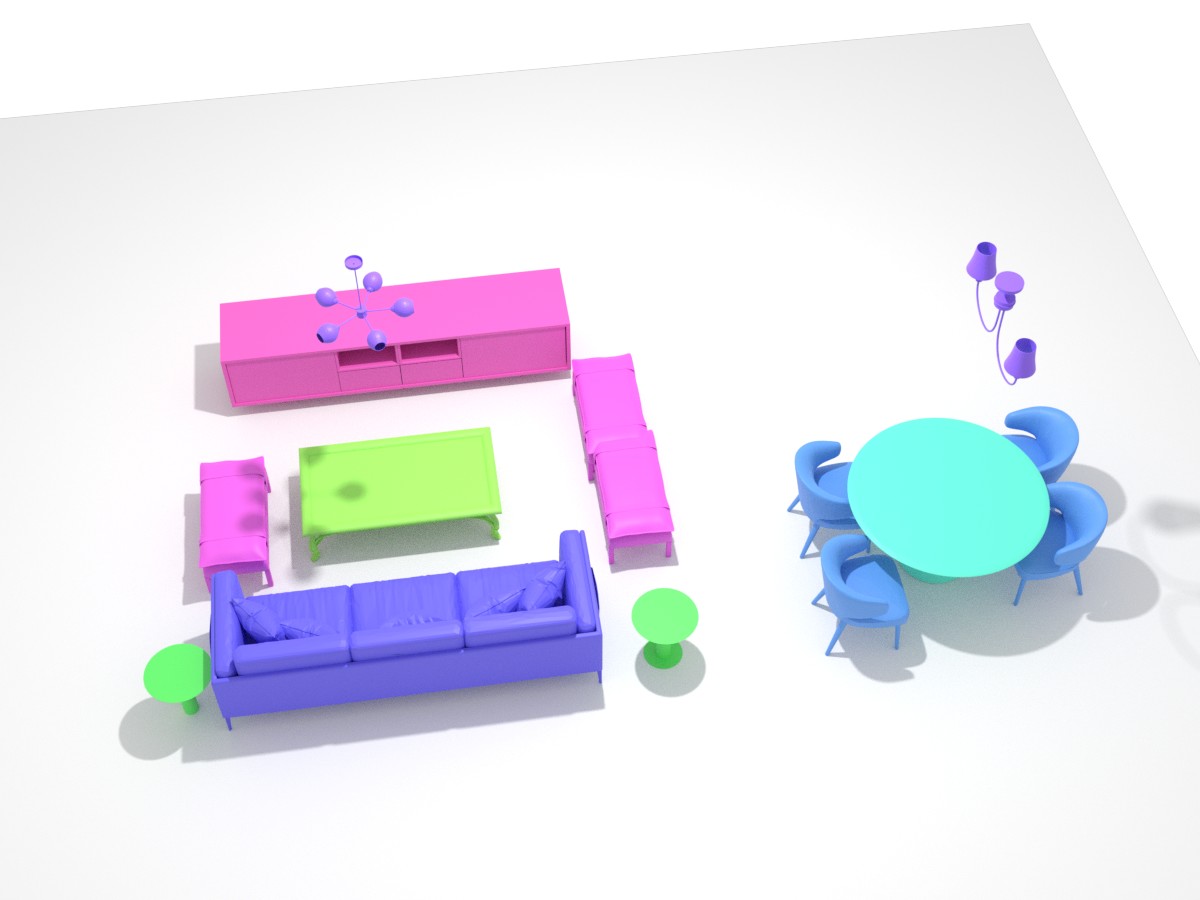}

            \includegraphics[width=\textwidth]{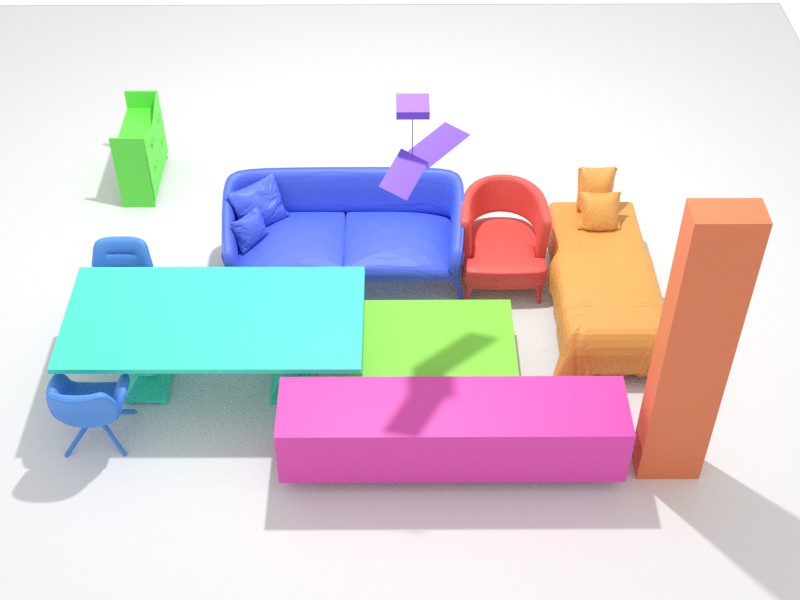}
    
	\end{subfigure}
	\rulesep
	\begin{subfigure}[t]{0.23\textwidth}
		\includegraphics[width=\textwidth]{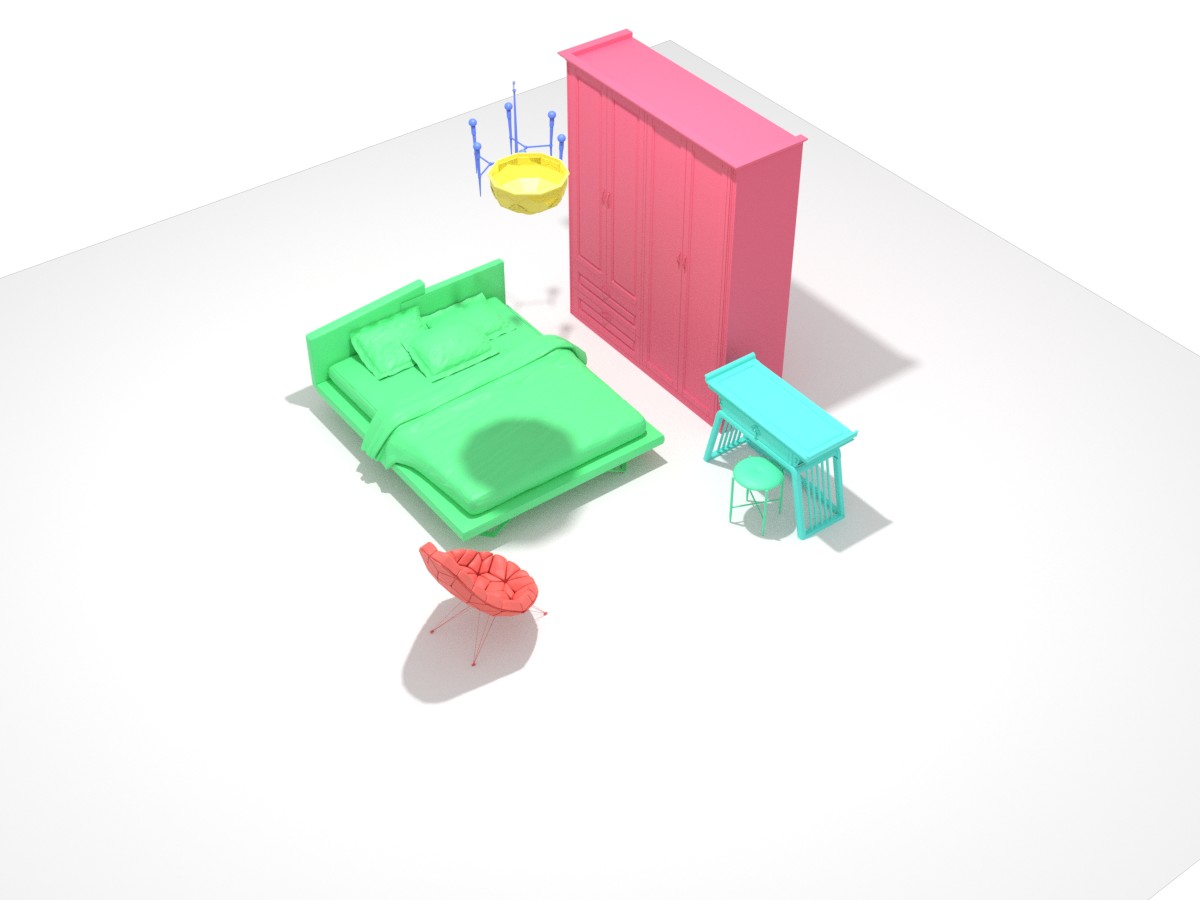}
 
  	\includegraphics[width=\textwidth]{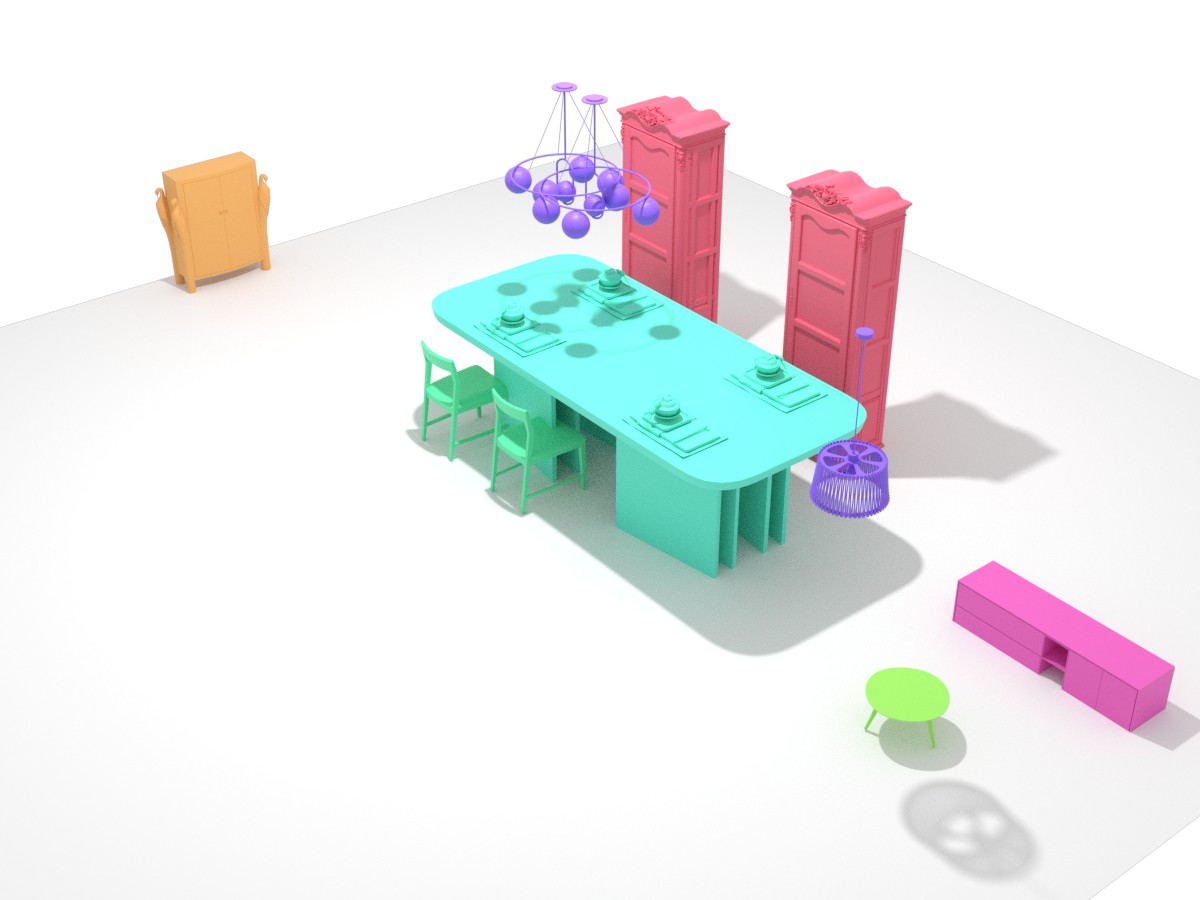}
   
		\includegraphics[width=\textwidth]{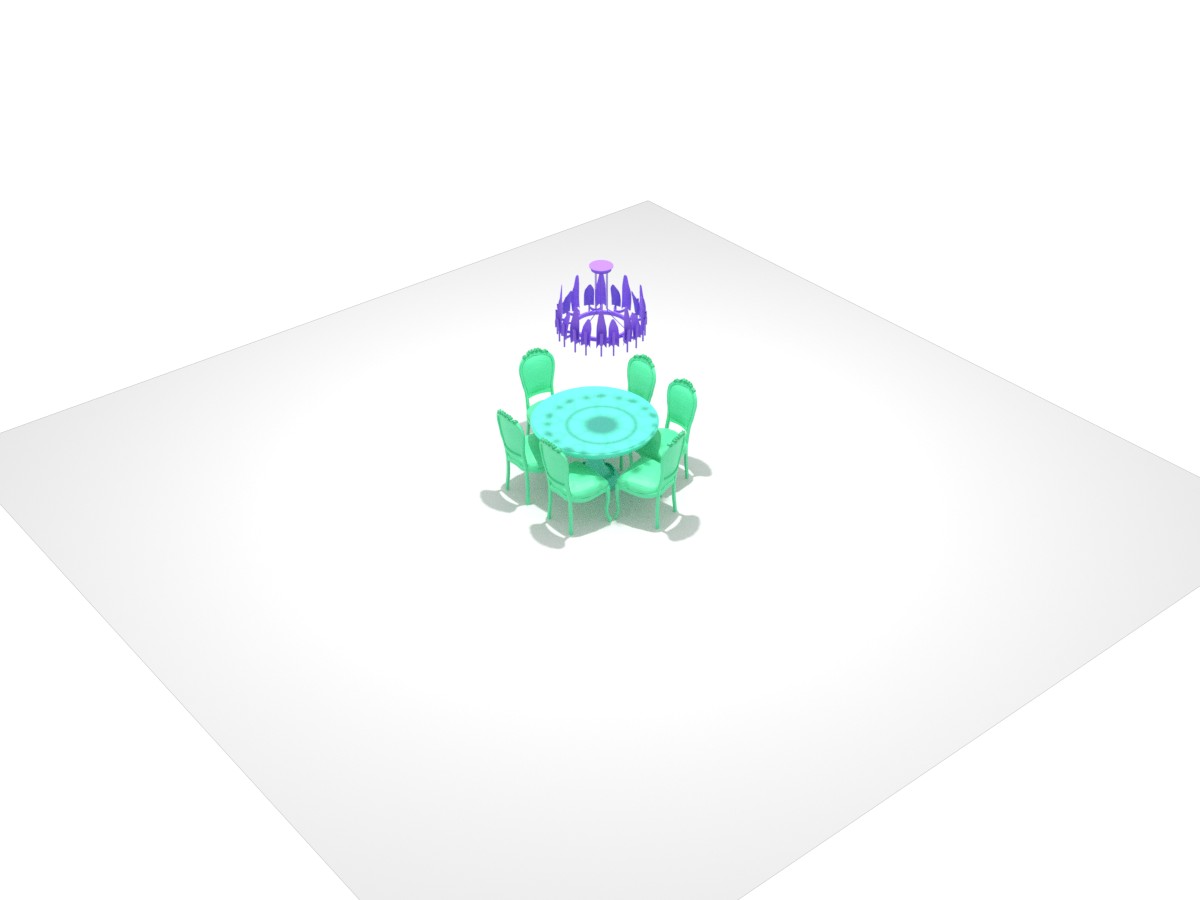}

            \includegraphics[width=\textwidth]{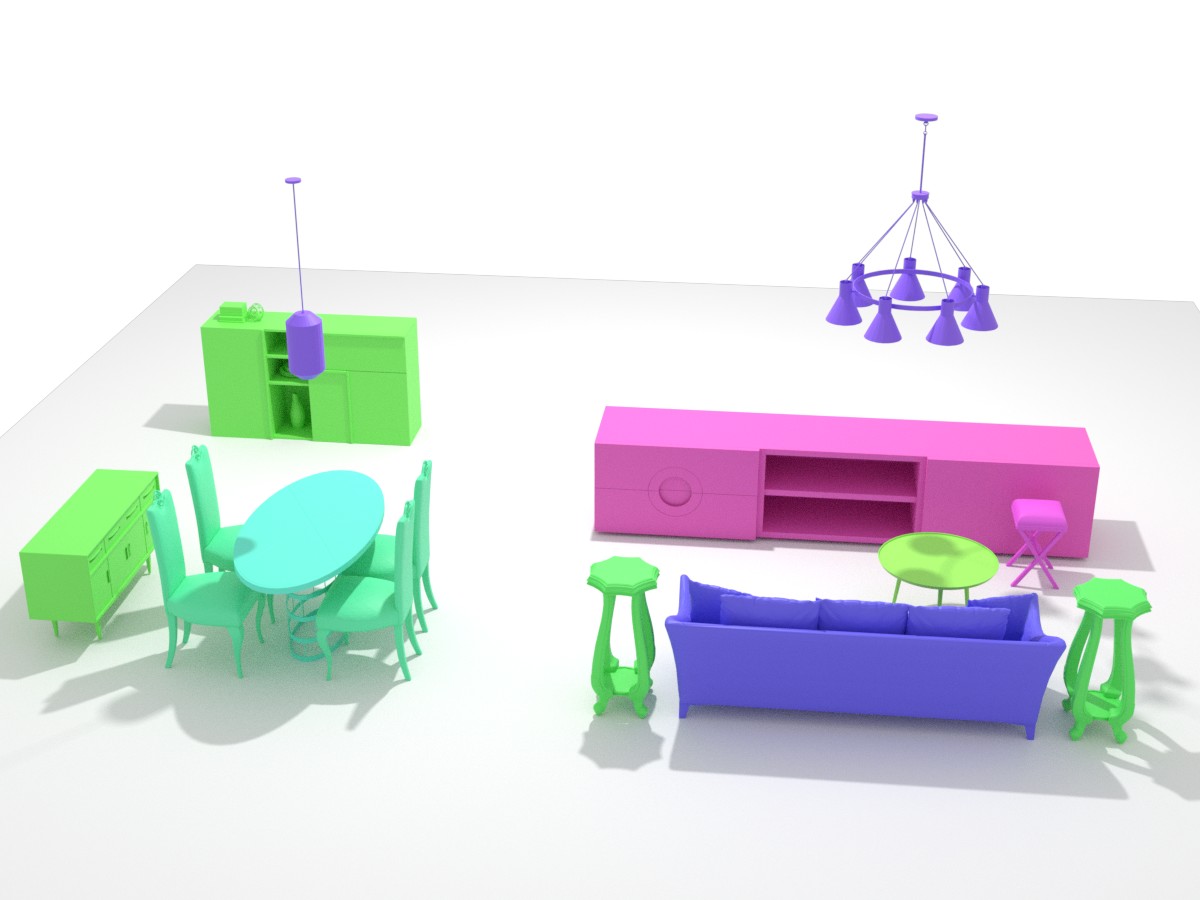}

	\end{subfigure}
	\rulesep
        \begin{subfigure}[t]{0.23\textwidth}
		\includegraphics[width=\textwidth]
	{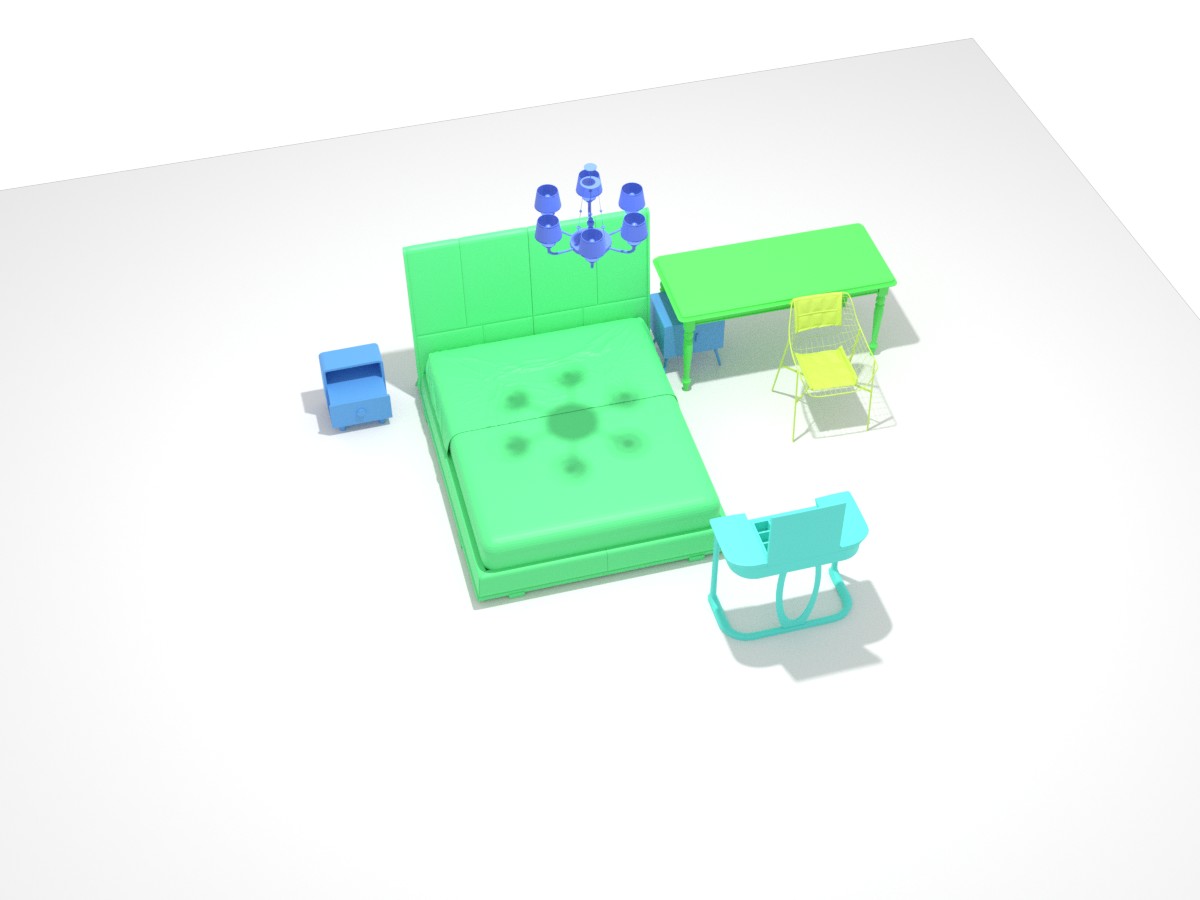}
 
  	\includegraphics[width=\textwidth]{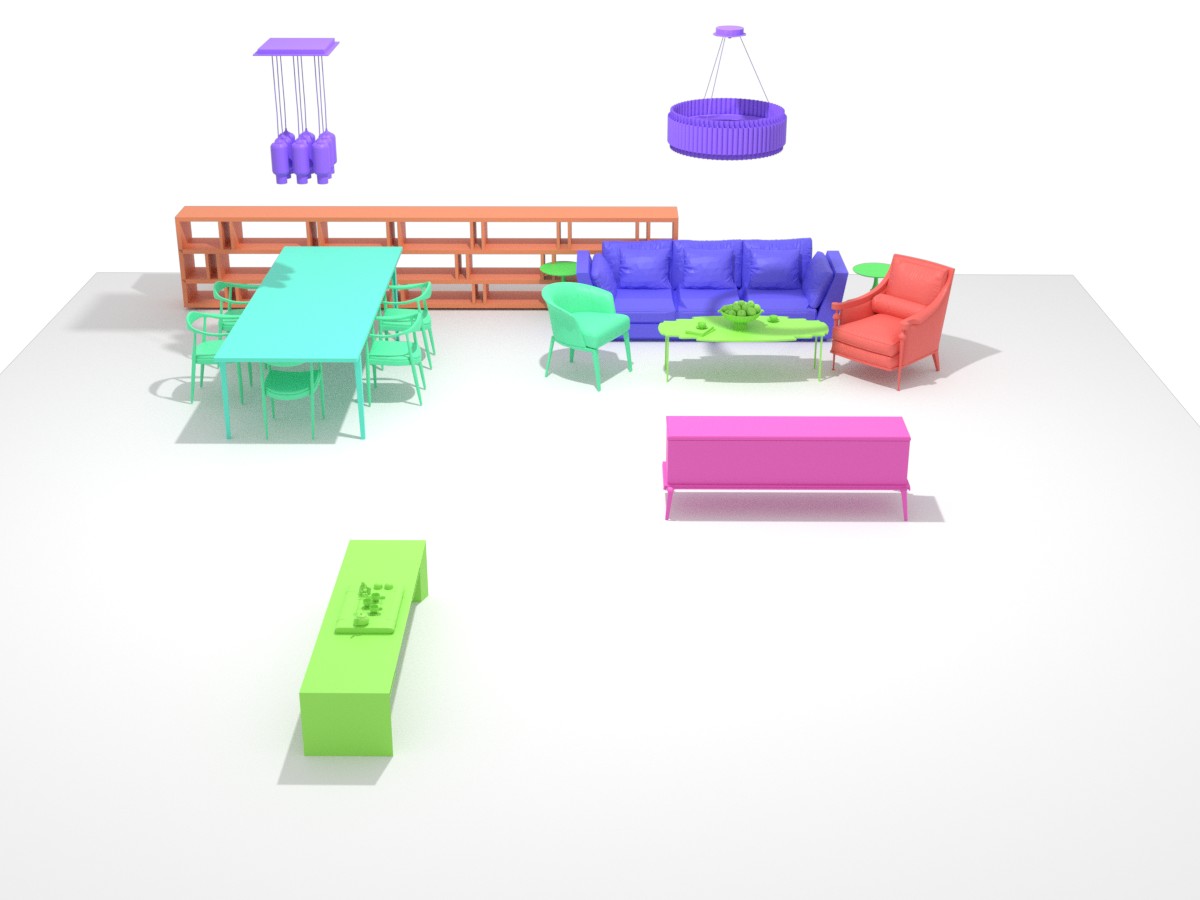}
   
		\includegraphics[width=\textwidth]{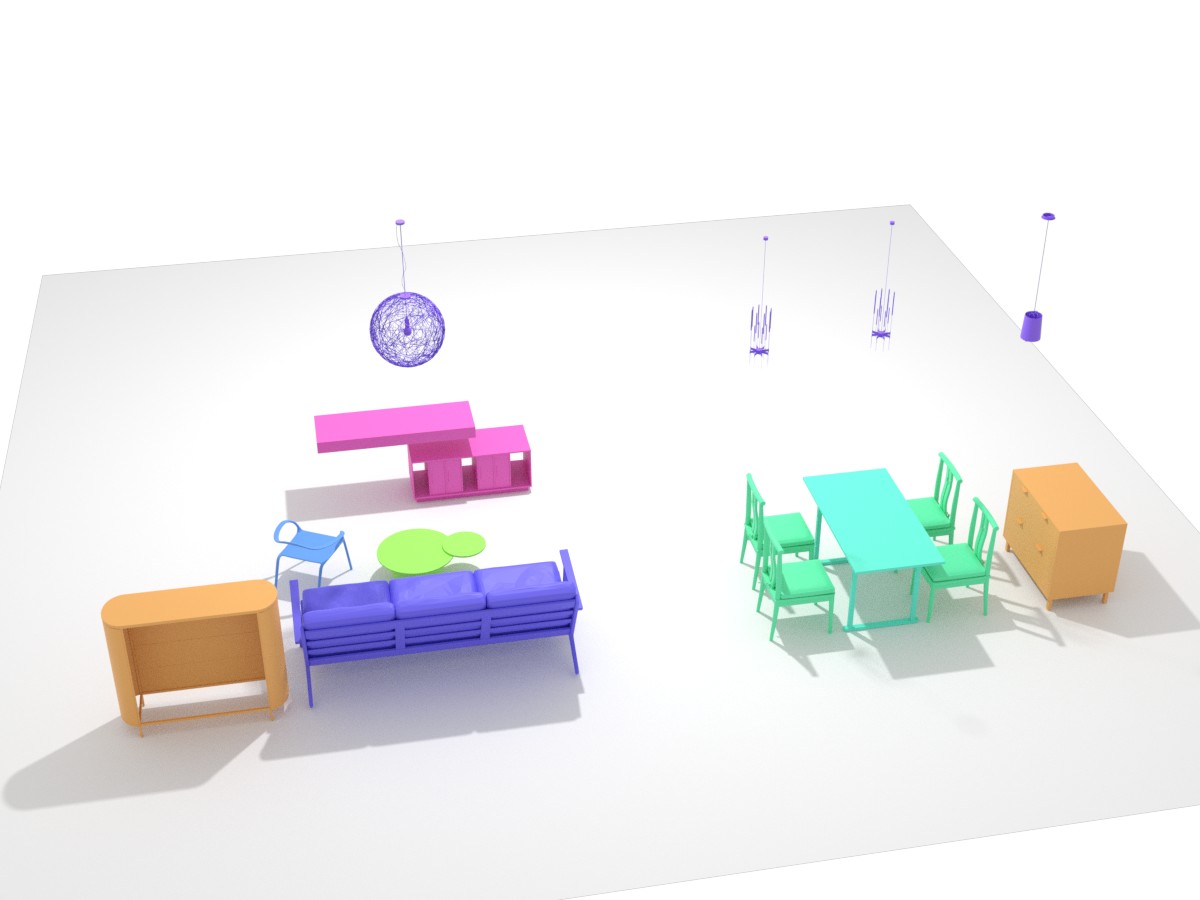}

        \includegraphics[width=\textwidth]{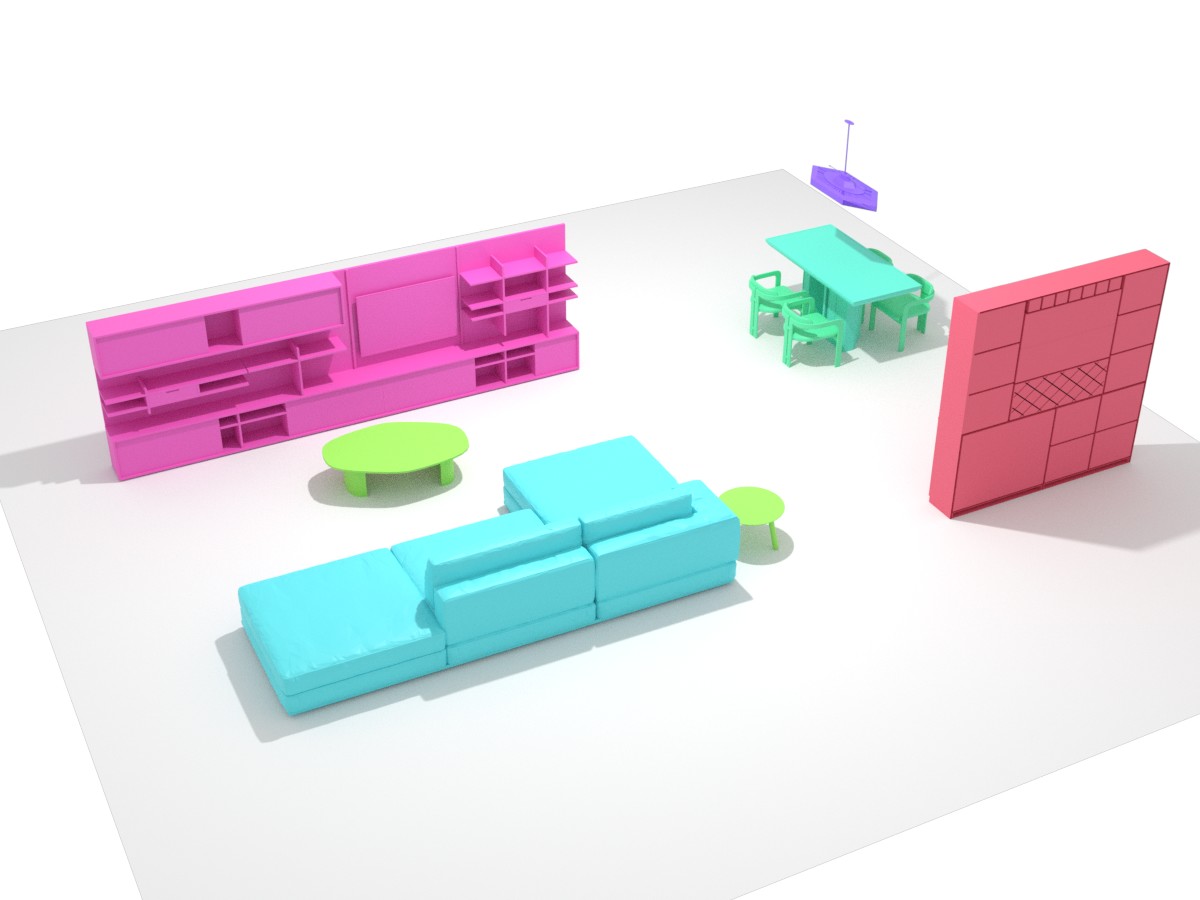}
        
	\end{subfigure}
	\rulesep
	\begin{subfigure}[t]{0.23\textwidth}
		\includegraphics[width=\textwidth]
	{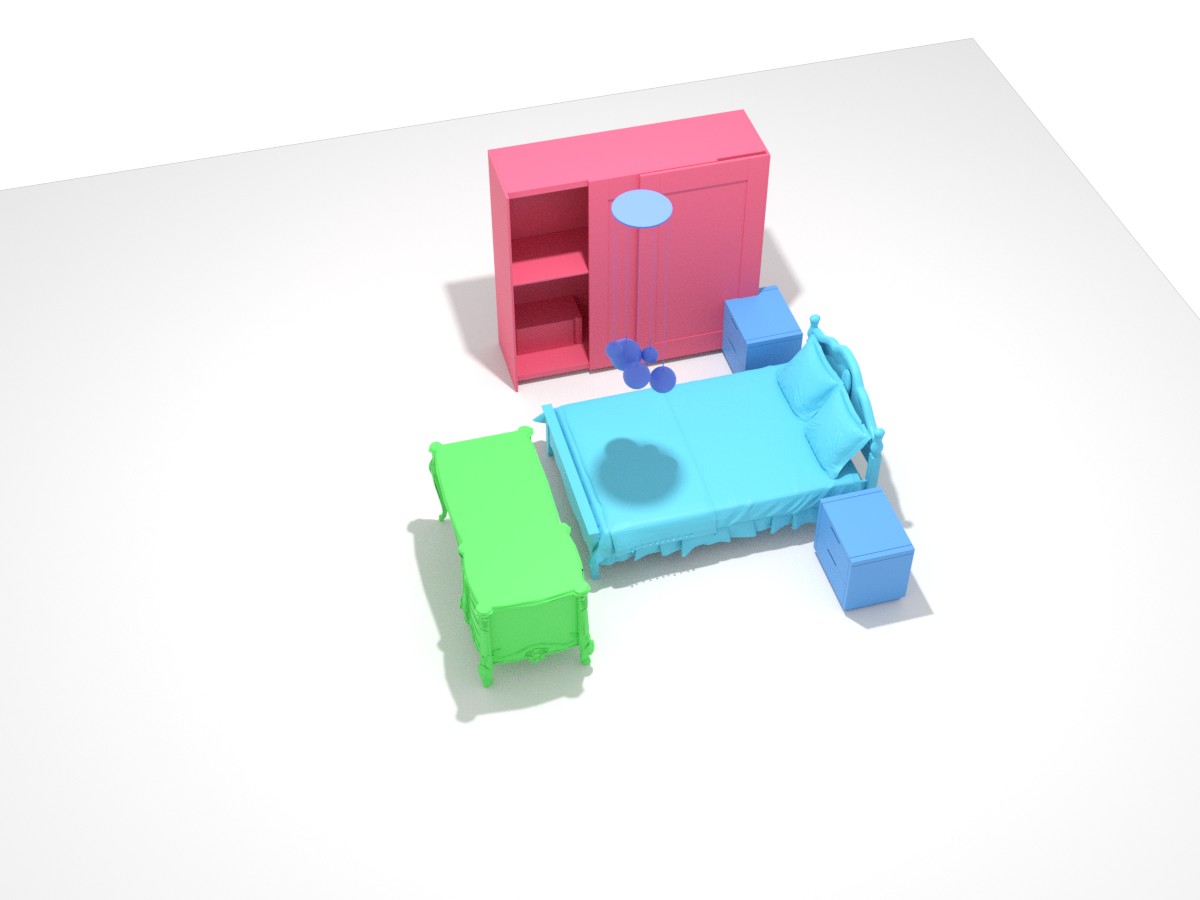}
 
  	\includegraphics[width=\textwidth]{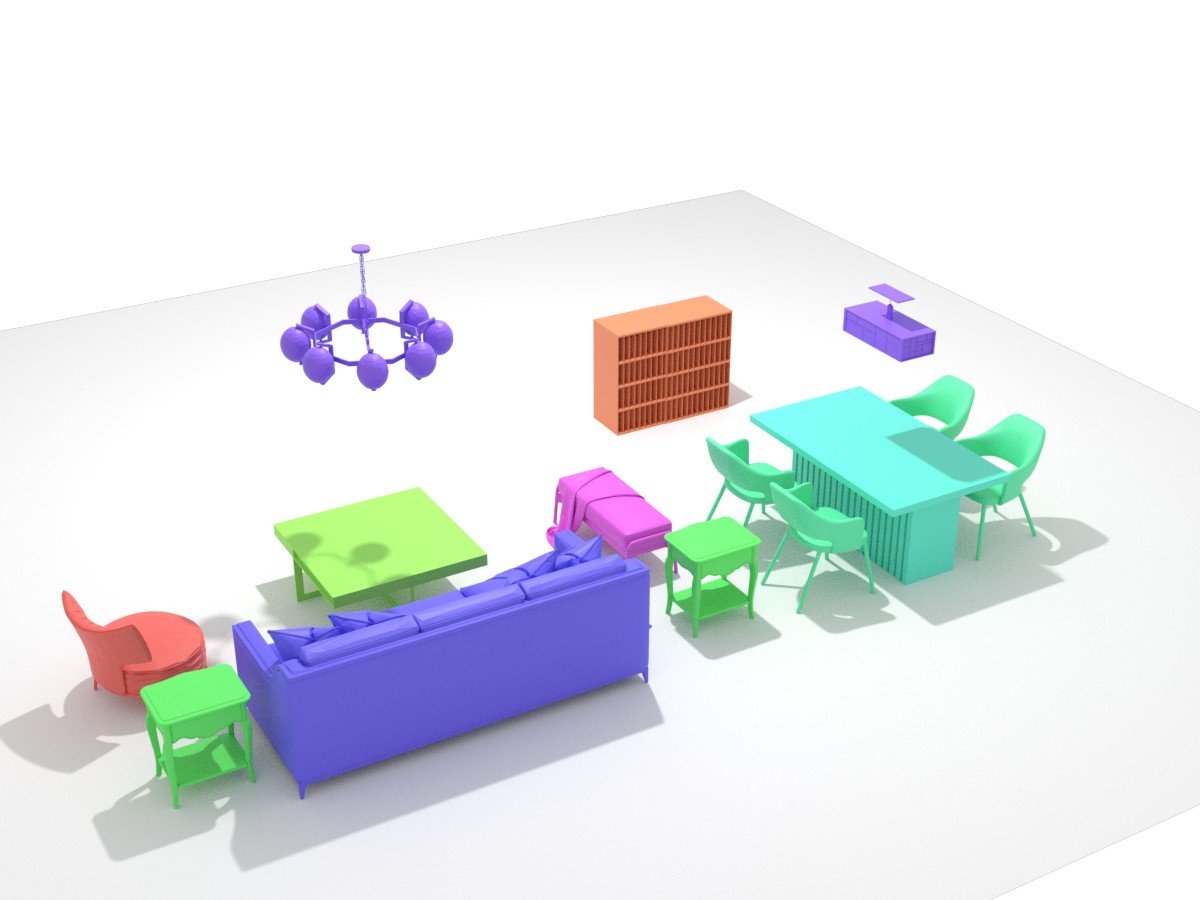}
   
		\includegraphics[width=\textwidth]{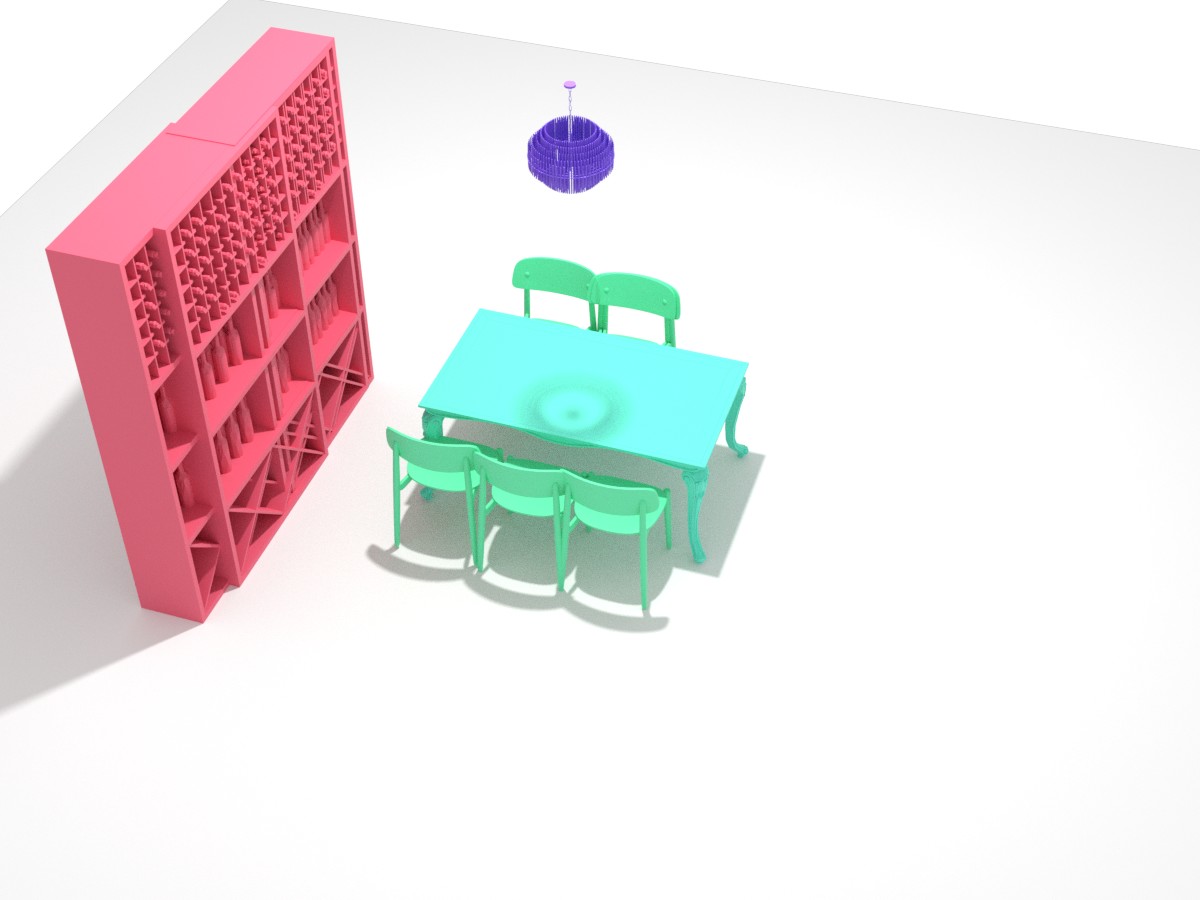}

  \includegraphics[width=\textwidth]{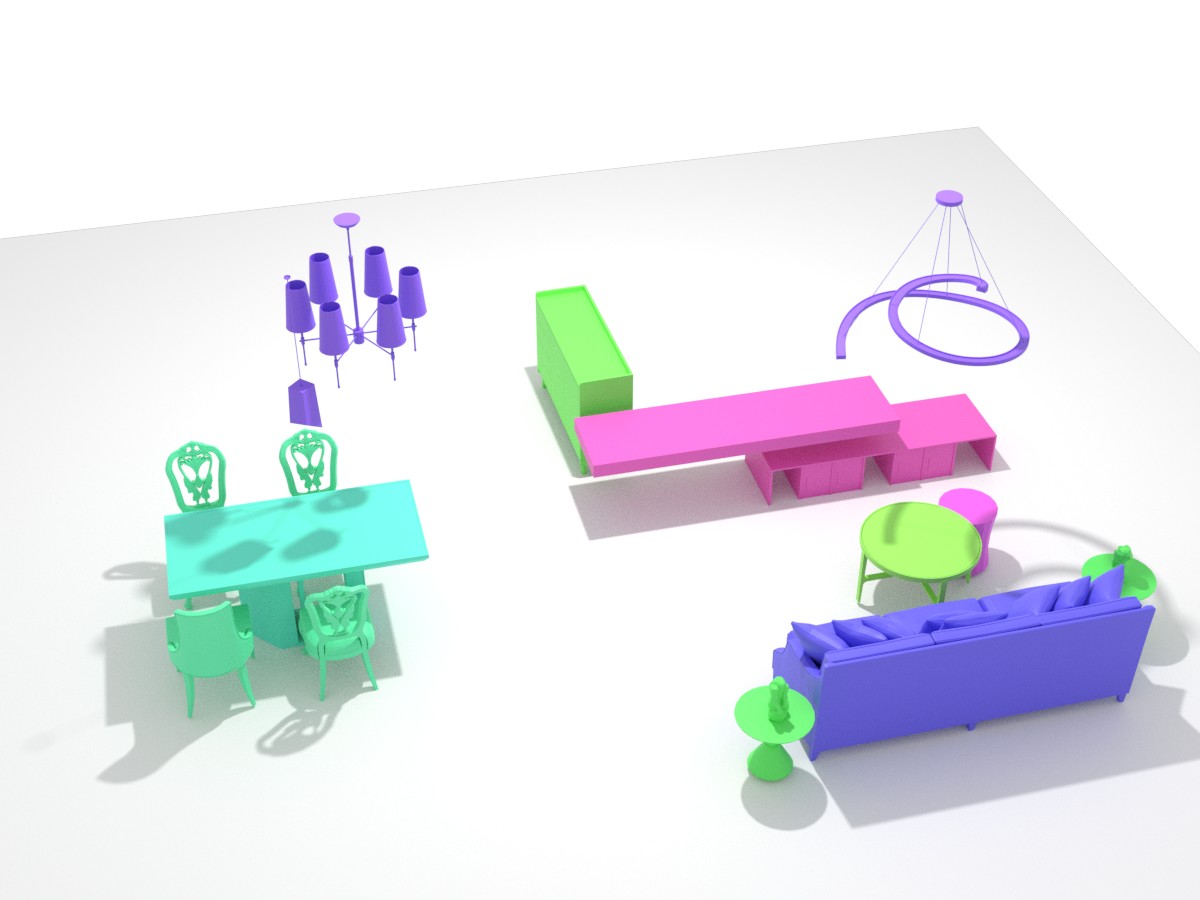}
	\end{subfigure}
	\caption{Diverse and plausible results of \textbf{unconditional scene synthesis from our method}. }
\label{fig:uncond_gallery}
\end{figure*}

\begin{figure*}[t]
	\centering
	\begin{subfigure}[t]{0.14\textwidth}
            \includegraphics[width=\textwidth]{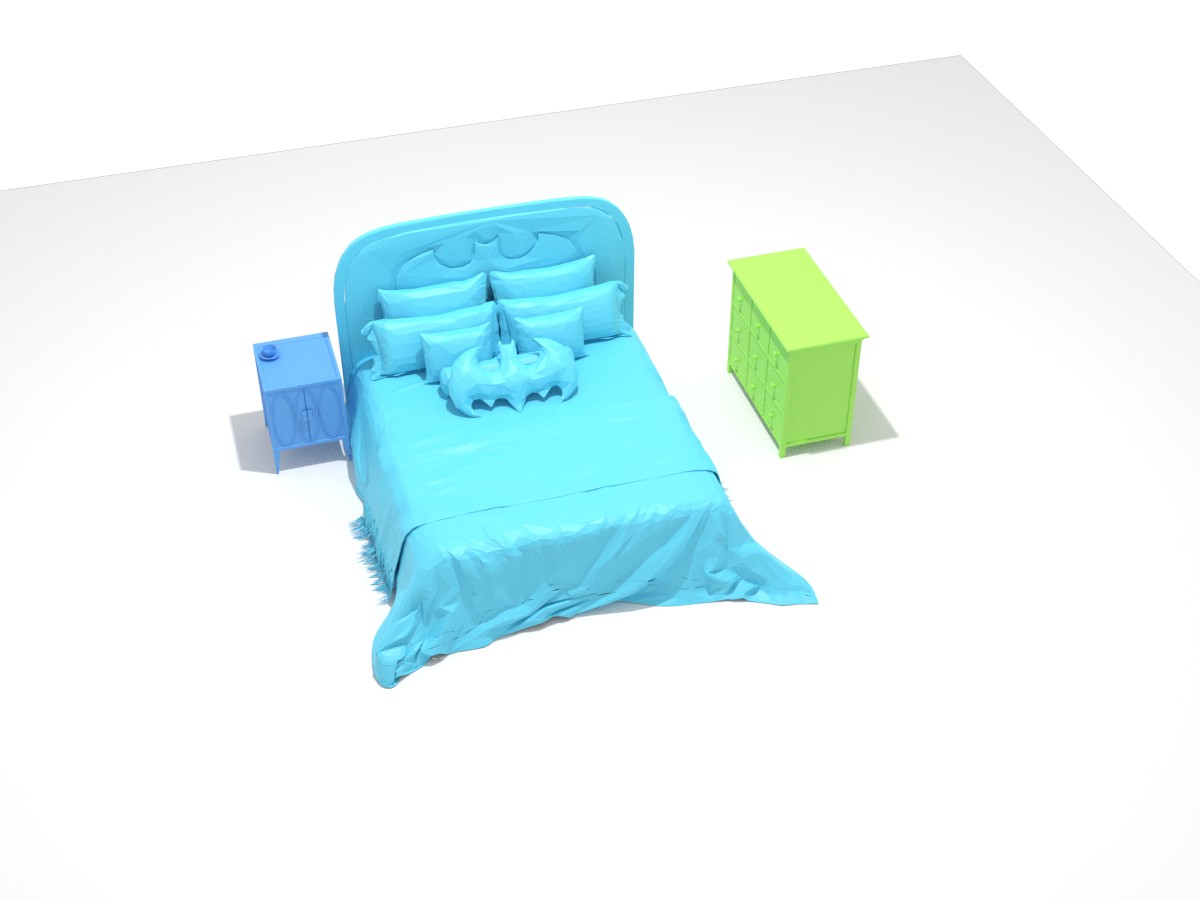}
            \includegraphics[width=\textwidth]{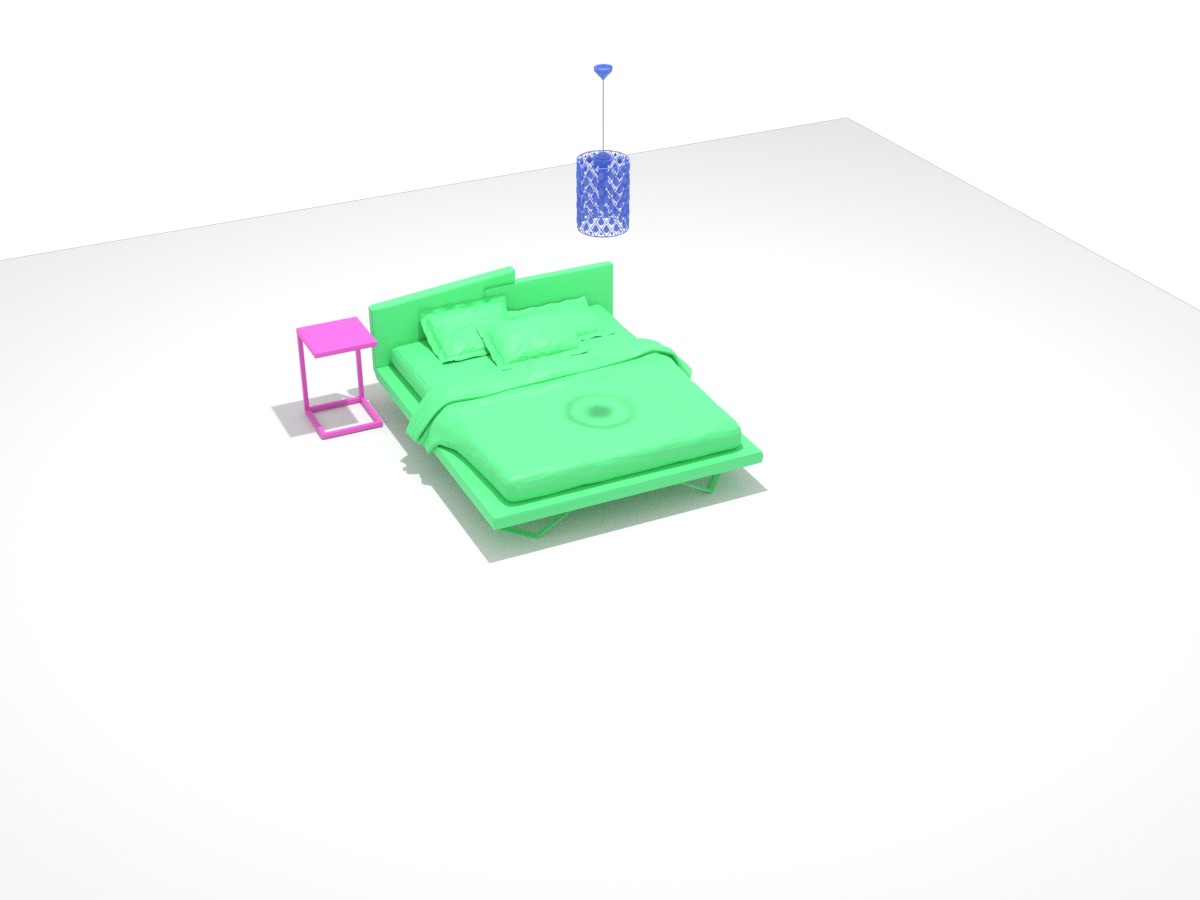}
            \includegraphics[width=\textwidth]{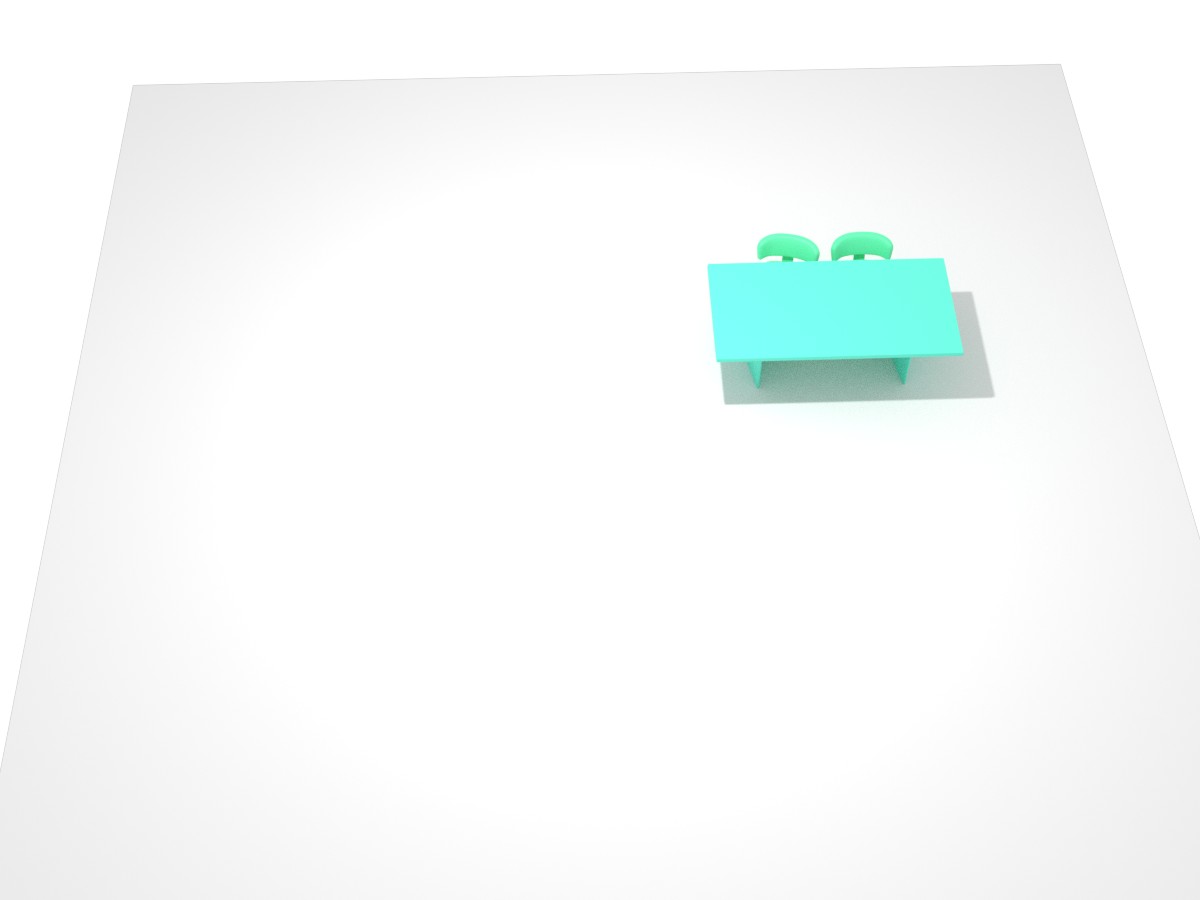}
            \includegraphics[width=\textwidth]{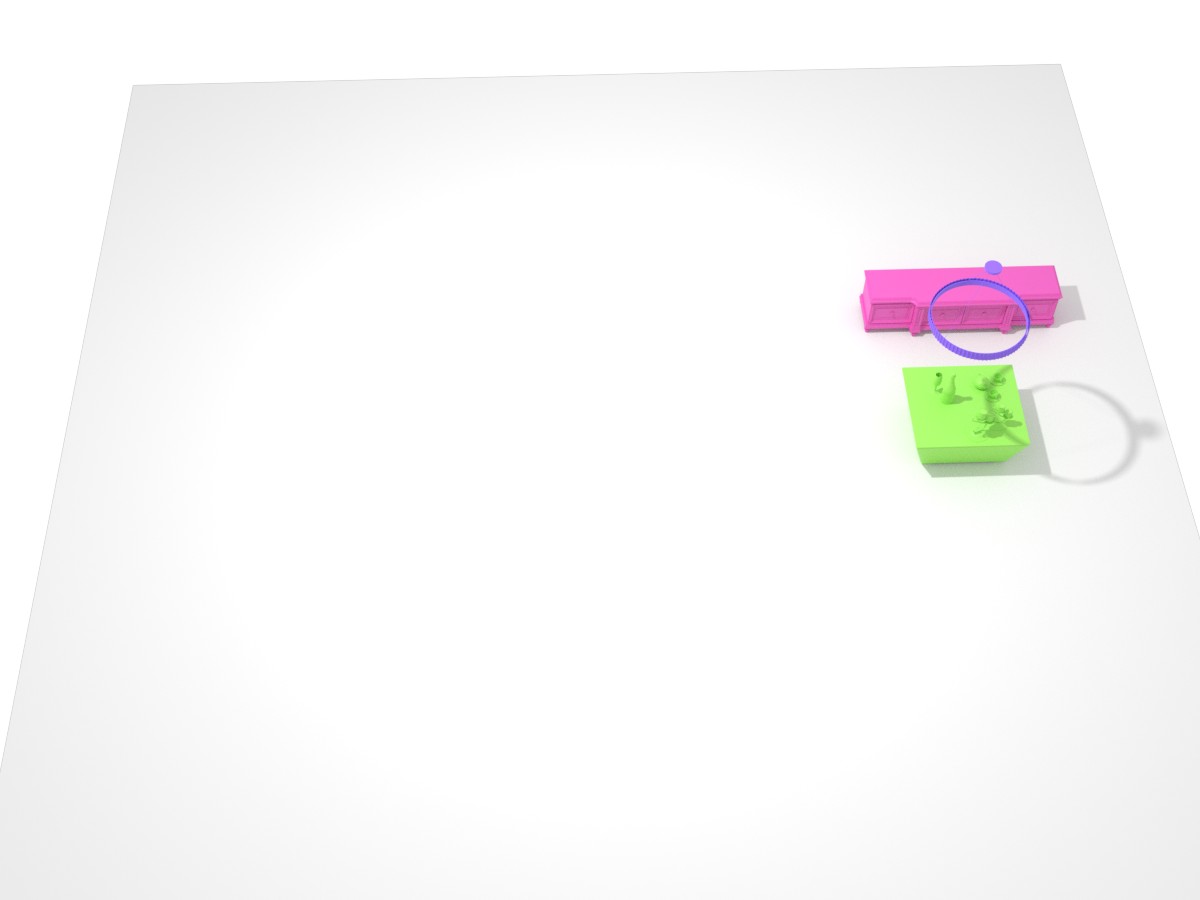}
        \caption{Partial Scenes}
	\end{subfigure}
        \rulesep
        \begin{subfigure}[t]{0.41\textwidth}
    	\includegraphics[width=0.33\textwidth]{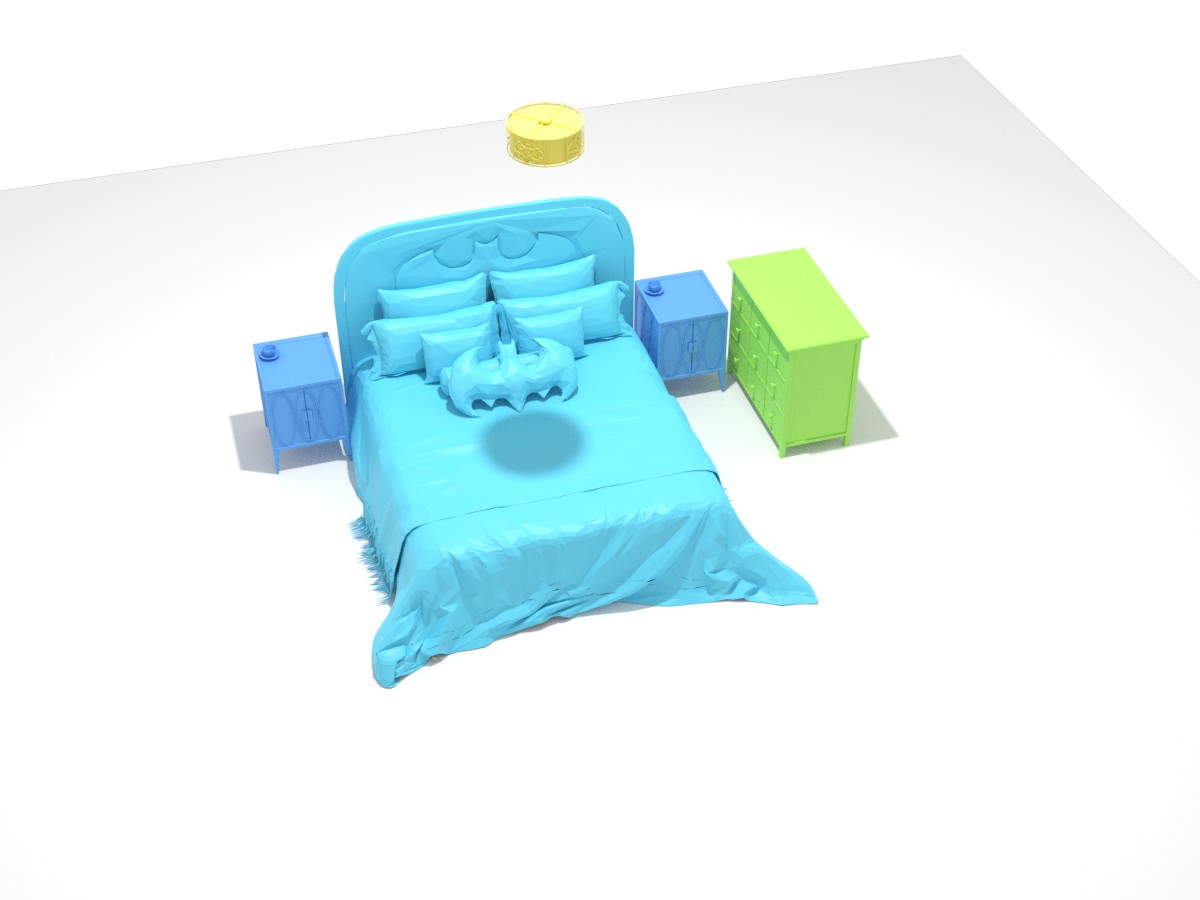}%
            \hfill
    	\includegraphics[width=0.33\textwidth]{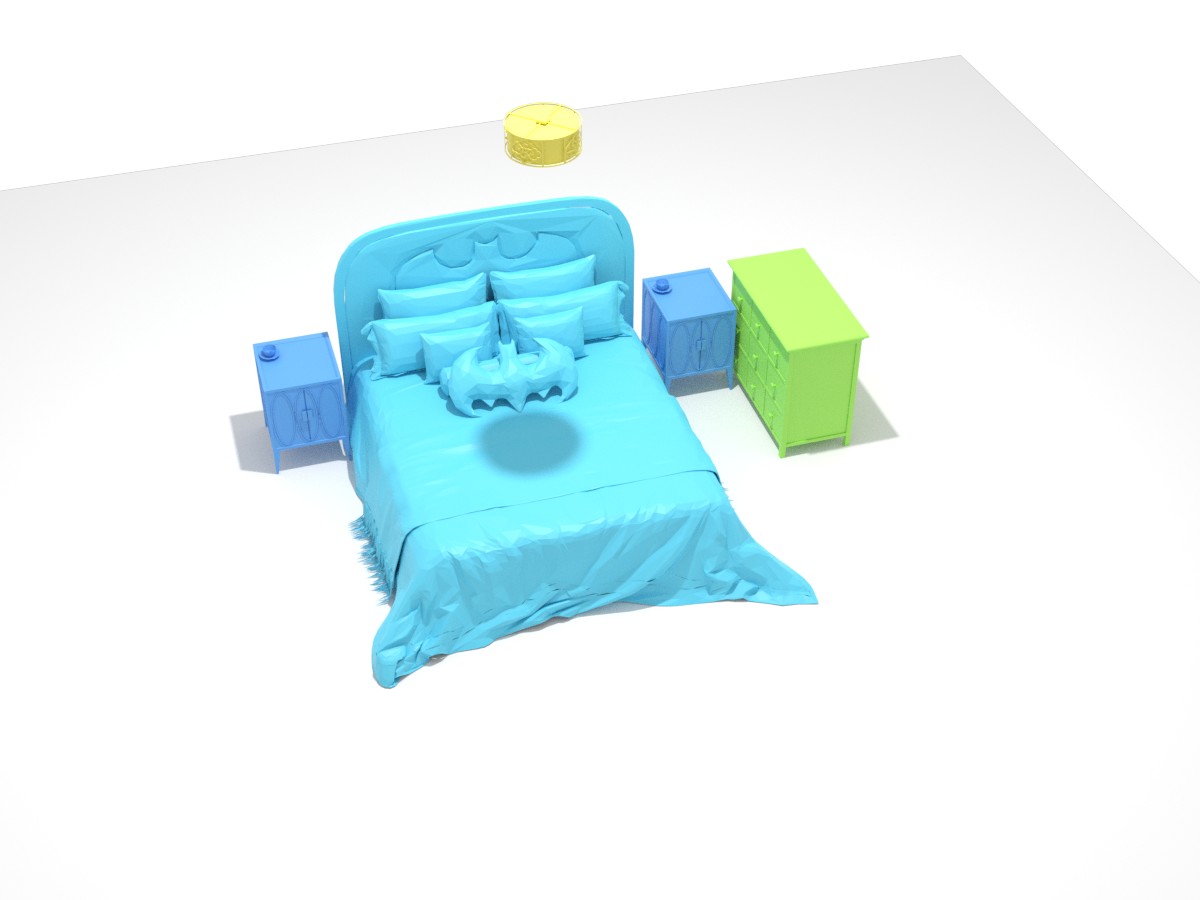}%
            \hfill
      	\includegraphics[width=0.33\textwidth]{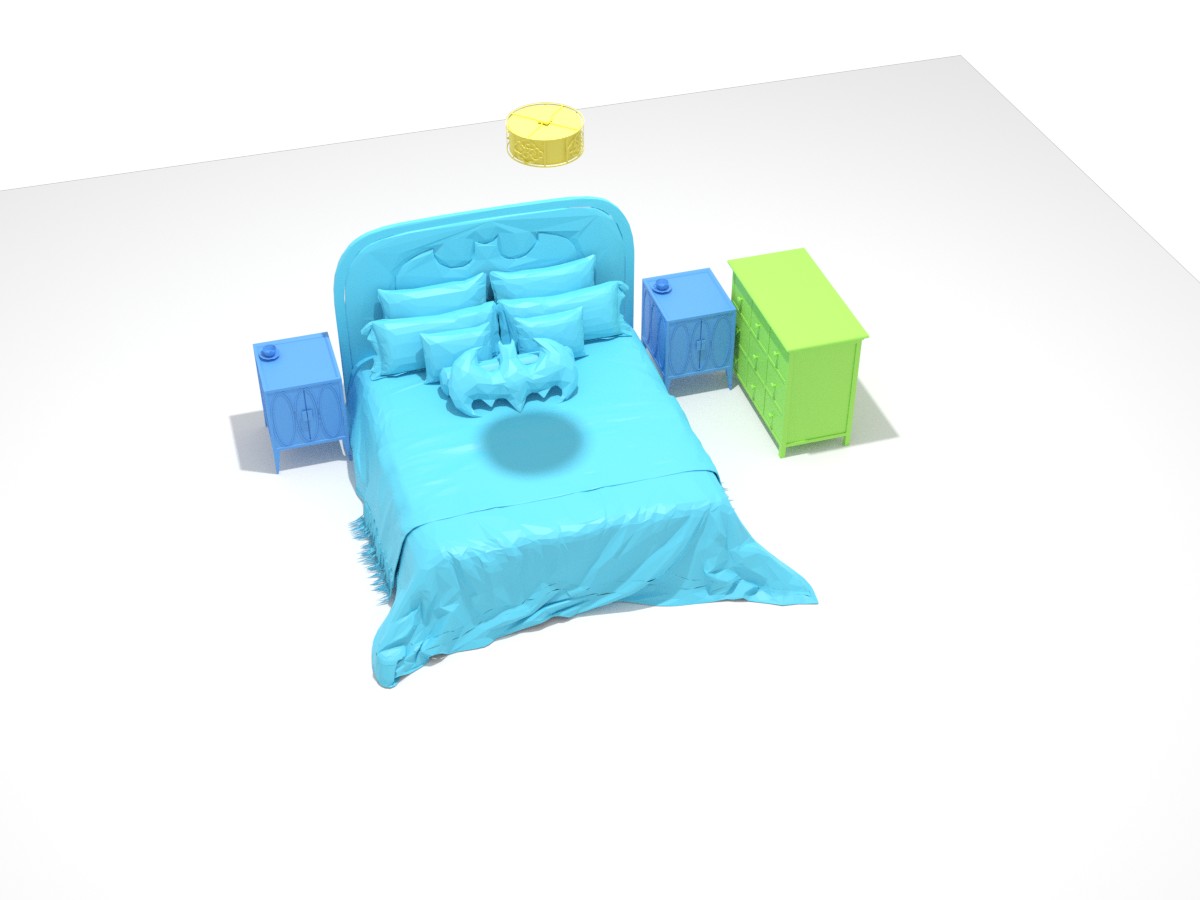} 
    	\includegraphics[width=0.33\textwidth]{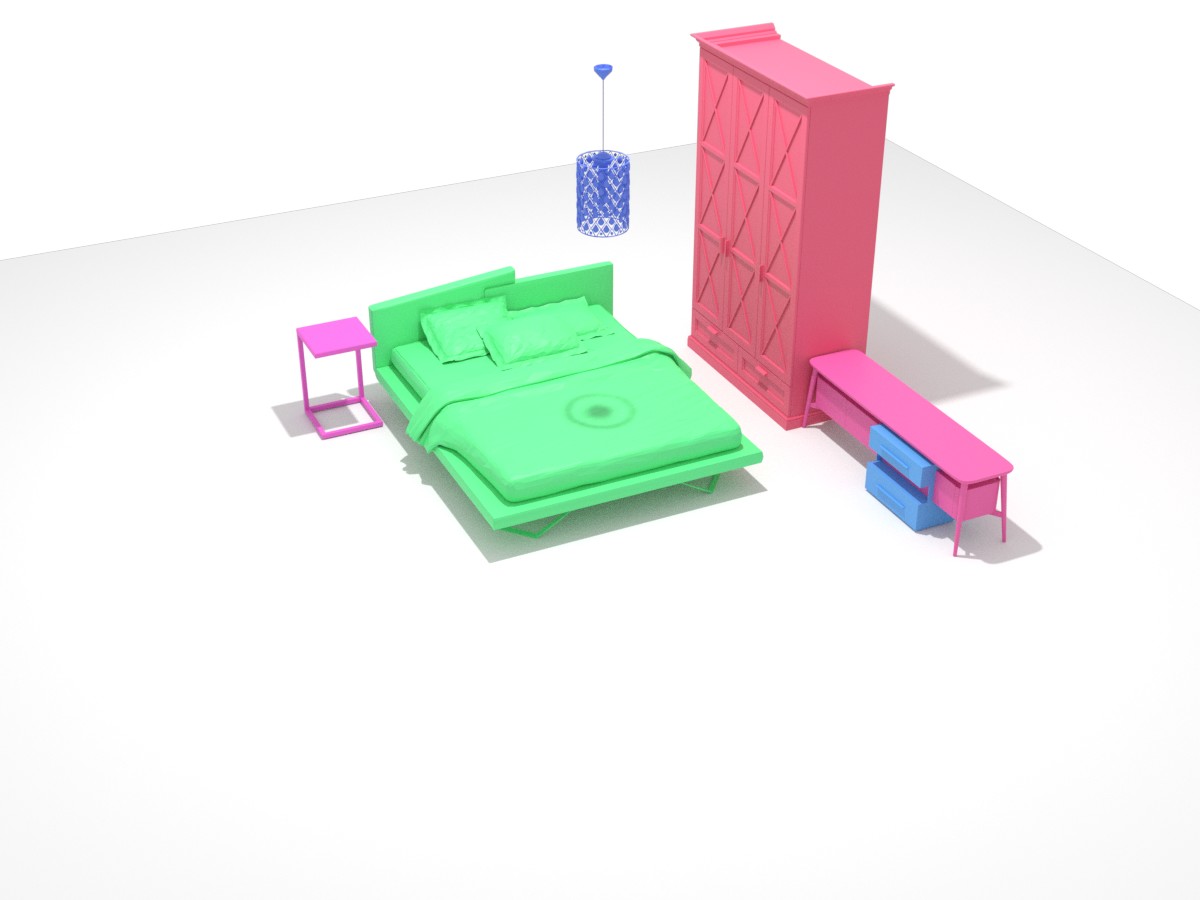}%
            \hfill
        \includegraphics[width=0.33\textwidth]{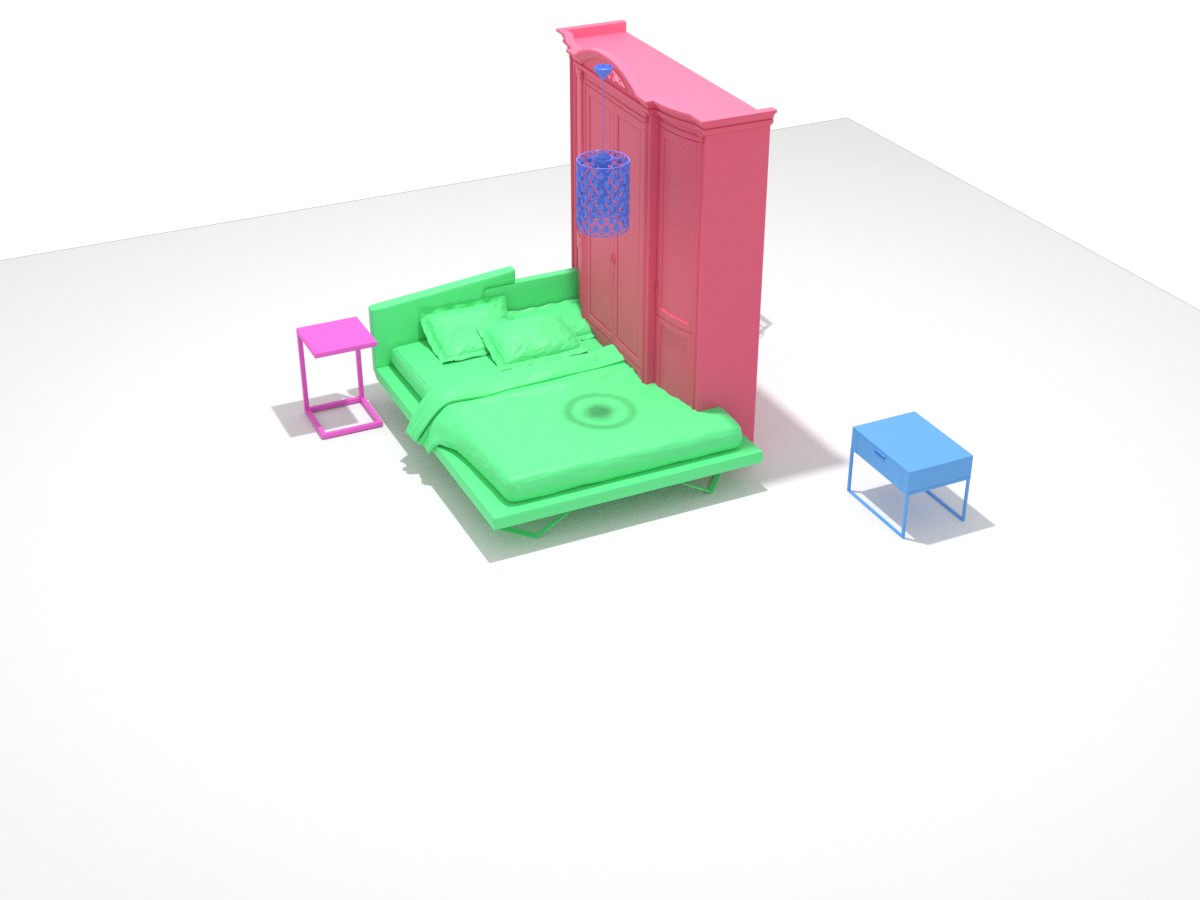}%
        \hfill
        \includegraphics[width=0.33\textwidth]{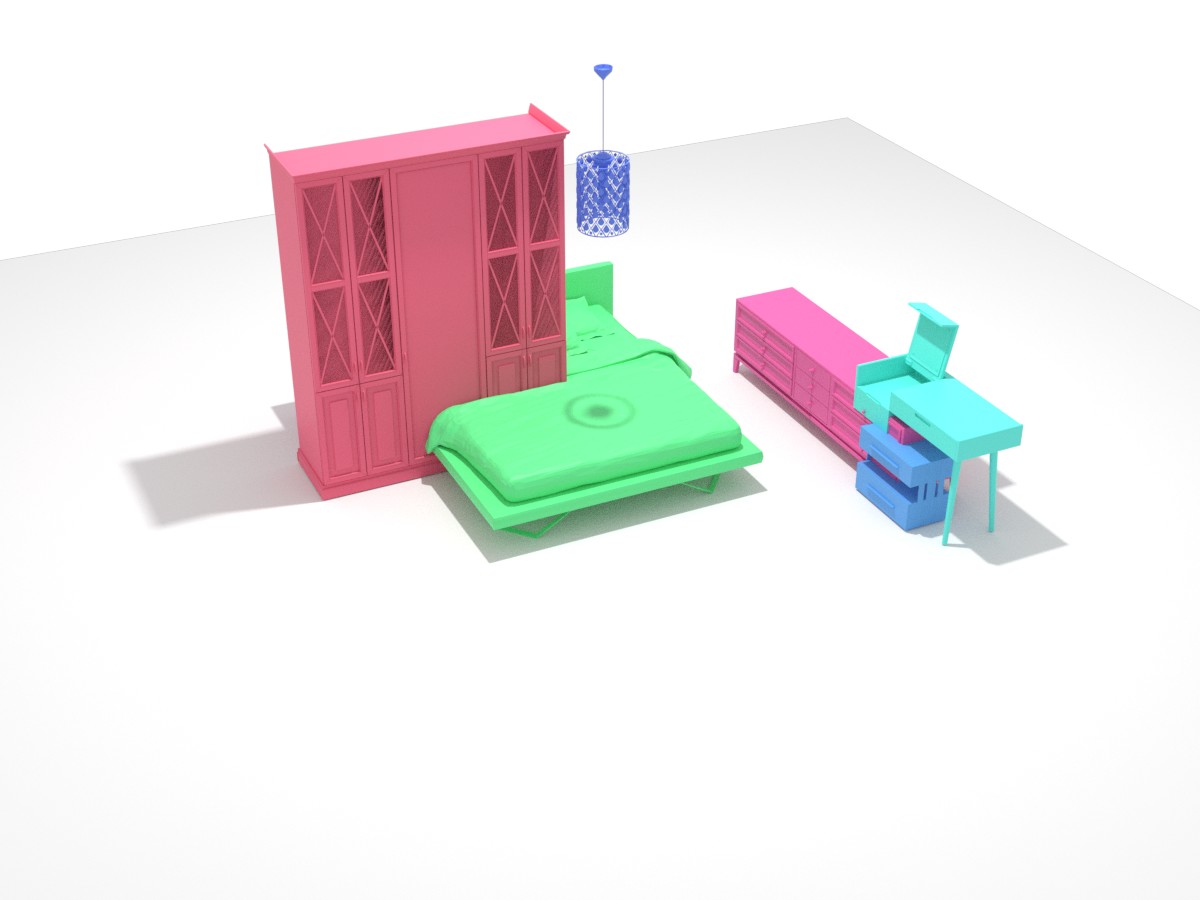}
        \includegraphics[width=0.33\textwidth]{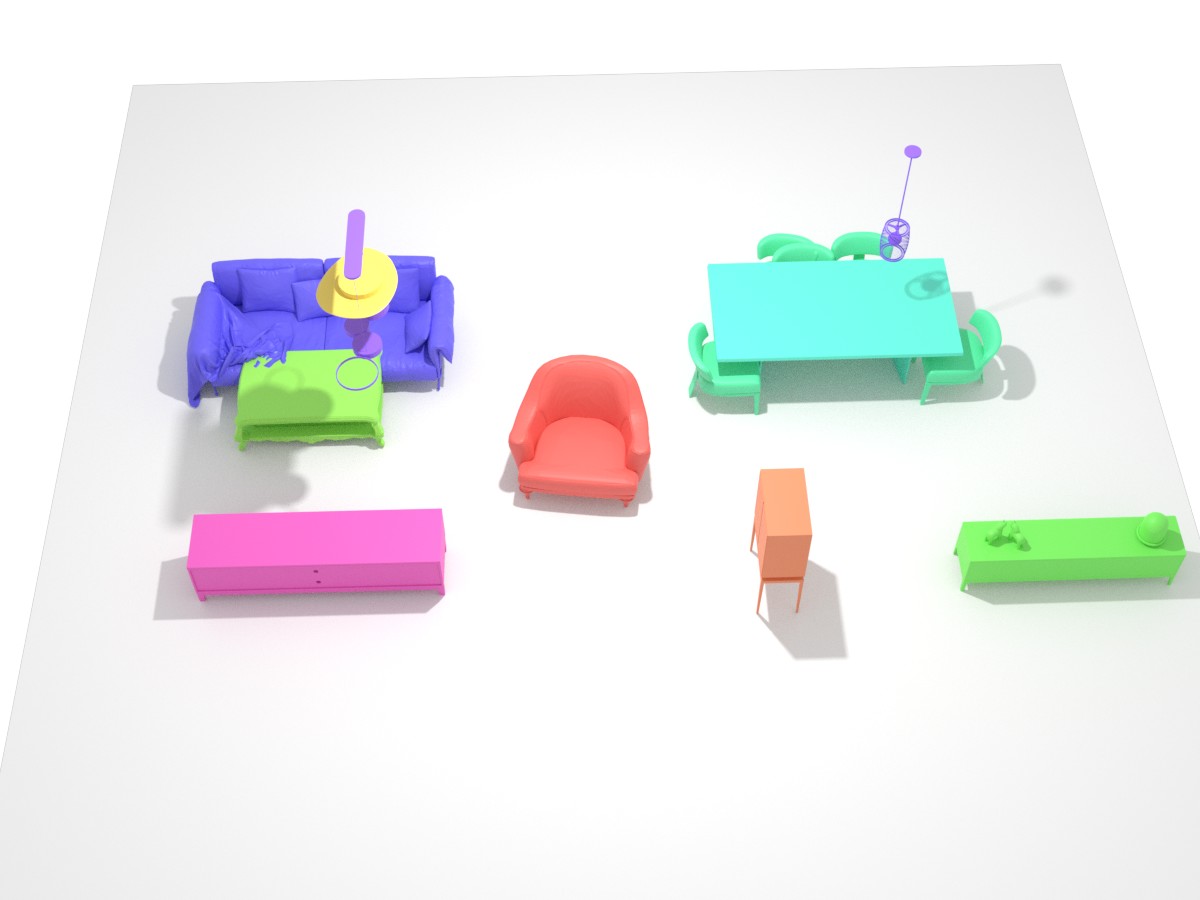}%
        \hfill
        \includegraphics[width=0.33\textwidth]{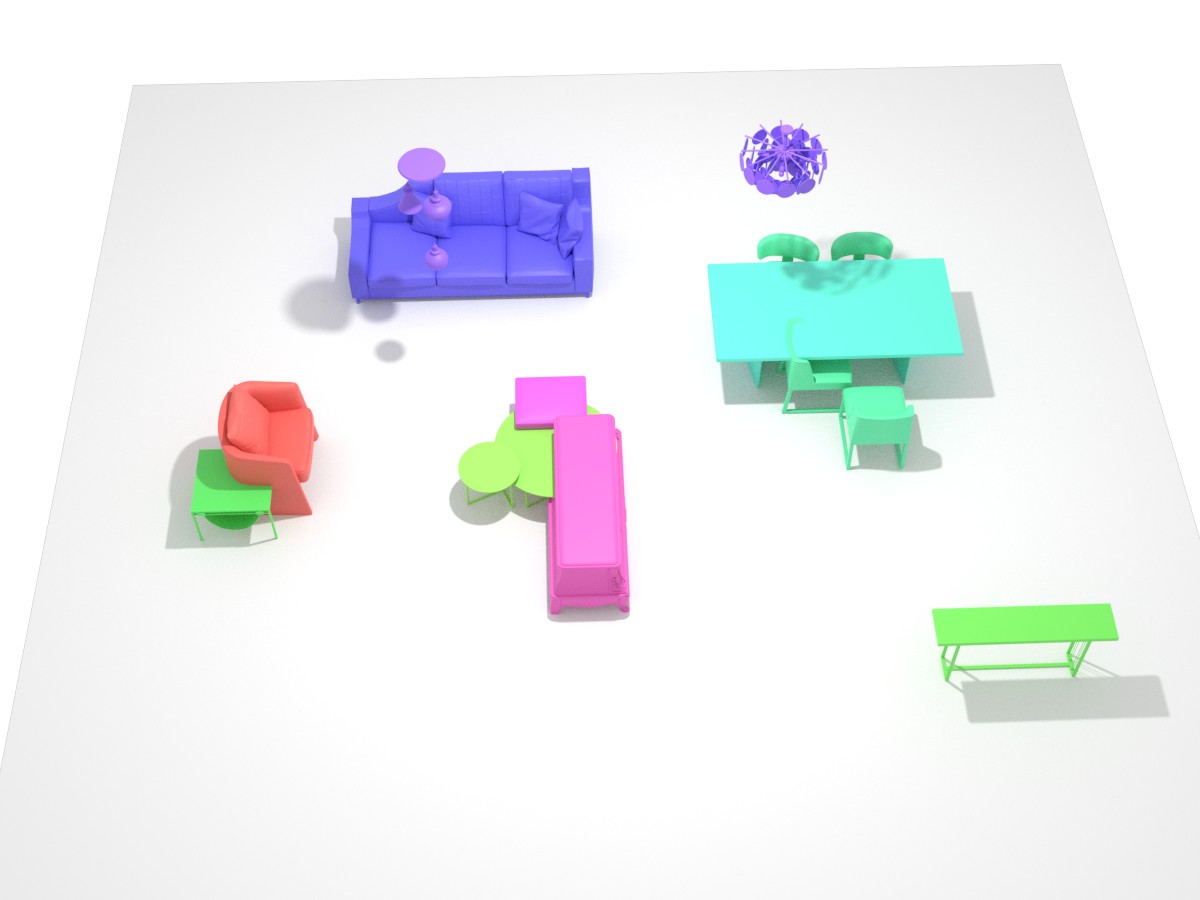}%
        \hfill
        \includegraphics[width=0.33\textwidth]{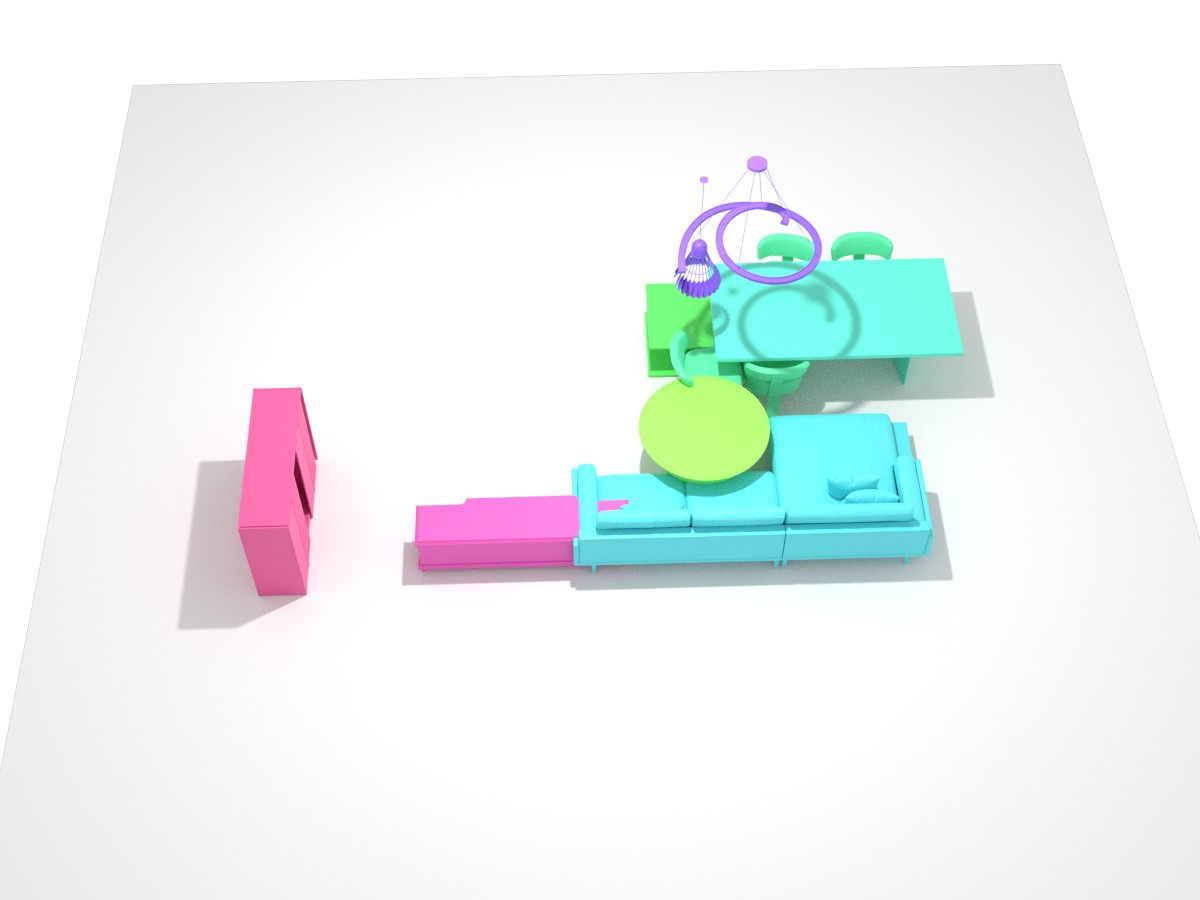}
        \includegraphics[width=0.33\textwidth]{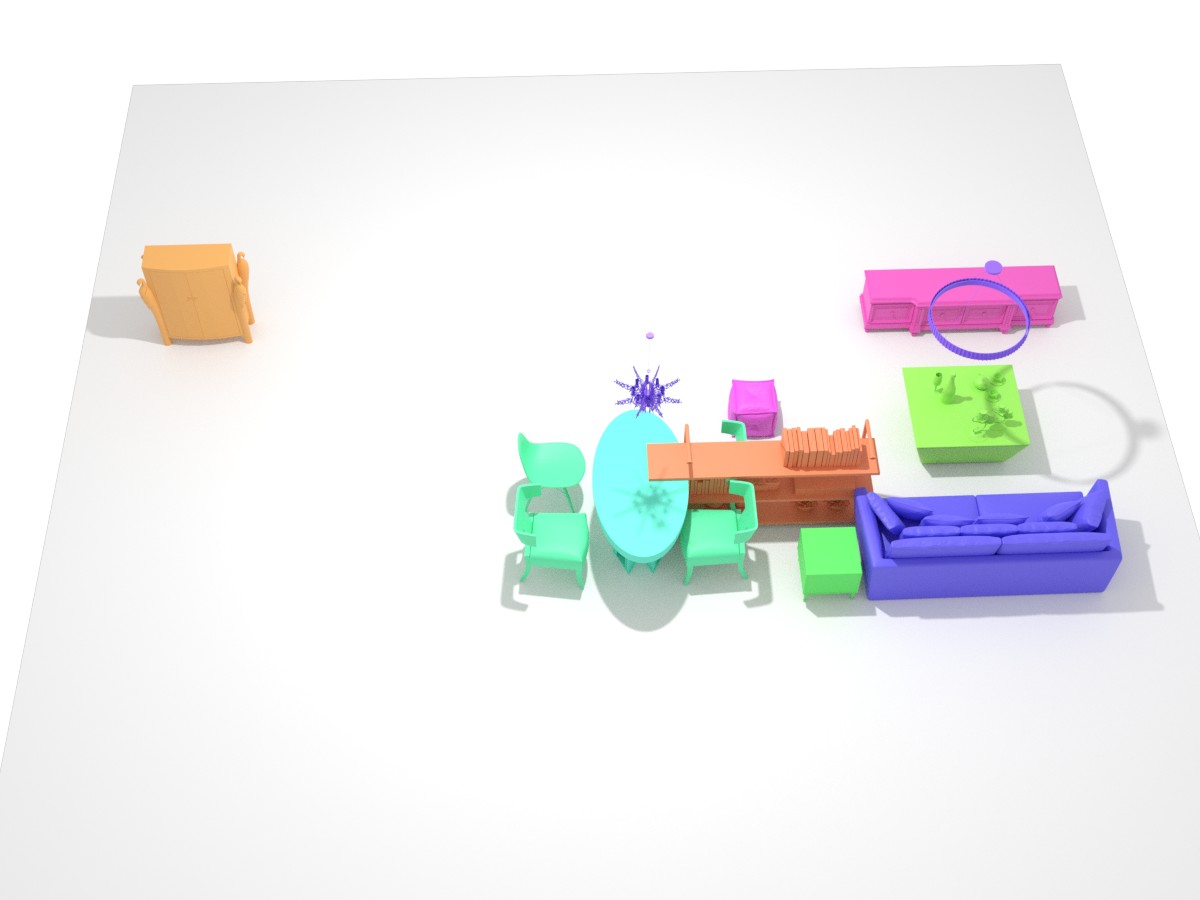}%
        \hfill
        \includegraphics[width=0.33\textwidth]{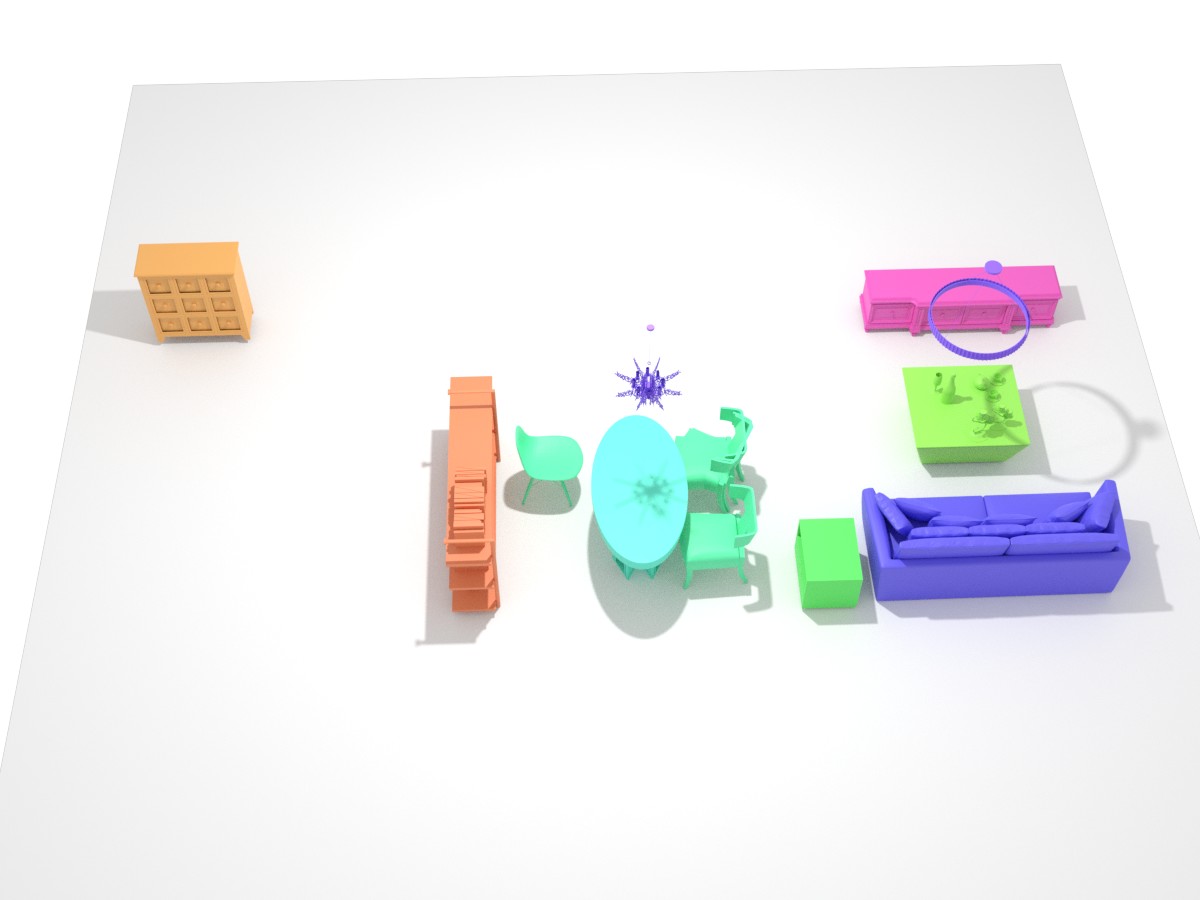}%
        \hfill
        \includegraphics[width=0.33\textwidth]{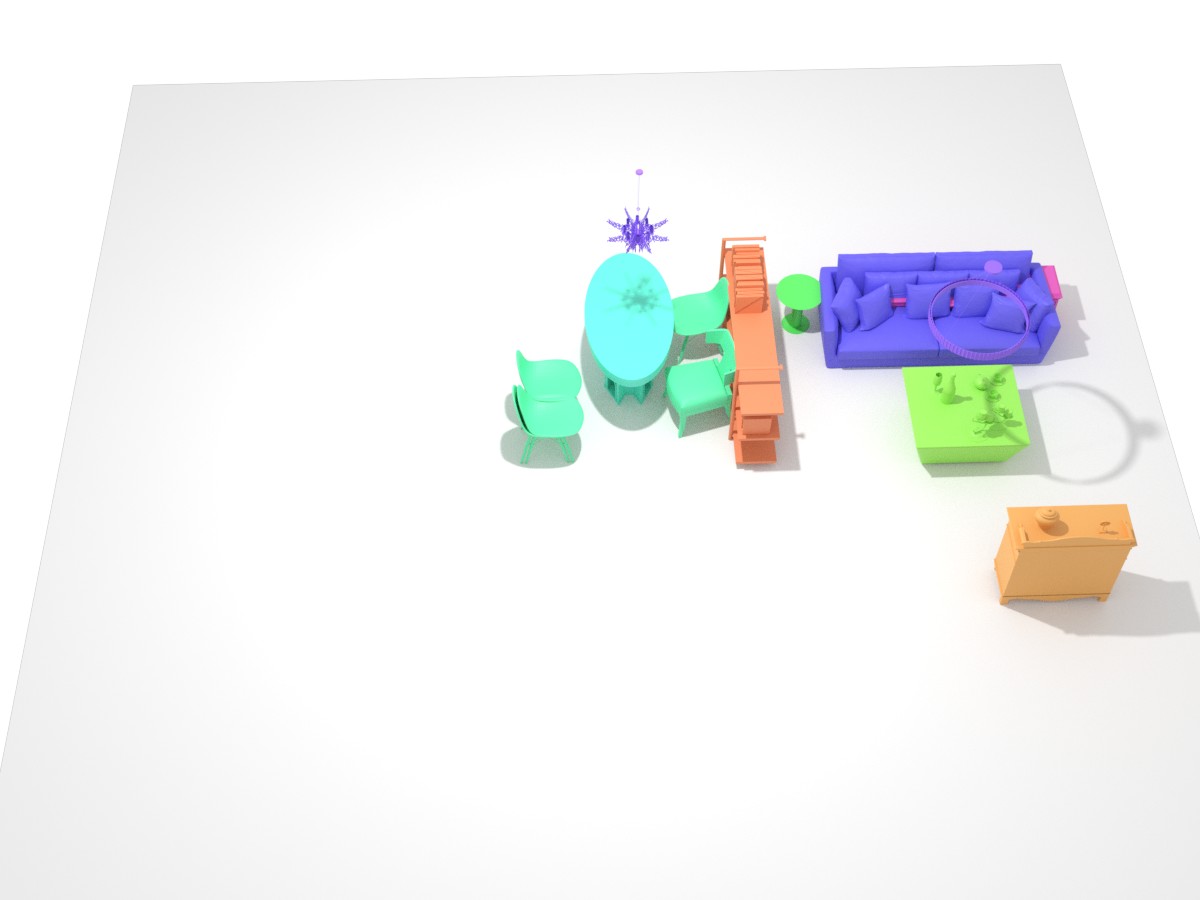}
        \caption{ATISS~\cite{paschalidou2021atiss}}
	\end{subfigure}
        \rulesep
	\begin{subfigure}[t]{0.41\textwidth}
    	\includegraphics[width=0.33\textwidth]{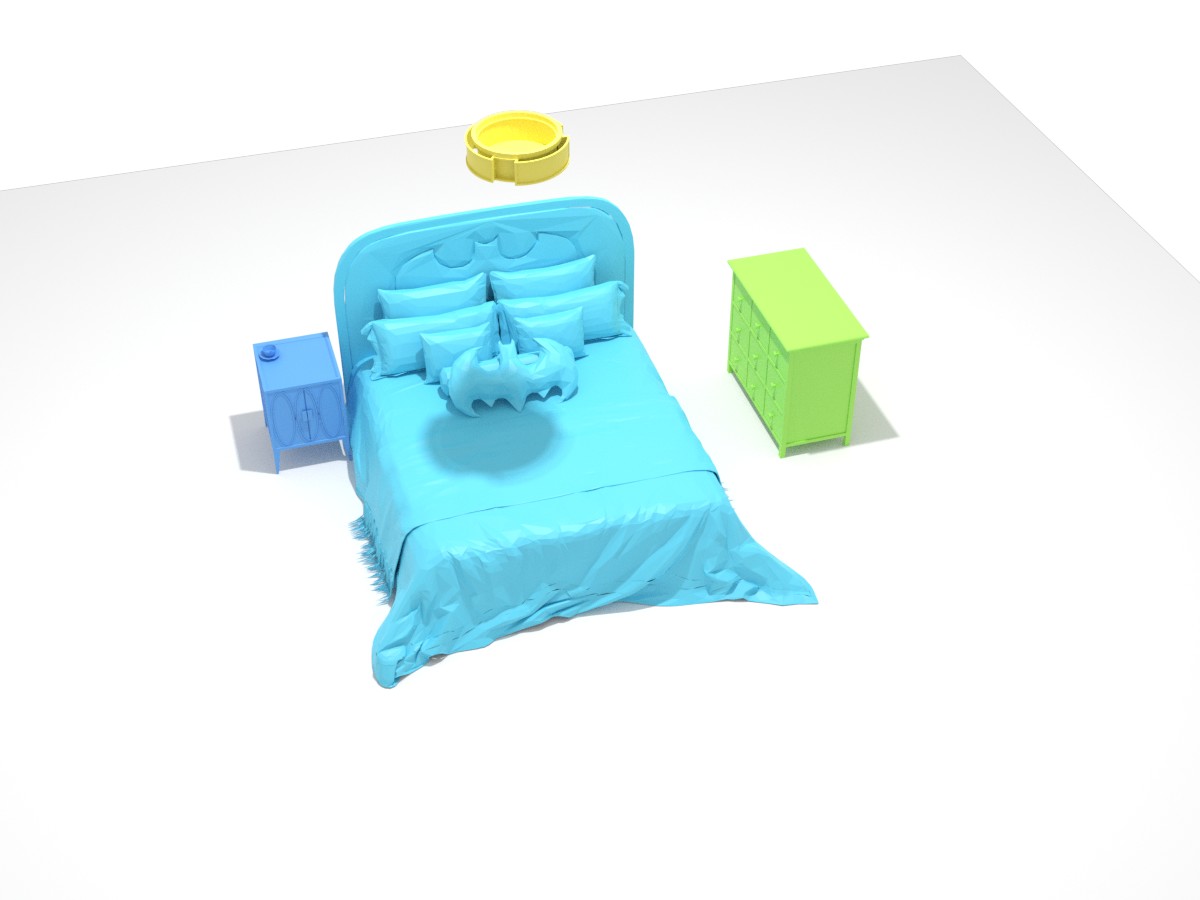}%
            \hfill
            \includegraphics[width=0.33\textwidth]{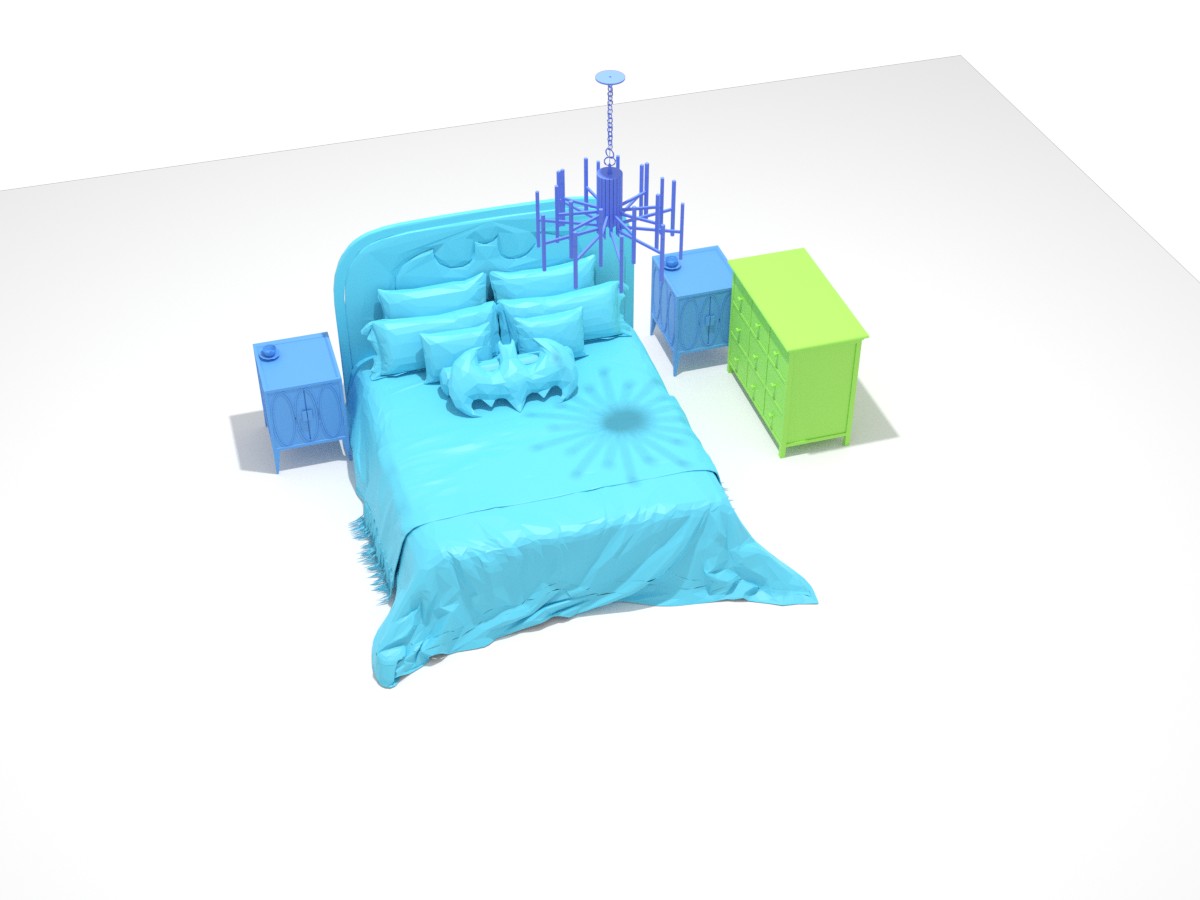}%
    	\hfill
    	\includegraphics[width=0.33\textwidth]{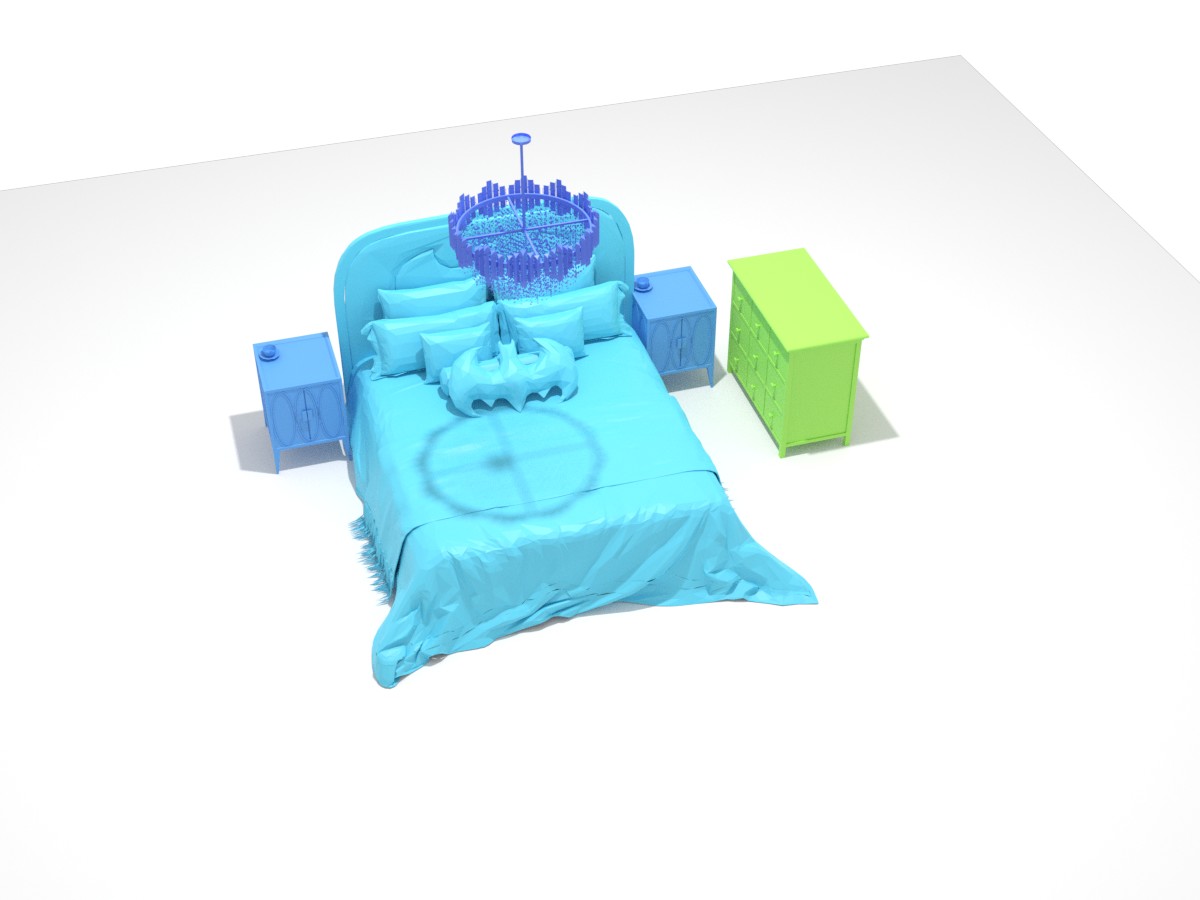}
    	\includegraphics[width=0.33\textwidth]{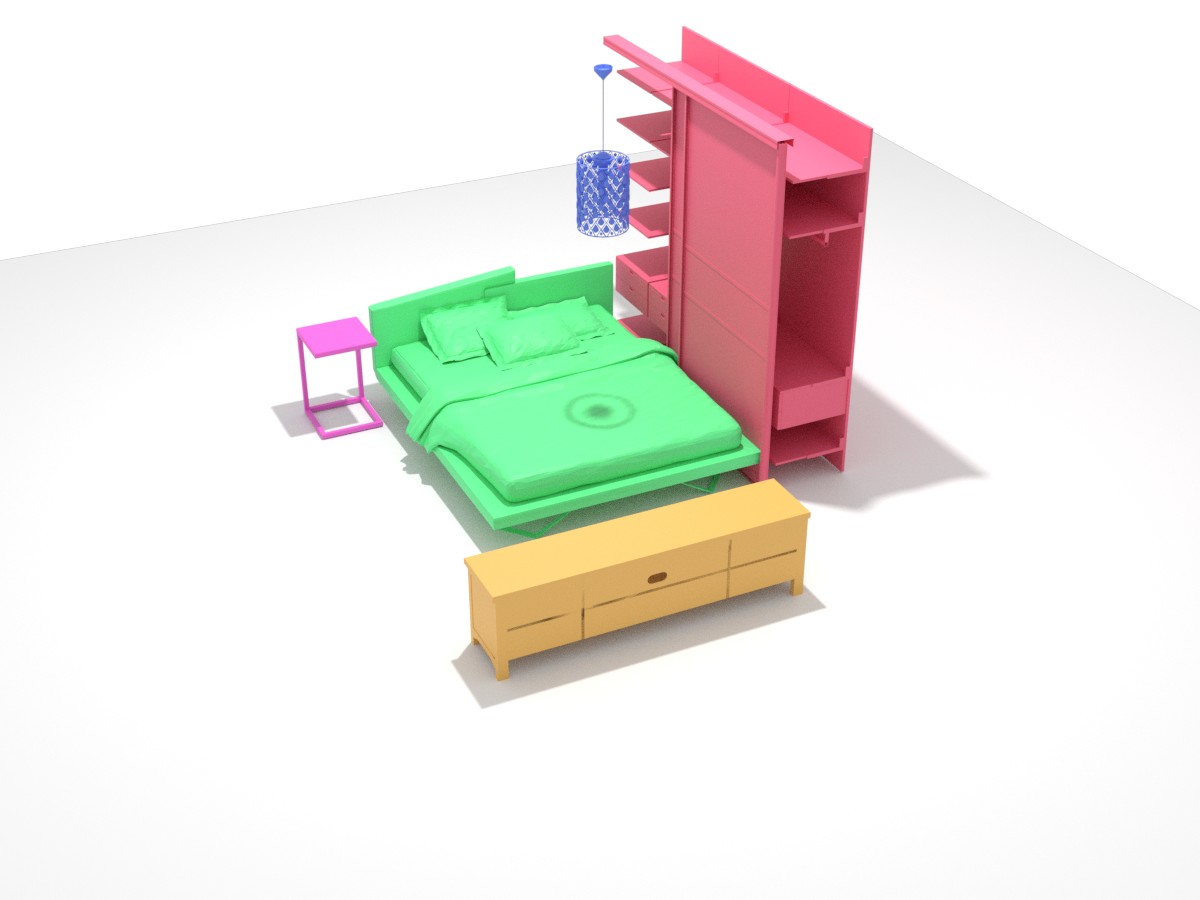}%
            \hfill
            \includegraphics[width=0.33\textwidth]{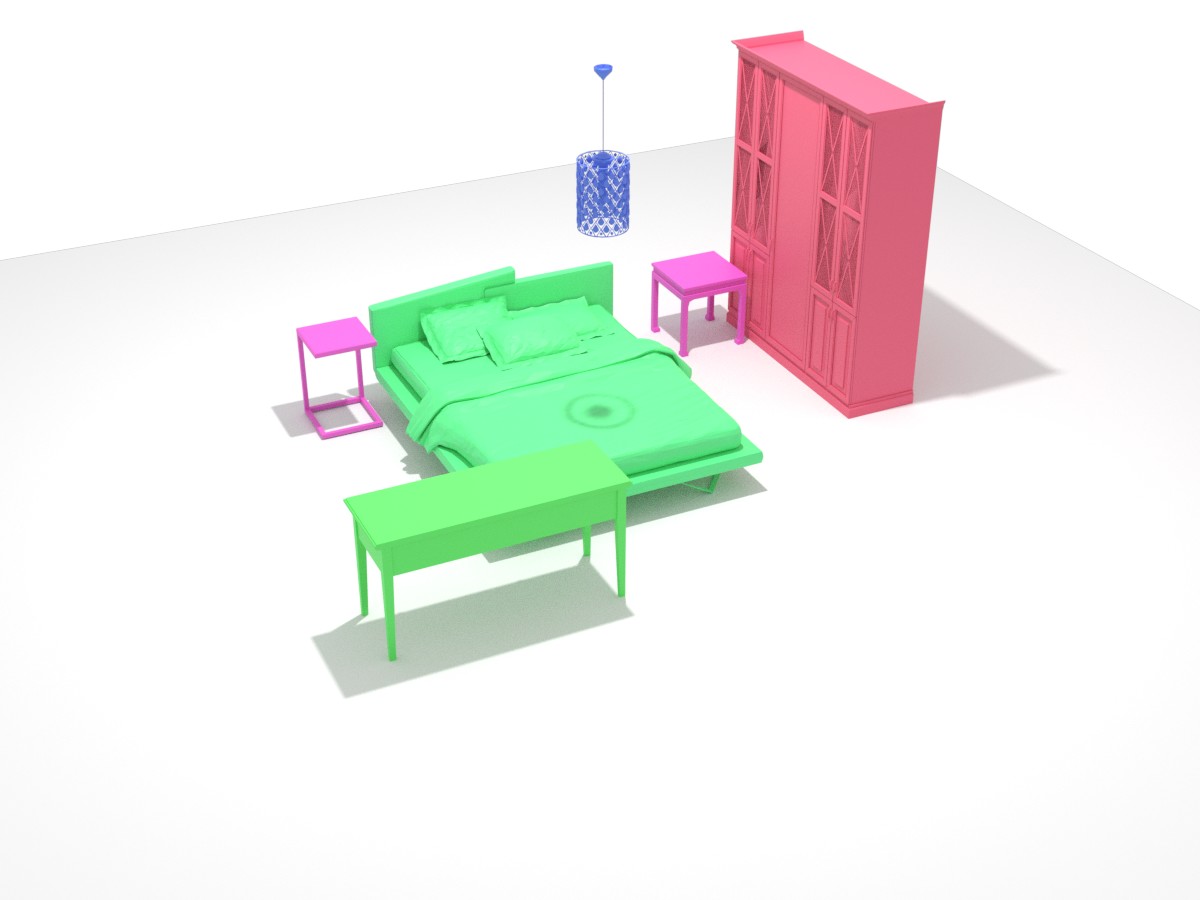}%
    	\hfill
    	\includegraphics[width=0.33\textwidth]{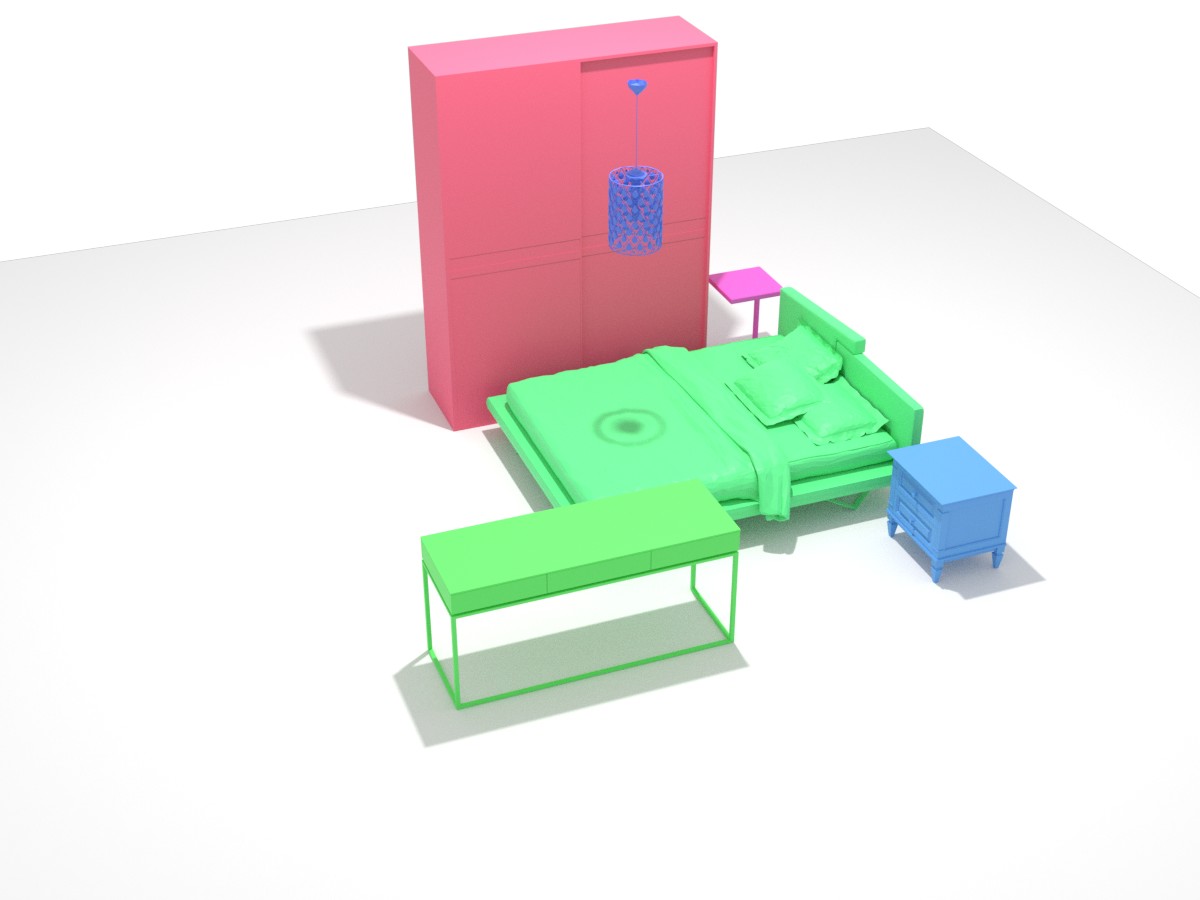}
    	\includegraphics[width=0.33\textwidth]{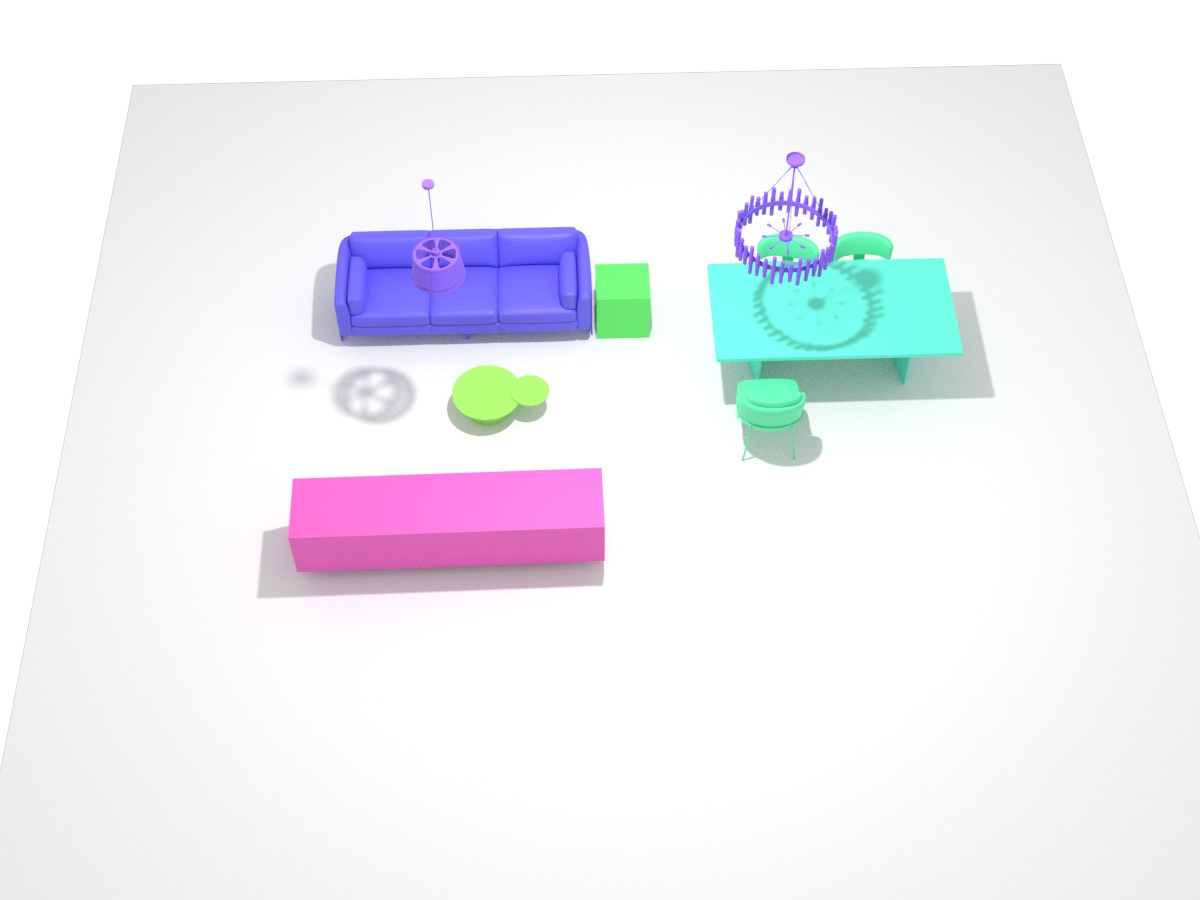}%
            \hfill
            \includegraphics[width=0.33\textwidth]{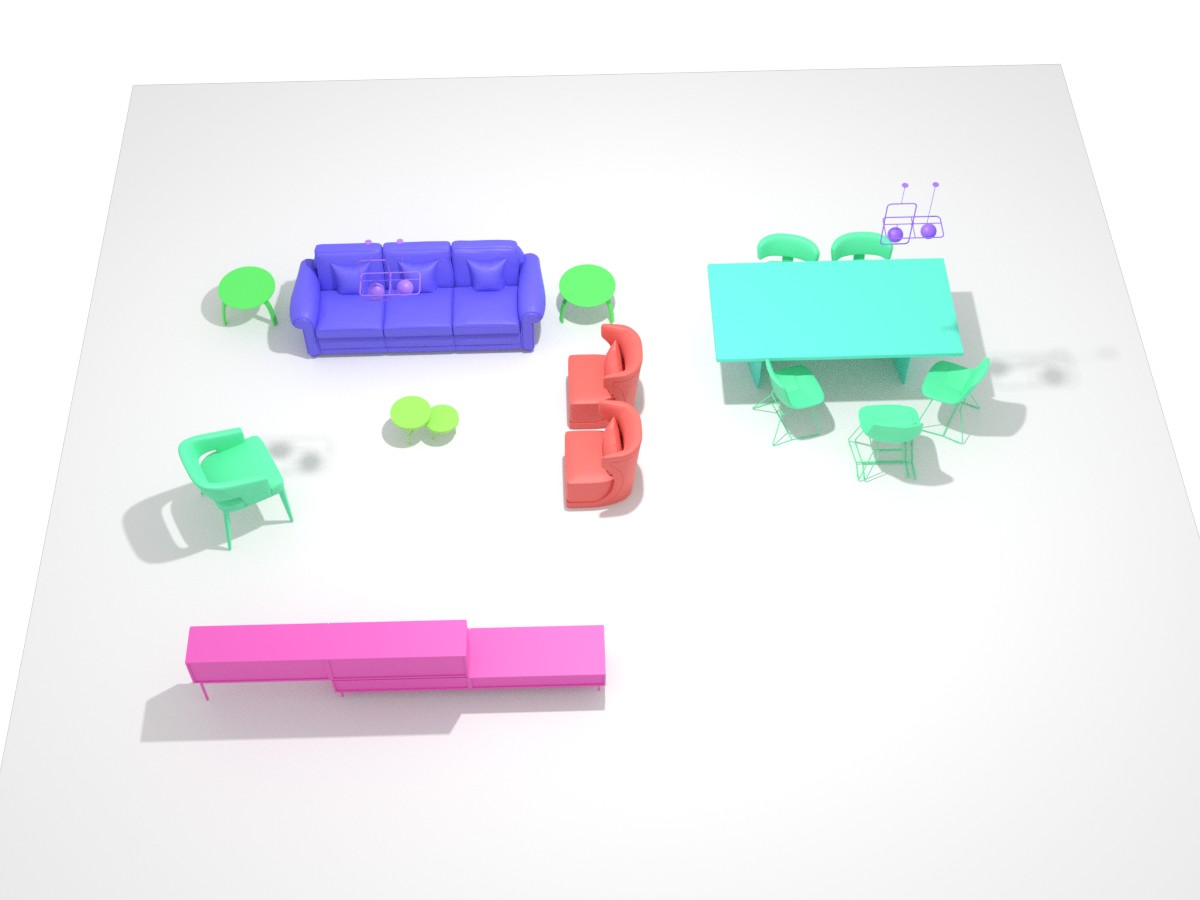}%
    	\hfill
    	\includegraphics[width=0.33\textwidth]{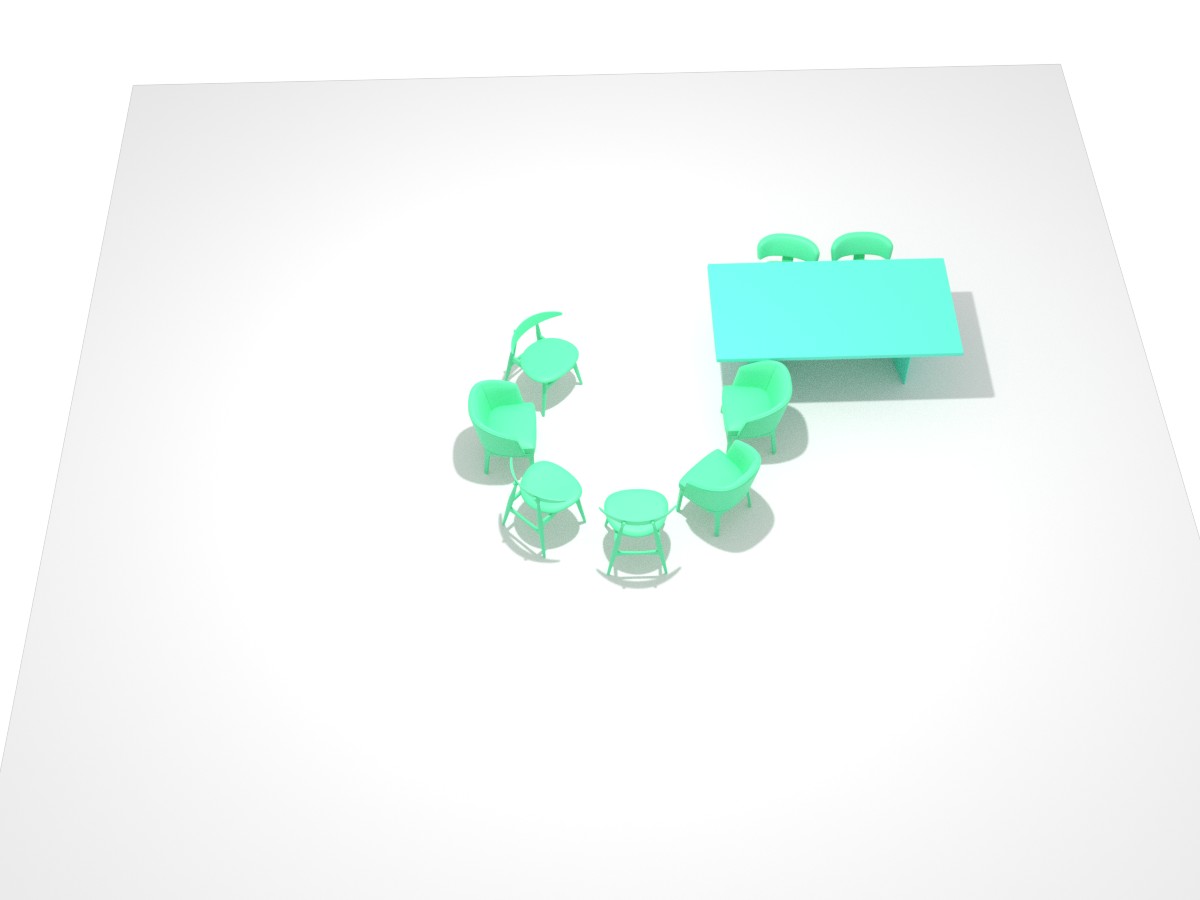}
    	\includegraphics[width=0.33\textwidth]{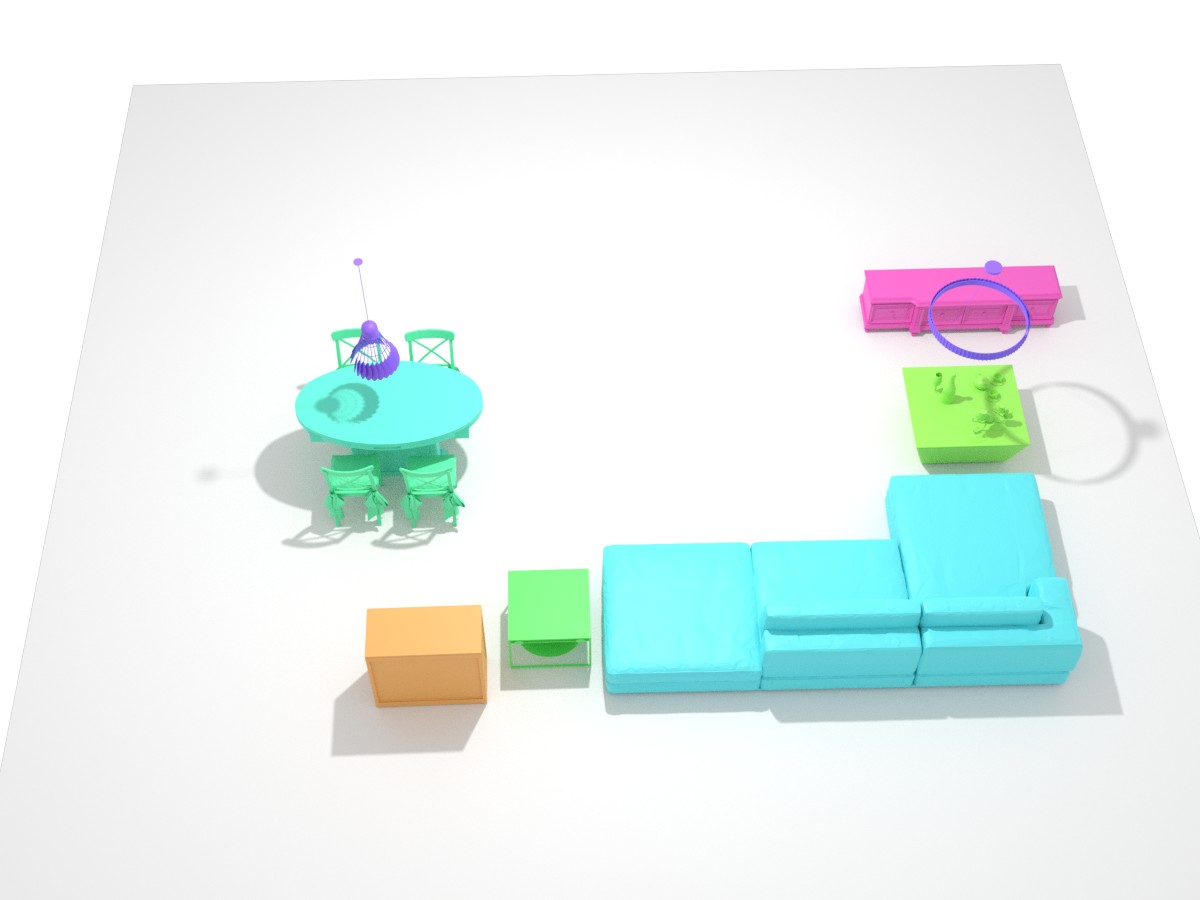}%
            \hfill
            \includegraphics[width=0.33\textwidth]{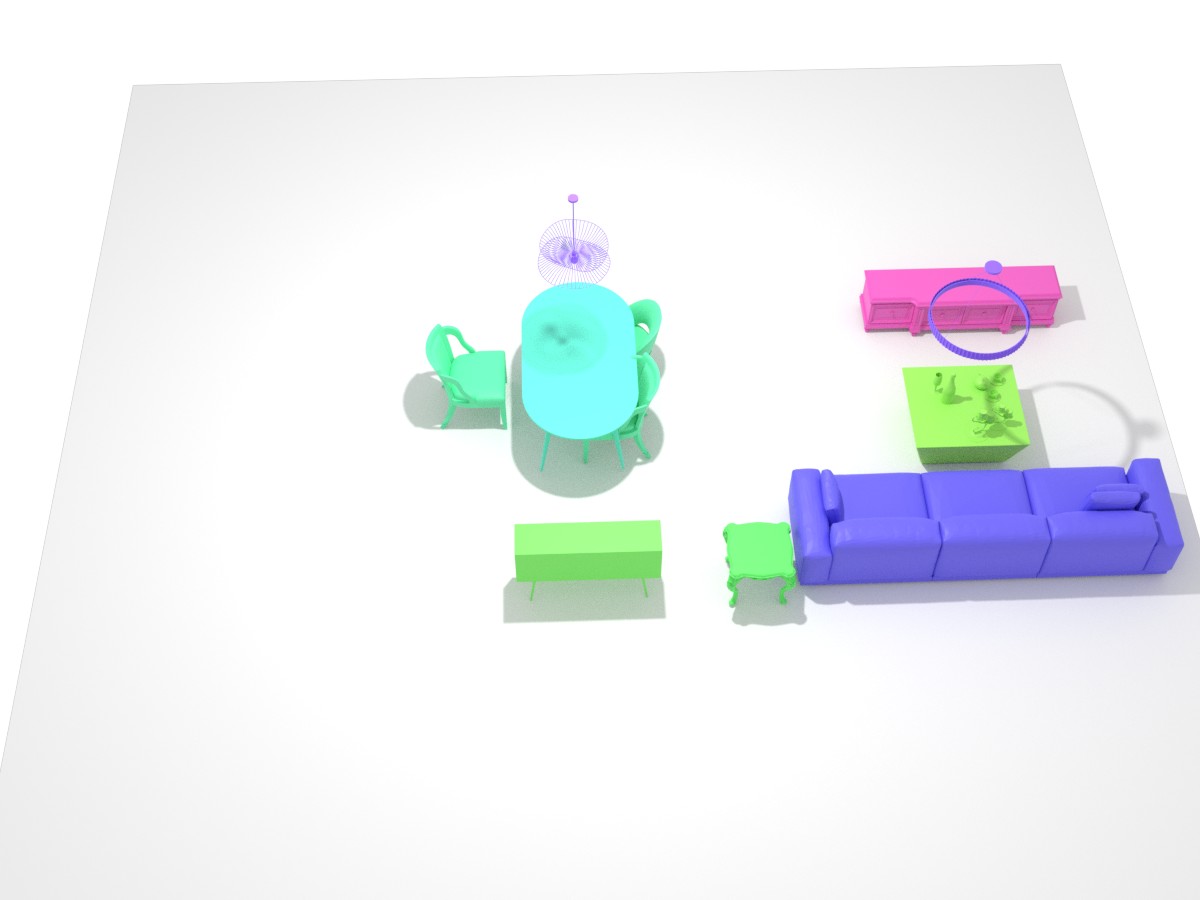}%
    	\hfill
    	\includegraphics[width=0.33\textwidth]{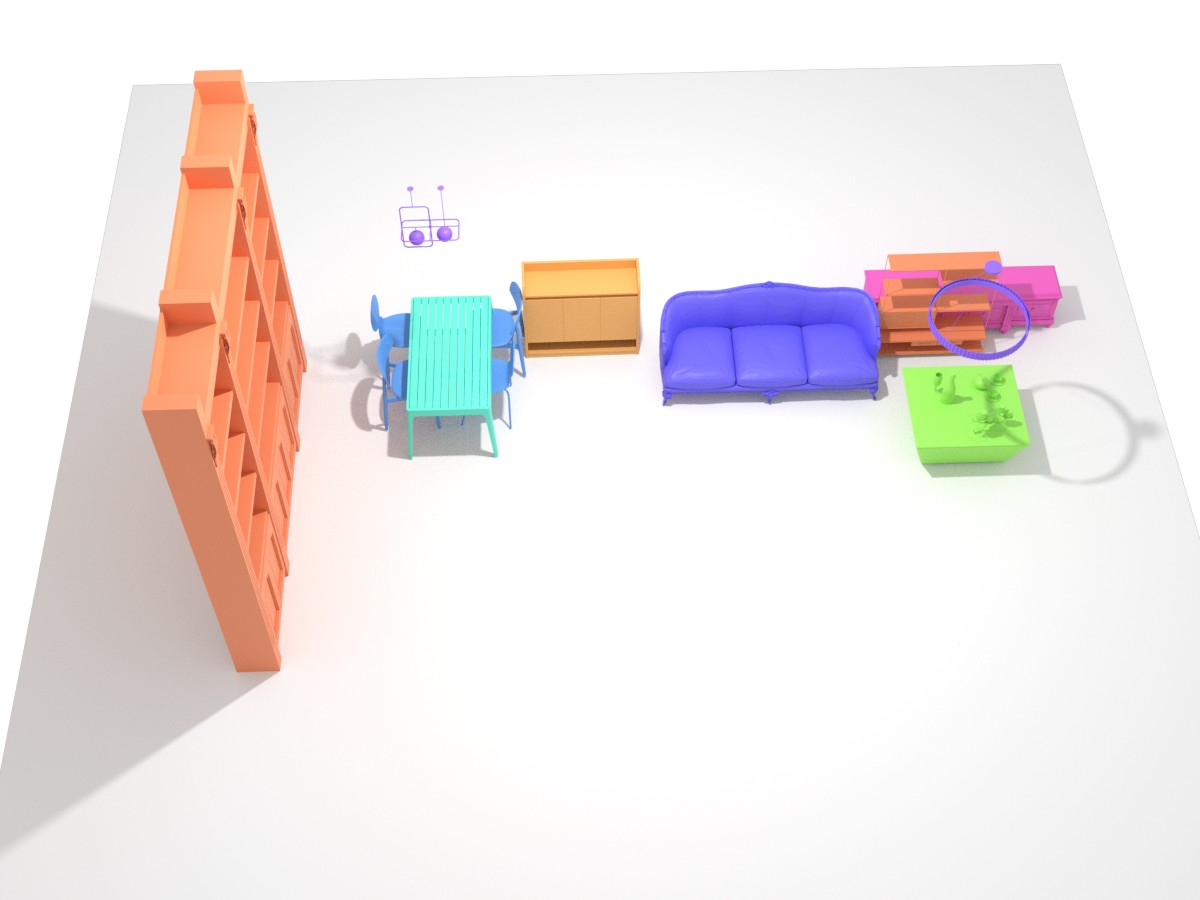}
		\caption{Ours}
	\end{subfigure}
	\caption{\textbf{Scene completion} from partial scenes with only three objects given as inputs. Compared to ATISS, our method produced more diverse completion results with higher fidelity.}
    \label{fig:completion_supple}
\end{figure*}
In Fig.~\ref{fig:uncond_comparison_supple}, we provide additional qualitative comparisons against state-of-the-art methods on the unconditional scene synthesis model. Also,  more visualization results of our unconditional scene synthesis model are presented in Fig.~\ref{fig:uncond_gallery}.

\begin{figure*}[b]
	\centering
 	\begin{subfigure}[t]{0.23\textwidth}
            \includegraphics[width=\textwidth]{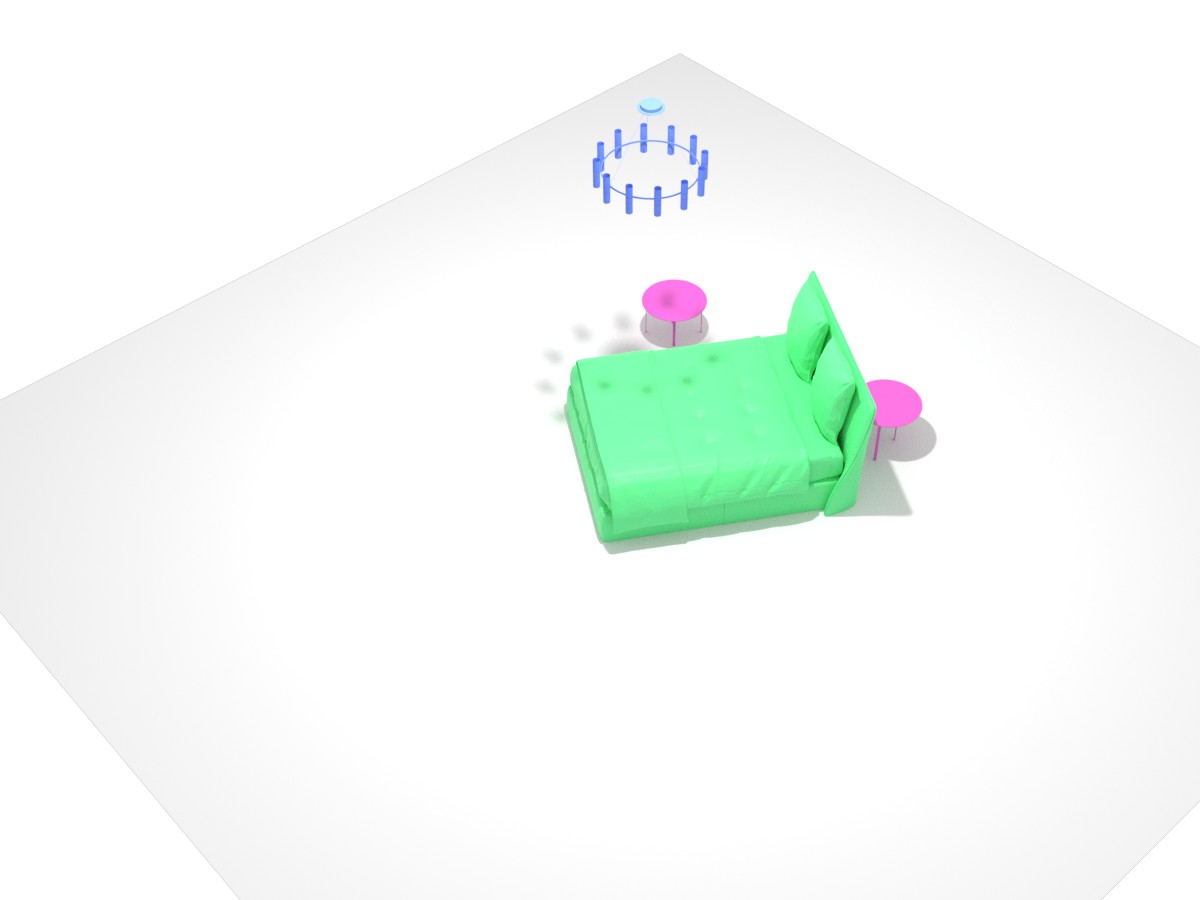}
            \includegraphics[width=\textwidth]{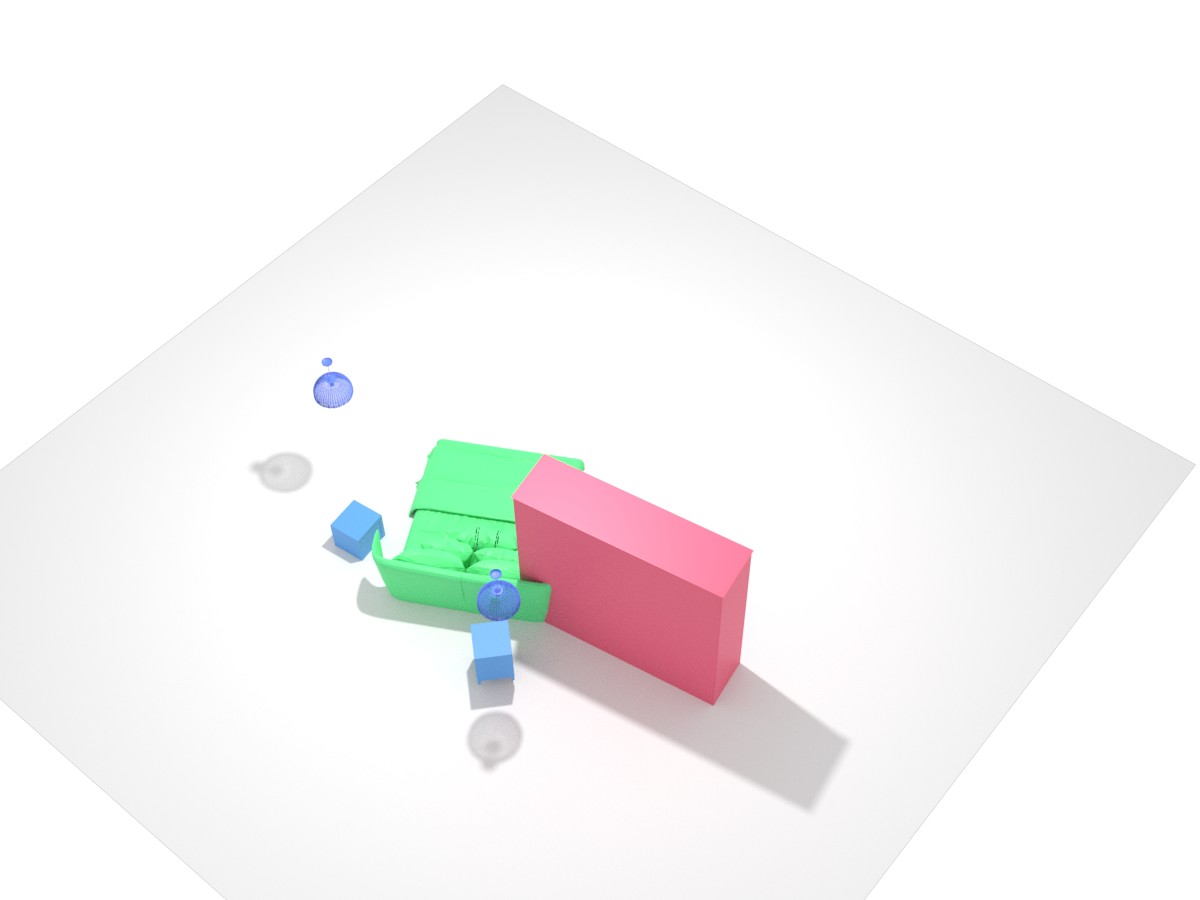}
            \includegraphics[width=\textwidth]{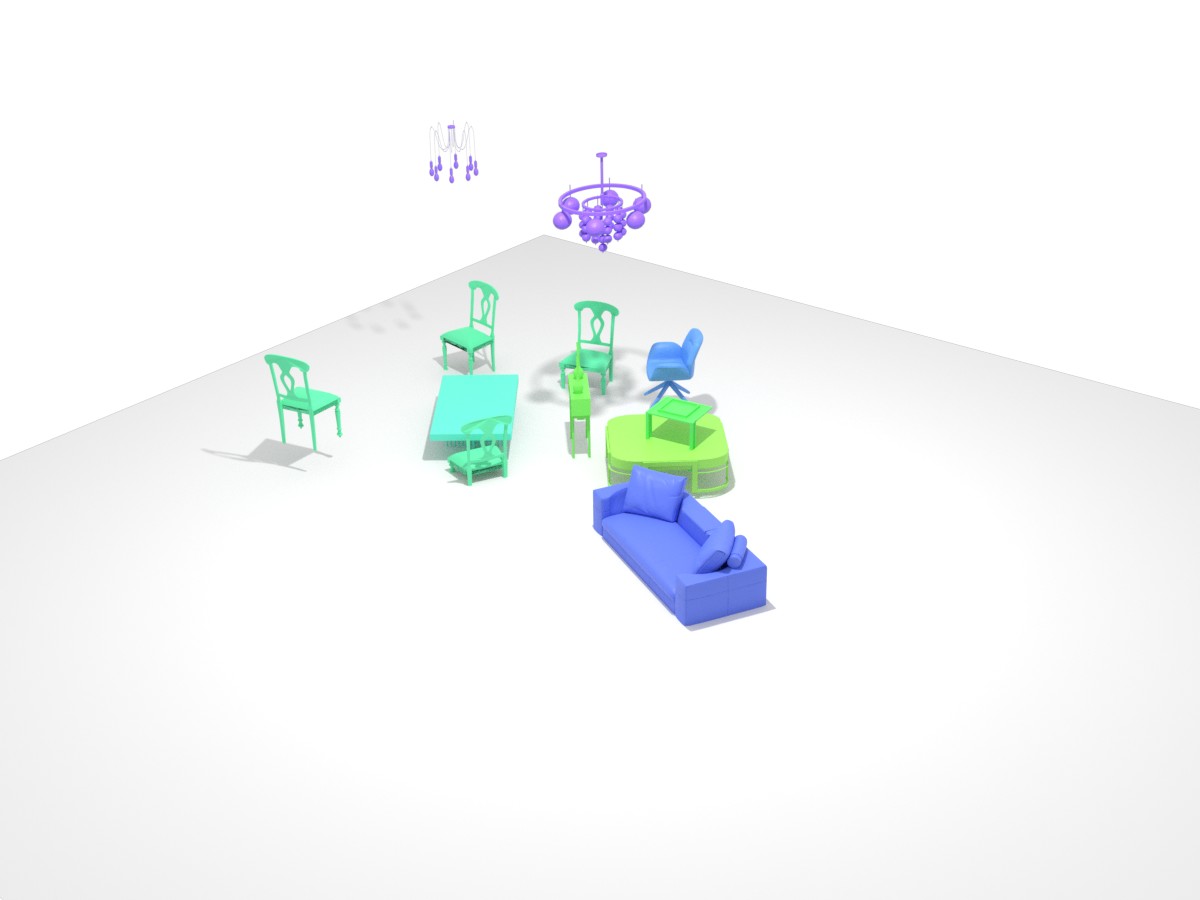}
            \includegraphics[width=\textwidth]{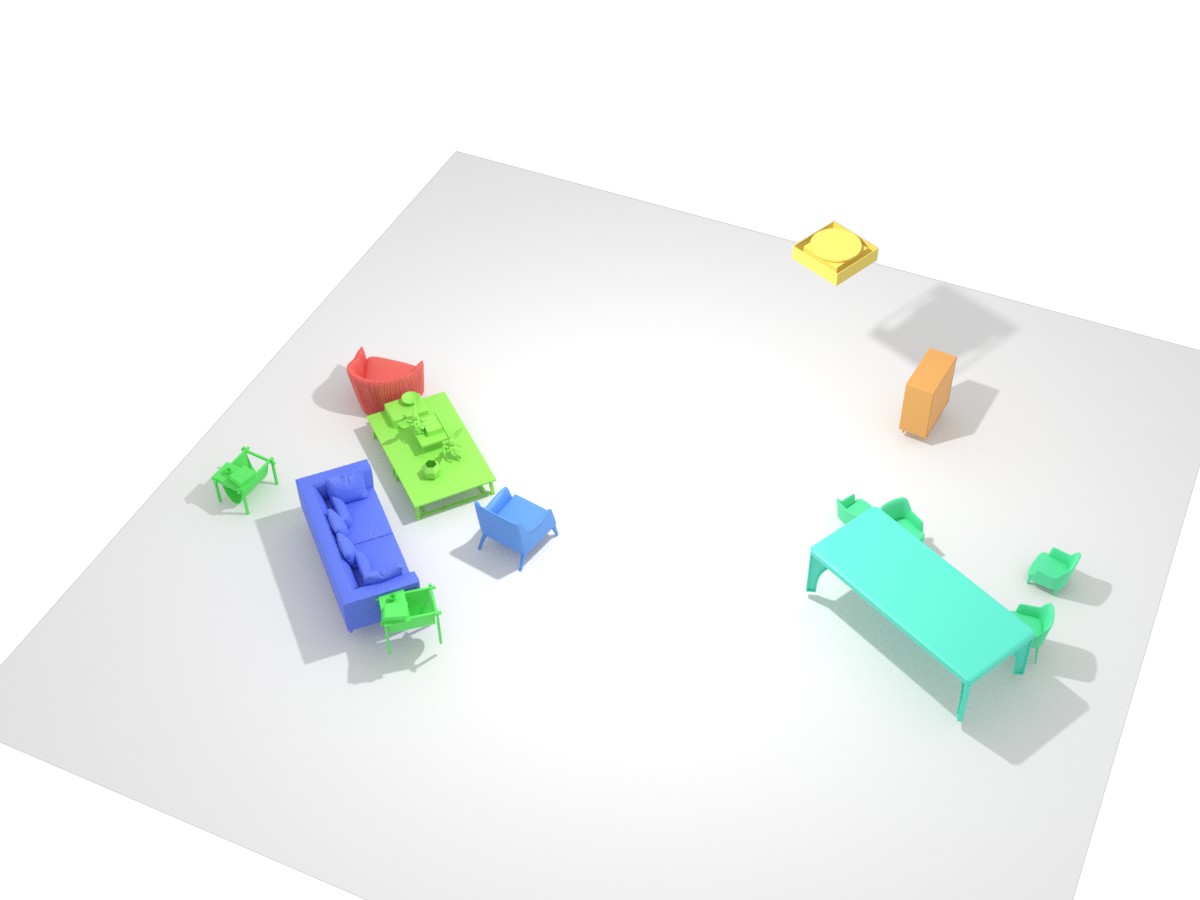}
            \includegraphics[width=\textwidth]{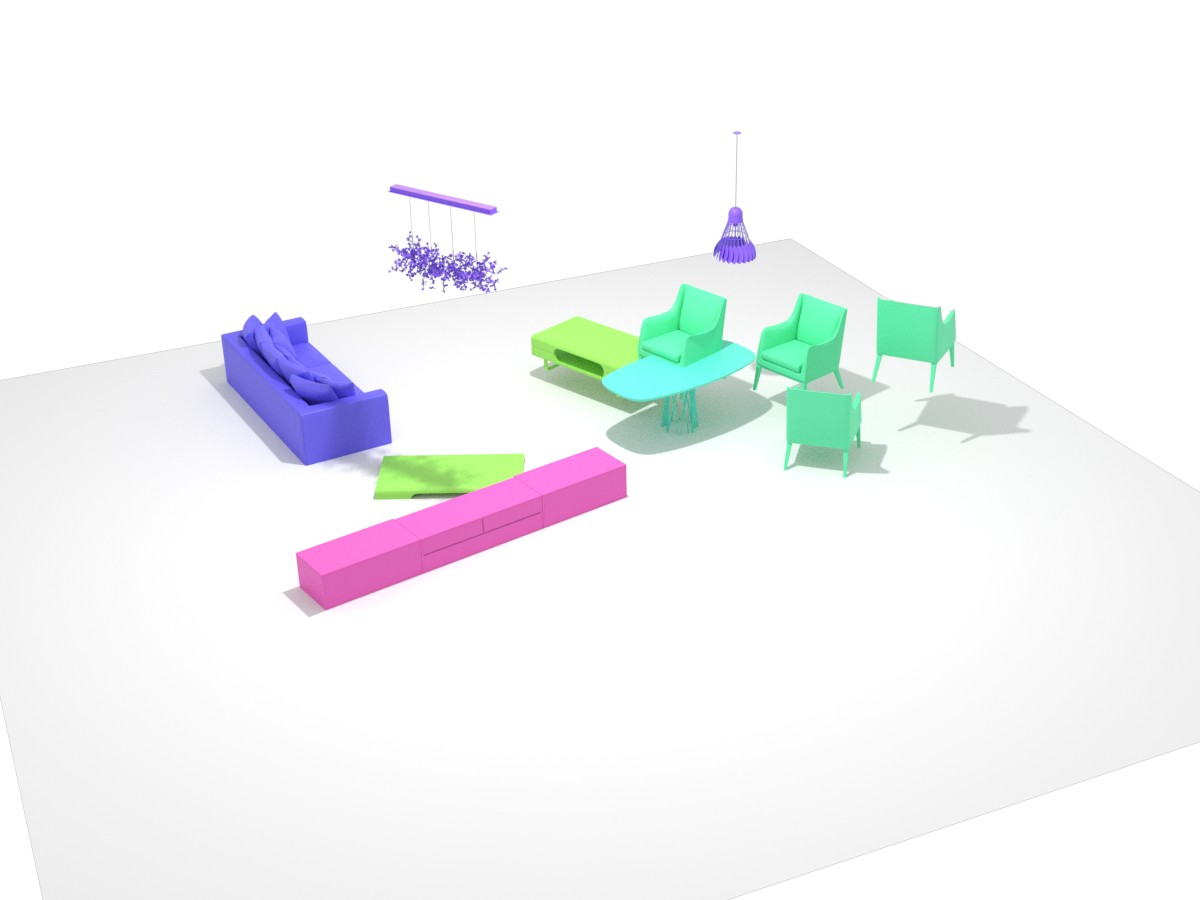}
            \includegraphics[width=\textwidth]{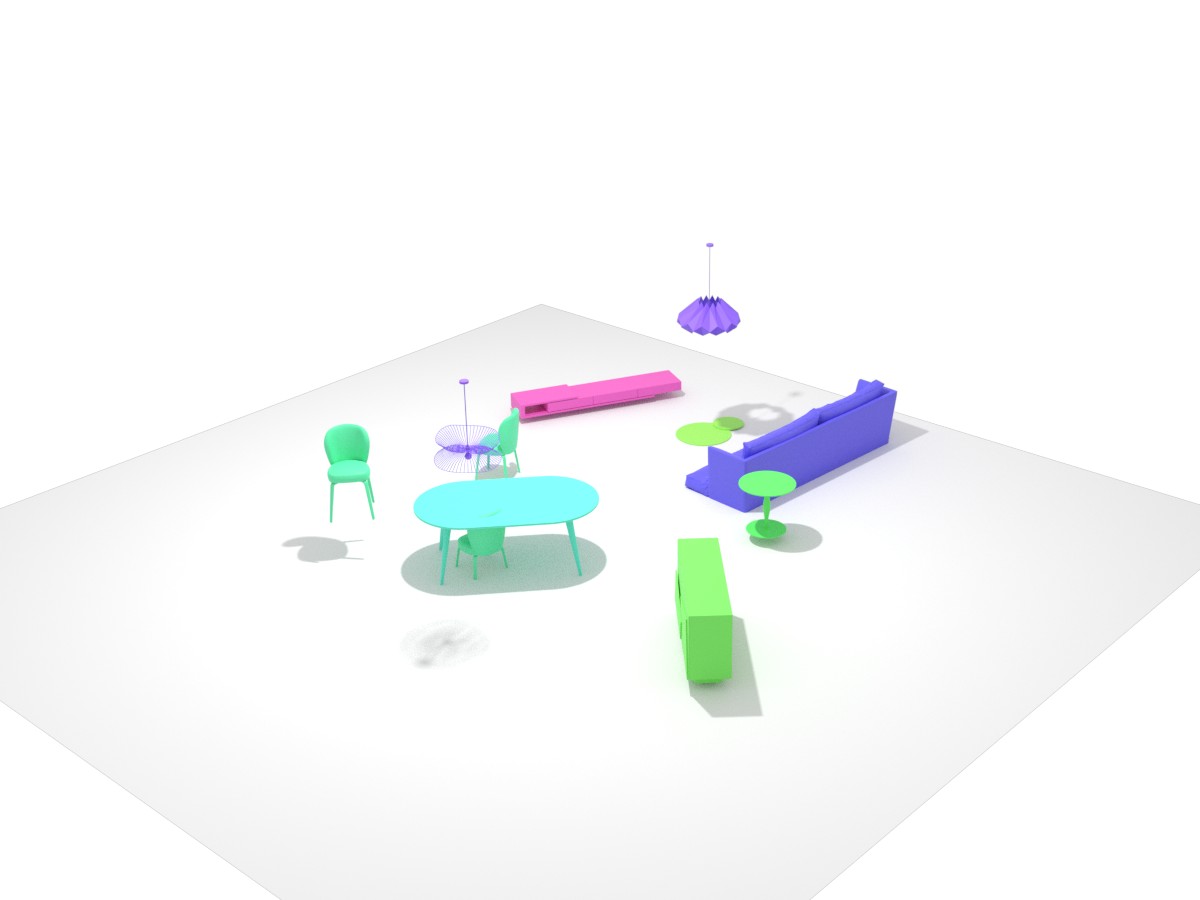}
		\caption{Noisy Scene}
	\end{subfigure}
	\rulesep
	\begin{subfigure}[t]{0.23\textwidth}
            \includegraphics[width=\textwidth]{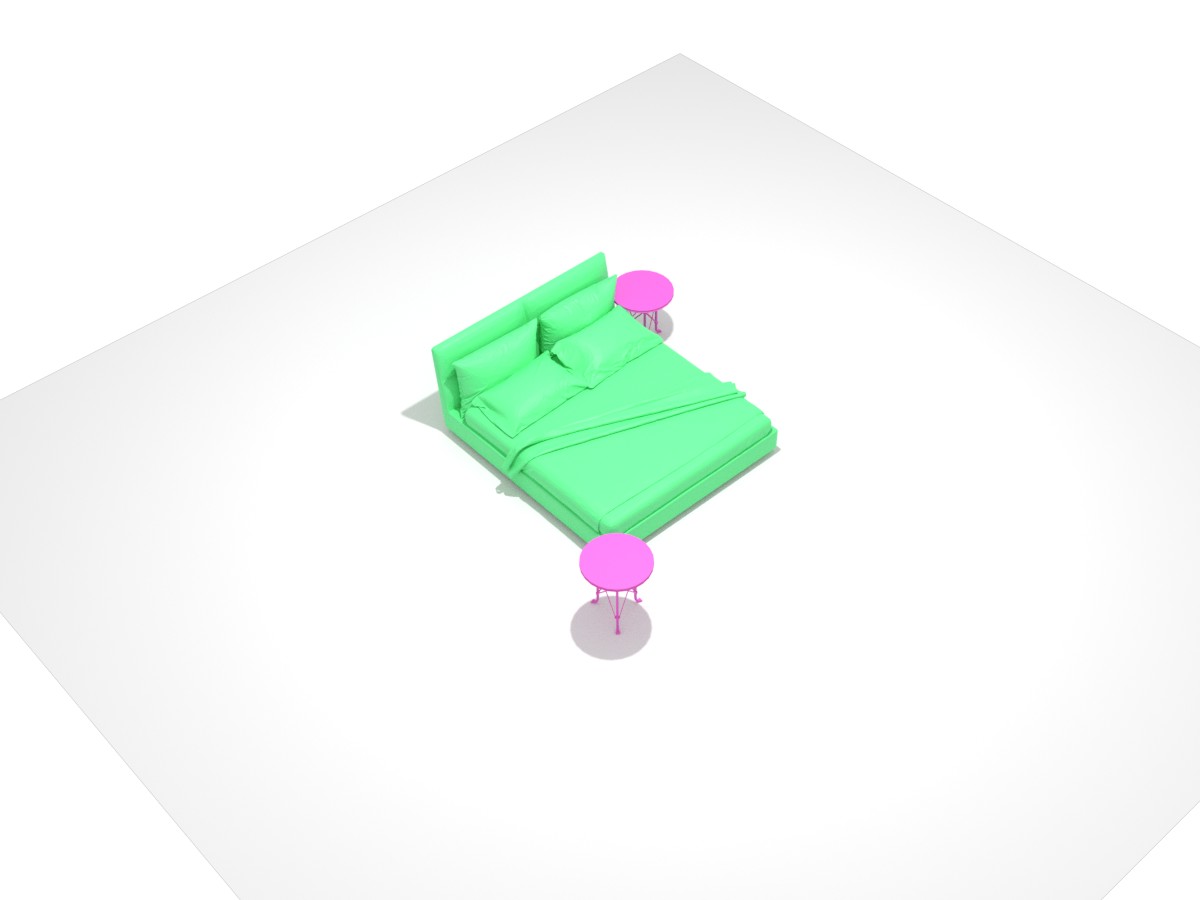}
            \includegraphics[width=\textwidth]{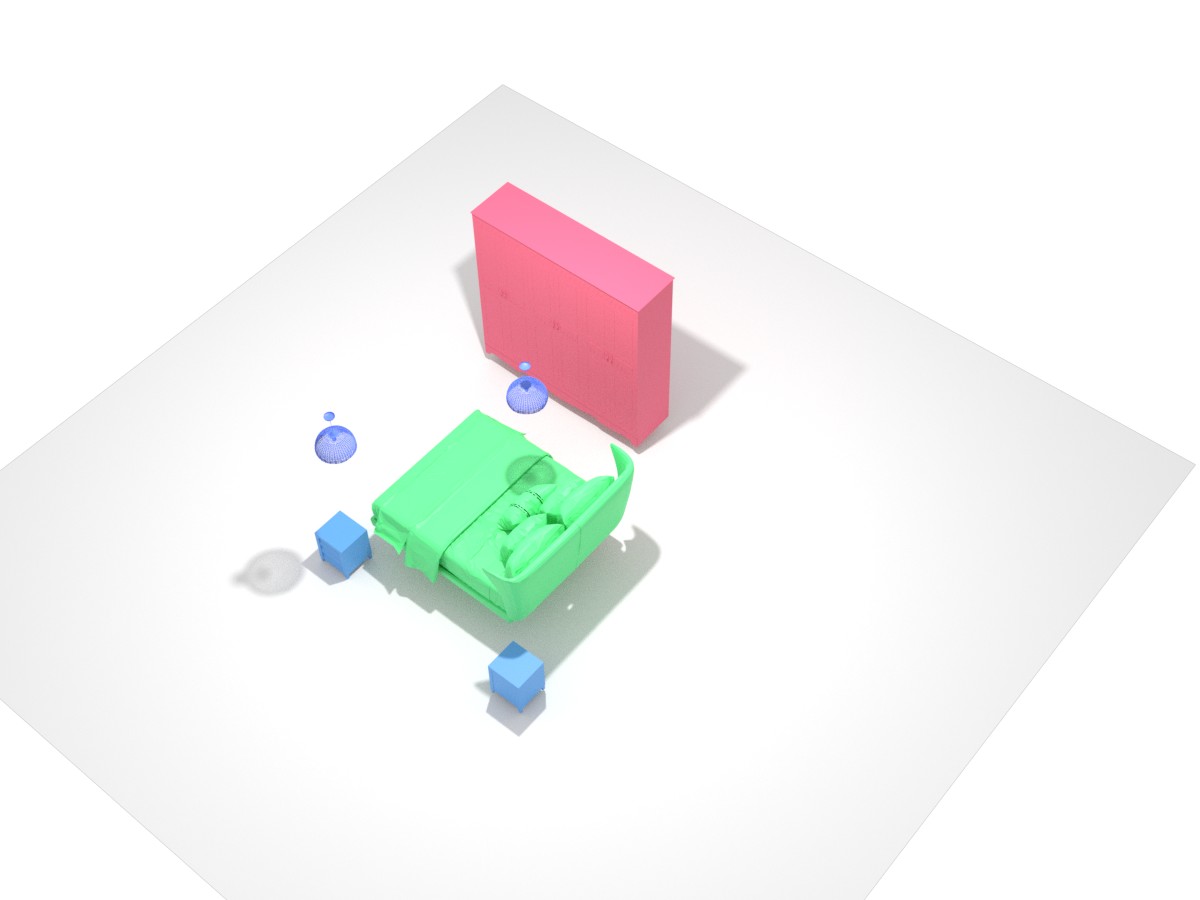}
		\includegraphics[width=\textwidth]{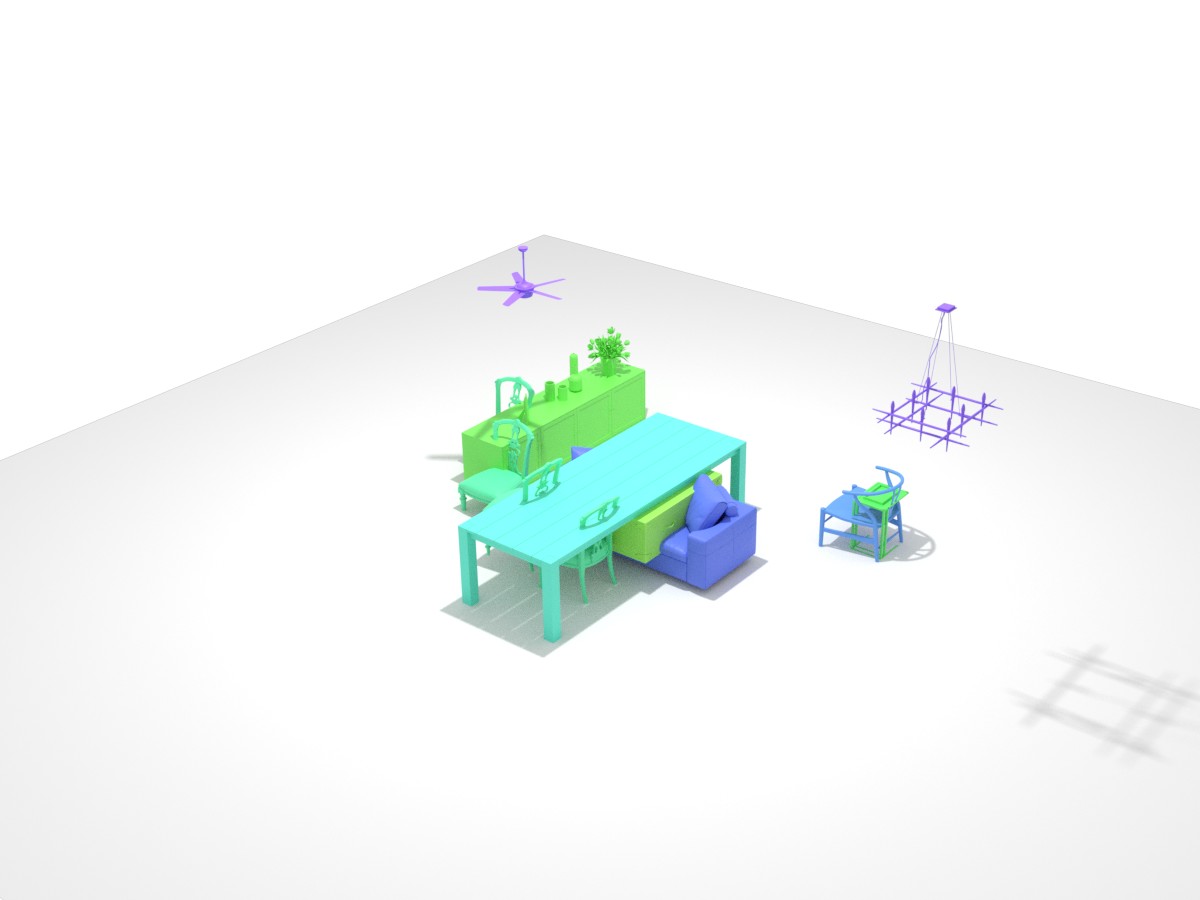}
            \includegraphics[width=\textwidth]{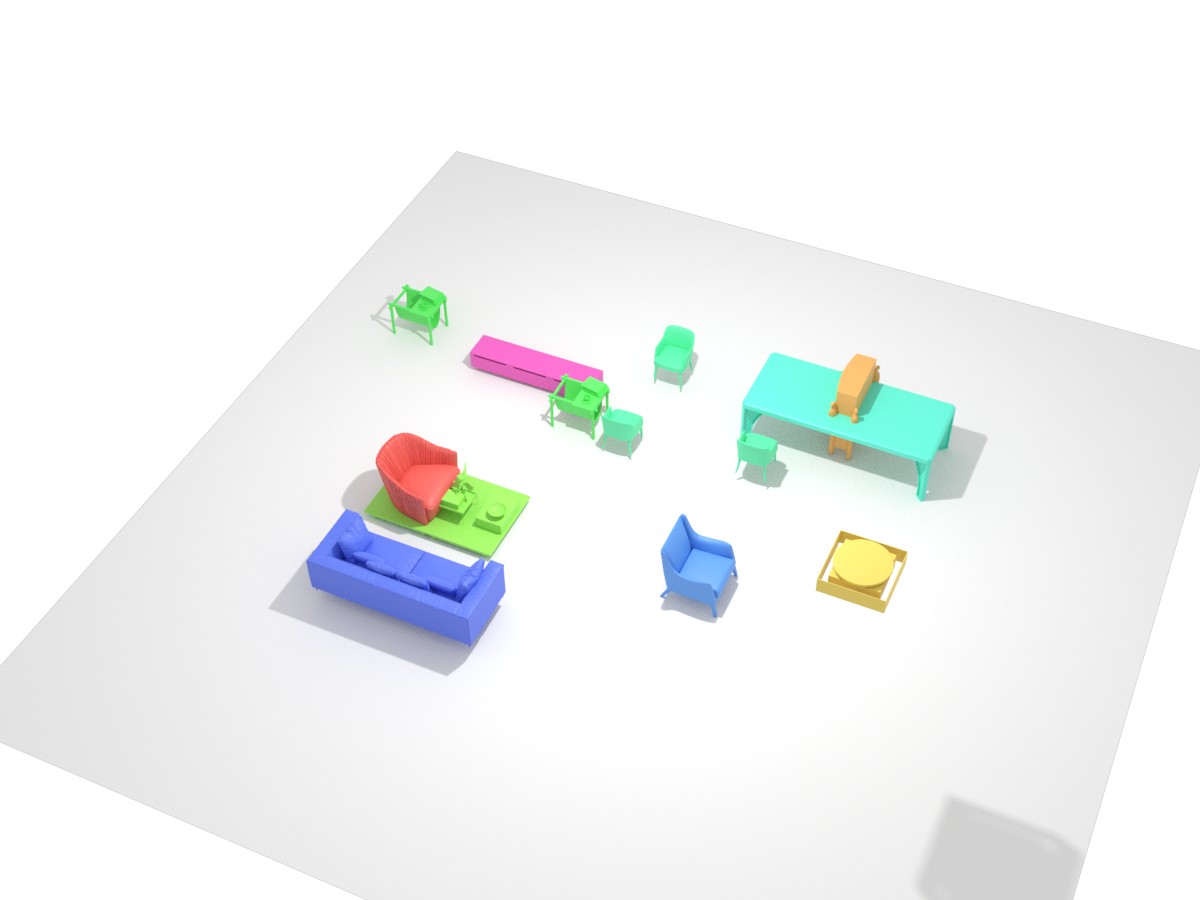}
            \includegraphics[width=\textwidth]{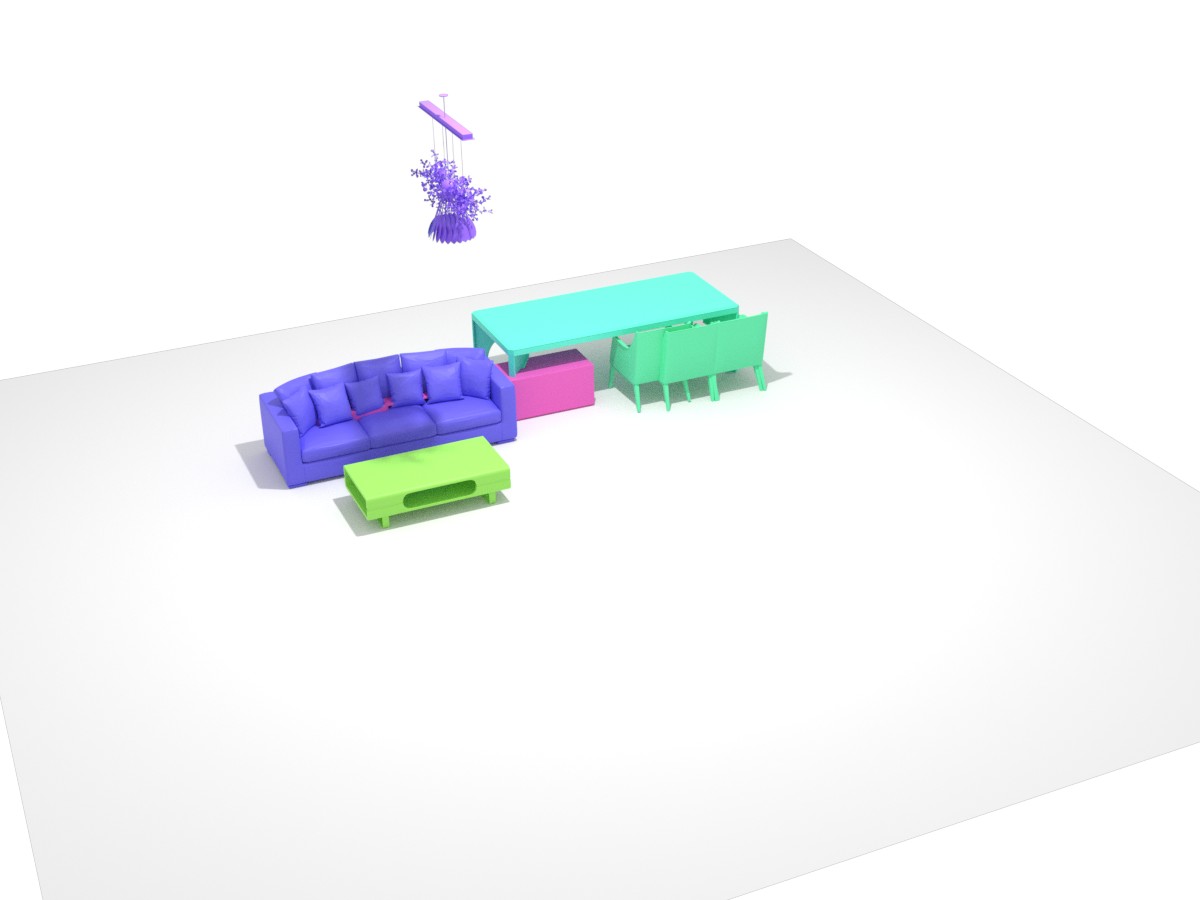}
            \includegraphics[width=\textwidth]{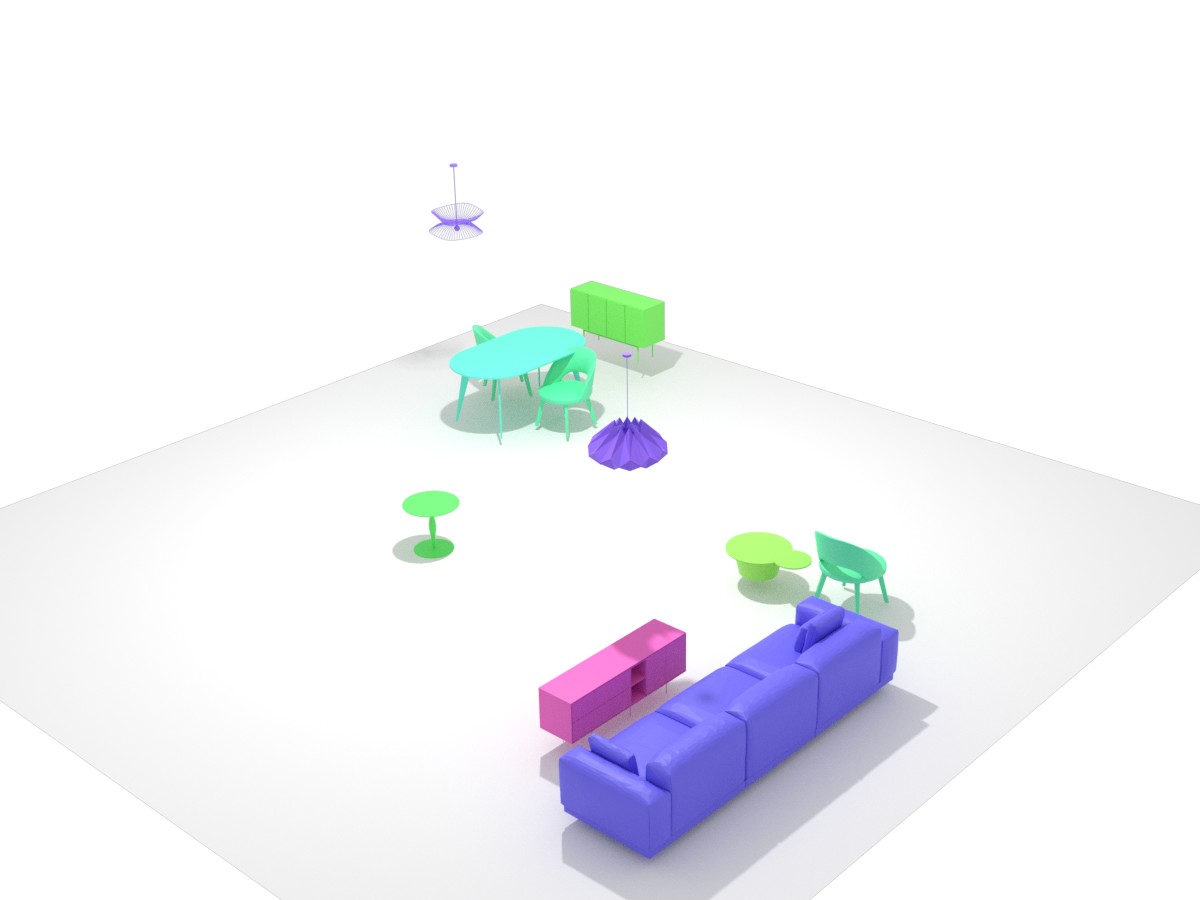}
        \caption{ATISS~\cite{paschalidou2021atiss}}
	\end{subfigure}
	\rulesep
        \begin{subfigure}[t]{0.23\textwidth}
            \includegraphics[width=\textwidth]{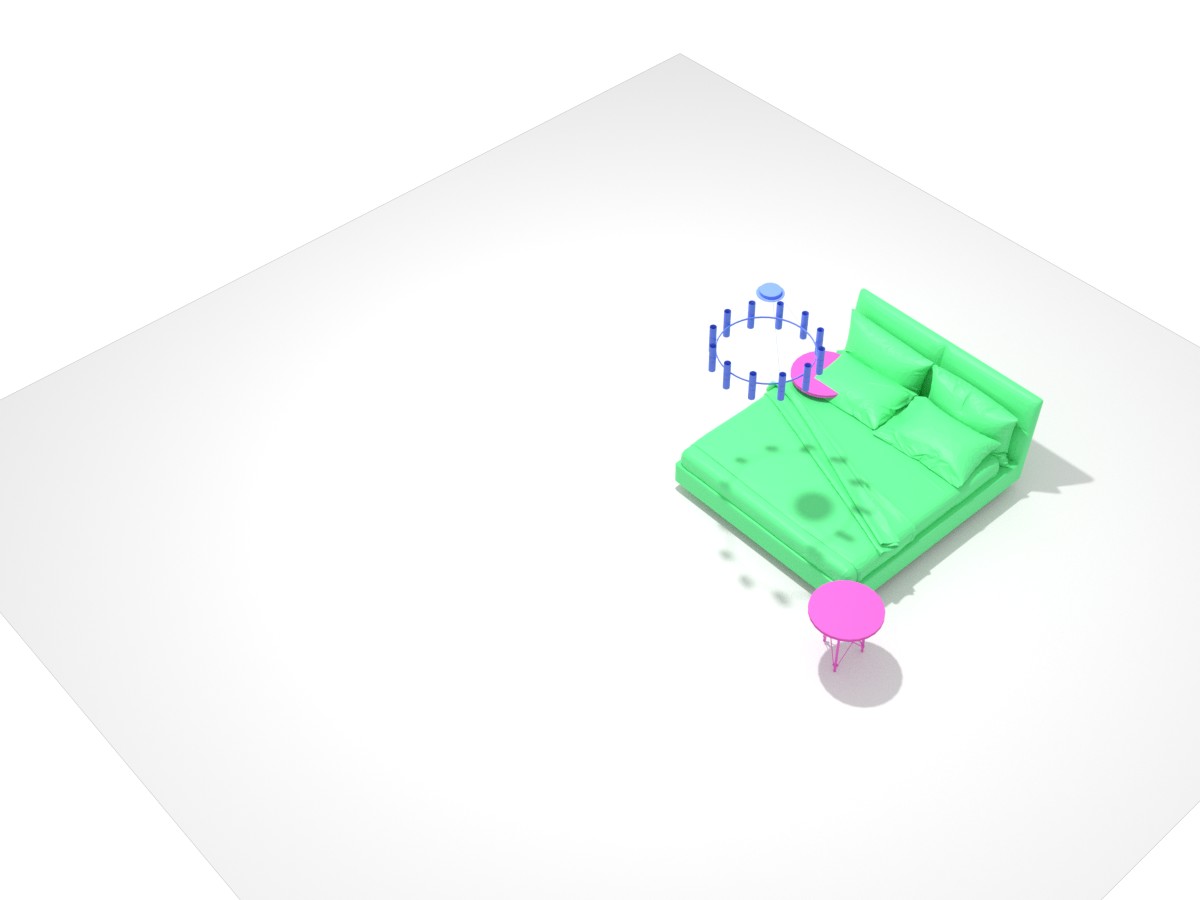}
		\includegraphics[width=\textwidth]{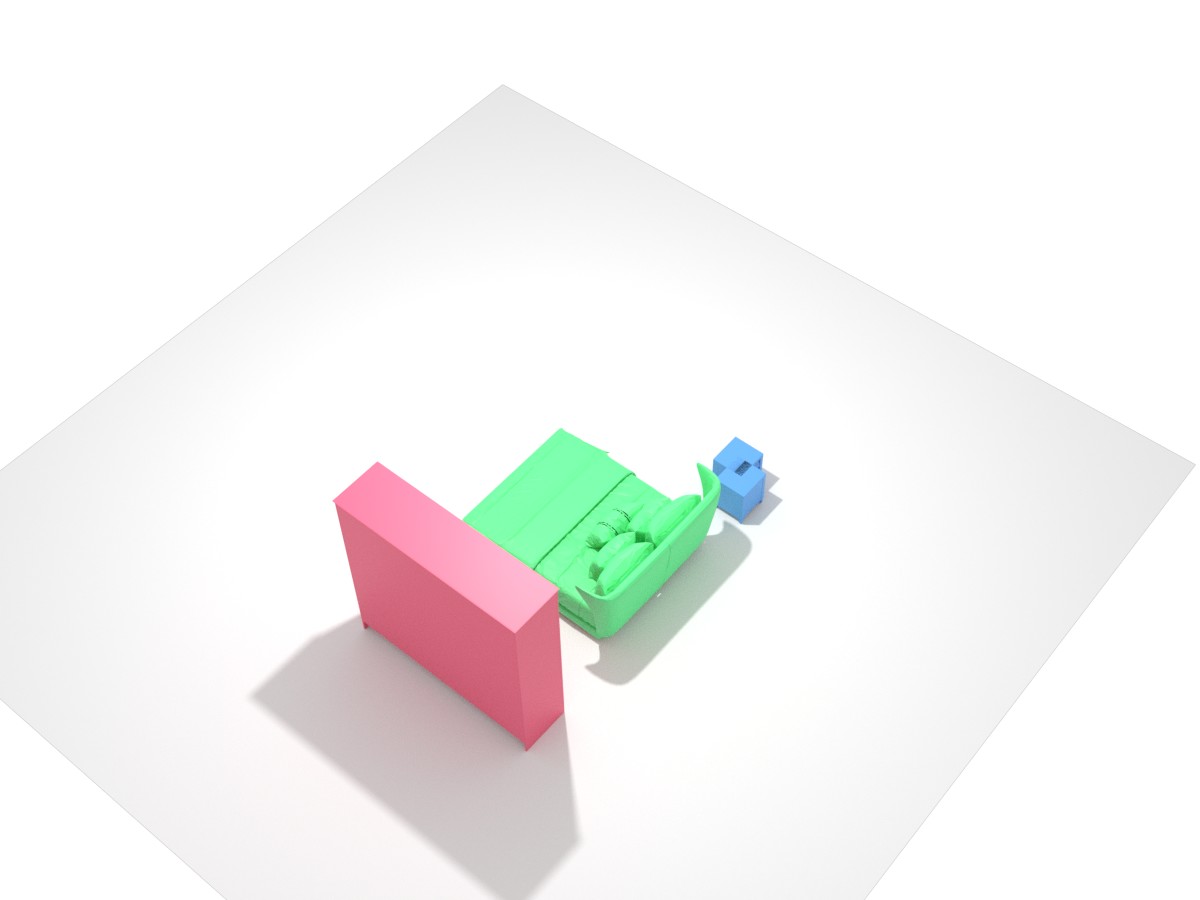}
            \includegraphics[width=\textwidth]{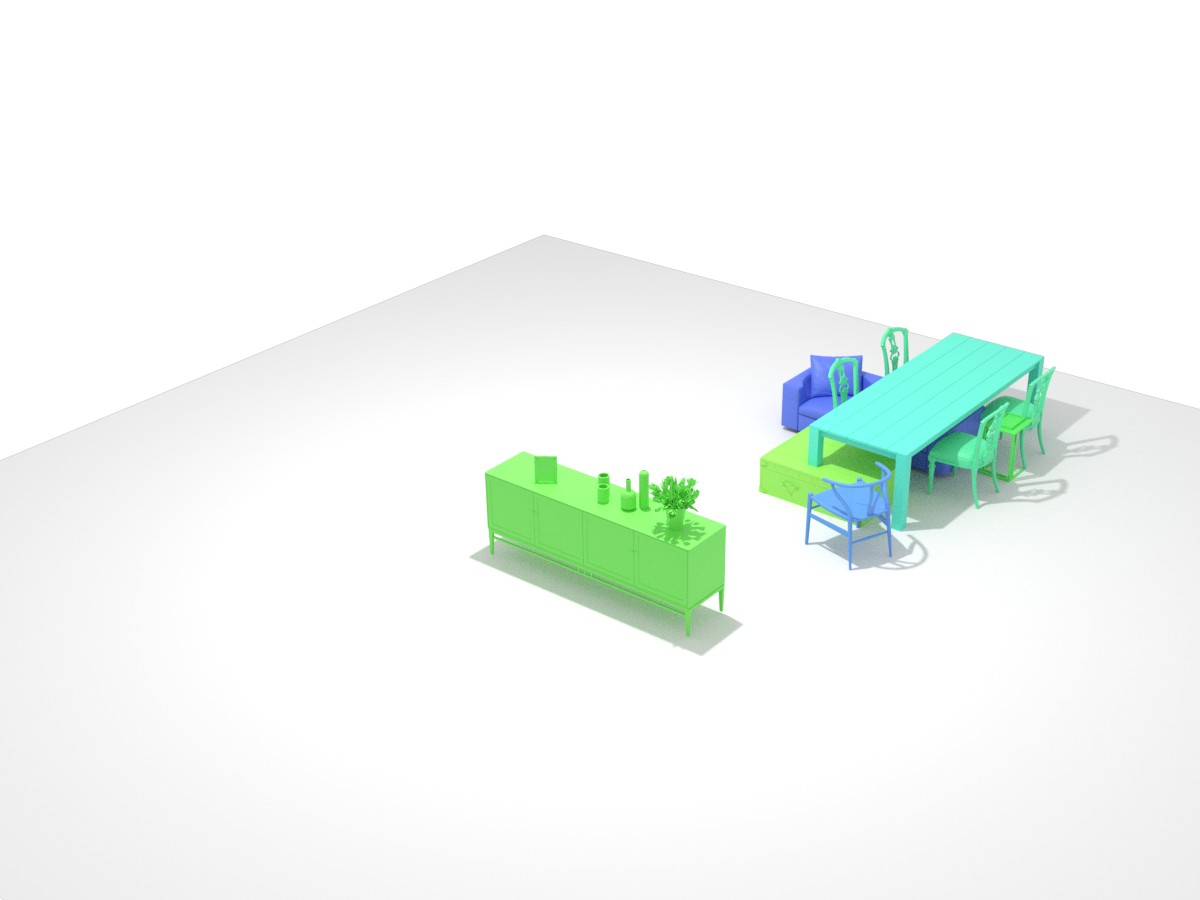}
            \includegraphics[width=\textwidth]{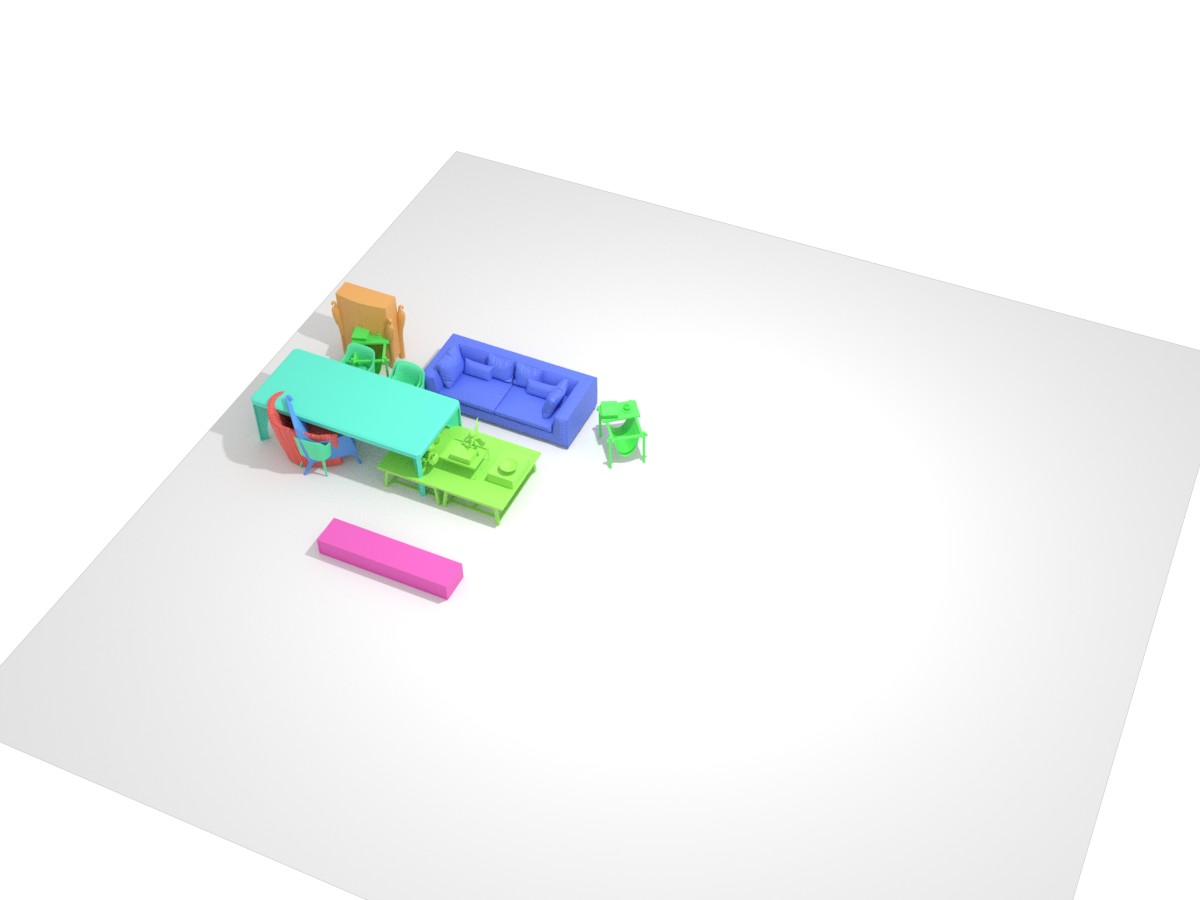}
            \includegraphics[width=\textwidth]{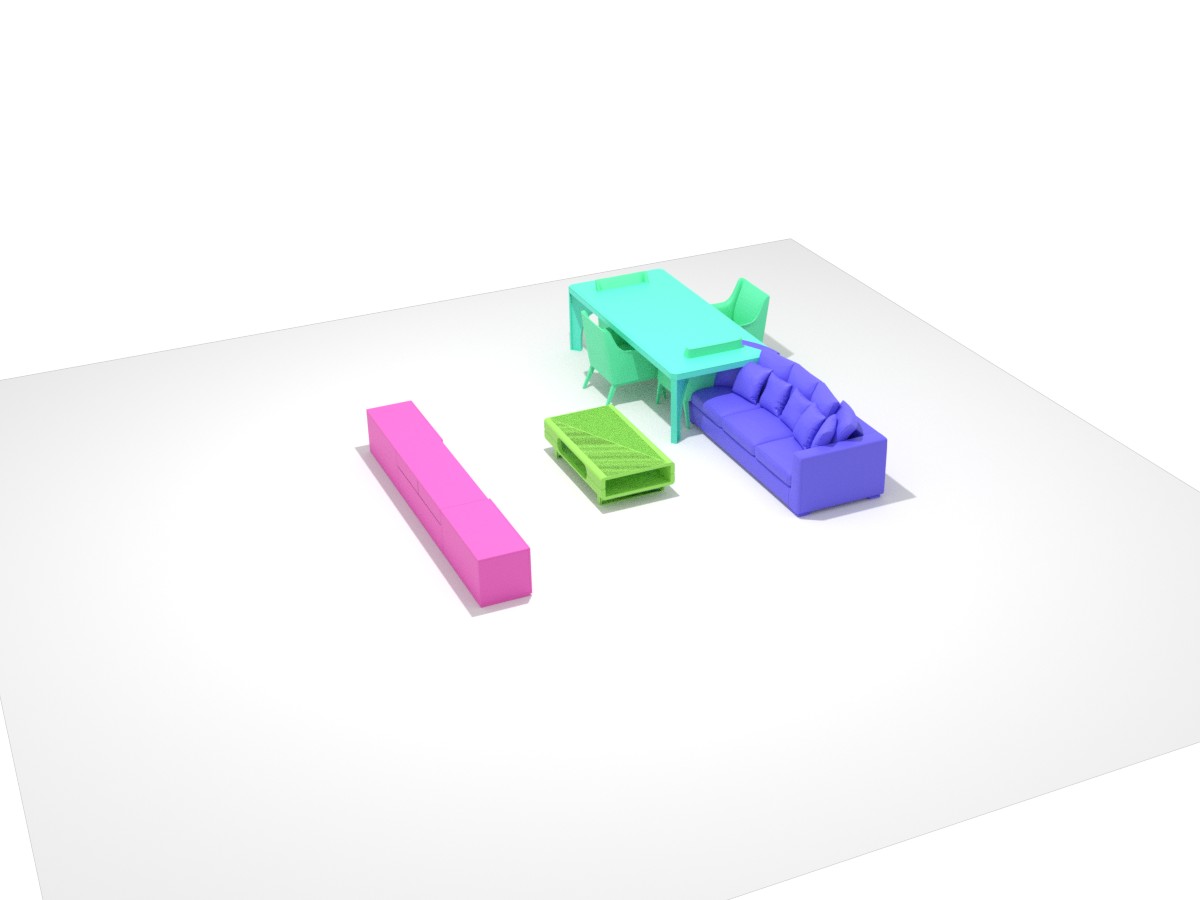}
            \includegraphics[width=\textwidth]{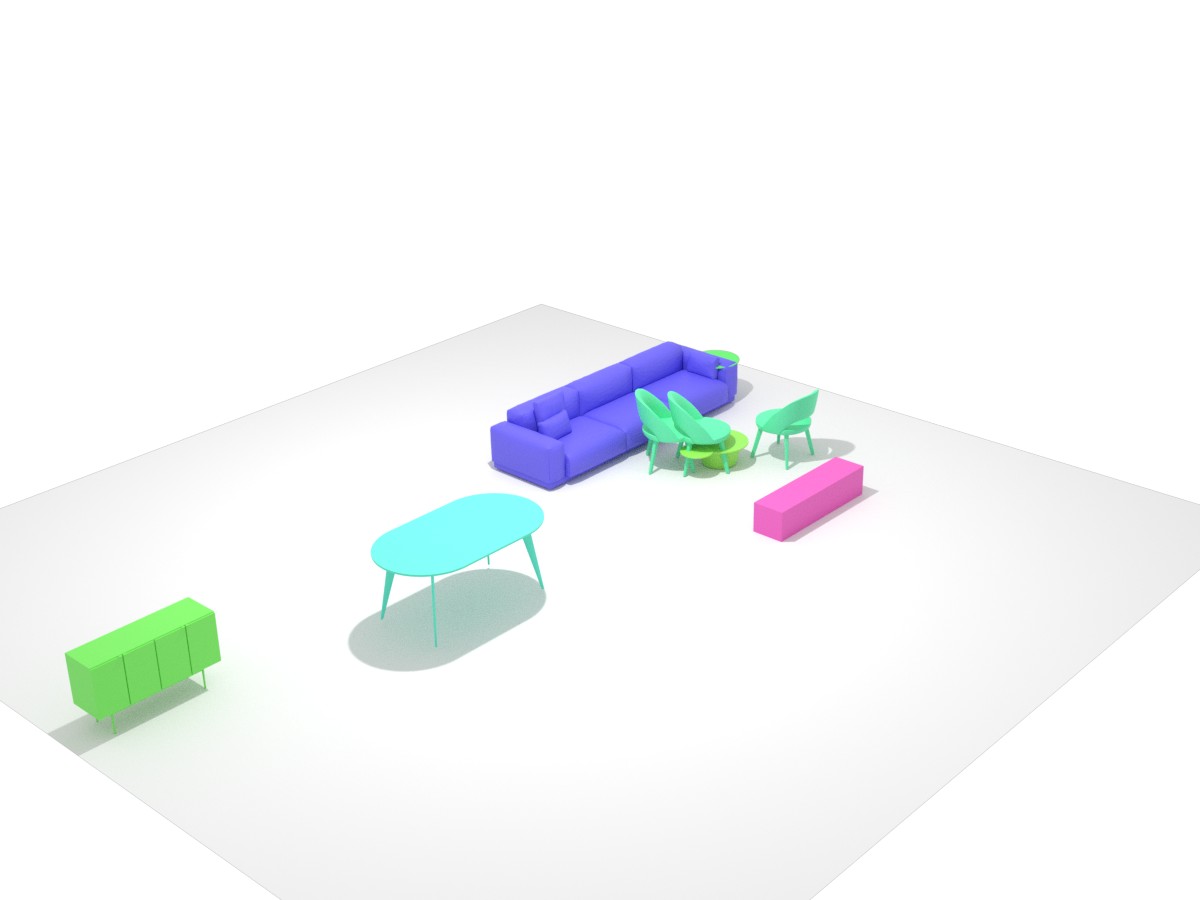}
		\caption{LEGO~\cite{wei2023lego}}
	\end{subfigure}
	\rulesep
	\begin{subfigure}[t]{0.23\textwidth}
            \includegraphics[width=\textwidth]{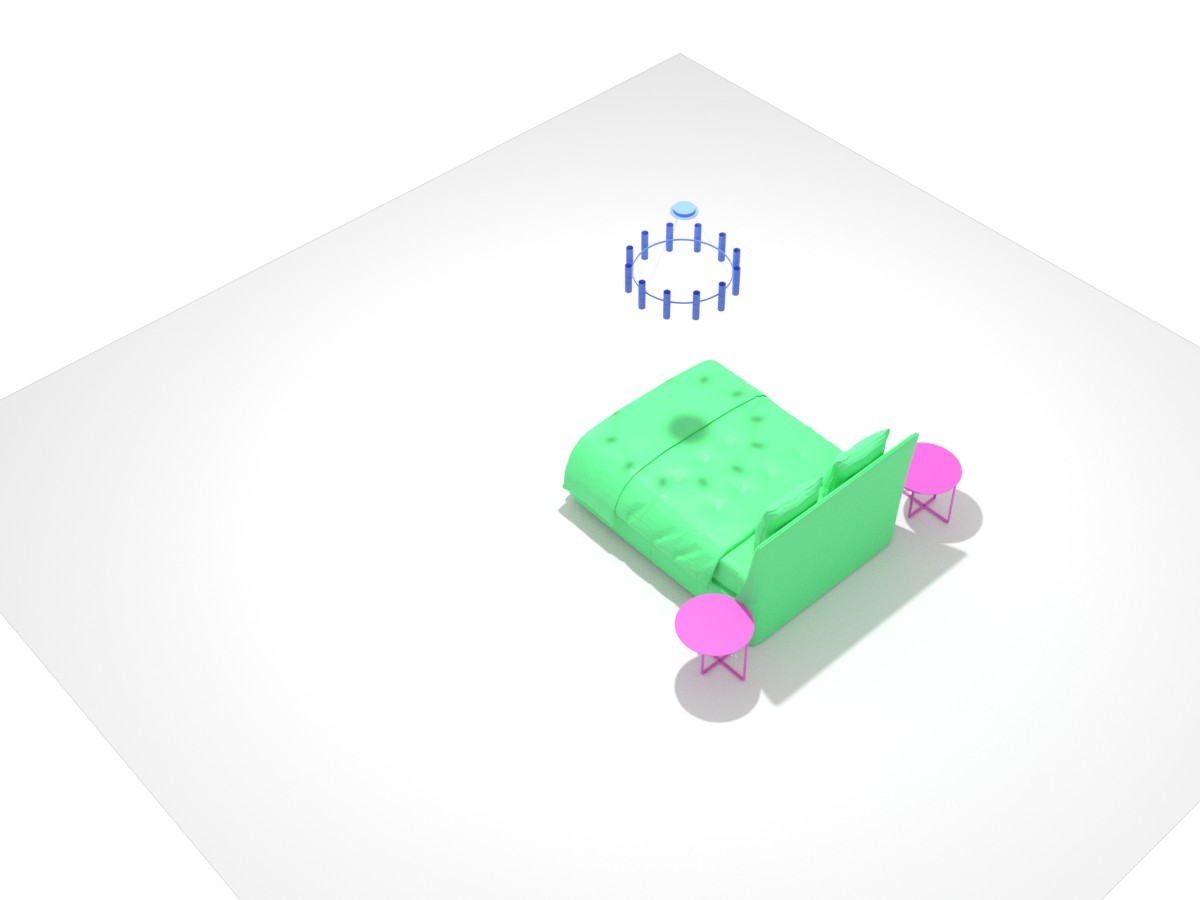}
        \includegraphics[width=\textwidth]{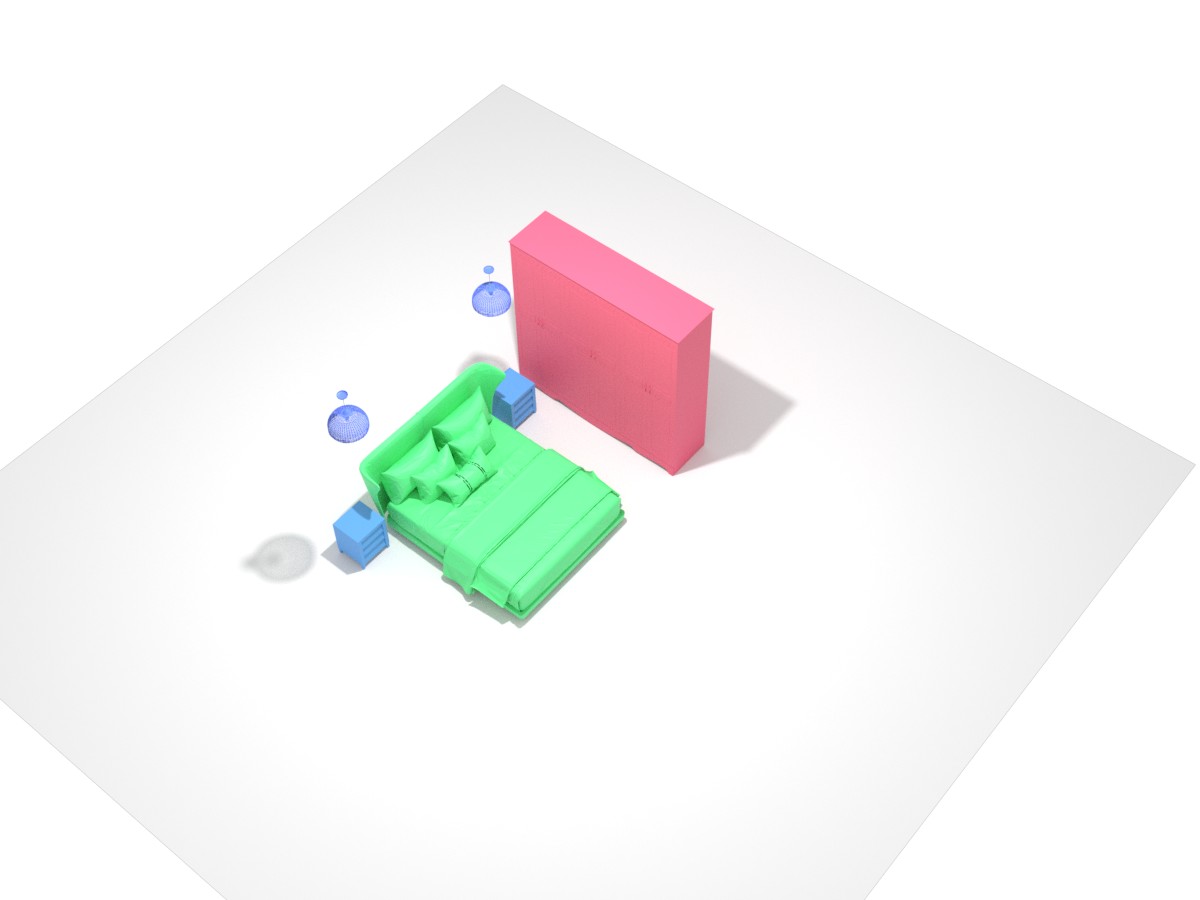}
        \includegraphics[width=\textwidth]{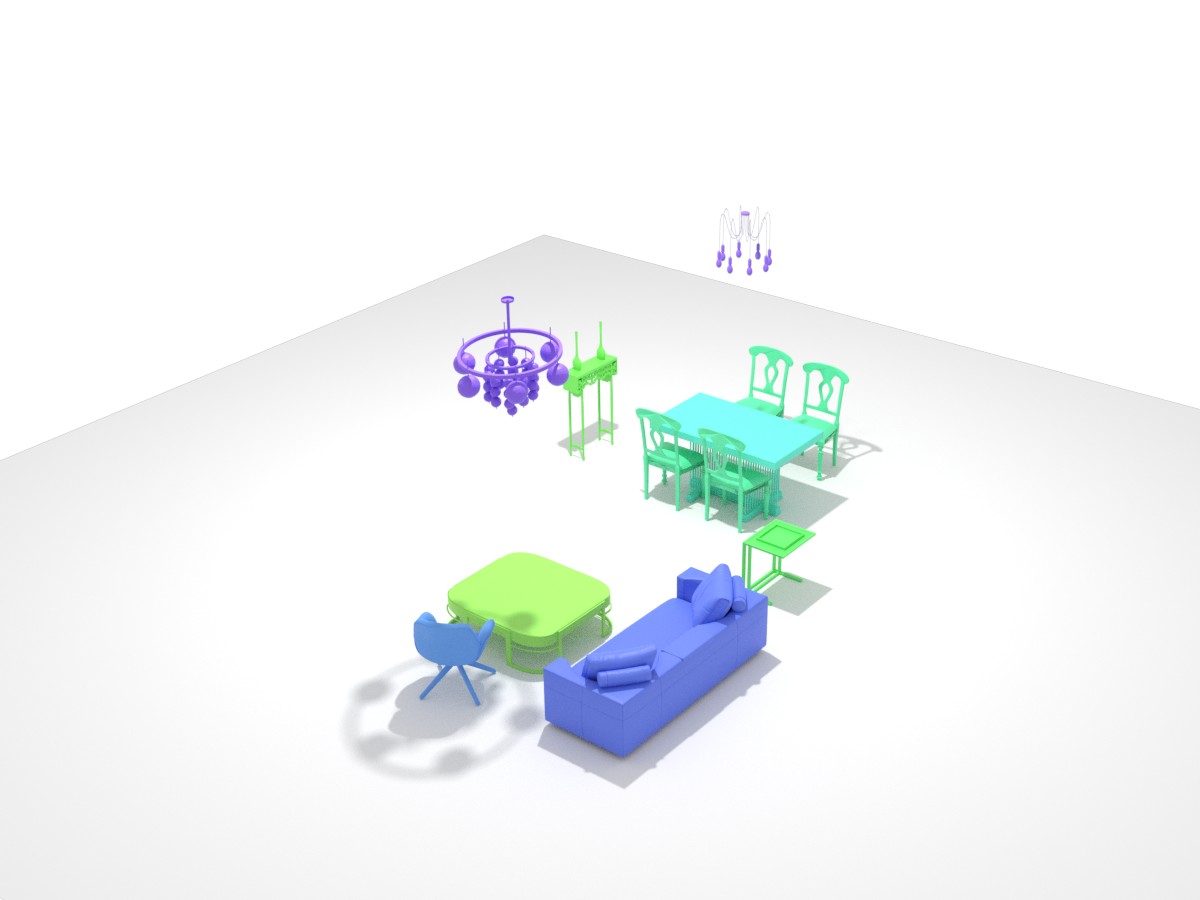}
		\includegraphics[width=\textwidth]{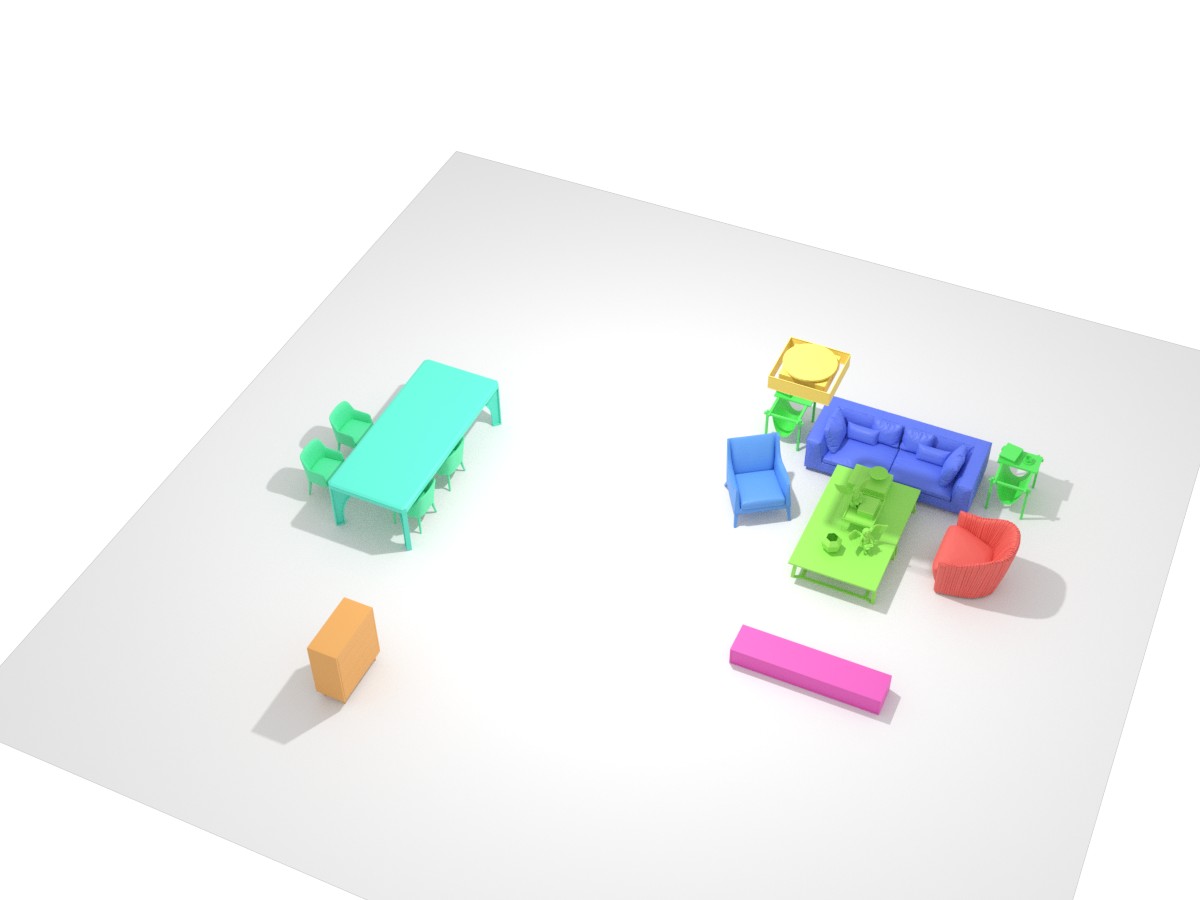}
        \includegraphics[width=\textwidth]{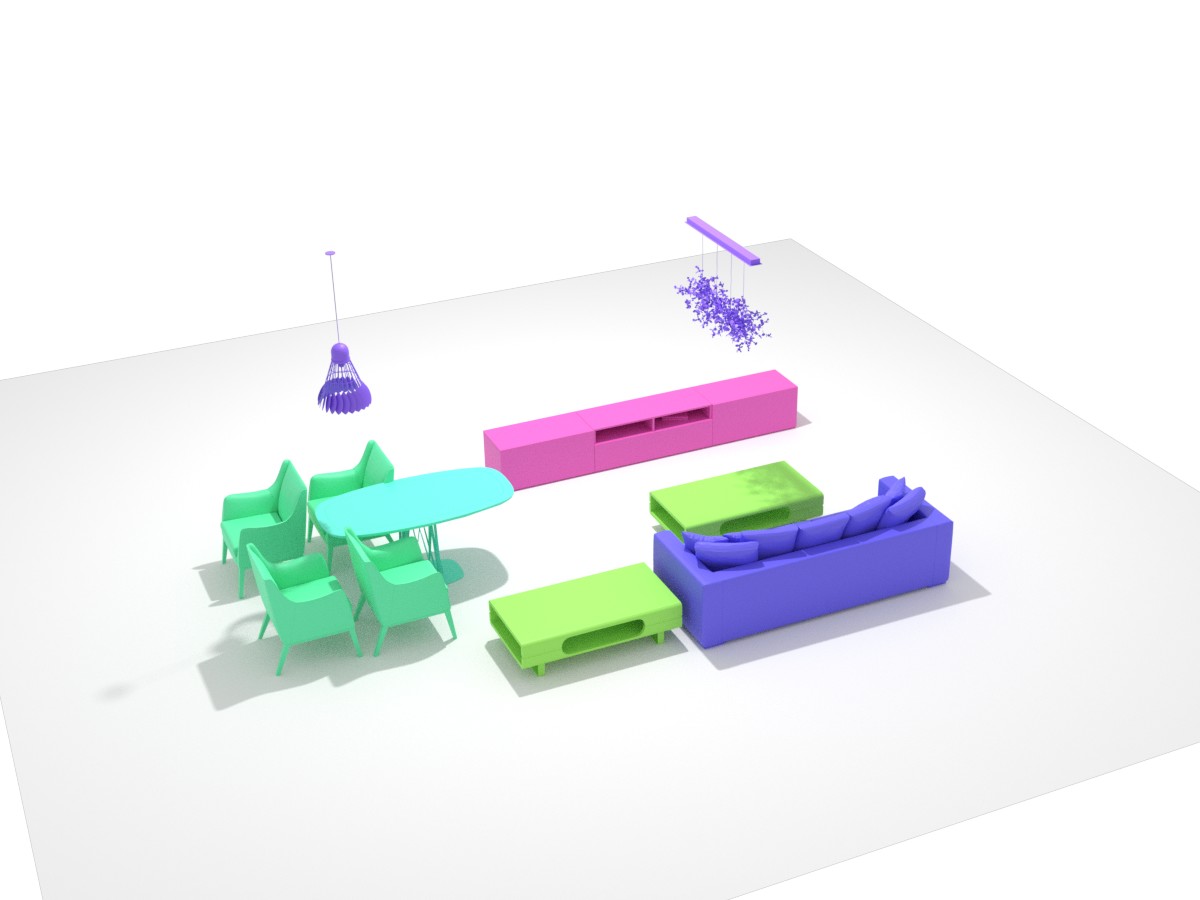}
		\includegraphics[width=\textwidth]{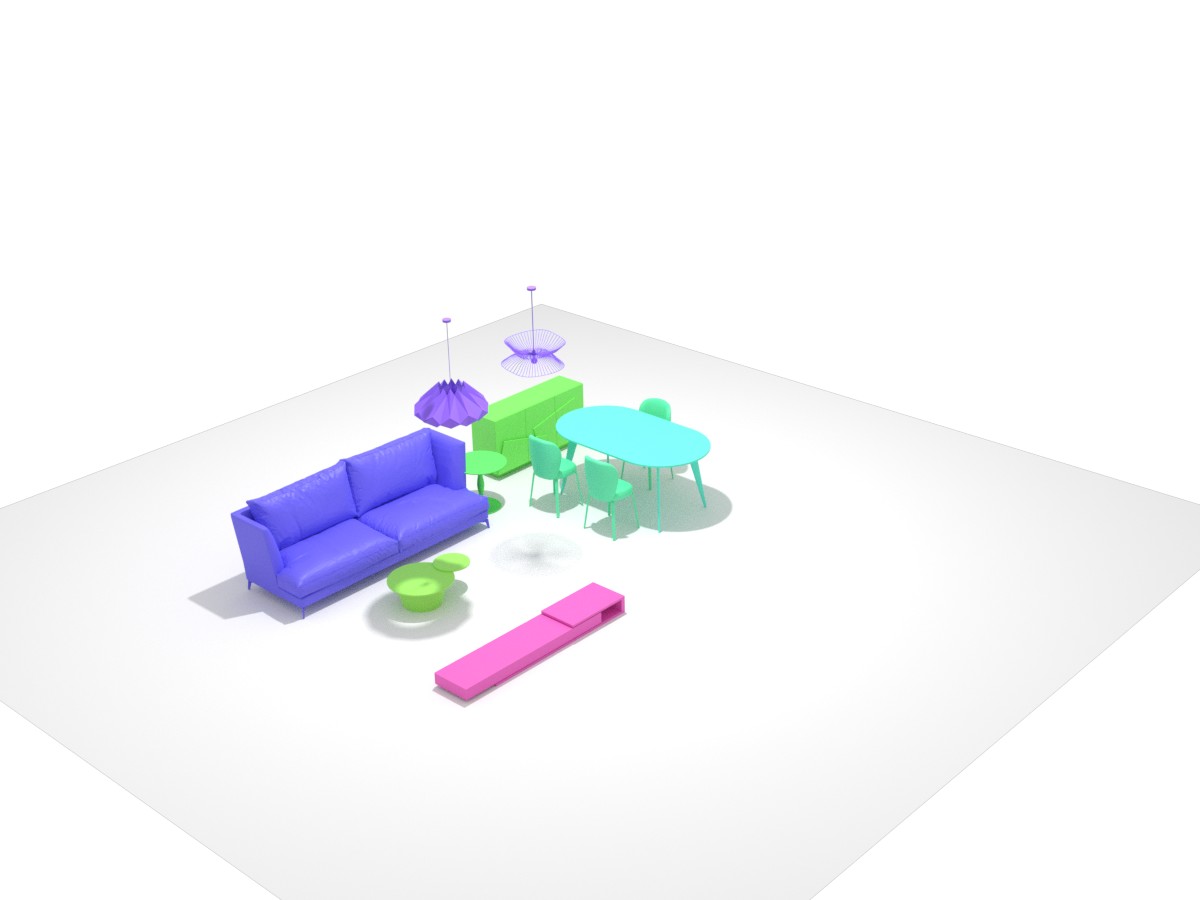}
		\caption{Ours}
	\end{subfigure}
	\caption{\textbf{Scene re-arrangements} of collections of random objects.  Compared to ATISS and LEGO, our method generates various object placement options
        with better plausibility and more symmetries.}
    \label{fig:arrangement_lego_supple}
\end{figure*}

\paragraph{Scene Arrangement}
We visualize additional qualitative comparisons on the task of scene arrangement in Fig.~\ref{fig:arrangement_lego_supple}.  
LEGO~\cite{wei2023lego} aims to predict 2D object locations and orientations, taking the input of a floor plane, object semantics and geometries. It does not handle objects like lamps that could hang from the ceiling. 
In contrast, DiffuScene is a scene-generative model that predicts 3D instance properties from random noise, including 3D locations and orientations, semantics, and geometries.  
Compared to ATISS and LEGO, our method generates various object placement options with better plausibility and more symmetries.

\paragraph{Scene Completion}
We present more qualitative comparisons on the task of scene completion in Fig.~\ref{fig:completion_supple}.
Also, the quantitative results are shown in Tab.~\ref{tab:completion}.  Compared to ATISS, our method produced more diverse completion results with higher fidelity. Our method can consistently outperform ATISS  in all listed metrics.
\begin{table}[!hbt]
    \renewcommand\arraystretch{1.2}
    \setlength{\tabcolsep}{2.4pt}
	\begin{center}
            \begin{tabular}{*{6}{c}}
			\toprule
			{Room} & {Method} & FID $\downarrow$ & KID $\downarrow$ 
                   & \#Sym. & PIoU \\  
			\midrule
			\midrule
            \multirow{2}*{Bed} &  ATISS & 30.54  & 2.38 
                                        &  0.01 & 0.84  \\  
                                  
                                  &  Ours  & \textbf{27.32} & \textbf{1.92} 
                                & \textbf{0.47} & \textbf{0.61} \\

        \midrule
        \multirow{2}*{Dining} & ATISS & 42.65  & 8.32 
                                        &  1.42  & 1.73 \\ 
                    &  Ours  & \textbf{40.99} & \textbf{6.31} 
                               & \textbf{2.57} & \textbf{0.84} \\
                \midrule
        \multirow{2}*{Living} & ATISS & 43.30  & 5.22  
                                        &  0.16  &  0.87\\ 
                              
                              &  Ours  & \textbf{40.49} & \textbf{4.59}  
                              & \textbf{2.24} & \textbf{0.58} \\ 

        \bottomrule
        \end{tabular}
        \caption{Quantitative comparisons on the task of \textbf{scene completion} on 3D-FRONT bedrooms, dining rooms, and living rooms. Only 3 objects are given in the partial scenes. }
        \label{tab:completion}
        \end{center}
        \vspace{-6mm}
\end{table}

\paragraph{Real-world Scene Generalization}
%
While trained on synthetic dataset, our method can be evaluated on real-world scenes without finetuning, e.g. for scene completion as shown in Fig.~\ref{fig:real_world_completion}.
Compared to ATISS, our method produces a more favourable scene.

\paragraph{Text-conditioned Scene Synthesis}
\begin{figure*}[!htbp]
	\centering
	\begin{subfigure}[t]{0.24\textwidth}
            \includegraphics[width=\textwidth]{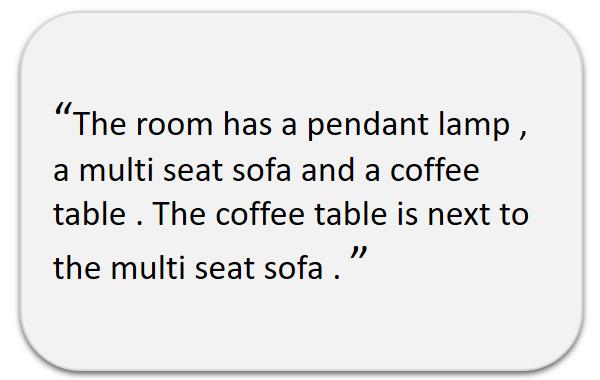}
            \vspace{2mm}
            \includegraphics[width=\textwidth]{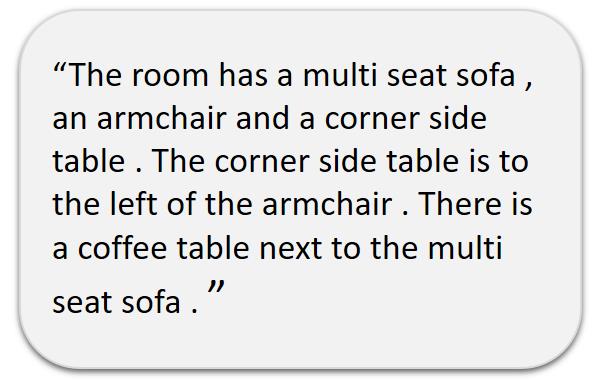}
            \vspace{2mm}
            \includegraphics[width=\textwidth]{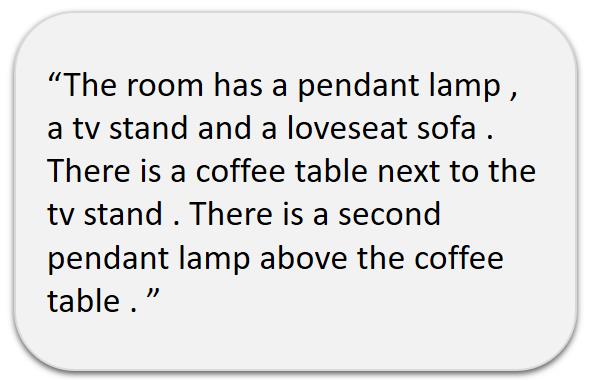}
            \vspace{2mm}
            \includegraphics[width=\textwidth]{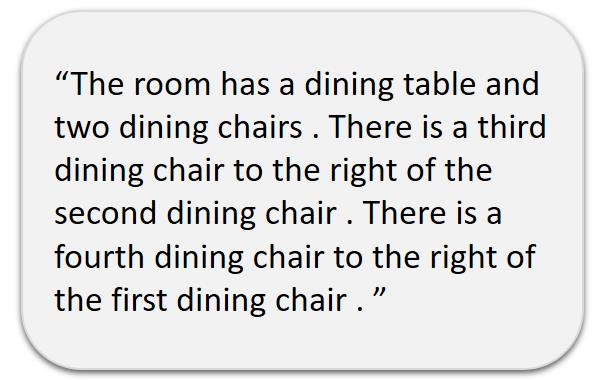}
        \caption{Input text}
	\end{subfigure}%
        \hfill
 	\begin{subfigure}[t]{0.215\textwidth}
            \includegraphics[width=\textwidth]{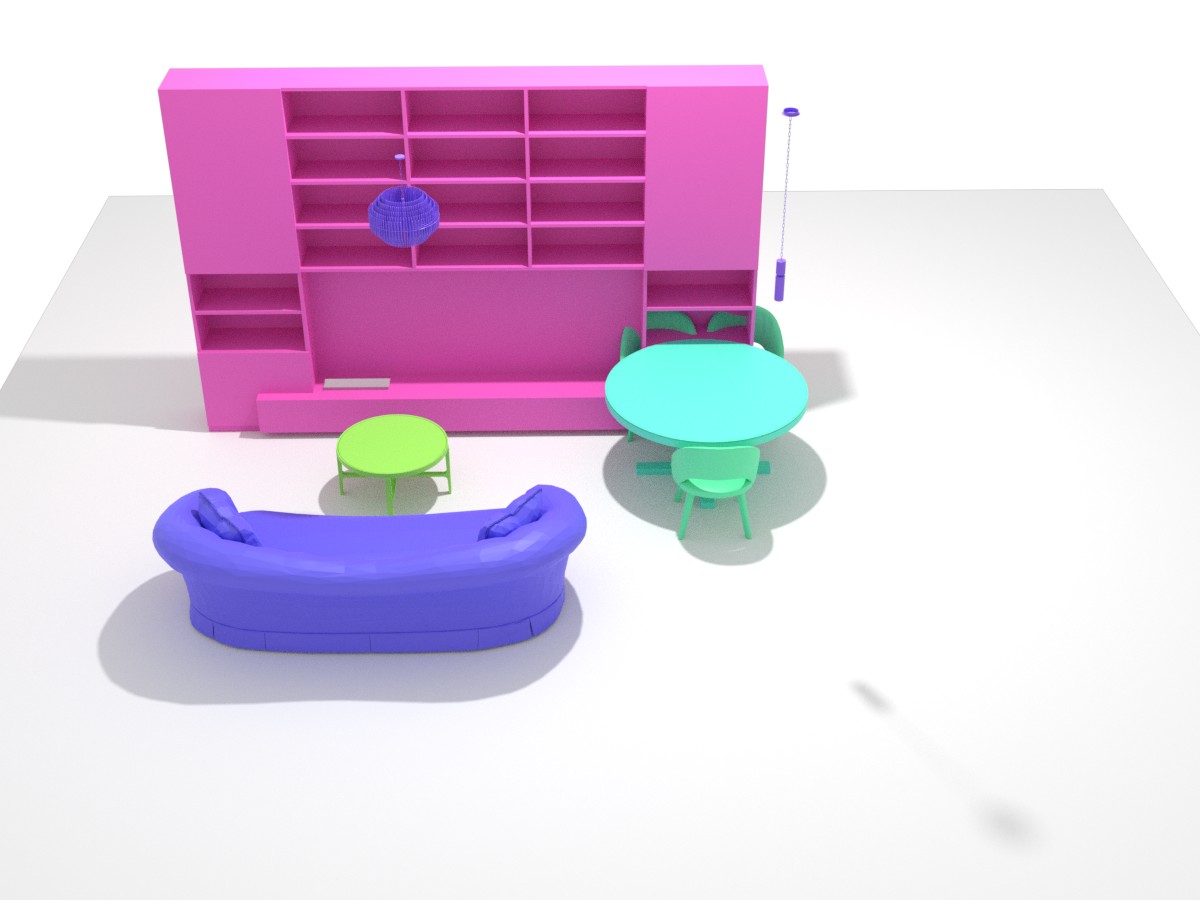}
            \includegraphics[width=\textwidth]{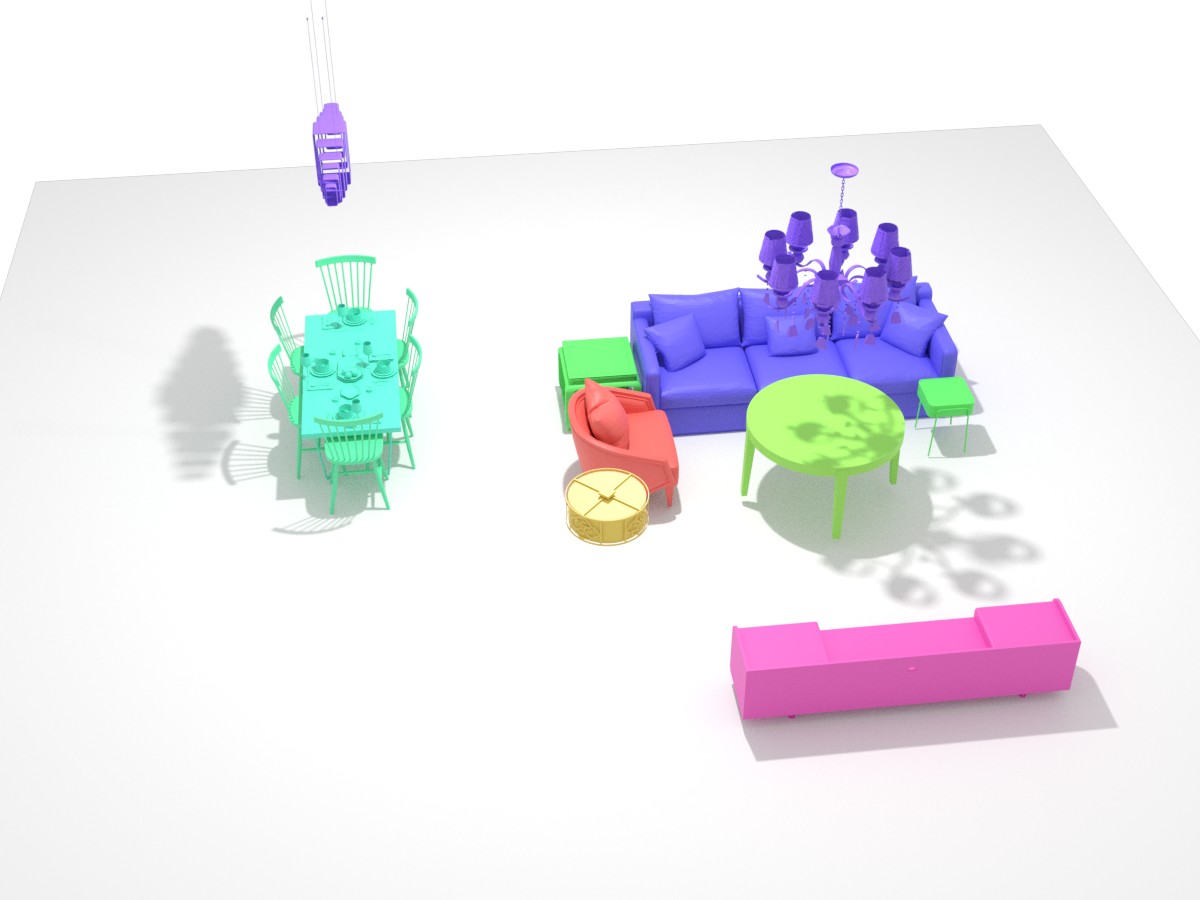}
            \includegraphics[width=\textwidth]{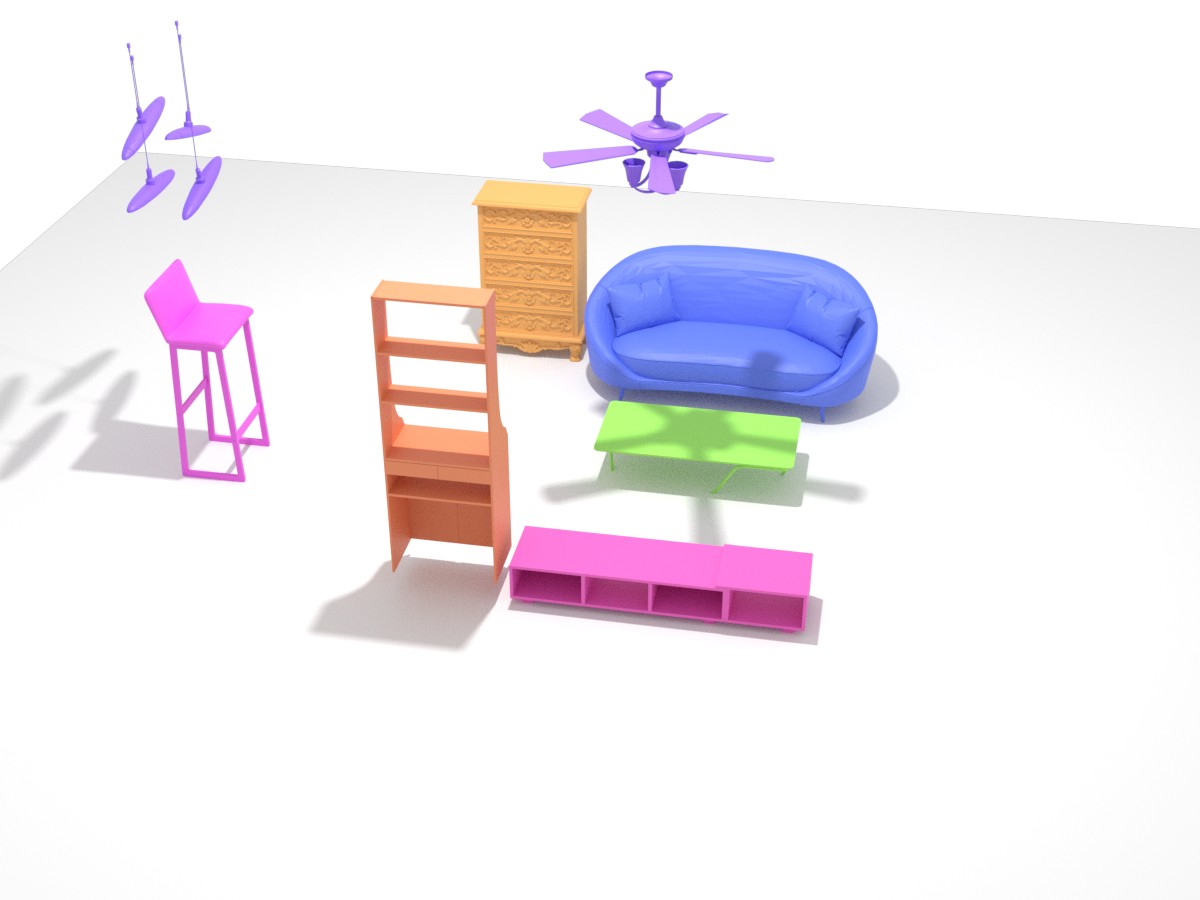}
            \includegraphics[width=\textwidth]{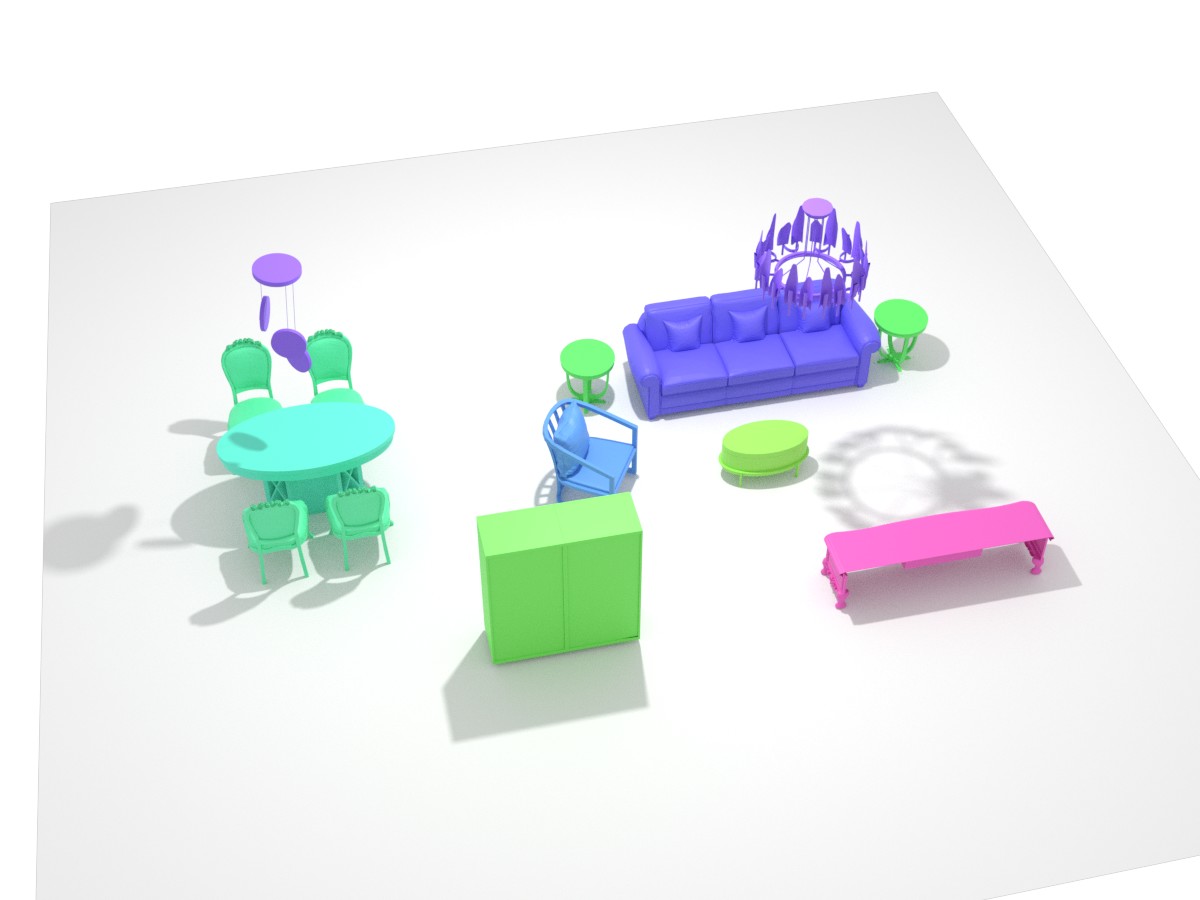}
        \caption{Reference}
	\end{subfigure}%
        \hfill
 	\begin{subfigure}[t]{0.215\textwidth}
            \includegraphics[width=\textwidth]{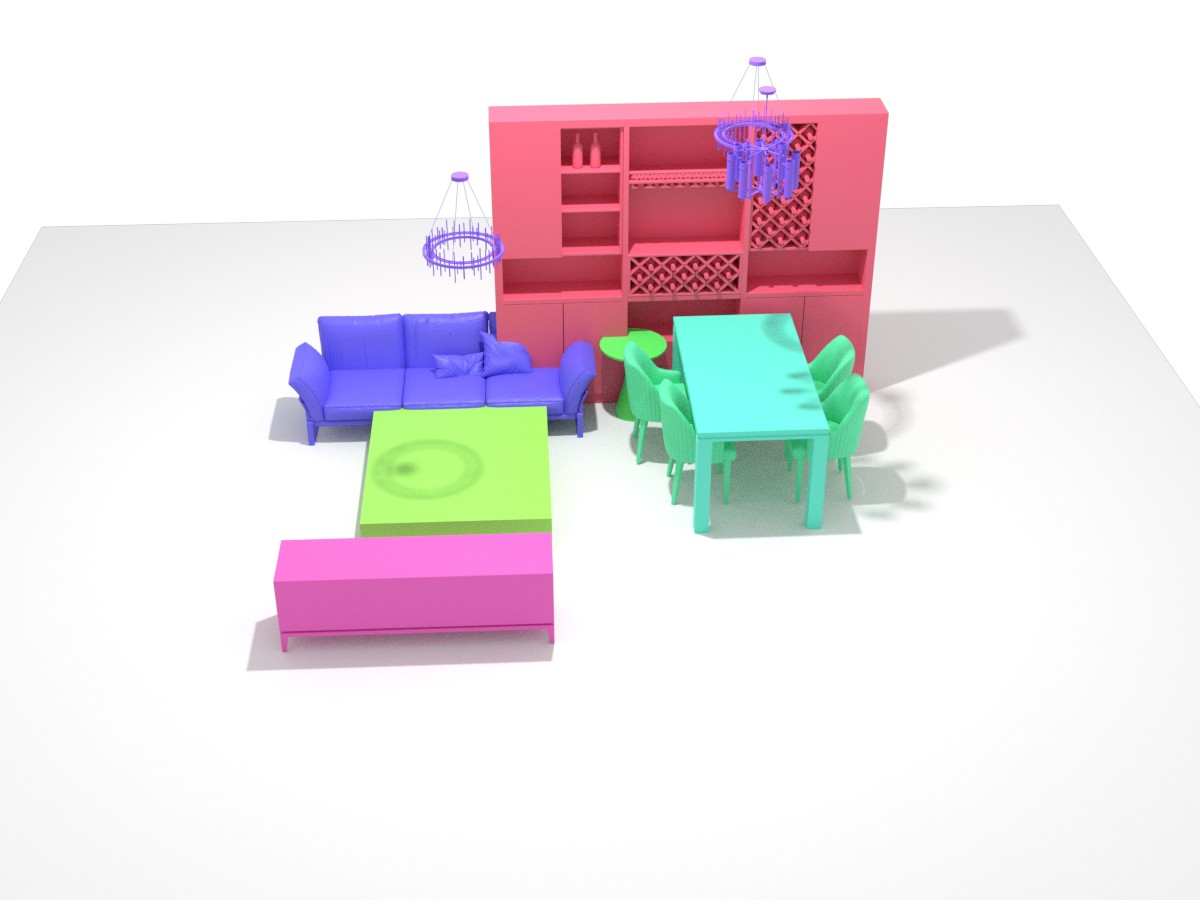}
            \includegraphics[width=\textwidth]{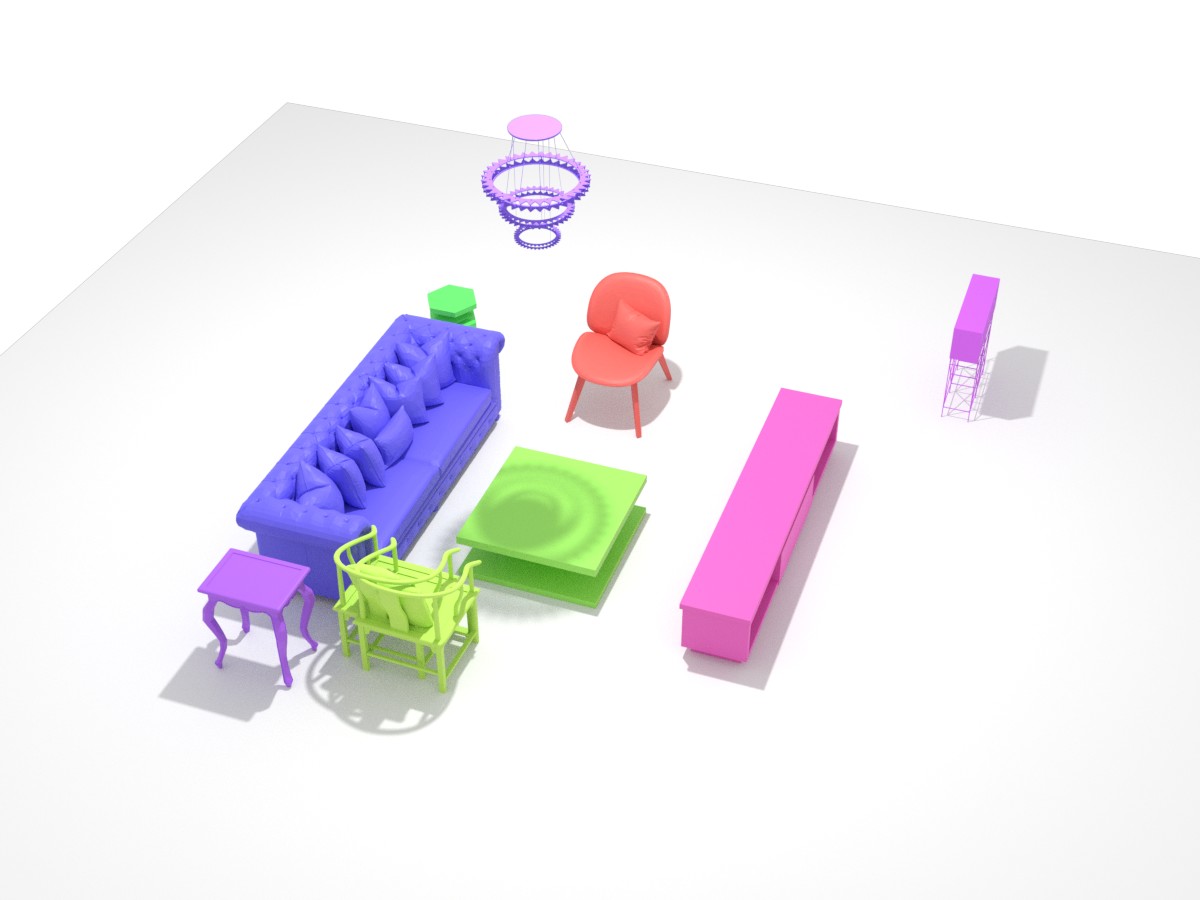}
            \includegraphics[width=\textwidth]{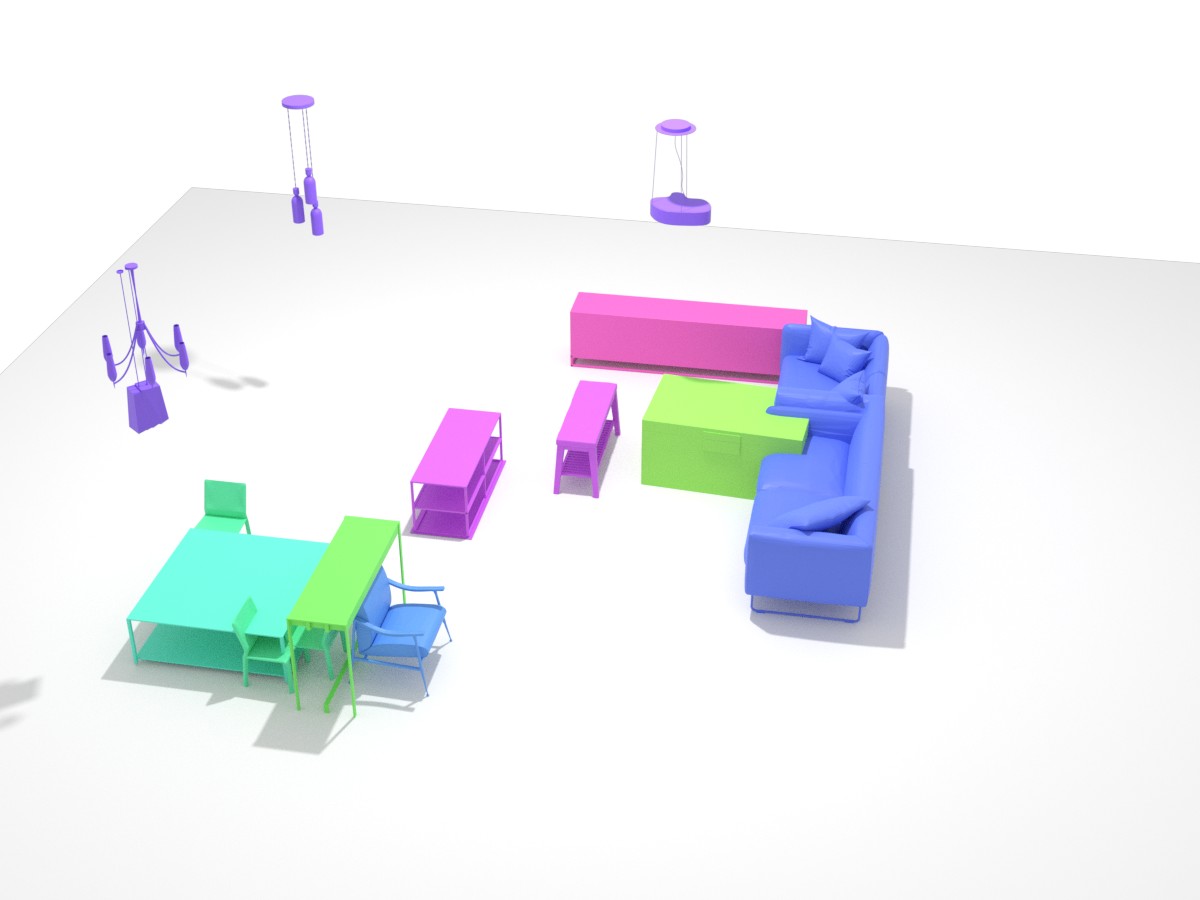}
            \includegraphics[width=\textwidth]{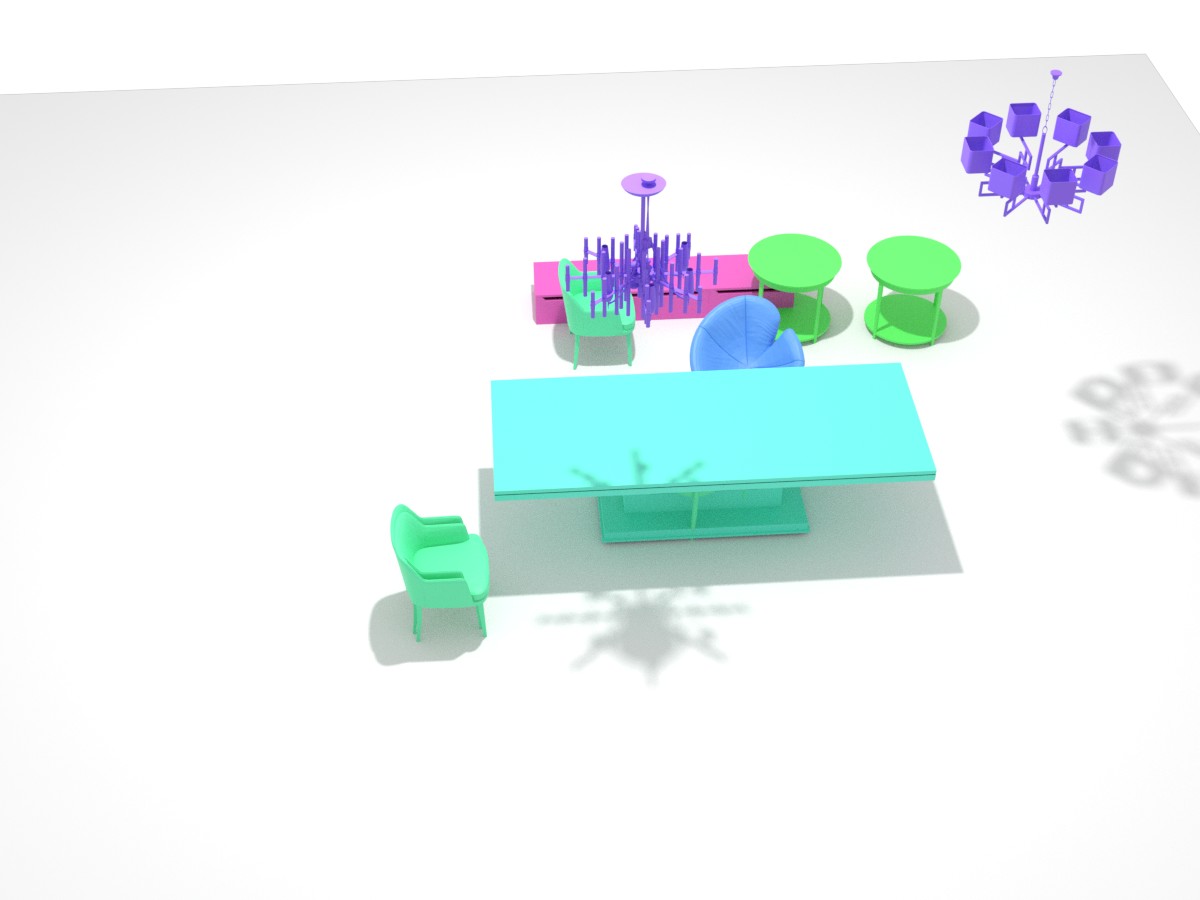}
        \caption{ATISS~\cite{paschalidou2021atiss}}
	\end{subfigure}%
        \hfill
 	\begin{subfigure}[t]{0.215\textwidth}
            \includegraphics[width=\textwidth]{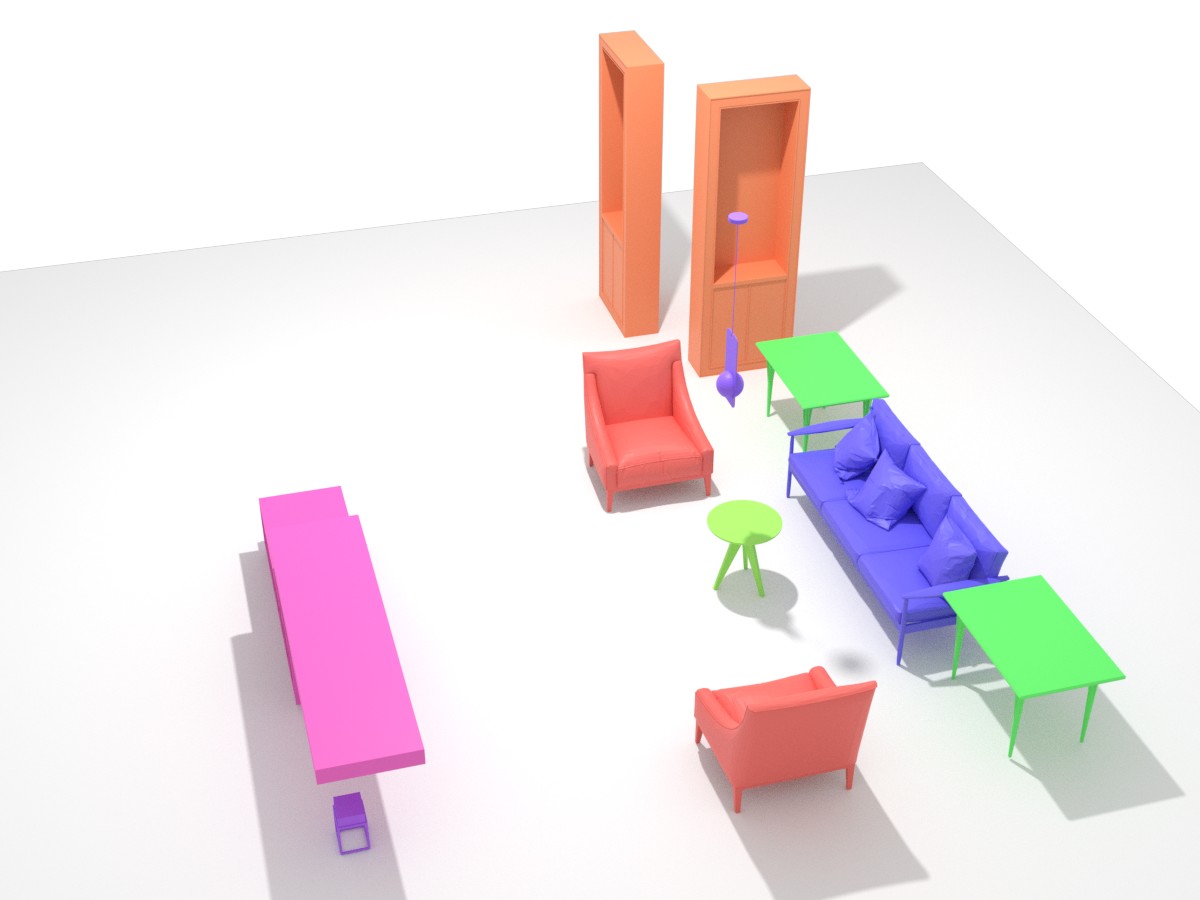}
            \includegraphics[width=\textwidth]{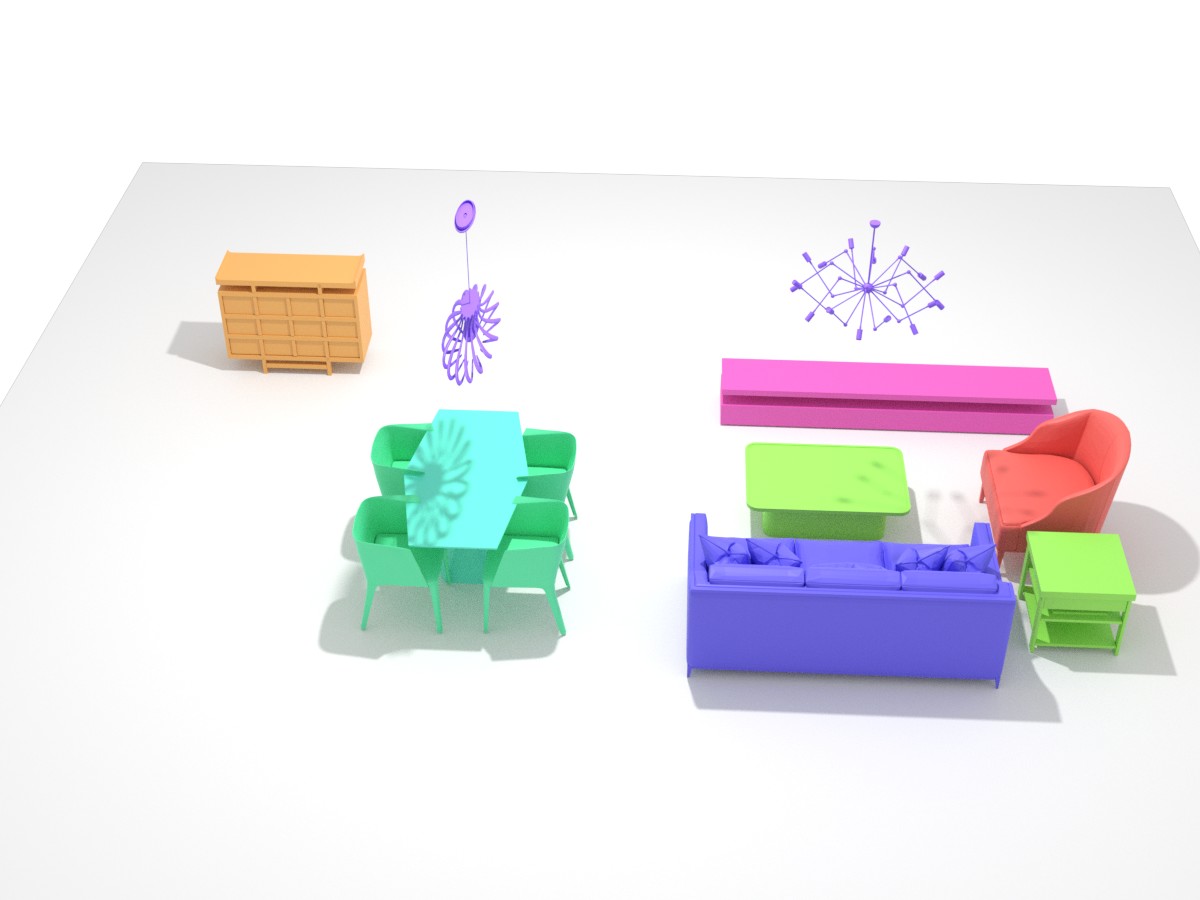}
            \includegraphics[width=\textwidth]{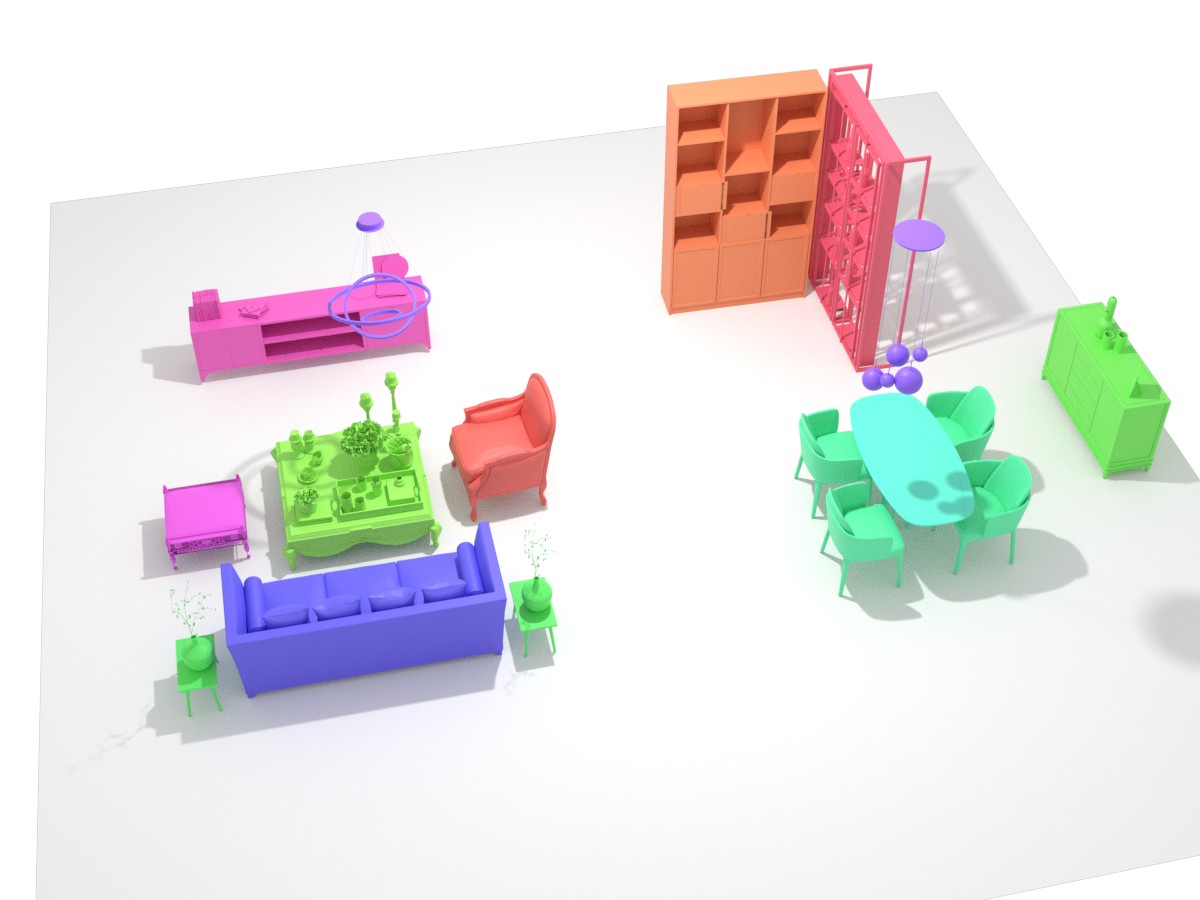}
            \includegraphics[width=\textwidth]{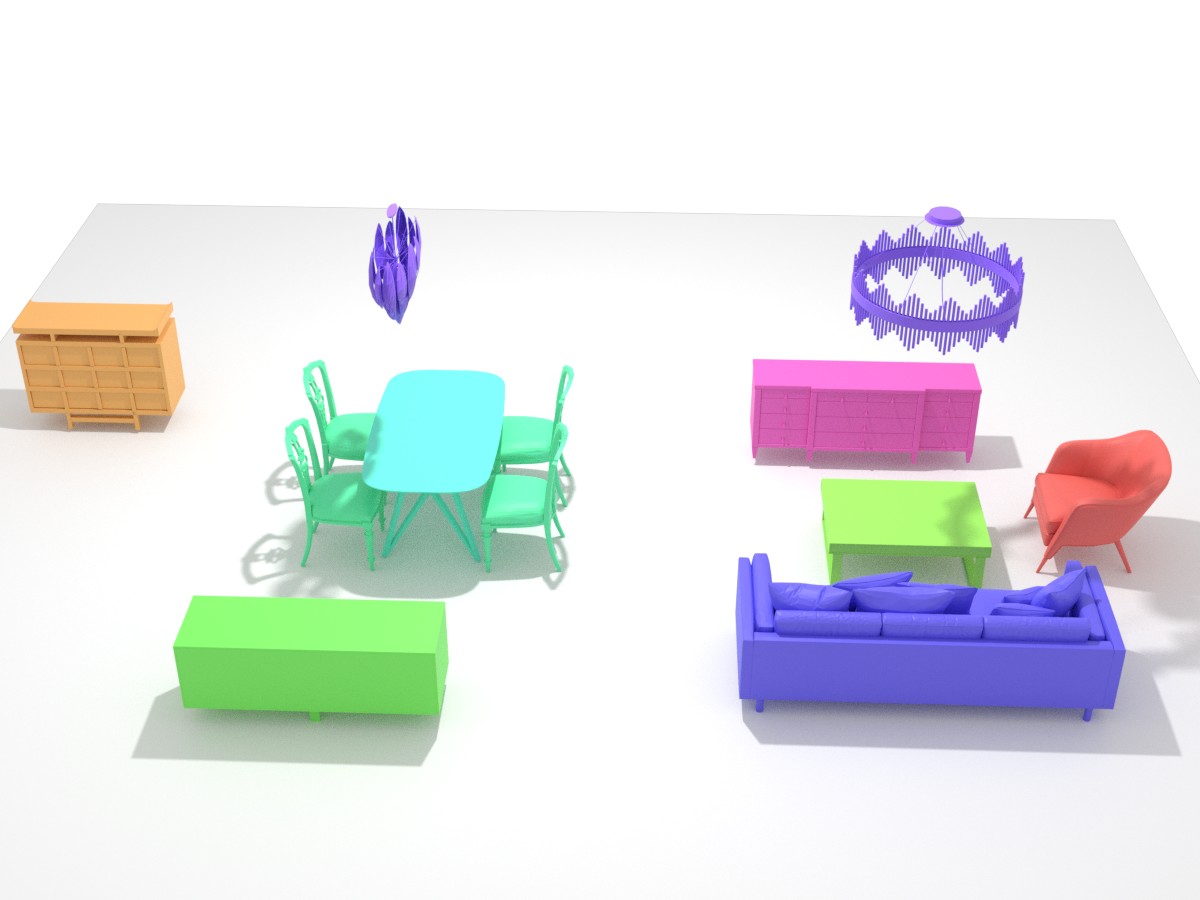}
        \caption{Ours}
	\end{subfigure}
	\caption{\textbf{Text-conditioned scene synthesis}. The input text describes only a partial scene configuration. Our method generates more plausible scenes matched with the texts.}
    \label{fig:text2scene_supple}
\end{figure*}
We provide additional qualitative comparisons on the text-conditioned scene synthesis in Fig.~\ref{fig:text2scene_supple}. 
As observed, in the first and third rows, ATISS has object intersection issues while ours does not. In the second row, our method can correctly generate a corner side table on the left of the armchair. However, ATISS generates a corner side table on the right of the armchair.
 In the fourth row, our method can generate four dining chairs that are consistent with the text description, but ATISS can only generate two dining chairs.
%
\paragraph{Scene editing via texts.} 
In Fig.~\ref{fig:text_editing}, we show that our method can support text-guided object suggestion and scene editing, without changing the attributes of other objects.

\section{User Study}
\label{SecUser}

\begin{figure*}[!htbp]
    \centering
    \includegraphics[width=\linewidth]{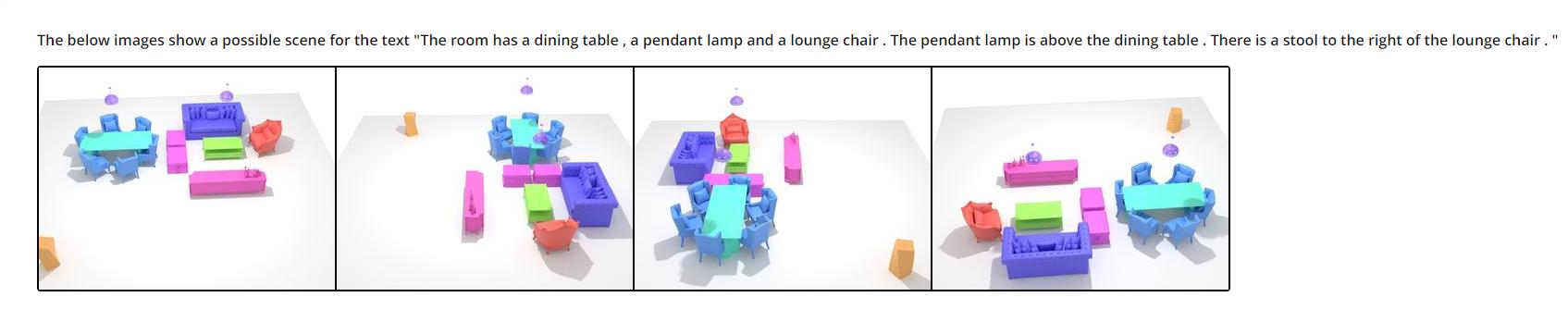}
    
    \includegraphics[width=\linewidth]{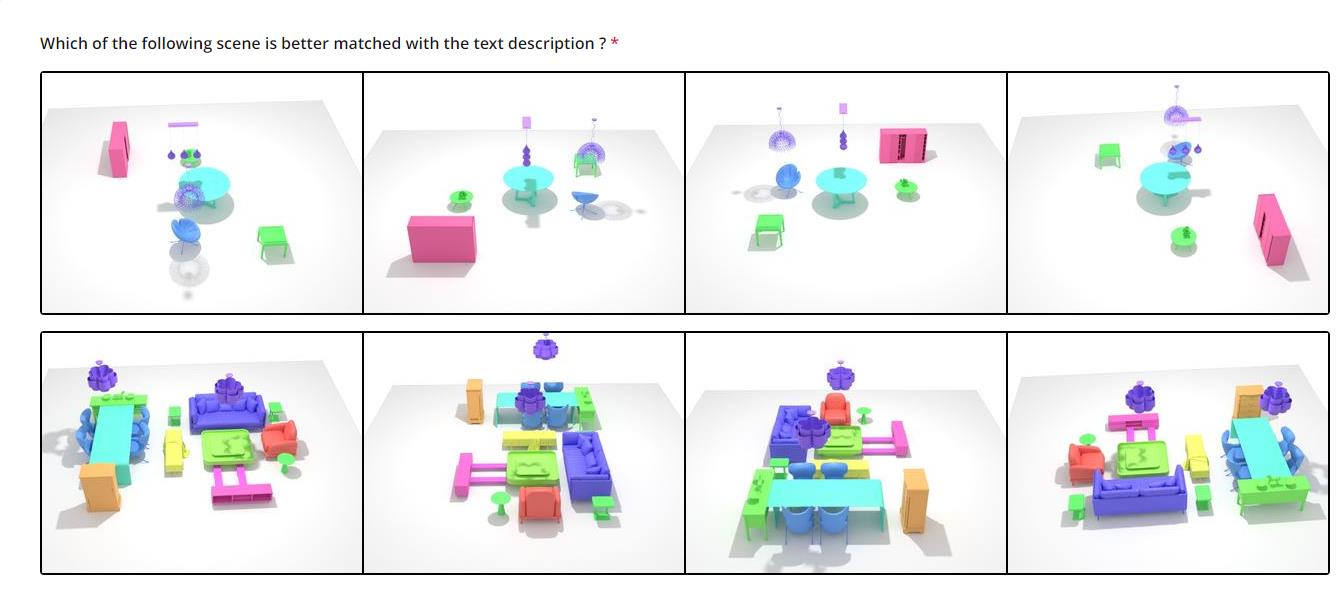}

    \includegraphics[width=\linewidth]{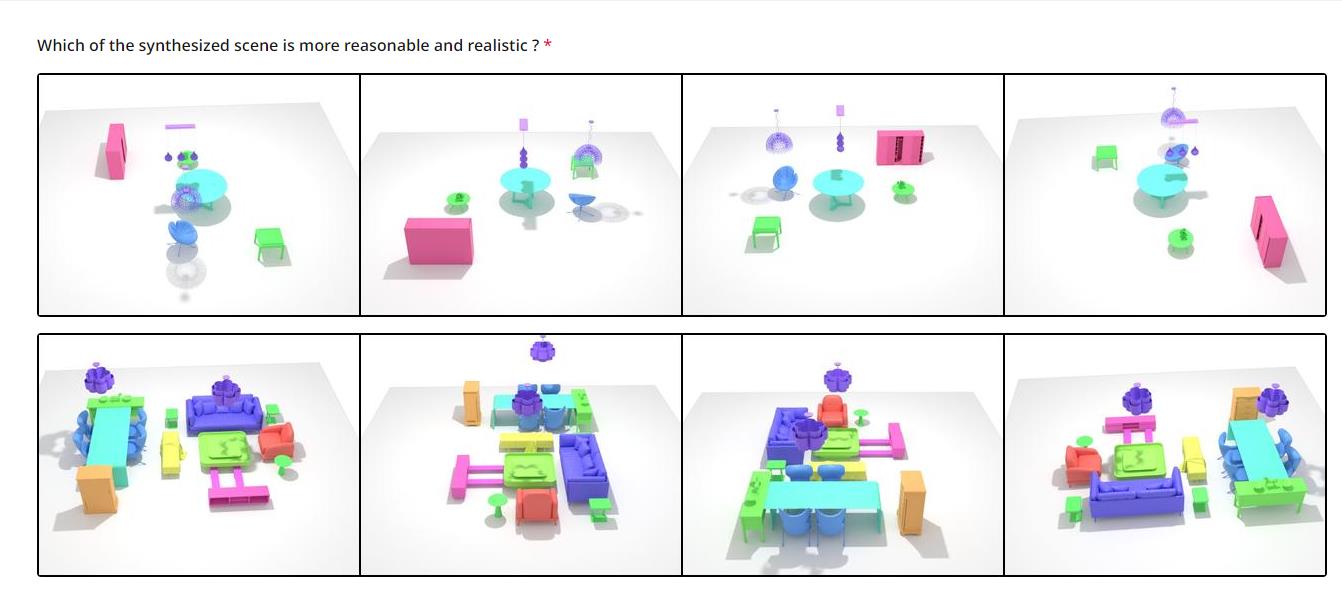}

    \caption{\textbf{User Study UI}. Based on the reference scene used to generate text prompts, users are asked which of the synthesized scene is more matched with the text prompt and more realistic. Note that the results from ATISS and our method are randomly shuffled to avoid bias.}
    \label{fig:user_study}
    
\end{figure*}

We conducted a perceptual user study to evaluate the quality of our method against ATISS on the application of text-conditioned scene synthesis.
As shown in Fig.~\ref{fig:user_study}, we provide the visualization of a ground-truth scene used to generate a text prompt as a reference. For each pair of results, a user needs to answer ``which of the generated scene can better match the text prompt?" and ``Which of the generated scene is more reasonable and realistic?".
We collect the answers of 225 scenes from 45 users and calculate the statistics. 62$\%$ of the user answers prefer our method to ATISS in realism.  55$\%$ of answers think our method is more consistent with the text prompt.

\end{document}